\documentclass[10pt,twocolumn,letterpaper]{article}

\usepackage[pagenumbers]{cvpr} 
\definecolor{cvprblue}{rgb}{0.21,0.49,0.74}
\usepackage[pagebackref,breaklinks,colorlinks,citecolor=cvprblue,linkcolor=cvprblue]{hyperref}
\usepackage{graphicx}
\usepackage{amsmath}
\usepackage{amssymb}
\usepackage{tabularx}
\usepackage{booktabs}
\usepackage{tikz}
\usepackage{adjustbox}
\usepackage{makecell}

\usepackage[ruled]{algorithm2e} 

\usepackage[pagebackref,breaklinks,colorlinks]{hyperref}

\usepackage[capitalize]{cleveref}
\crefname{section}{Sec.}{Secs.}
\Crefname{section}{Section}{Sections}
\Crefname{table}{Table}{Tables}
\crefname{table}{Tab.}{Tabs.}

\usepackage{todonotes}
\definecolor{Violet}{HTML}{dbaffc}
\definecolor{Green}{HTML}{d9fcaf}
\definecolor{Blue}{HTML}{afedfc}
\definecolor{Pink}{HTML}{FFCFCF}
\definecolor{Orange}{HTML}{fcd3af}

\def\paperID{12158} 
\def\confName{CVPR}
\def\confYear{2025}

\newcommand{\imageheight}{2.9cm}
\newcommand{\figureheight}{2.2cm}
\newcommand{\imageheightqual}{3.1cm}
\newcommand{\ablationheight}{2.5cm}
\newcommand{\extrafigureheight}{3.4cm}
\newcommand{\specialeffectheight}{4.3cm}
\title{SynthLight: Portrait Relighting with Diffusion Model \\ by Learning to Re-render Synthetic Faces}

\author{Sumit Chaturvedi$^{1 *}$ \quad Mengwei Ren$^2$ \quad Yannick Hold-Geoffroy$^2$ \quad Jingyuan Liu$^2$ \\ \quad Julie Dorsey$^1$ \quad Zhixin Shu$^{2 \dag}$ \vspace{0.2cm}\\
$^1$Yale University \quad $^2$Adobe Research}

\newcommand\blfootnote[1]{%
  \begingroup
  \renewcommand\thefootnote{}\footnote{#1}%
  \addtocounter{footnote}{-1}%
  \endgroup
}

\begin{document}


\twocolumn[{%
\renewcommand\twocolumn[1][]{#1}%
\maketitle
\begin{center}
\centering

    \begin{tabular}{@{\hskip 0mm}c@{\hskip 0.2mm}c@{\hskip 0.1mm}c@{\hskip 0.1mm}c@{\hskip 0.1mm}c@{\hskip 0.1mm}c@{\hskip 0.1mm}c@{\hskip 0.1mm}c@{\hskip 0mm}}
        \includegraphics[height=\imageheight]{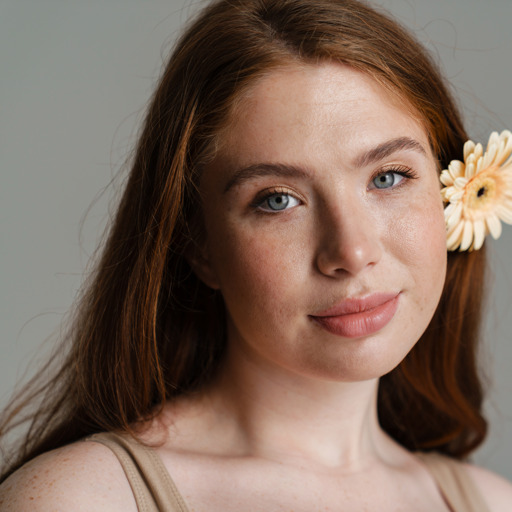} &
        \includegraphics[height=\imageheight]{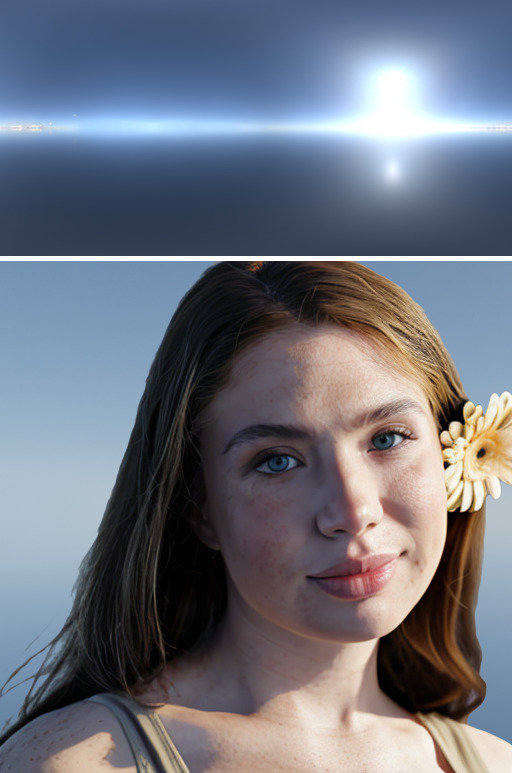} &
        \includegraphics[height=\imageheight]{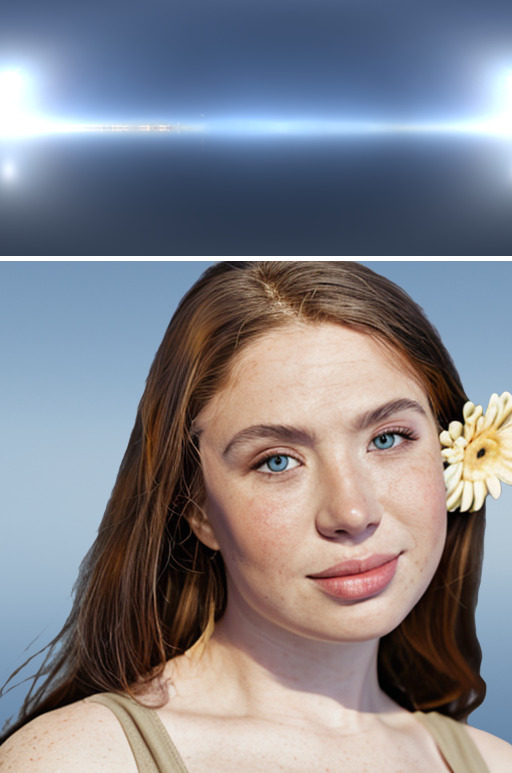} &
        \includegraphics[height=\imageheight]{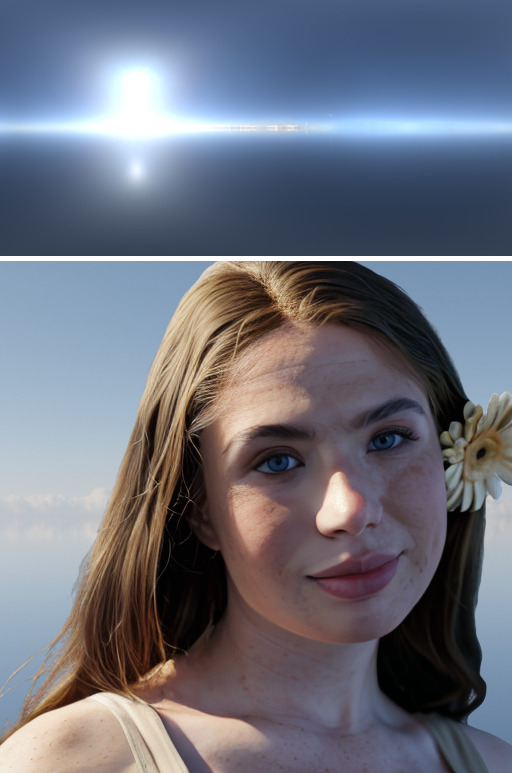} &
        \includegraphics[height=\imageheight]{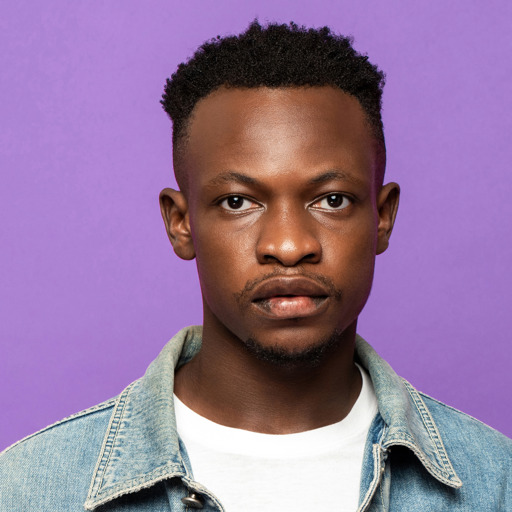} &
        \includegraphics[height=\imageheight]{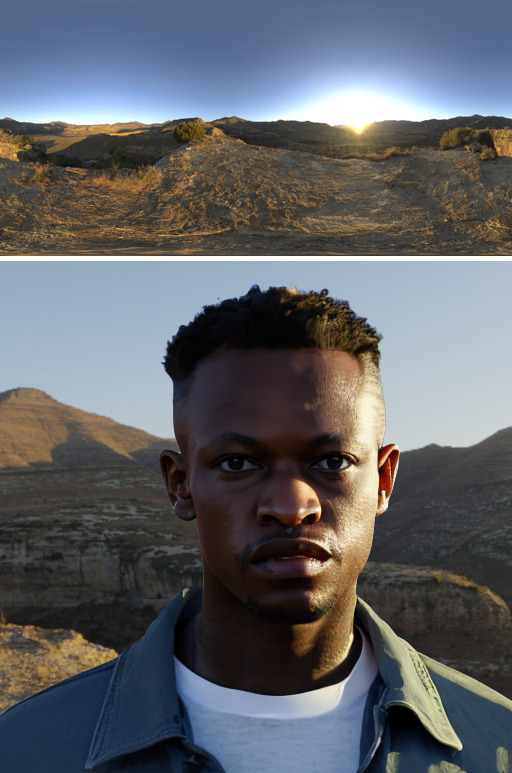} &
        \includegraphics[height=\imageheight]{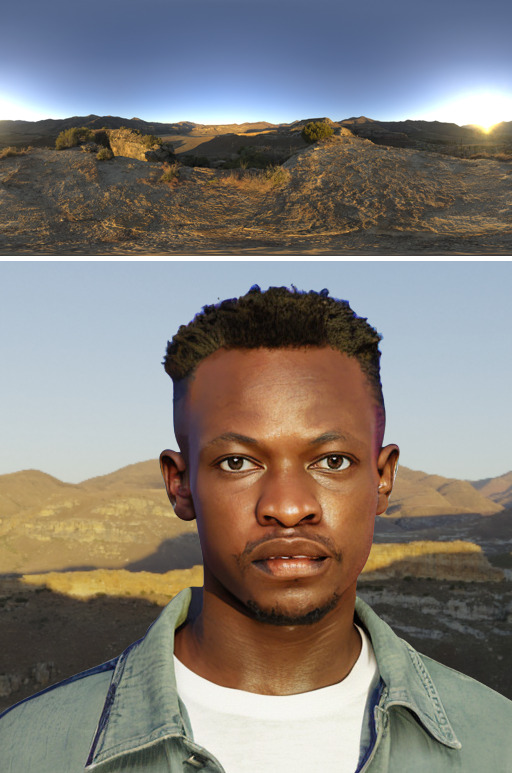} &
        \includegraphics[height=\imageheight]{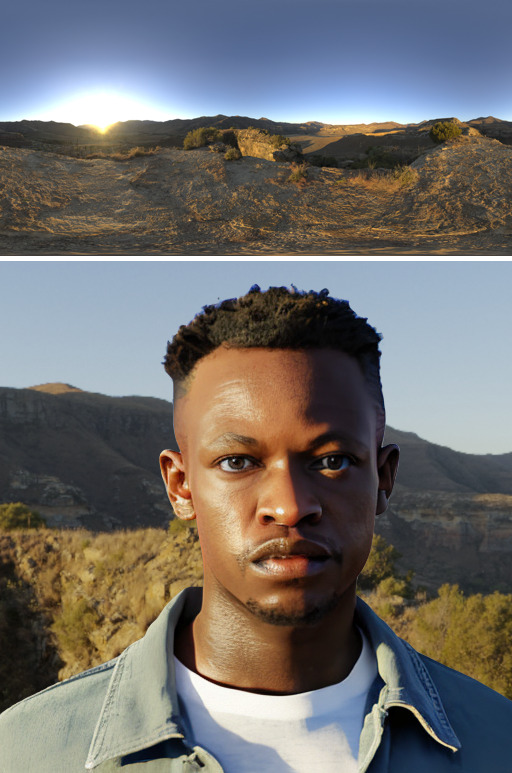}\\[5pt]
        \includegraphics[height=\imageheight]{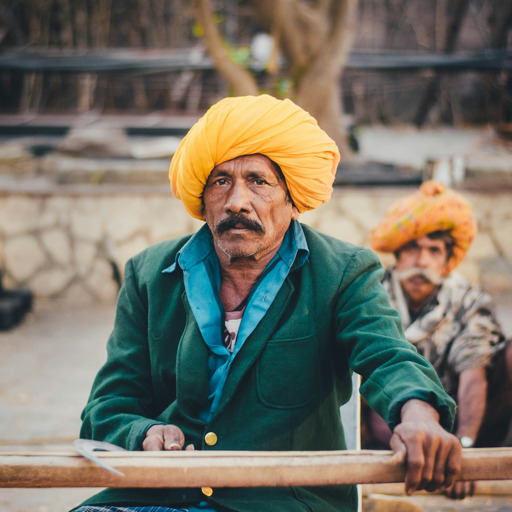} &
        \includegraphics[height=\imageheight]{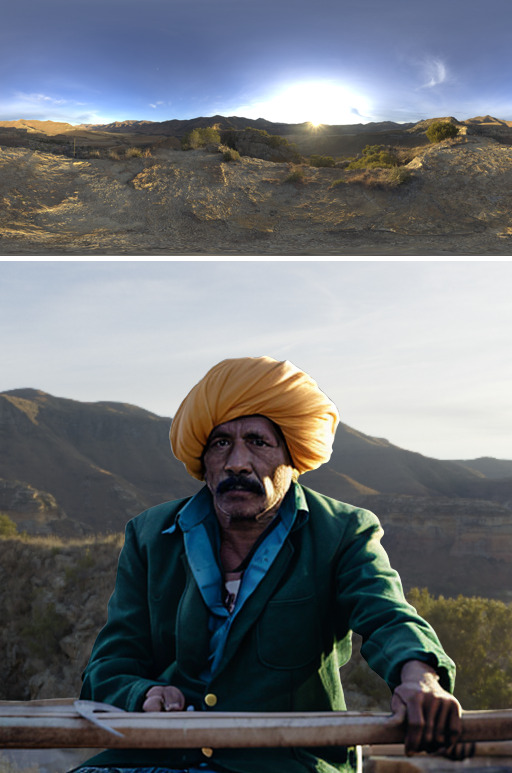} &
        \includegraphics[height=\imageheight]{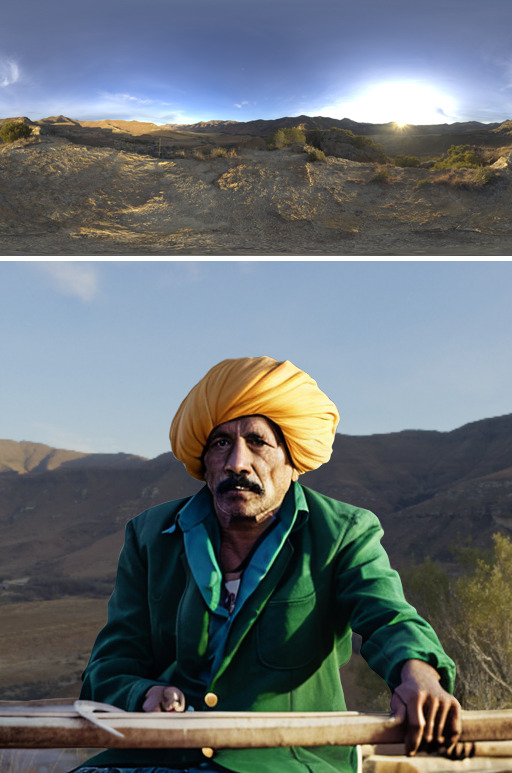} &
        \includegraphics[height=\imageheight]{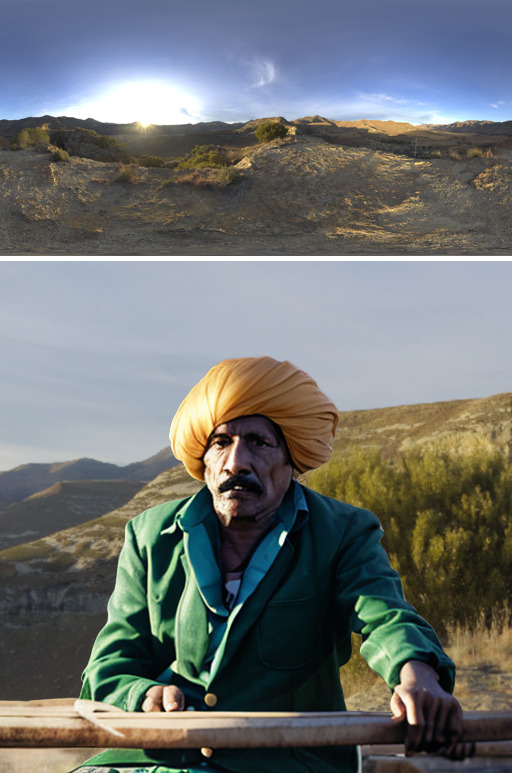}& 
        \includegraphics[height=\imageheight]{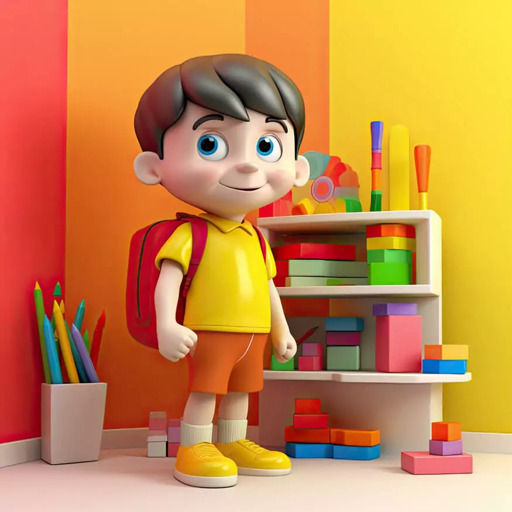} &
        \includegraphics[height=\imageheight]{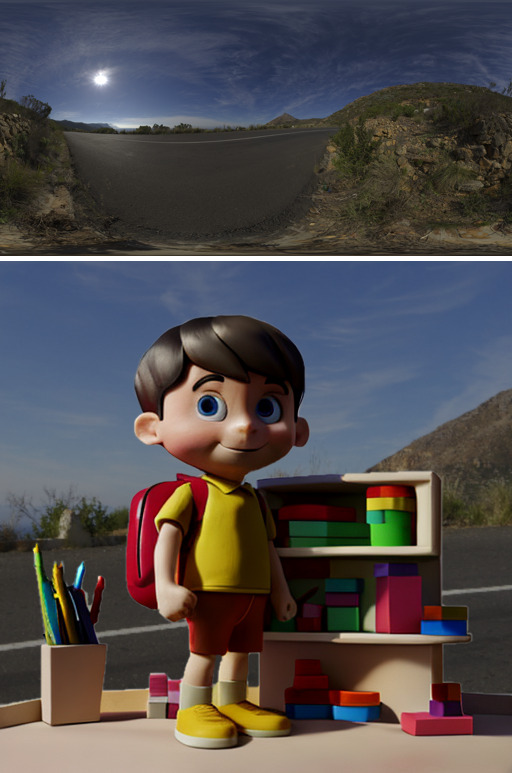} &
        \includegraphics[height=\imageheight]{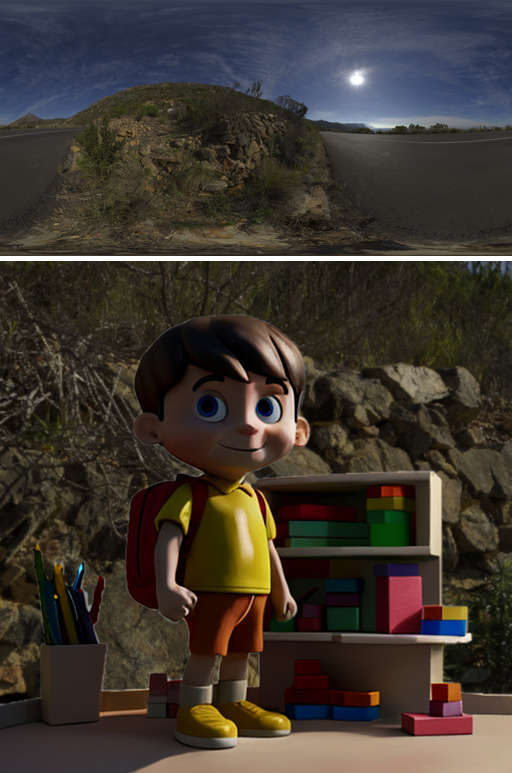} &
        \includegraphics[height=\imageheight]
        {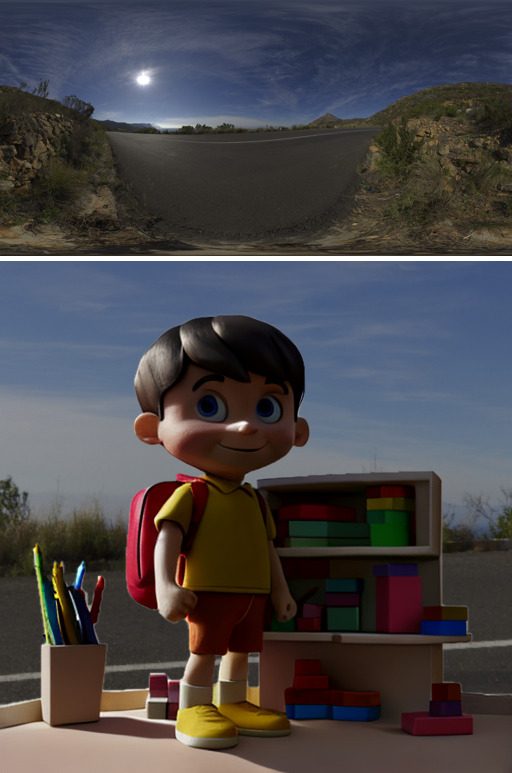} \\
    \end{tabular}
    \captionof{figure}{SynthLight performs relighting on portraits using an environment map lighting. By learning to re-render synthetic human faces, our diffusion model produces realistic illumination effects on real portrait photographs, including distinct cast shadows on the neck and natural specular highlights on the skin. Despite being trained exclusively on synthetic headshot images for relighting, the model demonstrates remarkable generalization to diverse scenarios, successfully handling half-body portraits and even full-body figurines. } 
    \label{fig:teaser}
\end{center}
}]

\blfootnote{*Work done as an intern at Adobe Research.}
\blfootnote{\dag Corresponding author.}

\begin{abstract}

We introduce SynthLight, a diffusion model for portrait relighting. Our approach frames image relighting as a re-rendering problem, where pixels are transformed in response to changes in environmental lighting conditions. Using a physically-based rendering engine, we synthesize a dataset to simulate this lighting-conditioned transformation with 3D head assets under varying lighting. We propose two training and inference strategies to bridge the gap between the synthetic and real image domains: (1) multi-task training that takes advantage of real human portraits without lighting labels; (2) an inference time diffusion sampling procedure based on classifier-free guidance that leverages the input portrait to better preserve details. Our method generalizes to diverse real photographs and produces realistic illumination effects, including specular highlights and cast shadows, while preserving the subject's identity. Our quantitative experiments on Light Stage data demonstrate results comparable to state-of-the-art relighting methods. Our qualitative results on in-the-wild images showcase rich and unprecedented illumination effects. Project Page: \url{https://vrroom.github.io/synthlight/}

\end{abstract}

\section{Introduction}

Lighting is fundamental to portrait photography, yet manipulating it after capture remains challenging. Recent advances in generative imaging models have demonstrated promising capabilities for controlling lighting in existing images \cite{he2024diffrelight, ren2024relightful, zeng2024dilightnet, jin2024neural_gaffer, iclight}. However, these approaches typically require labeled training data. For portrait relighting specifically, the most effective results have come from training on Light Stage data---portraits rendered with linear combinations of one-light-at-a-time (OLAT) captures. While powerful, Light Stage setups are constrained by physical limitations in light source density and require specialized artificial lighting equipment. In contrast, 3D workflows in VFX and gaming have long treated lighting as a relatively straightforward endeavor through modern physically based rendering engines, where light source control is nearly arbitrary. To relight a rendering, artists simply adjust the lighting configurations and re-render the scene.

Given a scene $S$ and lighting $L_1$, we denote the rendering as $I_1 = f_r(S,L_1)$.
The inverse graphics problem aims to find $S$ from $I_1$:
$S = f_{inv}(I_1,L_1)$
with known or unknown lighting.
To relight rendering $I_1$ to $I_2$ under lighting $L_2$, one aims to compute:
$I_2 = f_r(S,L_2)$.
Given only $I_1$, a relighting procedure seeks:
$I_2 = f_r(f_{inv}(I_1),L_2) = f_{re}(I_1,L_2)$, where $f_{re}$ is the relighting/re-rendering function.
Previous approaches \cite{pandey2021total, kim2024switchlight, yeh2022learning} have tackled this problem through inverse graphics, either explicitly or implicitly, by estimating lighting-invariant intrinsic scene representations such as depth, surface normals, and albedo. This imposes limitations on subsequent rendering functions and often fails to capture complex illumination effects like inter-reflections, occlusion shadows, and subsurface scattering. In this paper, we propose bypassing inverse rendering entirely by learning the relighting function using physically based 3D renderings of human heads. Specifically, we render pairs of portrait images using Blender (Cycles) $(I_1, L_1)$ and $(I_2, L_2)$ and train a diffusion model to directly learn to ``re-render'' $I_2$ from $I_1$ and $L_2$.

However, this approach introduces an inevitable domain gap between simulated 3D renders and real photographs. To address this challenge, we leverage a latent diffusion model pretrained on vast internet images for text-to-image generation. We propose to finetune the network with our face renderings and introduce simple yet effective training and testing schemes to narrow the gap between training data and in-the-wild images. During training, we propose \textbf{multi-task} training that incorporates in-the-wild images without ground truth relighting information. This allows the model to learn relighting from our synthetic dataset while maintaining knowledge of the real image domain, preventing distributional drift. We further observe that input portraits contain rich textural information. Leveraging the flexibility of diffusion model inference, we design an \textbf{inference time adaptation} scheme that effectively preserves input portrait details in the relit result.

We evaluate our methods on in-the-wild portrait images, demonstrating highly detailed illumination effects that accurately capture interactions between the portrait scene and lighting. Our results produce realistic cast shadows and specular highlights on the skin. For the first time, we demonstrate an end-to-end system capable of non-trivial lighting effects including catch lights in eyes, subsurface scattering in ears, and inter-reflections with clothing. Notably, despite training only on simple headshot renderings of 3D faces without accessories, facial hair, or hats, our network generalizes effectively to complex portrait images, including half-body shots and multi-person photographs.

We quantitatively evaluate our method on a test set of our synthetic faces dataset as well as on a Light Stage OLAT dataset. Despite using no Light Stage data for training, our method achieves comparable or superior results to state-of-the-art portrait relighting methods trained on OLAT data. User studies show that our results are preferred across all evaluated aspects, including perceptual lighting accuracy, identity preservation, and overall image quality.

We summarize our contributions as follows:
\begin{enumerate}
   \item We propose modeling portrait relighting as a task of learning to re-render a portrait scene in 3D. Using physically based renderings of human heads under varying lighting conditions, we train a diffusion model to learn pixel transformations conditioned on lighting.
   \item We introduce two techniques enabling synthetic data learning while minimizing domain gap with real images, through the use of a training-time multi-task strategy that incorporates real images through a text-to-image task, and an inference-time approach based on classifier-free guidance that preserves portrait details in the relit result.
   \item Through extensive qualitative and quantitative evaluations, we demonstrate state-of-the-art portrait relighting results, achieving high-quality lighting effects previously unattainable by existing methods.
\end{enumerate}

\section{Related Work} \label{section:relwork}

\subsection{Portrait Relighting}

Portrait relighting has been explored in both 2D ~\cite{pandey2021total, kim2024switchlight, iclight, ren2024relightful, jin2024neural_gaffer, zeng2024dilightnet, sun2019single, wang2020single, paris2003lightweight, shu2017neural} and 3D domains ~\cite{rao2024lite2relight, sun2021nelf, tan2022volux, bi2021deep, cai2024real, wang2023sunstage, zhou2019deep}, with 2D image-based approaches being more relevant to our work. Since 2D portrait relighting is under-constrained, various priors have been proposed, such as morphable models \cite{blanz2023morphable} as 3D face priors in \cite{shu2017portrait}, explicit inverse rendering in \cite{barron2014shape, sengupta2018sfsnet}, and a style transfer approach for relighting in \cite{shih2014style}.

Recently, deep learning methods \cite{sun2019single, nestmeyer2020learning} trained on light stage data \cite{debevec2000acquiring} have driven the state-of-the-art for relighting, with \cite{pandey2021total, kim2024switchlight} demonstrating a widely adopted physics-guided architecture for relighting based on image decomposition into intrinsics such as albedo, normals, diffuse, and specular reflectance maps, conditioned on an HDR environment map lighting representation \cite{debevec2008rendering}. However, this formulation presents two main shortcomings. First, the rendering model assumes a BRDF-based reflectance model \cite{cook1982reflectance, phong1998illumination}, where light is reflected directly from the surface point of incidence, thus neglecting other modes of light transport such as subsurface scattering, which are significant in certain types of human skin (e.g., fair skin) \cite{kim2022countering, mashita2011measuring, donner2006spectral}. Additionally, albedo estimation becomes challenging in the presence of face accessories, inter-reflections and face paint \cite{yeh2022learning, kim2024switchlight}. Second, light stage setups inherently limit the types of lighting that can be captured due to restricted light intensity \cite{sun2019single} and lighting resolution \cite{sun2020light}, hindering the ability to learn complex lighting effects such as specular reflections and subsurface scattering. Motivated by these constraints, we employ diffusion models to learn face relighting, without assuming any appearance model, from a synthetic dataset rendered with a physically based renderer that provides input and relit training pairs for supervision. This enables our method to synthesize interesting illumination effects for human portraits such as hard cast shadows, subsurface scattering and inter-reflections.

\subsection{Diffusion Models for Relighting}

Diffusion models~\cite{karras2024analyzing, rombach2021highresolution, balaji2022ediff, karras2022elucidating, dhariwal2021diffusion, saharia2022palette, saharia2022photorealistic, sohl2015deep, song2020denoising} have become the standard framework for tasks ranging from text-to-image generation to image-to-image translation and appearance editing. Their ability to scale well with large datasets, coupled with pretrained weights \cite{rombach2021highresolution} that can be readily adapted to new domains~\cite{hu2021lora, zhang2023adding}, makes them especially suited for these applications. They also offer flexible inference mechanisms, where improved sampling procedures can significantly boost image quality~\cite{kynkaanniemi2024applying, ho2022classifier}.

Several recent works employ diffusion models specifically for relighting. DiLightNet~\cite{zeng2024dilightnet} demonstrates fine-grained control of object lighting by incorporating radiance hints. However, their multi-step pipeline, depends on scene reconstruction \cite{wang2024dust3r}, which is error-prone. Similarly, Neural Gaffer~\cite{jin2024neural_gaffer} focuses on object relighting, leveraging HDR environment maps. For human portrait relighting, Relightful Harmonization~\cite{ren2024relightful} and IC-Light~\cite{iclight} train on high-quality datasets (including light stage captures, synthetic Objaverse renders, and composited shadow materials) to synthesize background-harmonized portraits. Both methods rely on the background as lighting condition. In contrast, our approach directly tackles portrait-based relighting, using a diffusion model that learns to re-render synthetic faces. By starting from a pretrained model, and through our multi-task training strategy, we retain rich facial priors, while classifier-free guidance~\cite{ho2022classifier} on the input portrait further improves the preservation of texture and detail in the final relit output.

\subsection{Domain Adaptation}

Naively training on synthetic data often creates a domain gap for in-the-wild portraits, causing poor identity preservation and reduced photo-realism. Prior diffusion-based domain adaptation approaches \cite{ruiz2023dreambooth, gal2022image, hu2021lora, zhang2023adding, ye2023ip} mainly target style transfer or focused editing, not relighting.

\cite{he2024diffrelight} propose training a personalized diffusion model per subject, preserving identity but require light-stage capture and dedicated training for each subject. Other methods leverage real data to mitigate the synthetic-to-real gap: SwitchLight \cite{kim2024switchlight} pre-trains with a masked-autoencoder \cite{he2022masked} on real images before training on light-stage data, learning visual features (e.g. structure, color, texture) that are essential for relighting; Relightful Harmonization \cite{ren2024relightful} bootstraps a relighting model learned from light-stage data to pseudo-label in-the-wild images, subsequently finetuning on these pseudo-labels for improved photorealism; IC-Light \cite{iclight} uses large-scale data augmentation; and Lumos \cite{yeh2022learning} finetunes its albedo-prediction branch on real images, though its decomposition approach can fail with face paint, accessories, or strong shadows.

We propose a multi-task training scheme that unifies text-to-image and relighting tasks, enabling the training of our diffusion model with real images along with our synthetic dataset. In addition, our inference scheme based on classifier-free guidance helps preserve fine details from the input portrait. Our user study shows that the resulting relit portraits exhibit superior visual quality, identity, and lighting compared to existing methods.

\section{Method} \label{section:method}

Given a portrait image $I$ captured under unknown illumination conditions, our goal is to synthesize a relit version $I_R$ under a target lighting environment specified by a panoramic environment map $E$. The relit portrait $I_R$ should simultaneously: (1) preserve the subject's facial identity and characteristics from the original image $I$; (2) accurately reflect the illumination effects defined by the target environment map $E$
and (3) maintain photorealism in the final rendering. We first simulate this re-rendering to build a synthetic dataset for human portraits using Blender.

\subsection{Synthetic Data for Relighting}
\label{sec:data}

\begin{figure}[!htbp]
    \centering
    \begin{tabular}{@{\hskip 0mm}c@{\hskip 0.5mm}cc@{\hskip 0.5mm}c@{\hskip 0mm}}
        \includegraphics[width=0.11\textwidth]{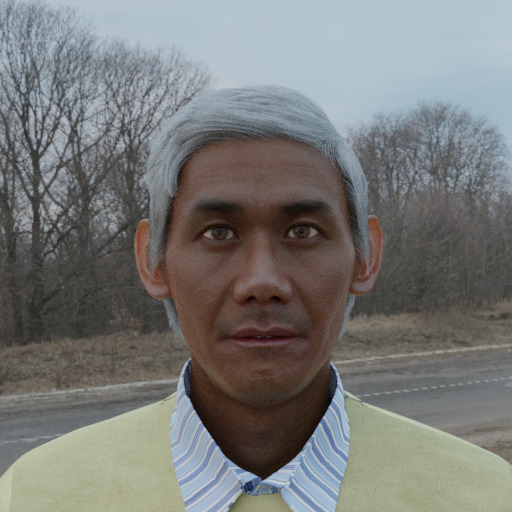} &
        \includegraphics[width=0.11\textwidth]{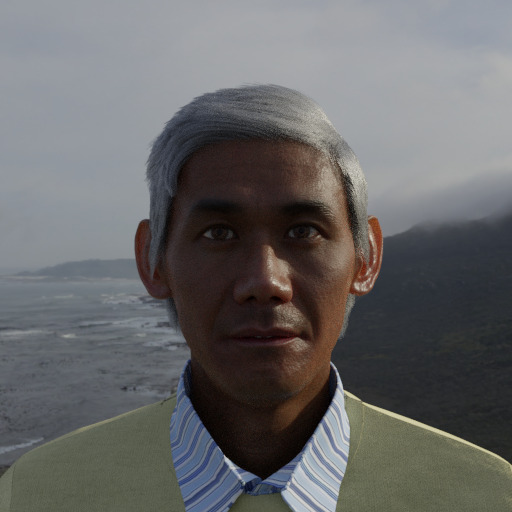} &
        \includegraphics[width=0.11\textwidth]{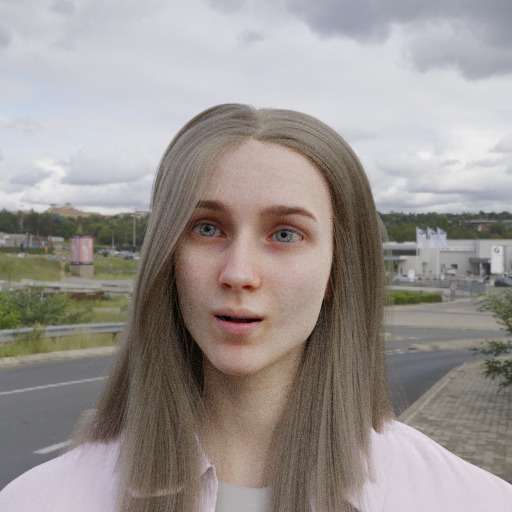} &
        \includegraphics[width=0.11\textwidth]{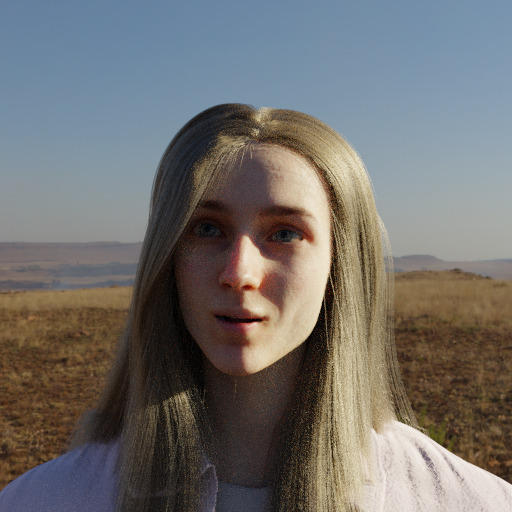} \\
    \end{tabular}
    \caption{\textit{Synthetic Faces}: Subjects are rendered under various lighting conditions (details in ~\cref{sec:data}). We show two examples, where each pair consists of a subject rendered using two different environment maps. The network is trained to re-render synthetic faces by transforming a subject rendered with one environment map into its counterpart rendered with the other environment map.}
    \label{fig:dataset}
\end{figure}

We build a 3D human portrait generation pipeline similar to ~\cite{wood2021fake}. Our system begins with a collection of high-quality, artist-created 3D head meshes, which we enhance by incorporating detailed facial components, including eyes, teeth, gums, and hair. We then augment these base models through rigging for pose variation and blendshape deformation for diverse facial expressions. To render realistic appearances, we incorporated a set of high quality PBR texture maps, including albedo, normal, roughness, specular, and subsurface scattering maps. We combine the head with random clothing meshes to build a portrait scene. The system is built with Blender and the images are rendered with the Cycles renderer.

To train our networks, we render images (samples shown in Fig.~\ref{fig:dataset}) at 512$\times$512 resolution from 350 subjects, each with roughly 10 varied appearance samples, including different hairstyles, skin tones, expressions, clothes, poses, \etc. We render each sample with 10 random HDR environment maps, each rotated 36 times evenly with a random initial rotation. In total, the dataset contains roughly 1.26 million images. See \cref{fig:more_examples_dataset} in the supplemental material for more examples from the dataset.



\subsection{Modeling Relighting with Diffusion Model}
\label{sec:model}

We build on top of Stable Diffusion~\cite{rombach2021highresolution}, a text-to-image foundation model pretrained with vast internet data. As shown in Fig.~\ref{fig:multi_task}, we incorporate the input portrait $I$, along with the target environment map $E$  to the input of the network backbone, by expanding the number of channels in the first convolutional layer of the Unet as per \cite{saharia2022palette}. 
To generate training samples $(I, E, T, I_R)$, where $T$ is a text prompt, we render portrait images from a subject $S$ with $n$ different HDR maps $E^{HDR}_1 \cdots E^{HDR}_n$ to obtain portraits $I^S_1 \cdots I^S_n$. We use an off-the-shelf image captioning model \cite{liu2024visual} to caption these images. Training samples are constructed by sampling two indices $i, j \in \{1 \cdots n\}$ and then using them to select input portrait, environment map, text prompt and target portrait as $(I^S_i, E^{HDR}_j, T^S_{j}, I^S_j)$. In the following, we drop the superscript $S$ for the subject to simplify notation. 
We use the sample to supervise our diffusion model in the following manner. First, we convert the HDR environment map $E^{HDR}_j$ into LDR $E^{LDR}_j$ by tone-mapping similar to~\cite{jin2024neural_gaffer}. The LDR environment map along with the input and target portraits are encoded using the encoder $Enc$ of Stable Diffusion's VAE, i.e., $\hat{I_i} = Enc(I_i), \hat{E^{LDR}_j} = Enc(E^{LDR}_j), \hat{I_j} = Enc(I_j)$.

Following the DDPM formulation \cite{ho2020denoising}, we randomly sample Gaussian noise $\epsilon$ and a diffusion timestep $t$ to add noise to the relit image latent $\hat{I_j}$ to obtain the noised latent $\hat{I^t_j}$.
We concatenate $\hat{I_i}, \hat{E^{LDR}_j}, \hat{I^t_j}$ along the channel axis and feed it to the Unet, following \cite{jin2024neural_gaffer}. The Unet $\epsilon_\theta$ is trained with the DDPM objective: 
\begin{equation}
    \min_{\theta} \mathbb{E}_{x \in Enc(I_R), t, \epsilon \in \mathcal{N}(0, I)} \|\epsilon_\theta(x_t, I, E, T) - \epsilon\| 
\end{equation}


\subsection{Multitask Training}
\label{sec:train}
Training or fine-tuning a diffusion model on a synthetic dataset creates a substantial domain gap when applied to in-the-wild images, resulting in degraded output quality. For instance, when applied to real-world images, the model fails to reproduce critical details, such as textures in clothing, jewelry, and accessories, which are absent in the synthetic data distribution (e.g., as seen in the 'Base' result in Fig.~\ref{fig:ablations}).
To address this, we propose a multitask training strategy to mitigate potential model distribution drifting to synthetic renderings. Similar techniques have been applied in the context of inpainting~\cite{xie2023smartbrush} to combat the lack of diversity in training data.

\begin{figure}[!htbp]
    \centering
    \includegraphics[width=0.95\linewidth]{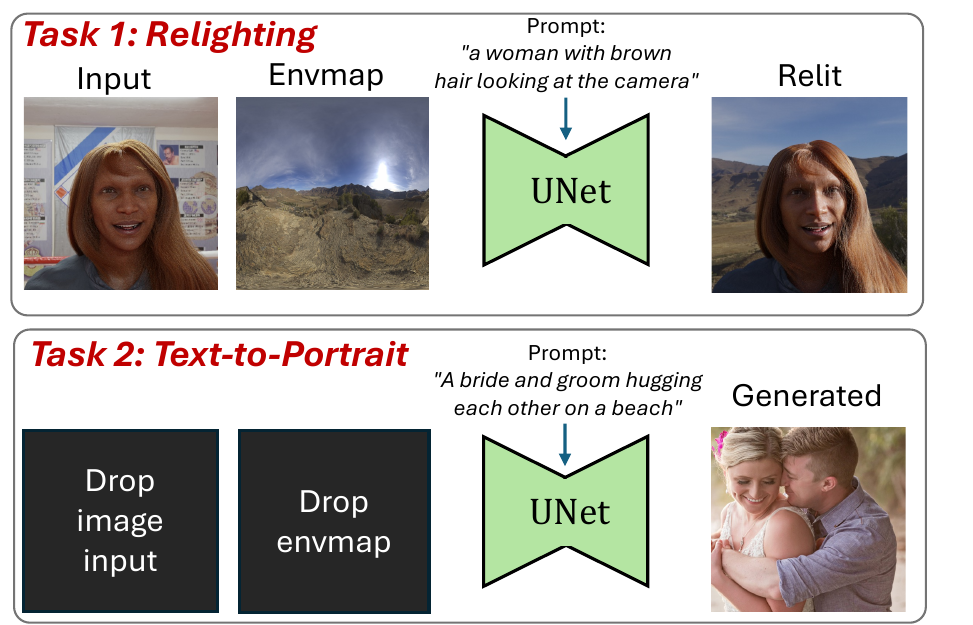}
    \caption{Training pipeline of SynthLight. We first enable the relighting modeling by training the diffusion backbone with synthetic relighting tuples (Task 1, top row), detailed in Sec.~\ref{sec:model}. To further alleviate the domain gap between synthetic and real image domain, we include a joint training of the text-to-image task (Task 2, bottom row), detailed in Sec.~\ref{sec:train}. Our model is based on LDM \cite{rombach2021highresolution} and is composed of a VAE and a UNet. For simplicity, VAE is omitted in the diagram.}
    \label{fig:multi_task}
\end{figure}

Specifically, we incorporate a text-to-portrait generation task, which constraints the diffusion model to produce a realistic portrait image given an input prompt.
This task is trained alongside the original relighting task, and this helps to improve the photorealism and generalization of the trained model. Since both tasks share the same network architecture, we simply replace the image and LDR inputs with two black images, as illustrated in Fig.~\ref{fig:multi_task}.

To obtain training samples for the text-to-portrait, we curate a subset of human portrait images from the LAION~\cite{schuhmann2022laion} dataset by sampling the images filtered by a face detector. Details on detection and filtering are provided in the supplementary material (see \cref{sec:appendix_dataset}). During training, we empirically set the sampling ratios of the synthetic dataset versus the real dataset as $0.7$ and $0.3$, respectively. We observe significant benefits from incorporating the real images during training in improving identity preservation and photorealism. This echoes the findings in~\cite{ren2024relightful}, where a bootstrapped dataset helps generalize of image harmonization, emphasizing the benefits of data diversity.





\begin{figure}[!htbp]
    \centering
    \includegraphics[width=0.85\linewidth]{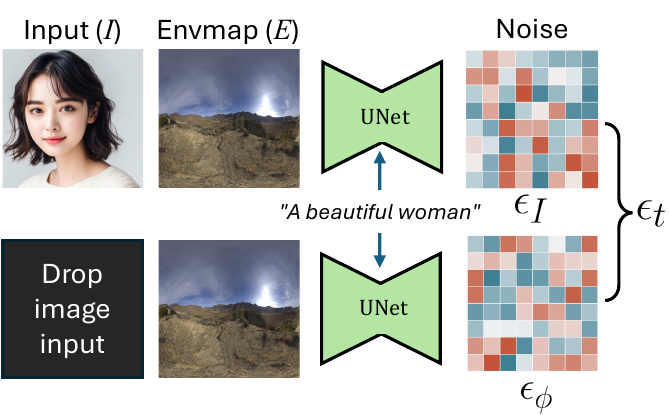}
    \caption{We employ the image-conditioning classifier-free guidance during inference to proportionally balance between identity preservation, and relighting effects. The final score estimate is computed as per \cref{eq:cfg}.}
    \label{fig:inference}
\end{figure}

\subsection{Inference Time Adaptation} 
\label{sec:inf}
We further employ a simple yet effective inference time adaptation scheme that proportionally balances between the identity preservation of the input portrait and the relighting strength. 
Inspired by the dual-conditioning classifier-free guidance \cite{ho2022classifier} proposed in InstructPix2Pix \cite{brooks2023instructpix2pix}, we define an analogous concept in our inference. As illustrated in Fig.~\ref{fig:inference}, at each step of the diffusion inference, the diffusion score is a composition of scores from both image-conditional and unconditional output. Specifically, for unconditional inference, we drop the input image while keeping the LDR and text-prompt conditioning identical. 
Formally, we apply the following score estimate at a particular timestep $t$: 

\begin{align}
    \epsilon_t &= \epsilon_\theta(x_{t+1}, \phi, E, \phi) \notag \\
    & + \lambda_T (\epsilon_\theta(x_{t+1}, I, E, T) - \epsilon_\theta(x_{t+1}, I, E, \phi)) \notag \\
    & + \lambda_I (\epsilon_\theta(x_{t+1}, I, E, \phi) - \epsilon_\theta(x_{t+1}, \phi, E, \phi)) \; . \label{eq:cfg}
\end{align}

Here, $\lambda_T$ and $\lambda_I$ are the \textit{guidance} parameters, where $\lambda_T$ is inherited from the original definition of CFG, which specifies the how much the model respects to the text prompts, while $\lambda_I$ specifies the strength of the input portrait guidance. 
With this score estimate, we use DDIM \cite{song2020denoising} to obtain the latent at current timestep $x_t = \text{DDIM}(x_{t + 1}, \epsilon_t)$. We empirically find that using a guidance value of $\lambda_I \in [2, 3]$ for the input portrait helps achieve a balance between the details and identity preservation while performing reasonable relighting. 

\begin{figure}[!htbp]
    \centering
    \includegraphics[width=0.48\textwidth]{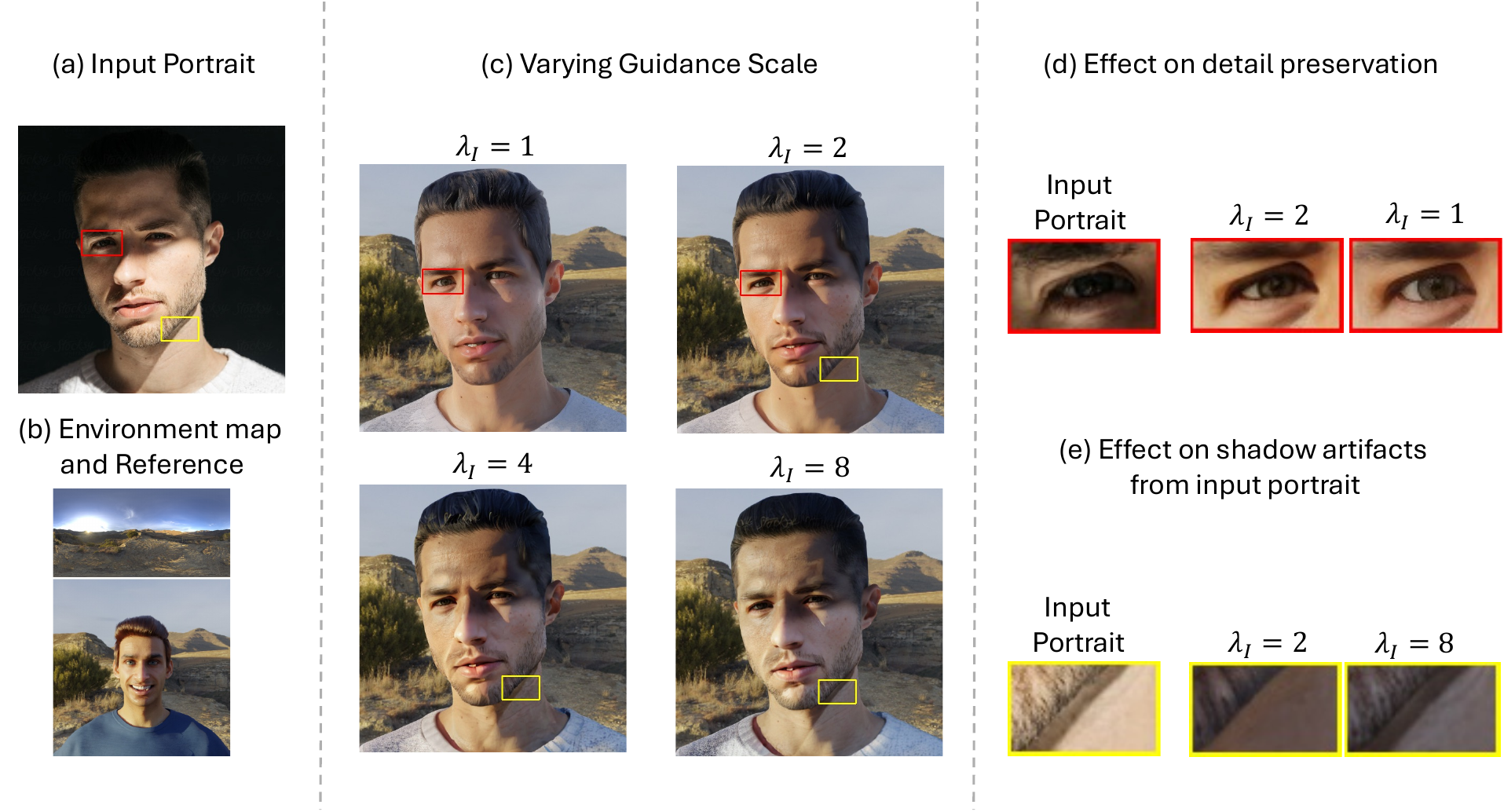}
    \caption{Effect of input portrait \textit{guidance parameter} $\lambda_I$: We show (a) the input portrait, (b) the lighting condition and a reference image rendered in Blender with the same lighting, and (c) outputs with varying $\lambda_I$. (d) highlights that $\lambda_I=1$, equivalent to removing inference-time adaptation, alters the eye shape (in red rectangle). (e) shows that higher $\lambda_I$ introduces undesired lighting artifacts, such as shadow artifacts from the input portrait (in yellow rectangle).}
    \label{fig:cfg}
\end{figure}

In Fig.~\ref{fig:cfg}, we illustrate the effects of varying $\lambda_I$. Smaller values provide the strongest relighting effect while sacrificing some visual quality and losing the facial details of the input. Large values provide much better identity preservation but weaken the relighting effects where lighting information, such as shadows, leaks from the input into the output.


\begin{figure*}[!htbp]
    \centering

    \begin{subfigure}{\textwidth}
        \centering
        \begin{tabular}{@{\hskip 0mm}c@{\hskip 1mm}c@{\hskip 0.5mm}c@{\hskip 0.5mm}c@{\hskip 2.5mm}c@{\hskip 1mm}c@{\hskip 0.5mm}c}
            \includegraphics[height=\imageheightqual]{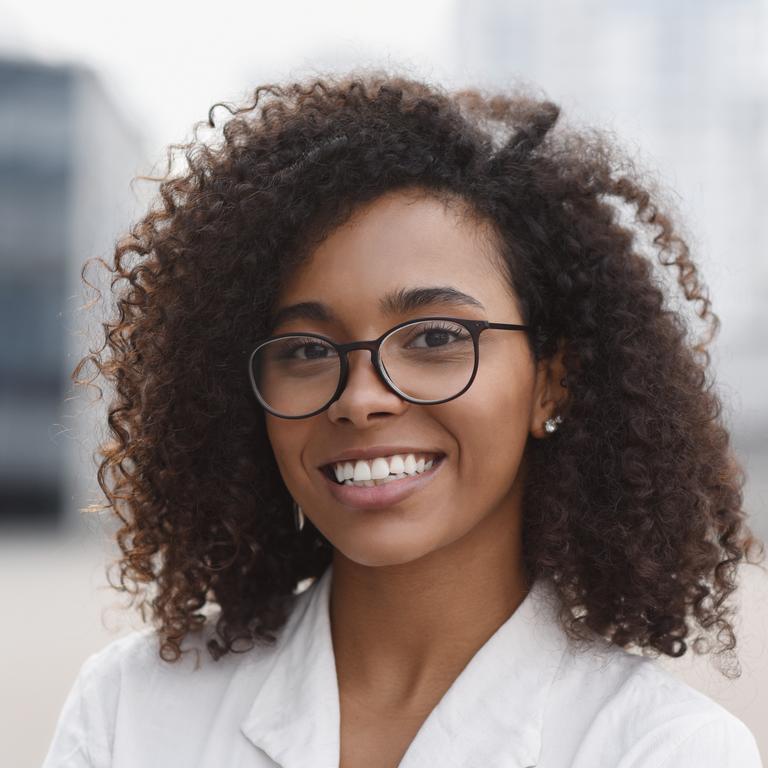} &
            \includegraphics[height=\imageheightqual]{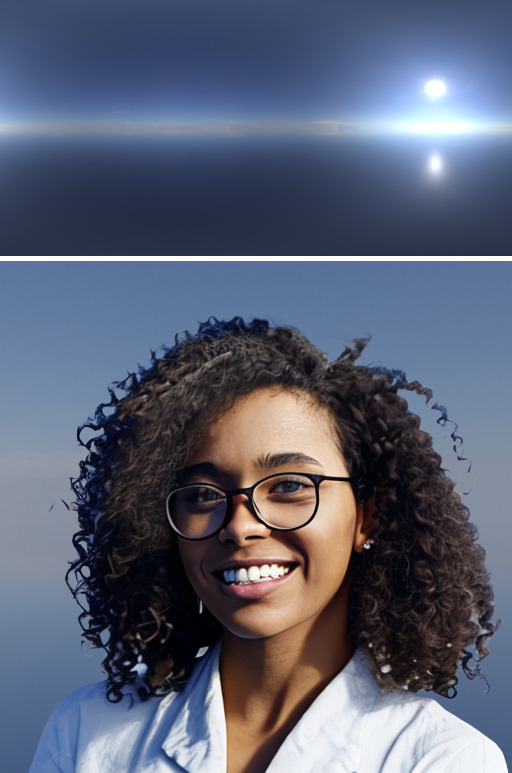} &
            \includegraphics[height=\imageheightqual]{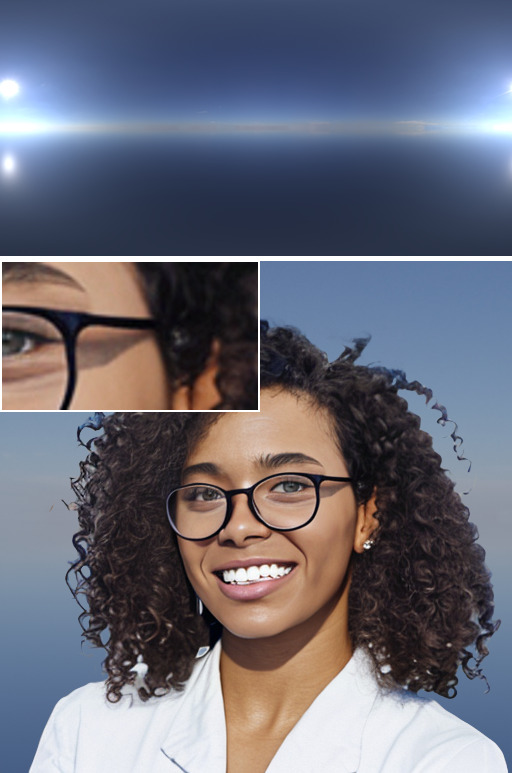} &
            \includegraphics[height=\imageheightqual]{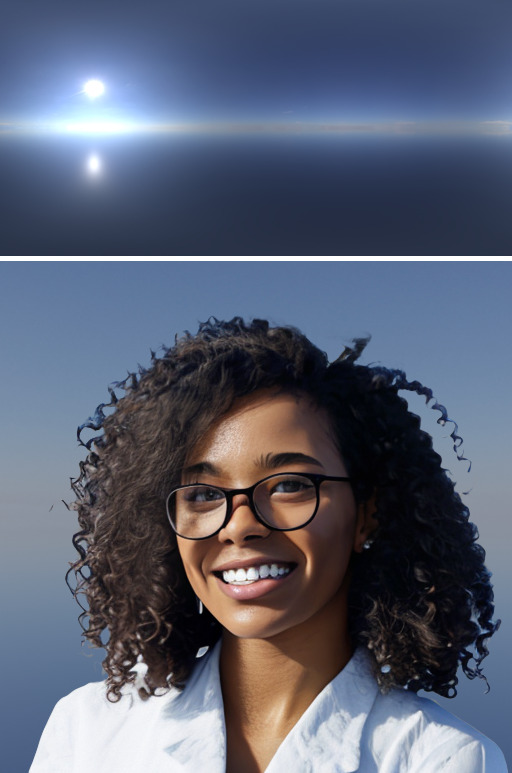} &
            \includegraphics[height=\imageheightqual]{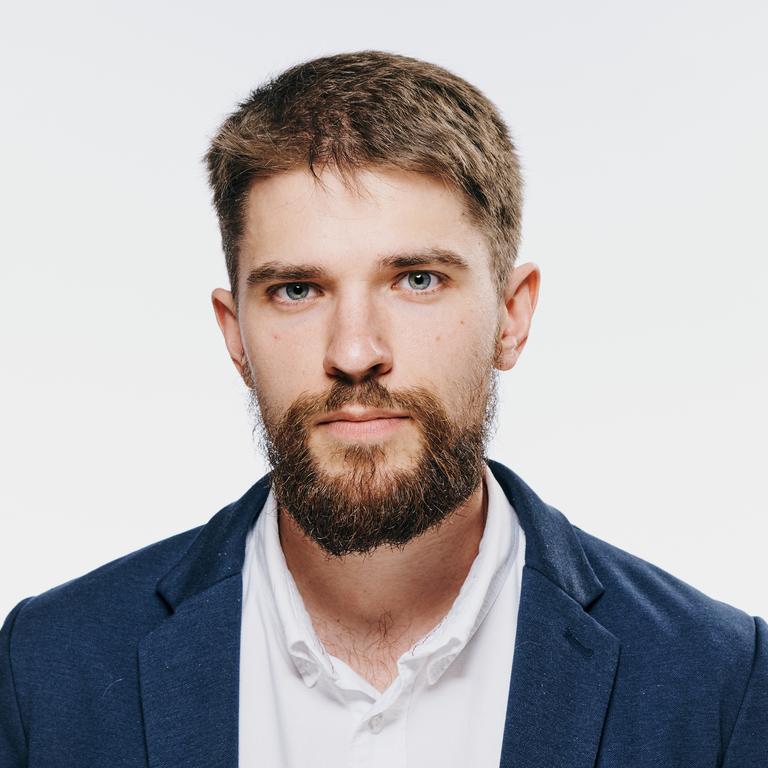} &
            \includegraphics[height=\imageheightqual]{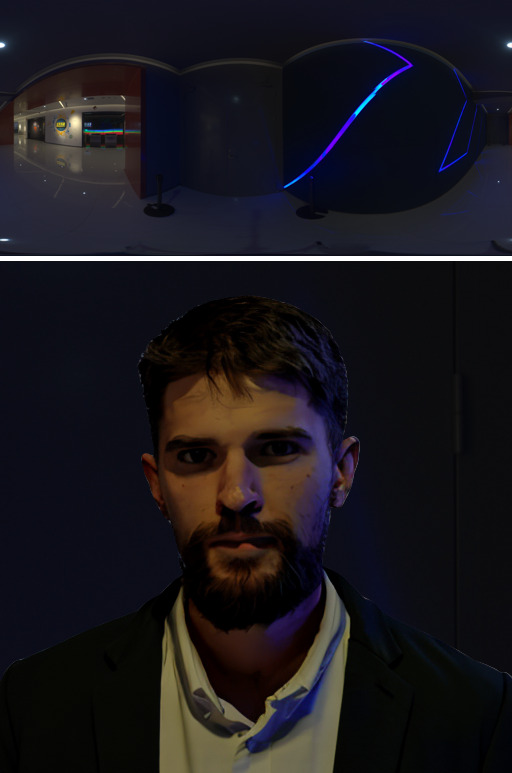} &
            \includegraphics[height=\imageheightqual]{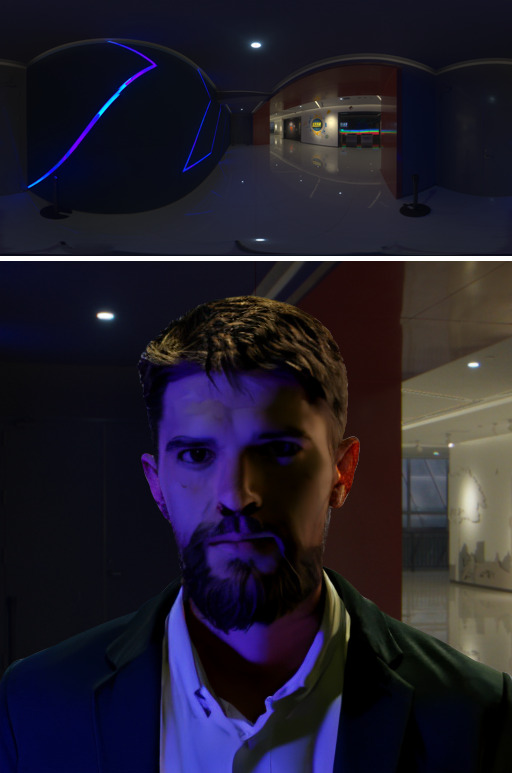} \\
        \end{tabular}
        \caption{Our method demonstrates the ability to relight subjects effectively in both \textbf{outdoor} (left) and \textbf{indoor} (right) settings. In outdoor scenarios, strong \textit{\textbf{cast shadows}} are produced due to self-occlusion from facial features and glasses (see \textbf{inset}). For indoor scenes, our method handles complex lighting conditions, such as casting neon lights on the input portrait.}
    \end{subfigure}

    \begin{subfigure}{\textwidth}
        \centering
        \begin{tabular}{@{\hskip 0mm}c@{\hskip 1mm}c@{\hskip 1.5mm}c@{\hskip 1mm}c@{\hskip 0.5mm}c@{\hskip 1.5mm}c@{\hskip 1mm}c}
            \includegraphics[height=\imageheightqual]{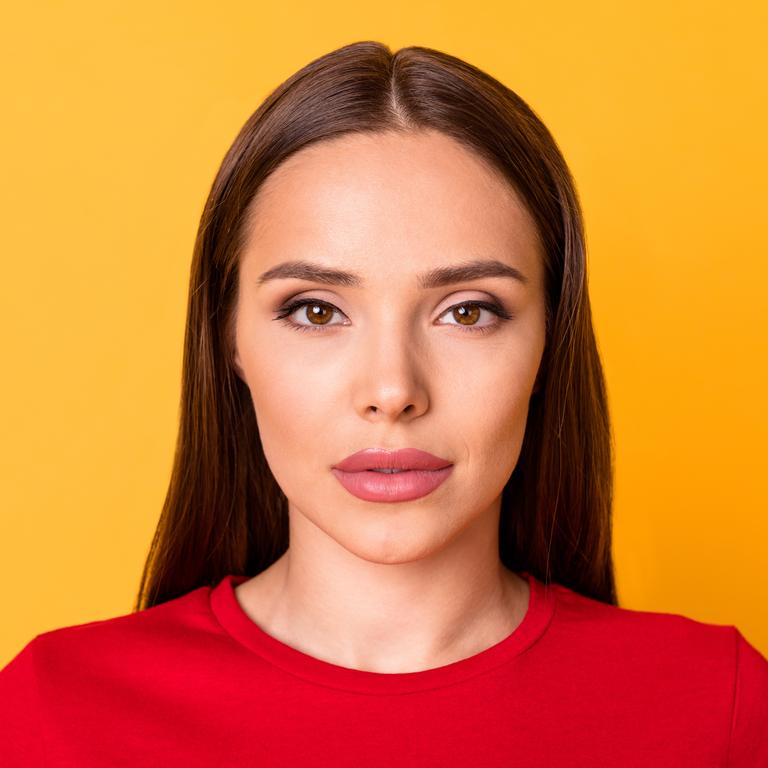} &
            \includegraphics[height=\imageheightqual]{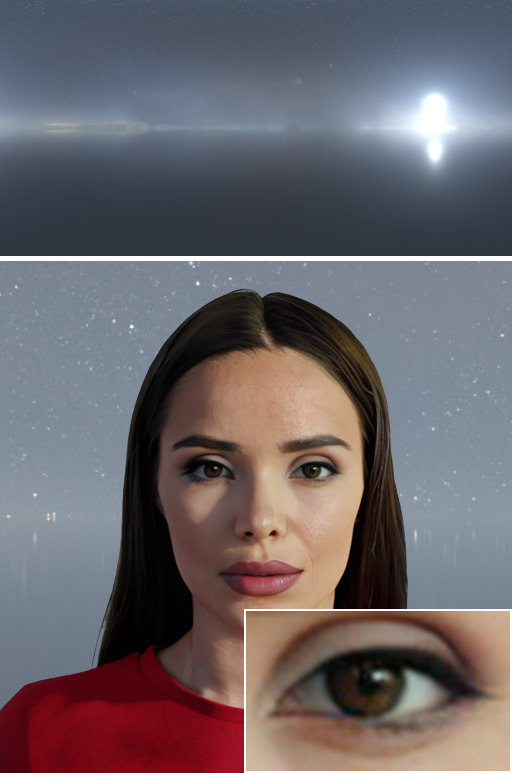} &
            \includegraphics[height=\imageheightqual]{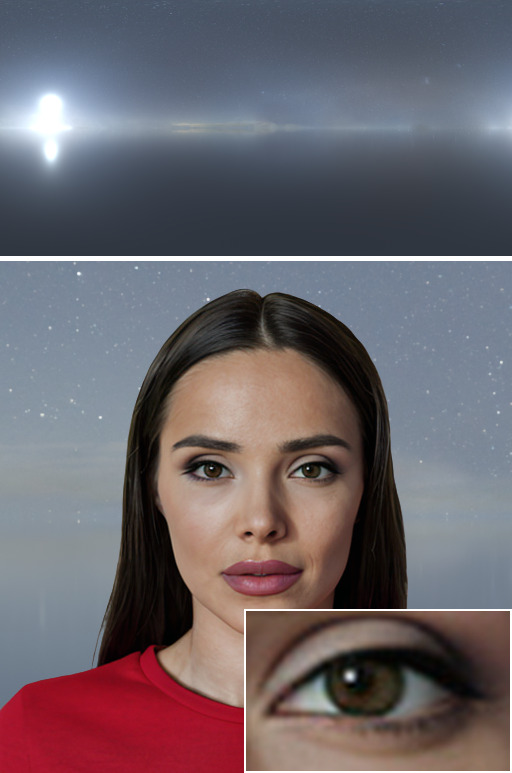} &
            \includegraphics[height=\imageheightqual]{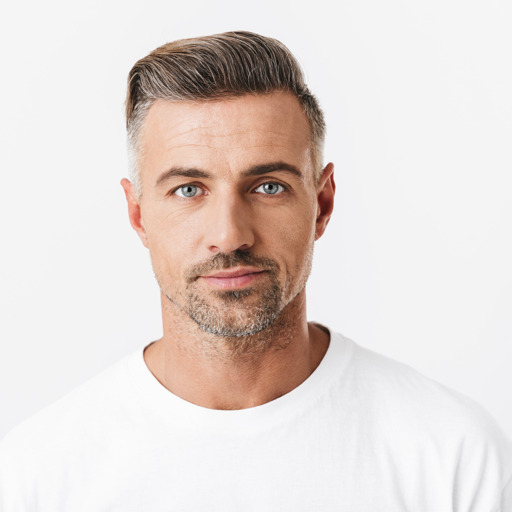} &
            \includegraphics[height=\imageheightqual]{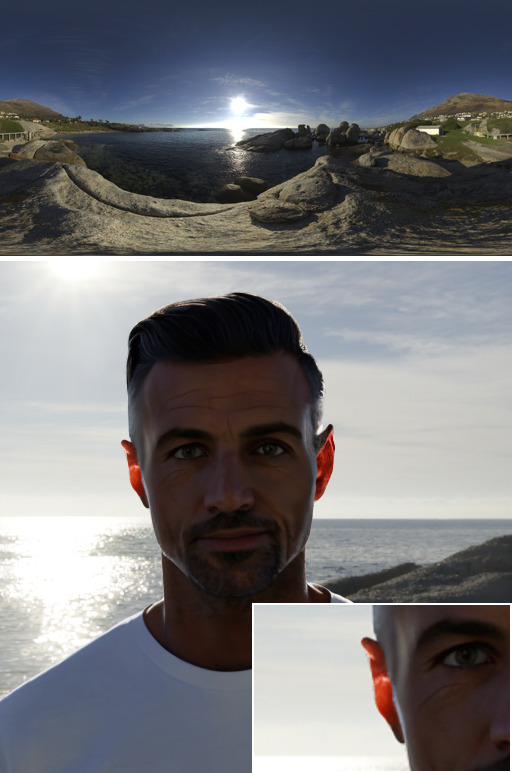} &
            \includegraphics[height=\imageheightqual]{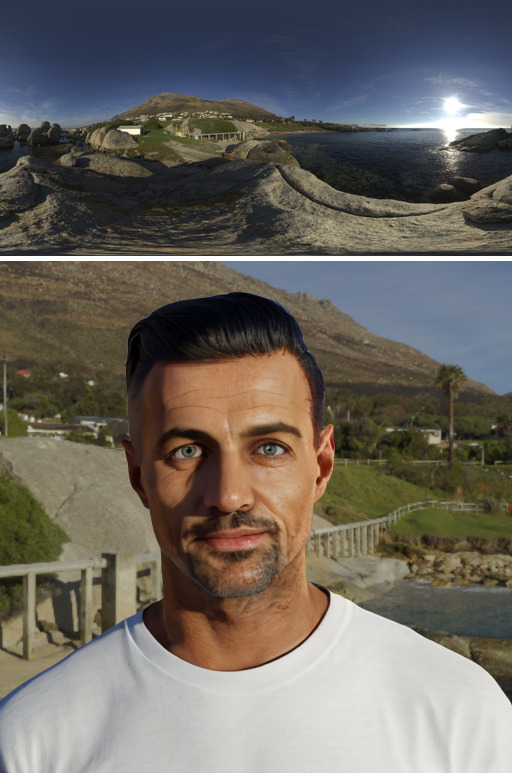} &
            \includegraphics[height=\imageheightqual]{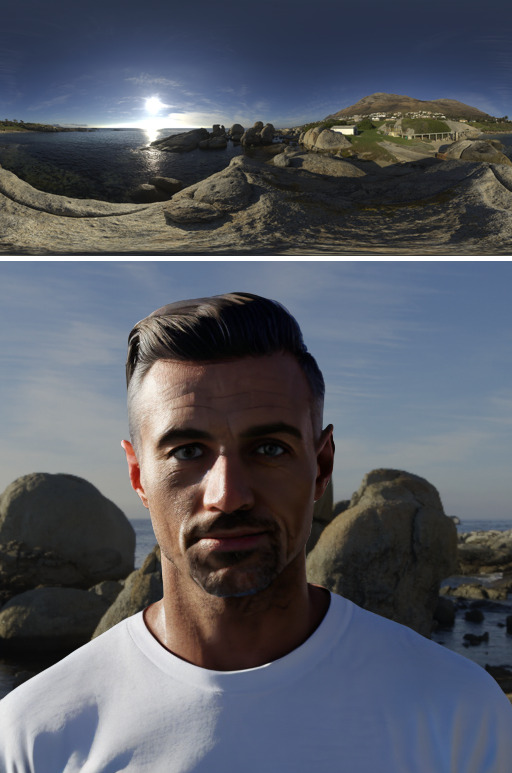} \\
        \end{tabular}
        \caption{Our method captures interesting lighting effects for portraits, synthesizing fine details like \textbf{\textit{catch light}} in the eye for realistic relighting (left, see \textbf{inset}) and \textbf{subsurface scattering} in the ear under strong backlight conditions, such as sunlight (right, see \textbf{inset}).} 
    \end{subfigure}

    \begin{subfigure}{\textwidth}
        \centering
        \begin{tabular}{@{\hskip 0mm}c@{\hskip 1mm}c@{\hskip 0.5mm}c@{\hskip 0.5mm}c@{\hskip 2.5mm}c@{\hskip 1mm}c@{\hskip 0.5mm}c}
            \includegraphics[height=\imageheightqual]{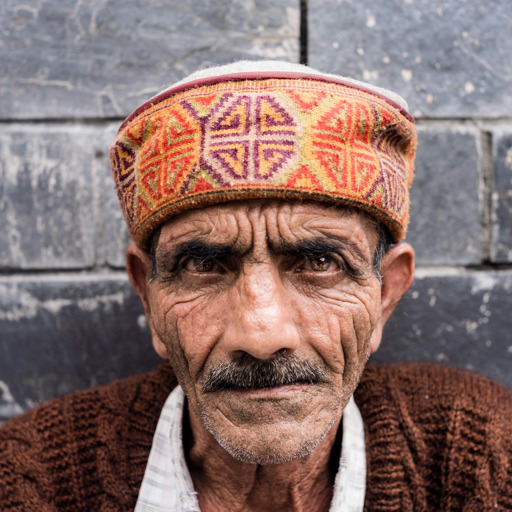} &
            \includegraphics[height=\imageheightqual]{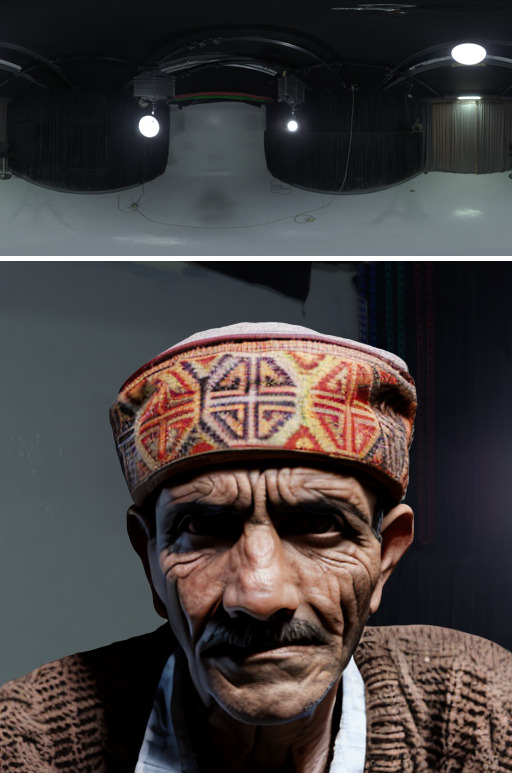} &
            \includegraphics[height=\imageheightqual]{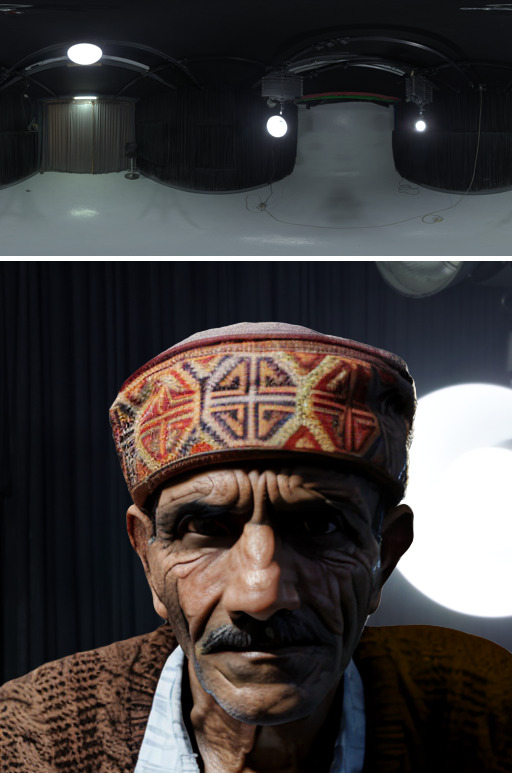} &
            \includegraphics[height=\imageheightqual]{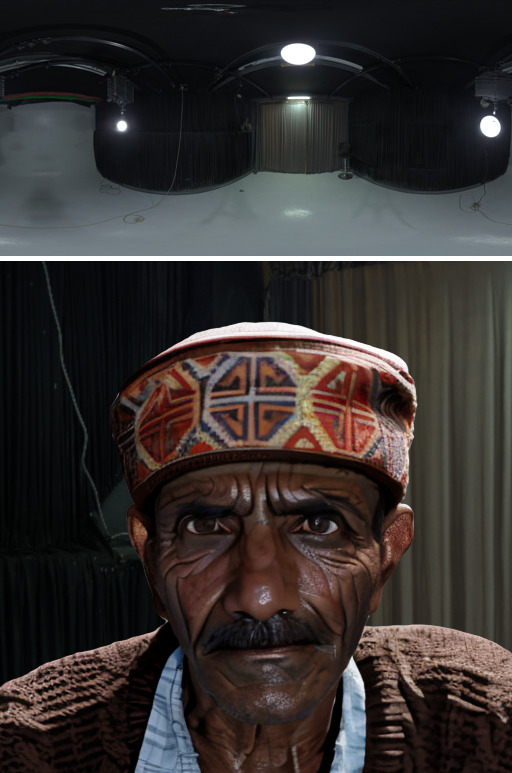} &
            \includegraphics[height=\imageheightqual]{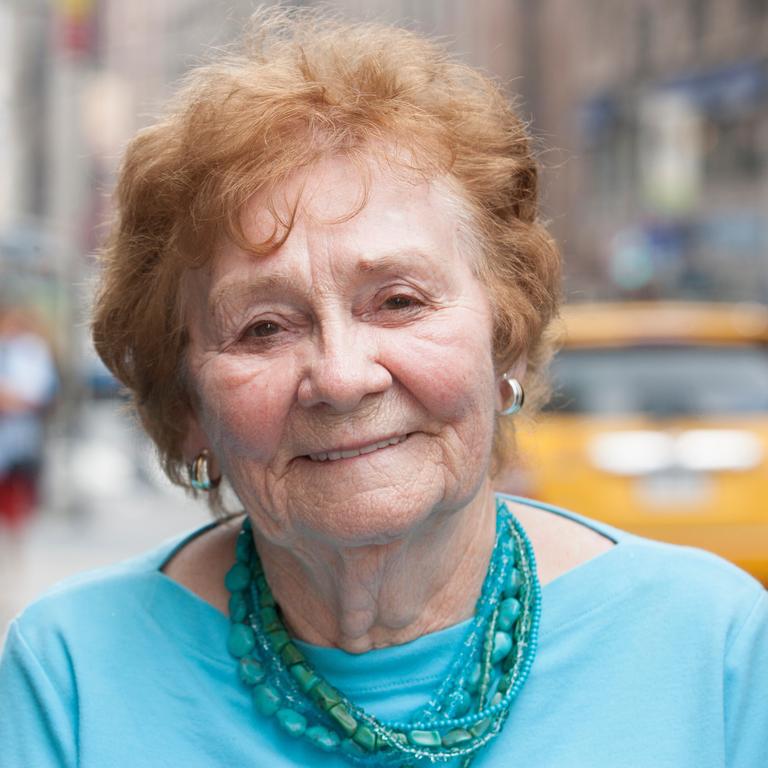} &
            \includegraphics[height=\imageheightqual]{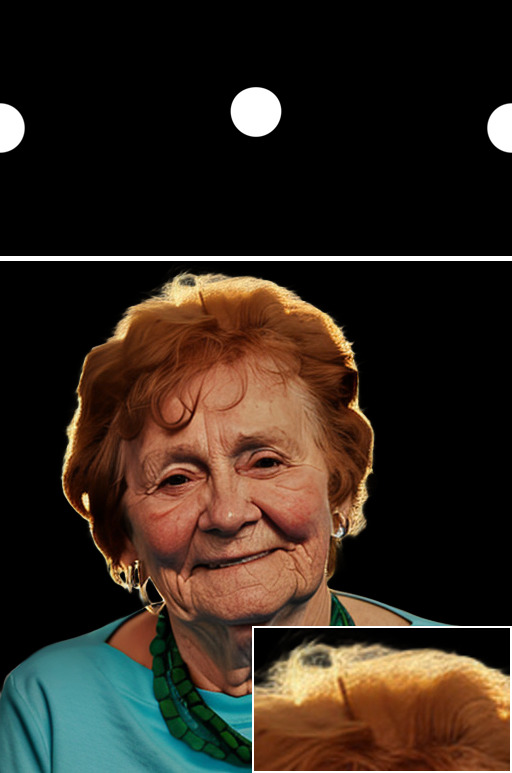} &
            \includegraphics[height=\imageheightqual]{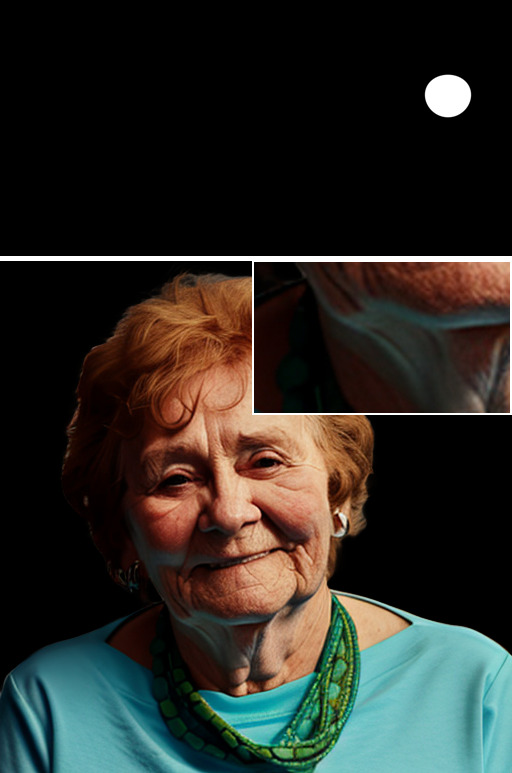} \\
        \end{tabular}
        \caption{Our method enables \textit{\textbf{studio-style lighting}} for portraits, creating dramatic effects in studio-like environments (left). Using hand-designed environment maps, we relight with two presets (right): \textbf{Backlight}, which uses a light behind the subject to define edges and produce a distinctive rim effect (see \textbf{inset}); and \textbf{Rembrandt}, where light comes from an angle, illuminating one portion of the face while casting the other in shadow to create depth and contrast. The Rembrandt image also highlights \textbf{inter-reflections} from clothing (rightmost, see \textbf{inset}).}
    \end{subfigure}

    \begin{subfigure}{\textwidth}
        \centering
        \begin{tabular}{@{\hskip 0mm}c@{\hskip 1mm}c@{\hskip 1.5mm}c@{\hskip 1mm}c@{\hskip 0.5mm}c@{\hskip 1.5mm}c@{\hskip 1mm}c}
            \includegraphics[height=\imageheight]{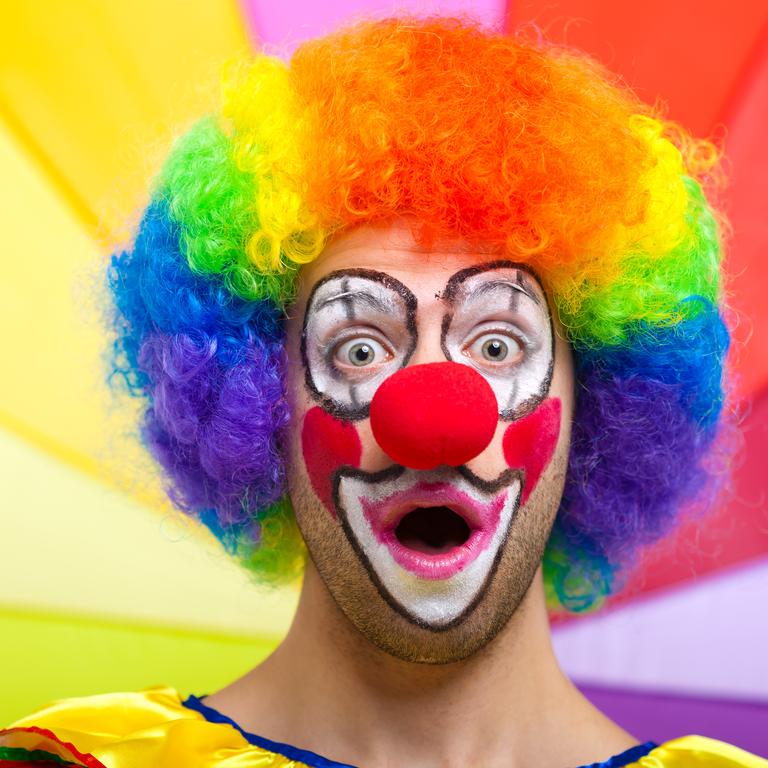} &
            \includegraphics[height=\imageheight]{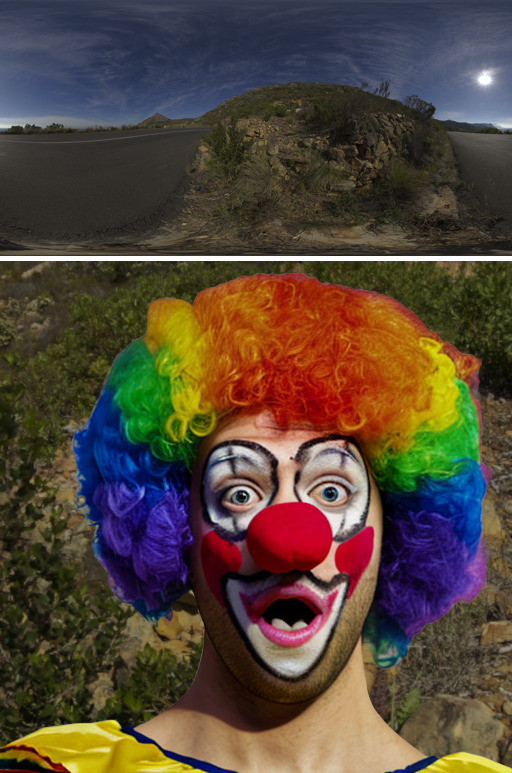} &
            \includegraphics[height=\imageheight]{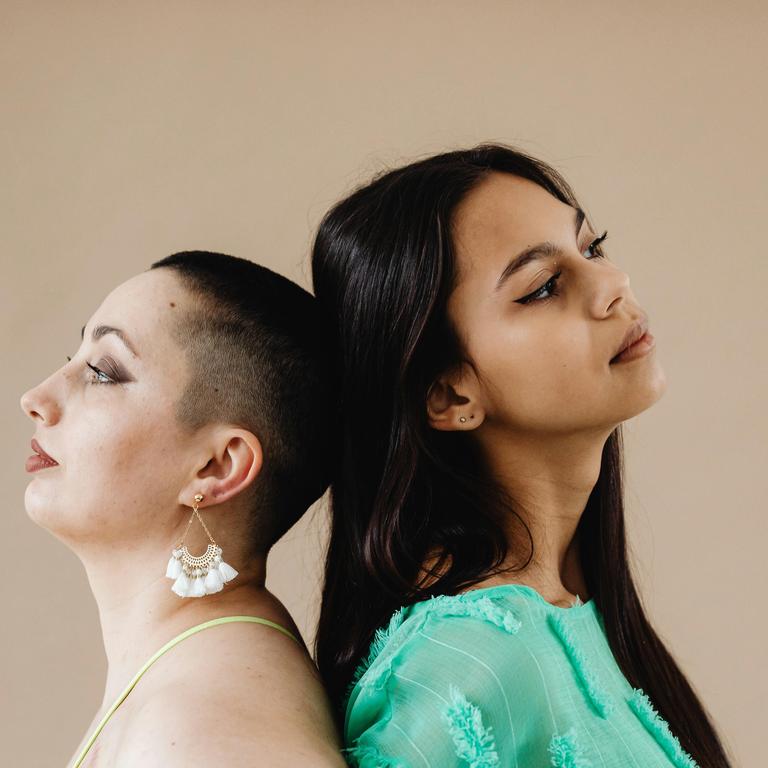} &
            \includegraphics[height=\imageheight]{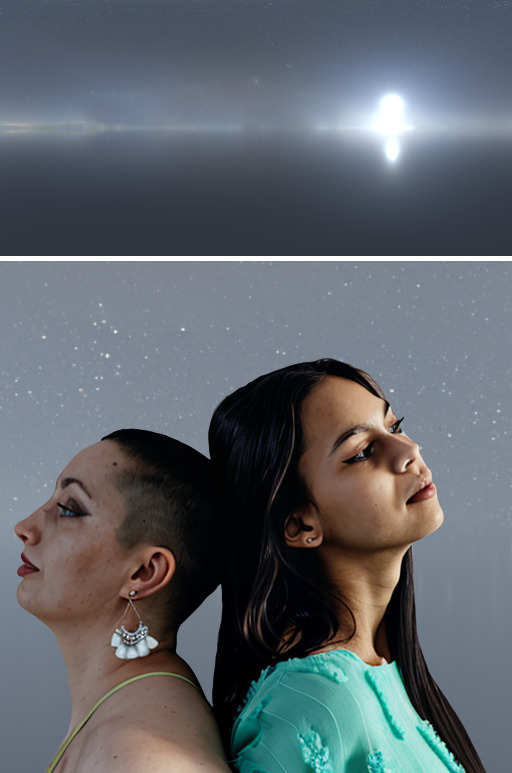} &
            \includegraphics[height=\imageheight]{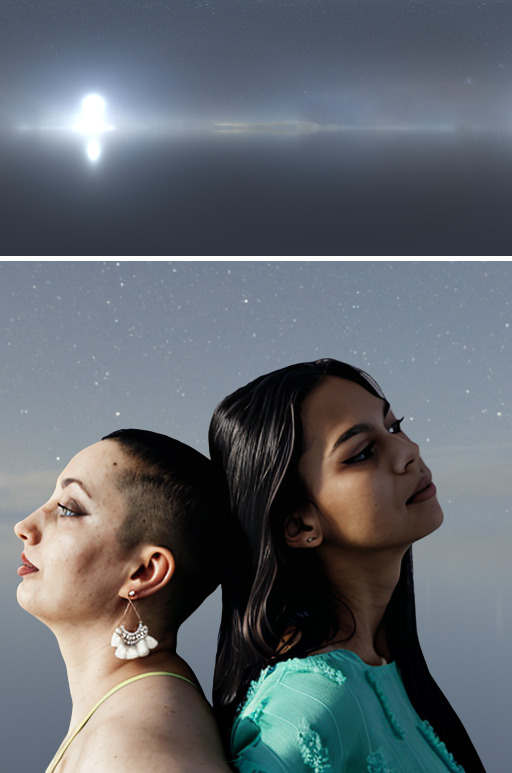} &
            \includegraphics[height=\imageheight]{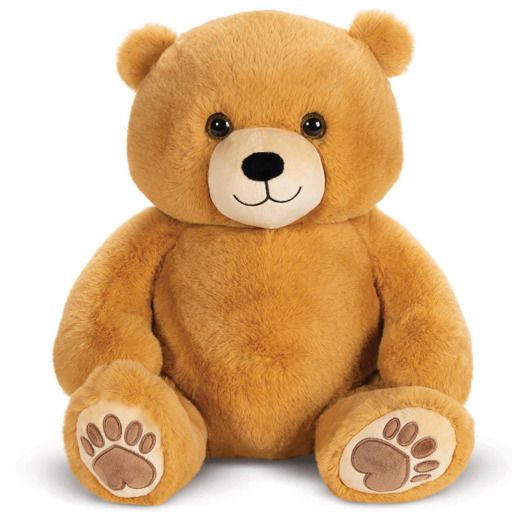} &
            \includegraphics[height=\imageheight]{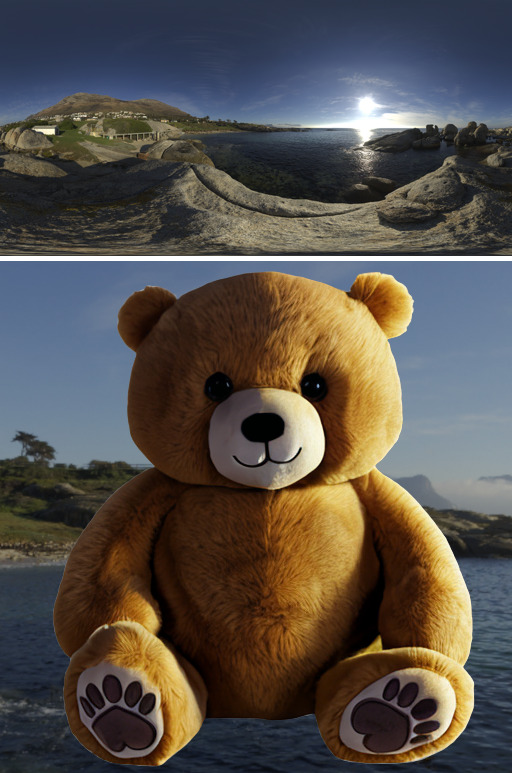} \\
        \end{tabular}
        \caption{While trained only on a synthetic dataset, our method \textit{\textbf{generalizes}} to \textbf{\textit{unseen}} image categories such as a clown (left), a photograph of two people (middle), and a teddy-bear (right).}
    \end{subfigure}

    \caption{Real-world results showcasing our method’s ability to handle diverse lighting scenarios. Each example includes the input portrait (left), the environment map used for relighting (top right), and the relit output (bottom right). The subfigures highlight: (a) relighting under indoor and outdoor environments, (b) capturing interesting lighting effects such as catch lights in eyes and sub-surface scattering on ears, (c) studio-style lighting setups, and (d) generalization across various challenging scenarios.}
    \label{fig:lighting_results}
\end{figure*}

\section{Experiments} \label{section:exps}

\subsection{Setup and Metrics}

We create three test sets for evaluating our method: (a) 300 Light Stage rendered relighting pairs, (b) a held out subset of our synthetic faces dataset consisting of 500 images, (c) in-the-wild portraits for qualitative evaluation of visual quality.
For test sets (a) and (b), we use standard quantitative metrics such as SSIM, PSNR, LPIPS \cite{zhang2018unreasonable} to evaluate image fidelity and face embedding distance such as FaceNet \cite{schroff2015facenet} for evaluating identity preservation. We train on the entire synthetic dataset but withhold 20\% of the environment maps to create the Light Stage test set. We also hold out 10\% of the subject identities and 10\% of the environment maps for the synthetic test set, ensuring they remain unseen during training.

\begin{figure*}[!htbp]
    \centering

    \begin{tabular}{@{\hskip 0mm}c@{\hskip 2mm}c@{\hskip 0.5mm}c@{\hskip 0.5mm}c@{\hskip 0.5mm}c@{\hskip 0.5mm}c@{\hskip 0.5mm}c@{\hskip 0.5mm}c}
        & & DiLightNet & IC-Light & Neural Gaffer & Total Relighting & SwitchLight & Ours \\
        
        \includegraphics[height=\figureheight]{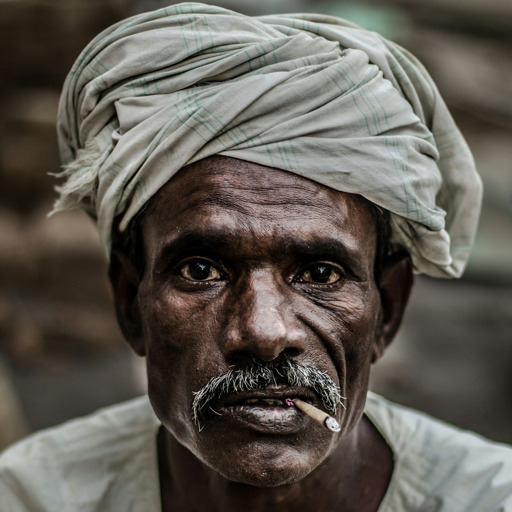} &
        \includegraphics[height=\figureheight]{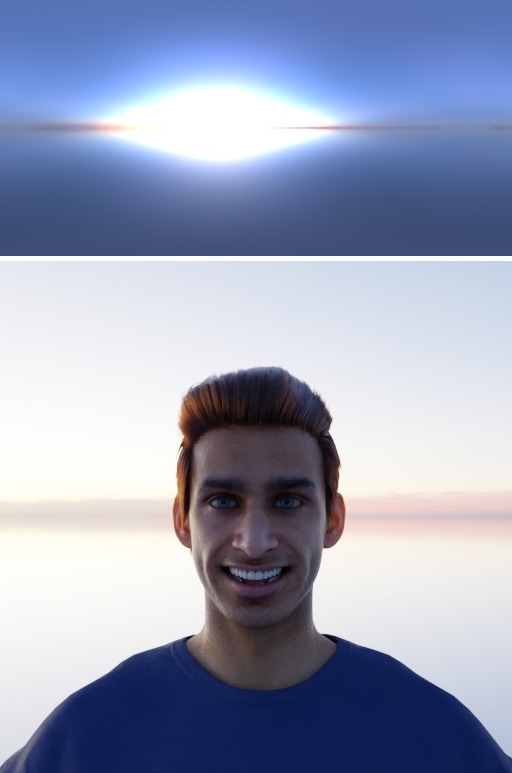} &
        \includegraphics[height=\figureheight]{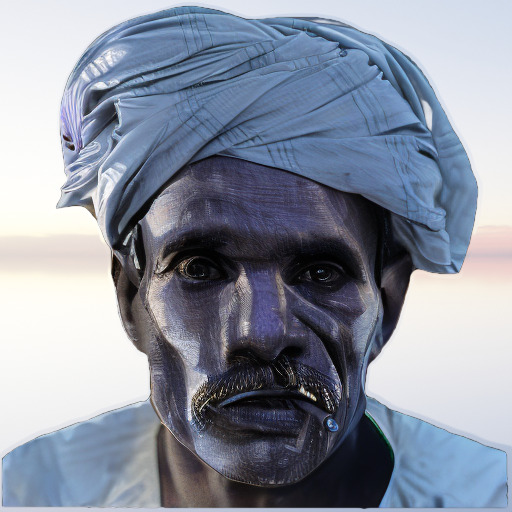} &
        \includegraphics[height=\figureheight]{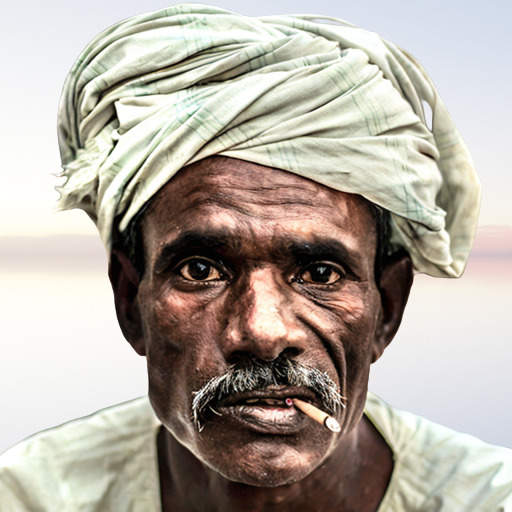} &
        \includegraphics[height=\figureheight]{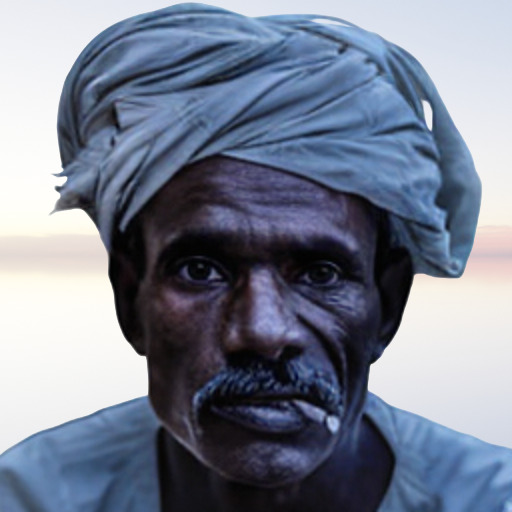} &
        \includegraphics[height=\figureheight]{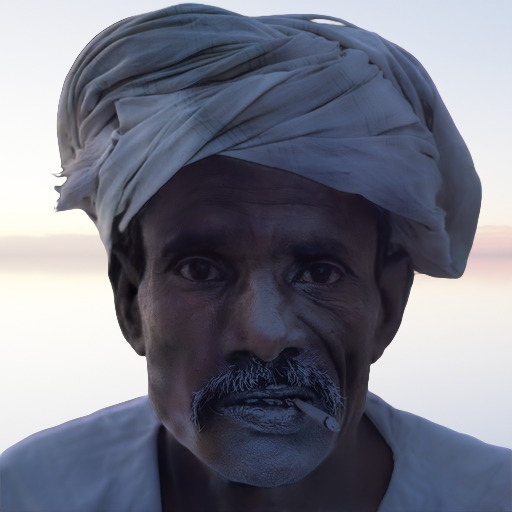} &
        \includegraphics[height=\figureheight]{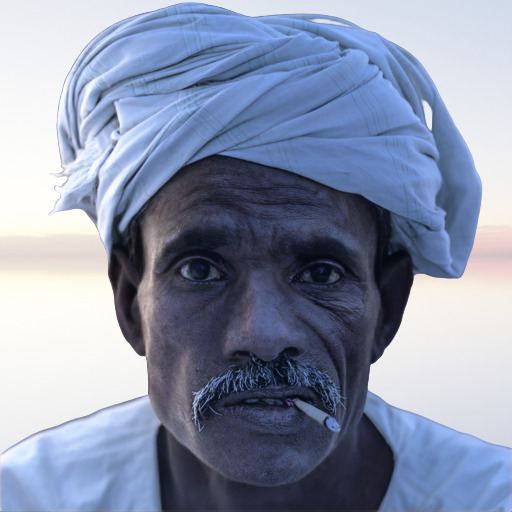} &
        \includegraphics[height=\figureheight]{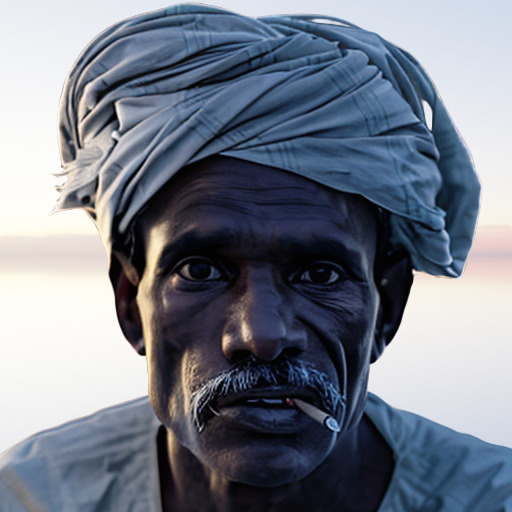} \\

        \includegraphics[height=\figureheight]{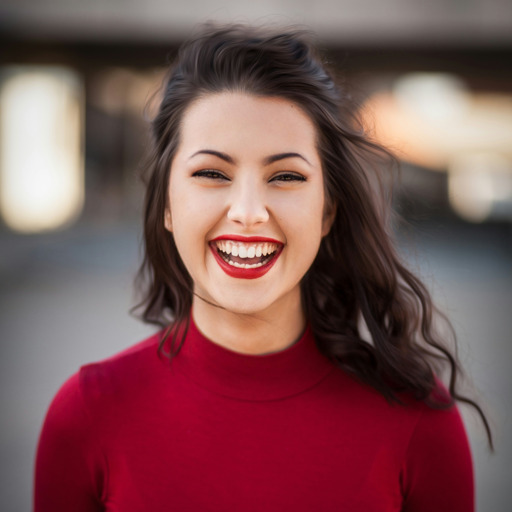} &
        \includegraphics[height=\figureheight]{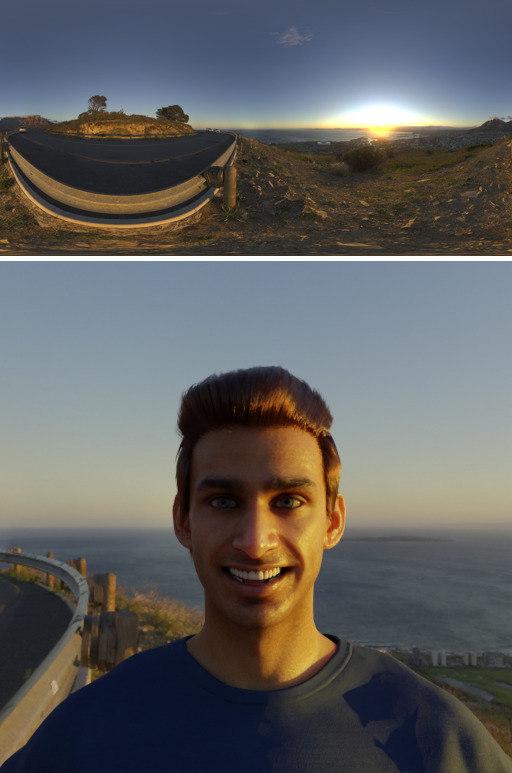} &
        \includegraphics[height=\figureheight]{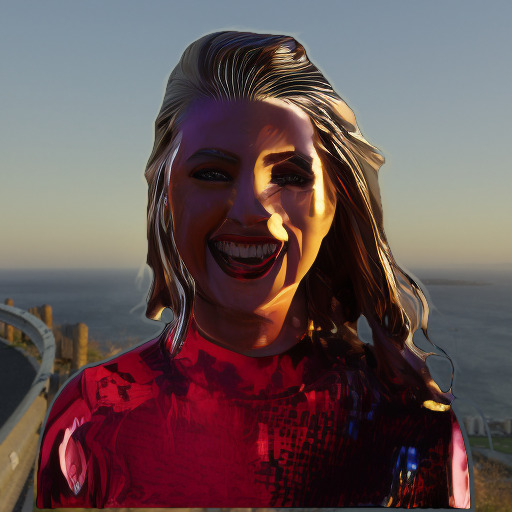} &
        \includegraphics[height=\figureheight]{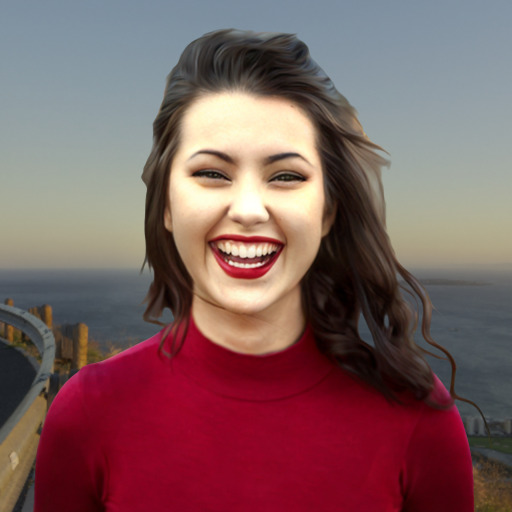} &
        \includegraphics[height=\figureheight]{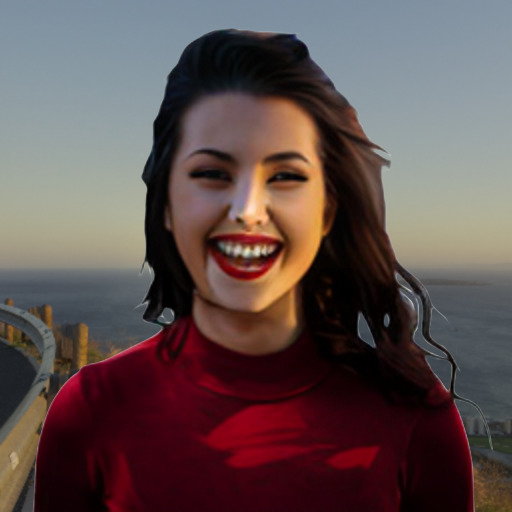} &
        \includegraphics[height=\figureheight]{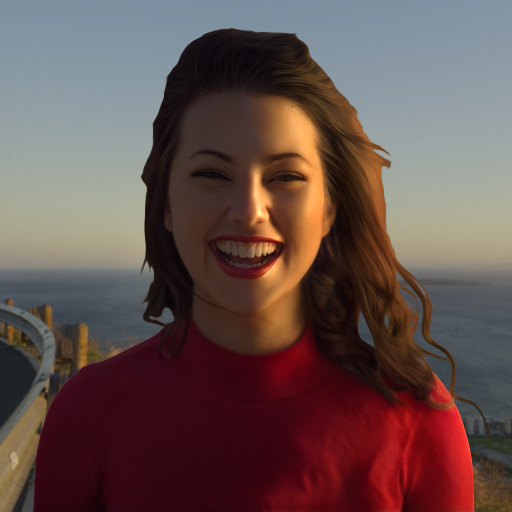} &
        \includegraphics[height=\figureheight]{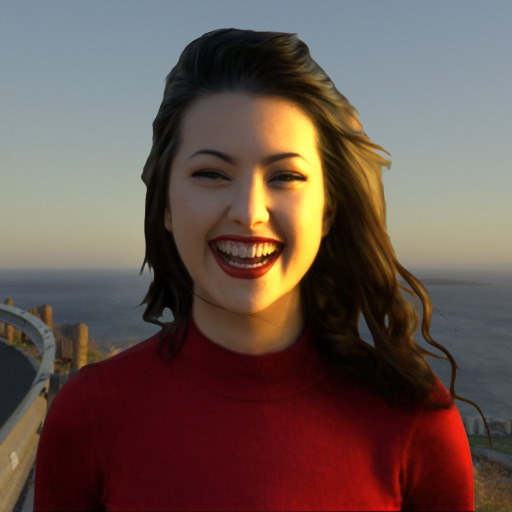} &
        \includegraphics[height=\figureheight]{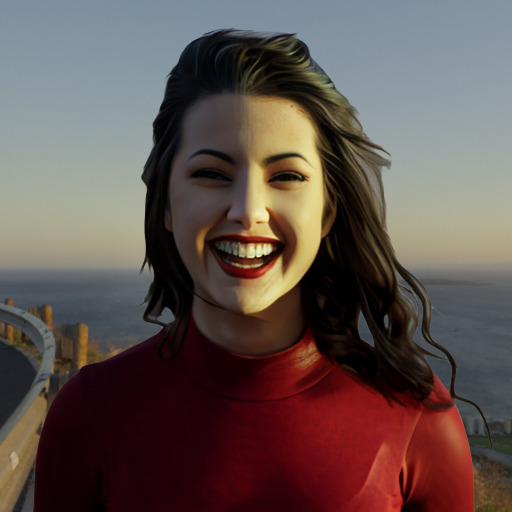} \\

        \includegraphics[height=\figureheight]{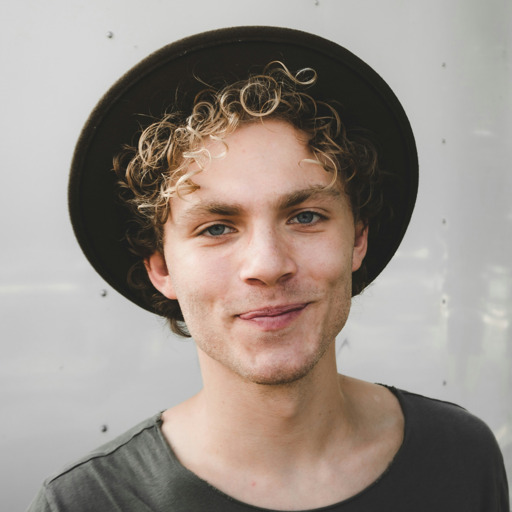} &
        \includegraphics[height=\figureheight]{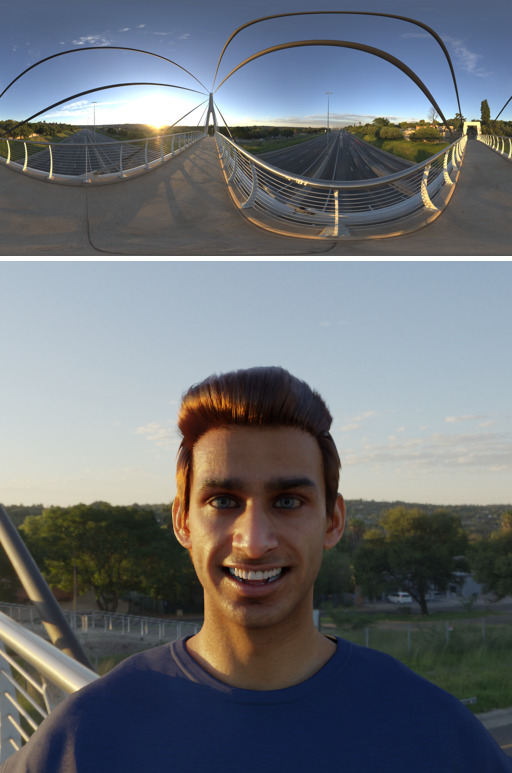} &
        \includegraphics[height=\figureheight]{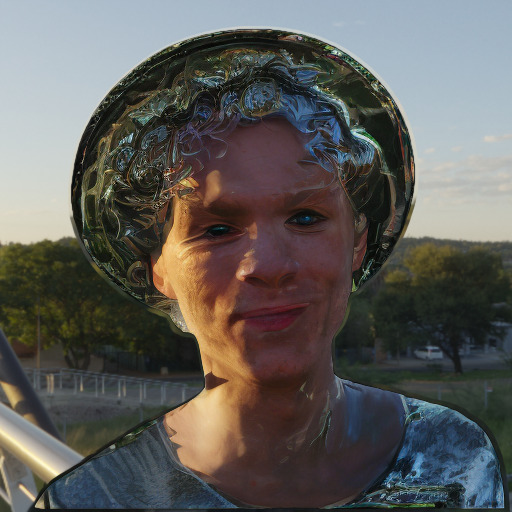} &
        \includegraphics[height=\figureheight]{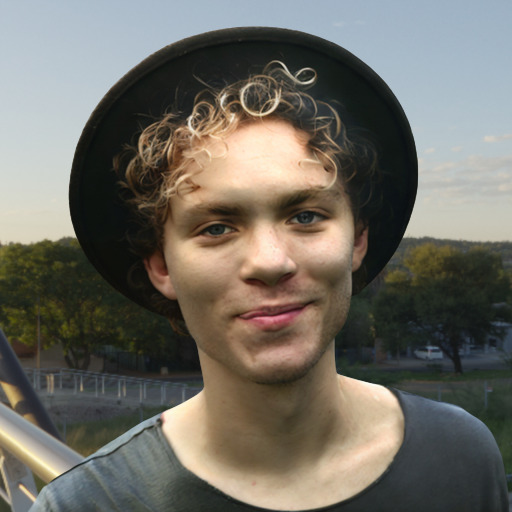} &
        \includegraphics[height=\figureheight]{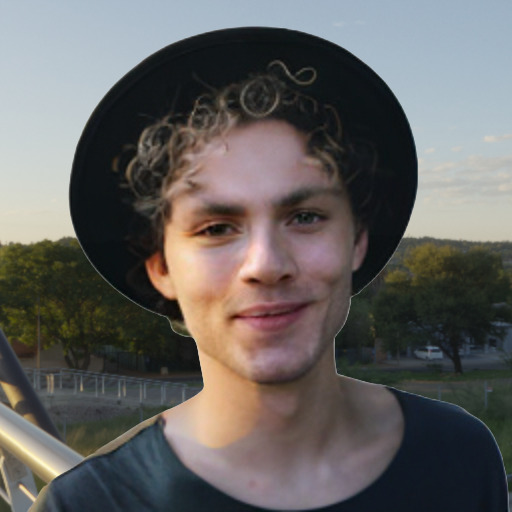} &
        \includegraphics[height=\figureheight]{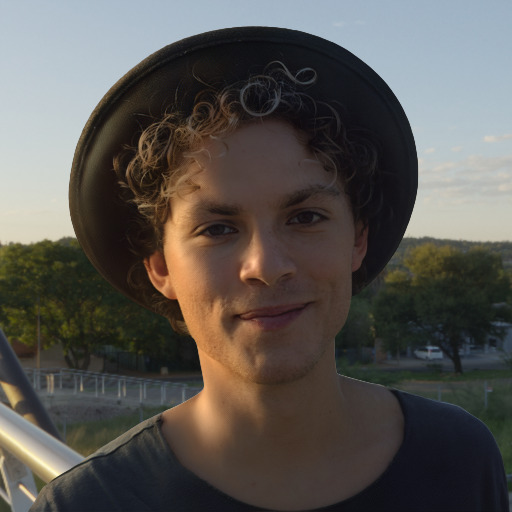} &
        \includegraphics[height=\figureheight]{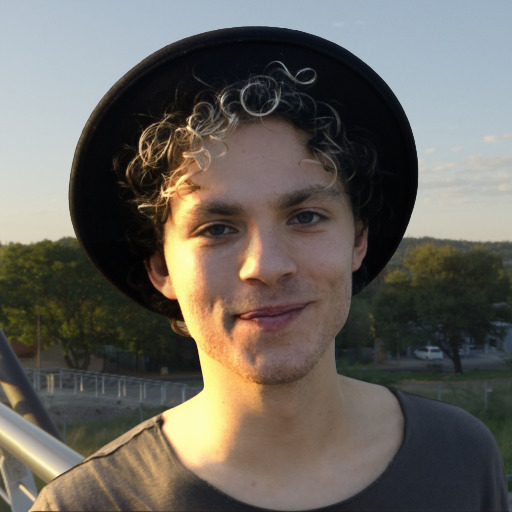} &
        \includegraphics[height=\figureheight]{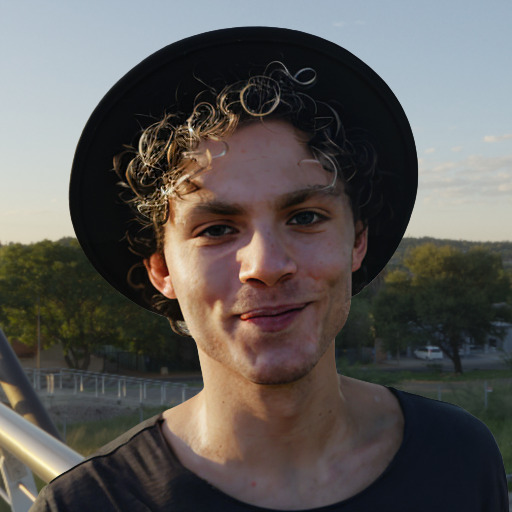} \\

        \includegraphics[height=\figureheight]{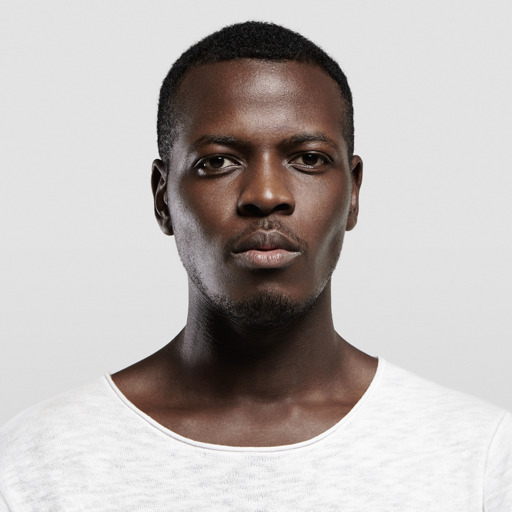} &
        \includegraphics[height=\figureheight]{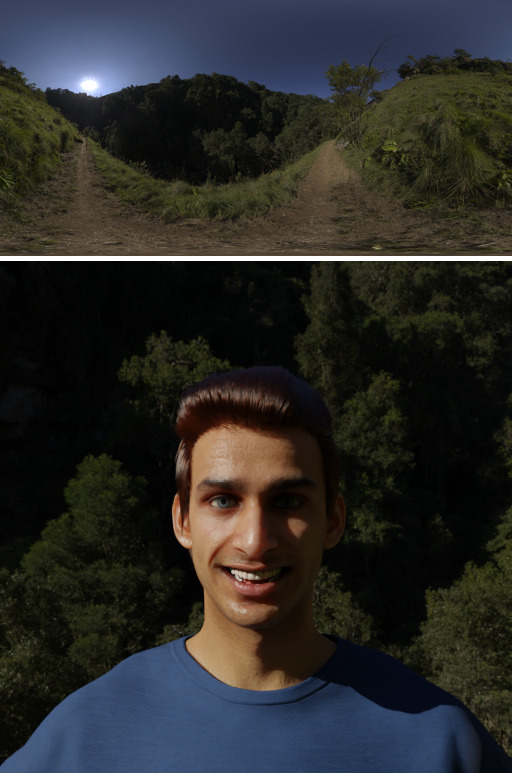} &
        \includegraphics[height=\figureheight]{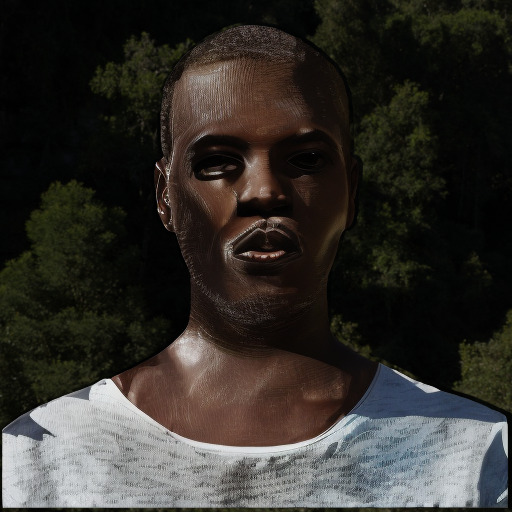} &
        \includegraphics[height=\figureheight]{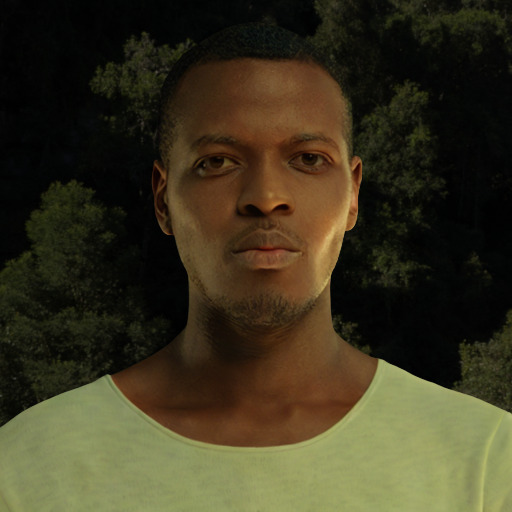} &
        \includegraphics[height=\figureheight]{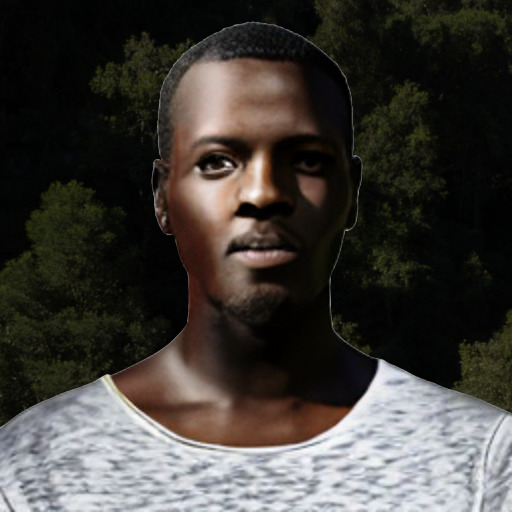} &
        \includegraphics[height=\figureheight]{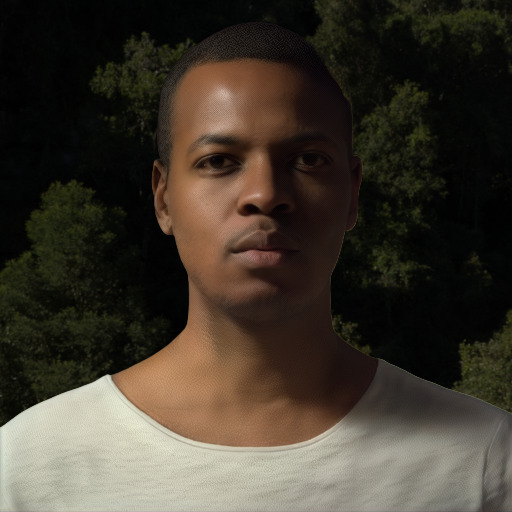} &
        \includegraphics[height=\figureheight]{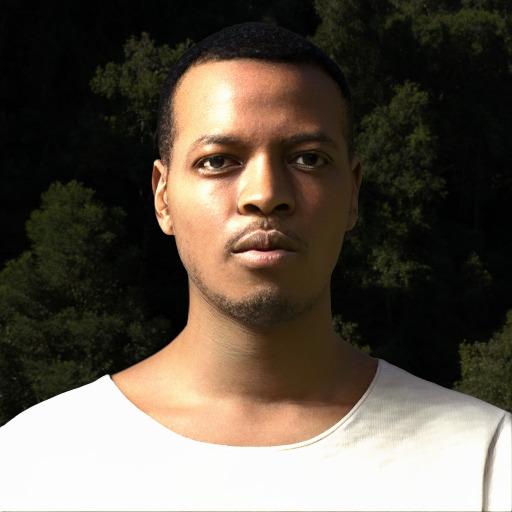} &
        \includegraphics[height=\figureheight]{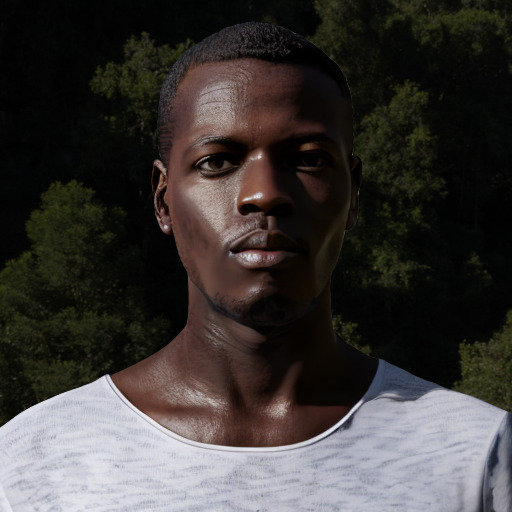} \\

    \end{tabular}
    \caption{\textit{In-the-wild portrait results}: We display the input portrait, environment map, a reference image, rendered in Blender, and baseline comparisons. DiLightNet \cite{zeng2024dilightnet} shows artifacts from 3D reconstruction failures central to its pipeline. Neural Gaffer \cite{jin2024neural_gaffer} generates inaccurate shadow contours on relit faces since it isn’t trained on human portraits. IC-Light \cite{iclight} struggles with relighting due to its choice of background as the lighting condition. Total Relighting and SwitchLight \cite{pandey2021total, kim2024switchlight}, trained on light stage data, produce soft shadows even under strong sunlight and alter skin tones. In contrast, our method achieves superior relighting while preserving subject identity.}
    \label{fig:comparison_all}
\end{figure*}

\begin{figure*}[!htbp]
    \centering

    \begin{tabular}{@{\hskip 0mm}c@{\hskip 1mm}c@{\hskip 0.5mm}c@{\hskip 0.5mm}c@{\hskip 0.5mm}c@{\hskip 0.5mm}c@{\hskip 0.5mm}c@{\hskip 0.5mm}c}
          Inputs & DiLightNet & IC-Light & Neural Gaffer & Total Relighting & SwitchLight & Ours & GT \\

         \includegraphics[height=\figureheight]{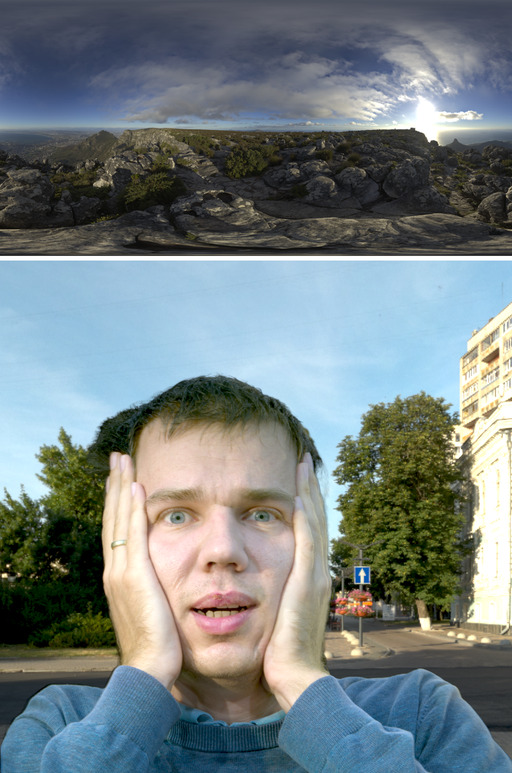} & \includegraphics[height=\figureheight]{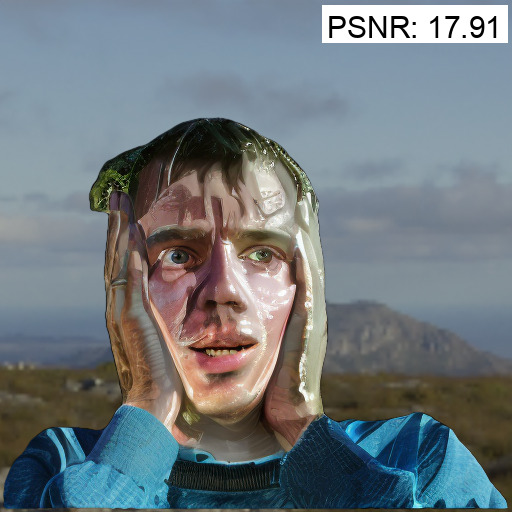} & \includegraphics[height=\figureheight]{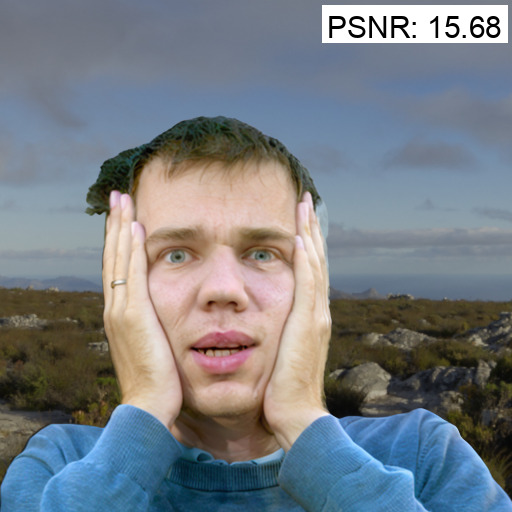} & \includegraphics[height=\figureheight]{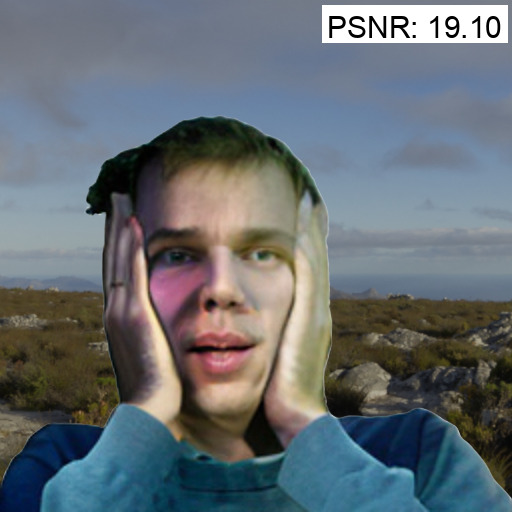} & \includegraphics[height=\figureheight]{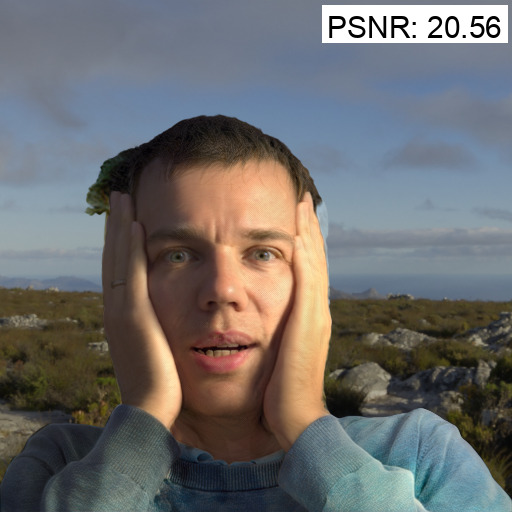} & \includegraphics[height=\figureheight]{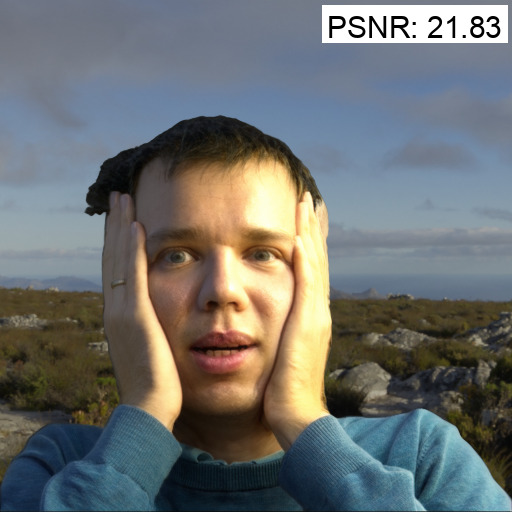} & \includegraphics[height=\figureheight]{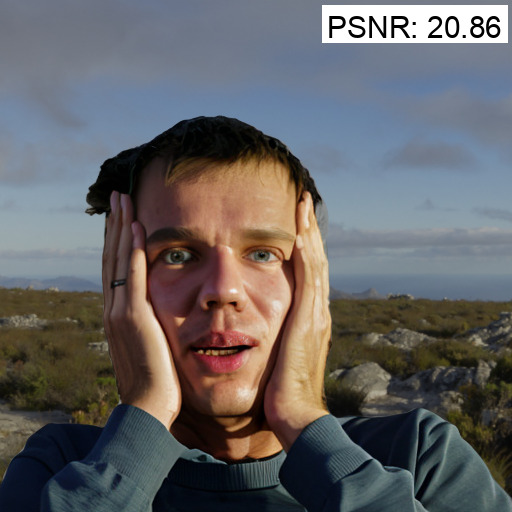}
         & \includegraphics[height=\figureheight]{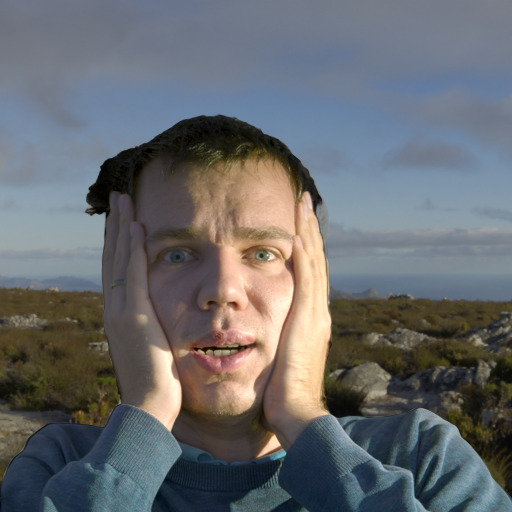} \\

         \includegraphics[height=\figureheight]{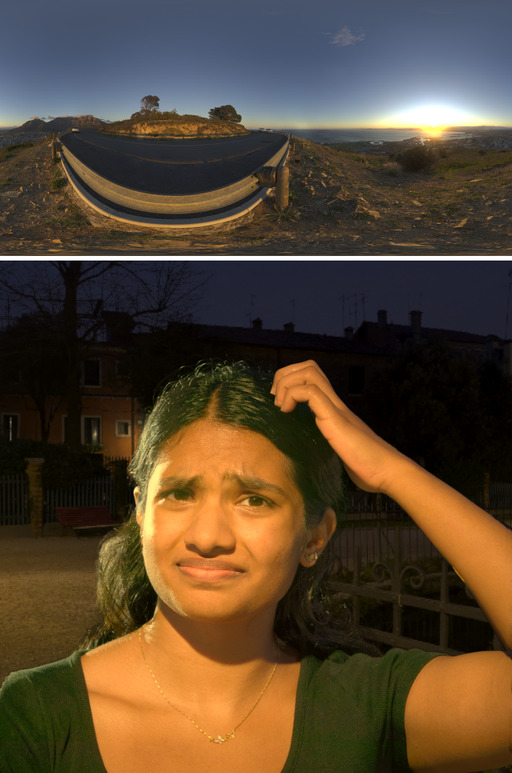} & 
         \includegraphics[height=\figureheight]{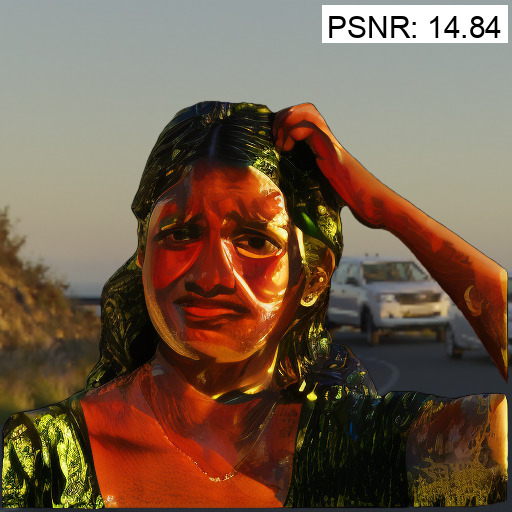} & 
         \includegraphics[height=\figureheight]{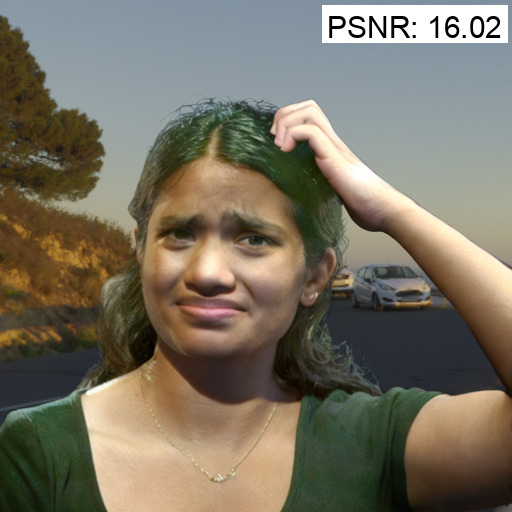} & 
         \includegraphics[height=\figureheight]{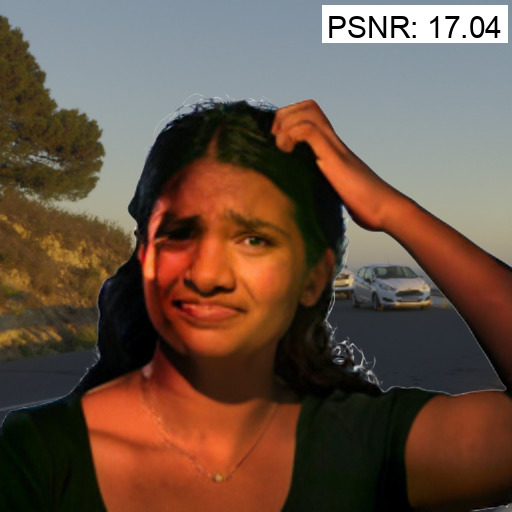} & 
         \includegraphics[height=\figureheight]{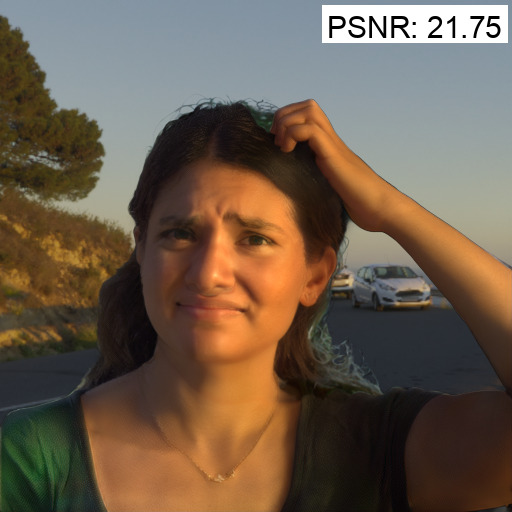} & 
         \includegraphics[height=\figureheight]{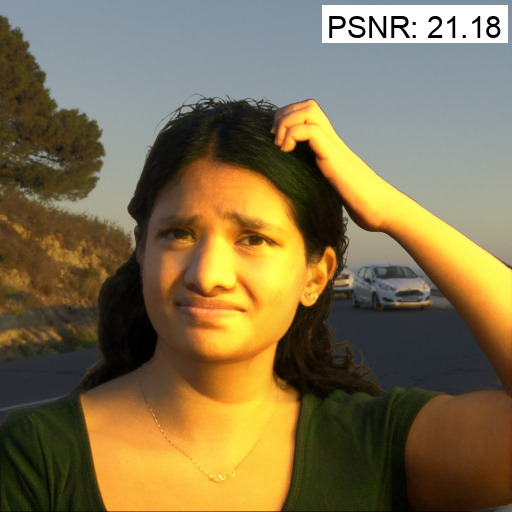} & 
         \includegraphics[height=\figureheight]{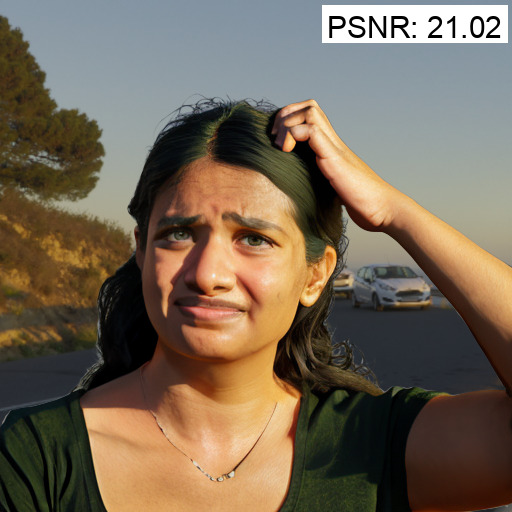} & 
         \includegraphics[height=\figureheight]{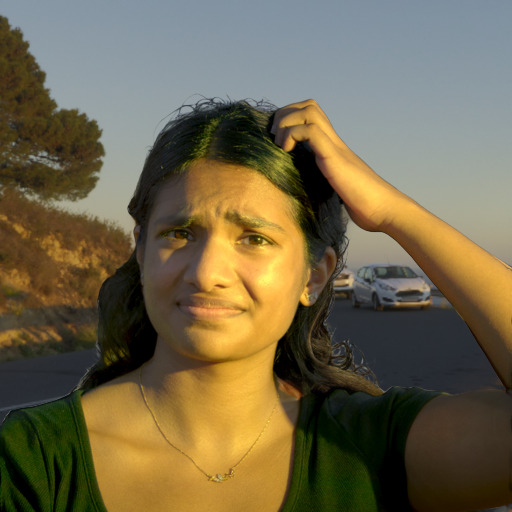} \\

         \includegraphics[height=\figureheight]{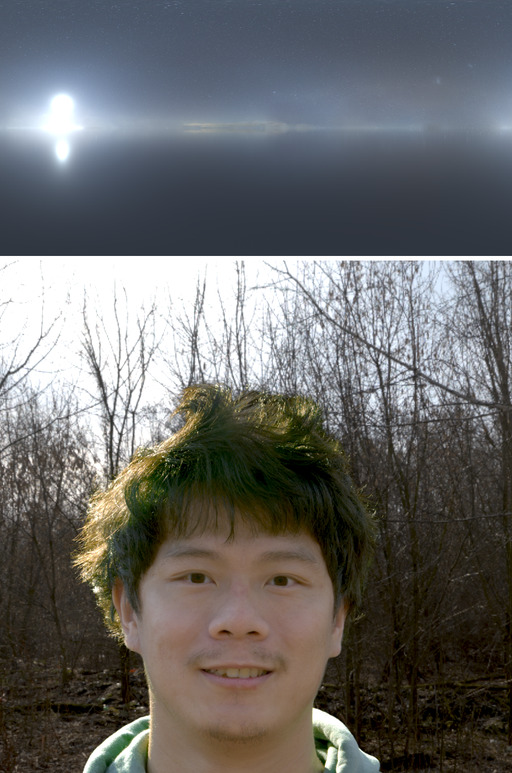} & 
         \includegraphics[height=\figureheight]{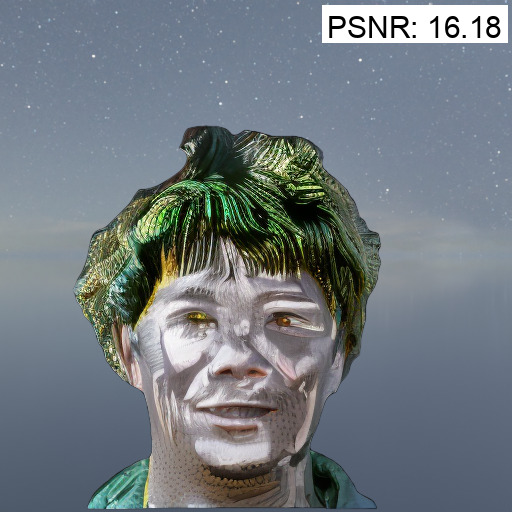} & 
         \includegraphics[height=\figureheight]{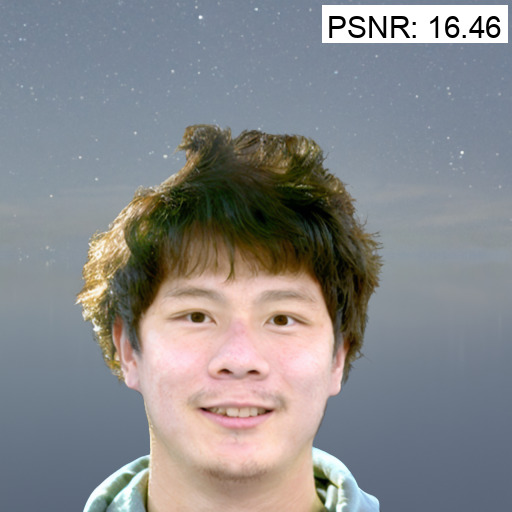} & 
         \includegraphics[height=\figureheight]{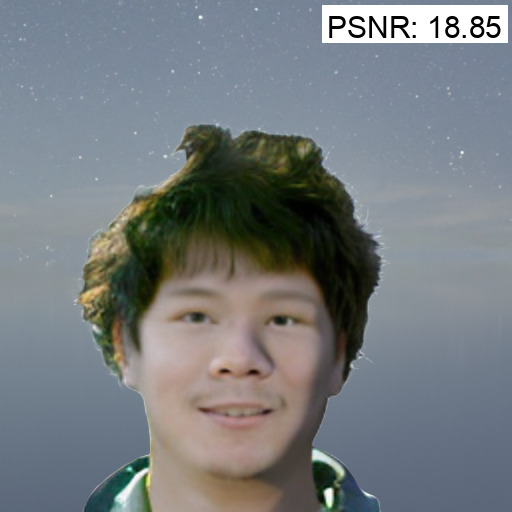} & 
         \includegraphics[height=\figureheight]{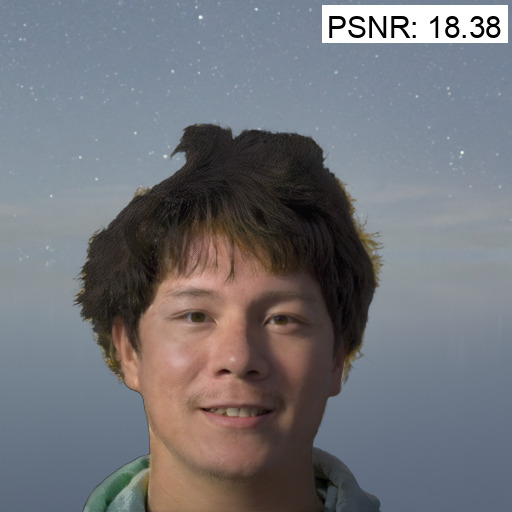} & 
         \includegraphics[height=\figureheight]{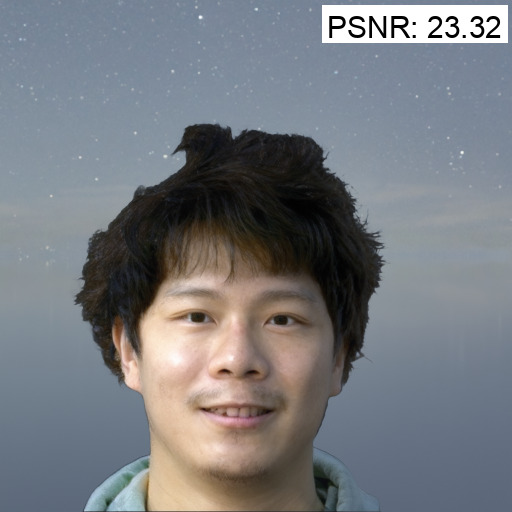} & 
         \includegraphics[height=\figureheight]{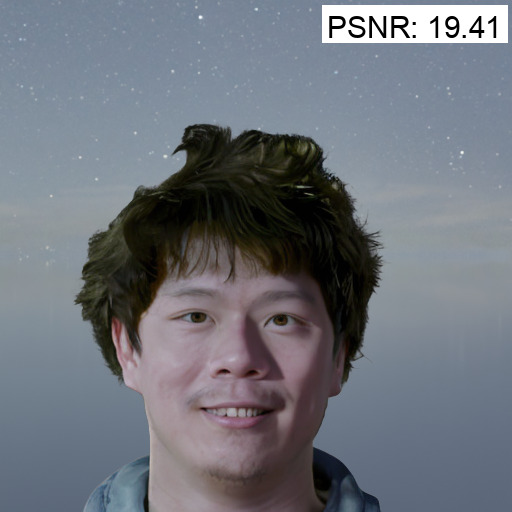} & 
         \includegraphics[height=\figureheight]{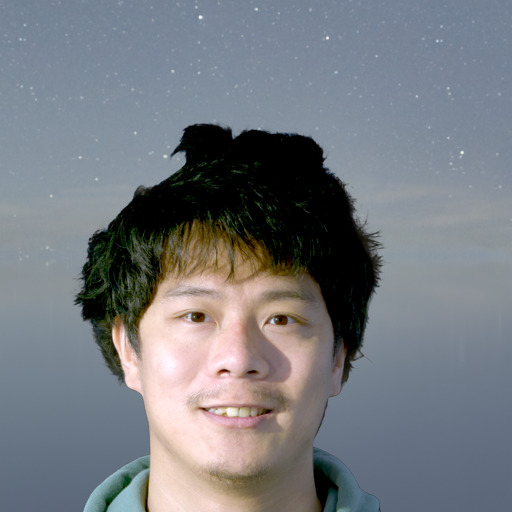} \\

    \end{tabular}
    \caption{\textit{Light Stage test results}: We compare our method against baselines on the input portrait (bottom left) from the Light Stage test set relit with a target environment map (top left).}
    \label{fig:olat_comparison}
\end{figure*}

\begin{table*}[!htbp]
    \centering
    \begin{tabularx}{\textwidth}{l|*{4}{>{\centering\arraybackslash}X}|*{4}{>{\centering\arraybackslash}X}}
        \toprule
        & \multicolumn{4}{c|}{Test Synthetic} & \multicolumn{4}{c}{Test Light Stage} \\
        \midrule
        Method         & LPIPS$\downarrow$ & SSIM$\uparrow$ & PSNR$\uparrow$  & FaceNet$\downarrow$ & LPIPS$\downarrow$ & SSIM$\uparrow$ & PSNR$\uparrow$ & FaceNet$\downarrow$ \\
        \midrule
        Ours           & \textbf{0.063}  & \textbf{0.945}  & \textbf{29.572}  & \textbf{0.165}    & 0.165  & 0.813  & 19.698  & 0.173    \\
        SwitchLight    & 0.088  & 0.911  & 21.432  & 0.198 & \textbf{0.141}  & \textbf{0.853}  & \textbf{20.299}  &  \textbf{0.152} \\
        IC-Light       & 0.108  & 0.874  & 20.283  & 0.284 & 0.172  & 0.789  & 17.440  &  0.195   \\
        DiLightNet     & 0.128  & 0.860  & 22.991  & 0.333  & 0.245  & 0.703  & 16.619  & 0.576    \\
        Neural Gaffer  & 0.102 & 0.900  & 25.327 &  0.357  &  0.196      &  0.788   &      19.311      &  0.247      \\
        \bottomrule
    \end{tabularx}
    \caption{\textit{Comparisons}: We compare against baselines on a held-out set of our synthetic dataset and data rendered through a Light Stage. While trained only on synthetic data, our model performs comparably to SwitchLight, a commercial relighting method trained with Light Stage data.} 
    \label{tab:comparison}
\end{table*}

\begin{figure*}[!htbp]
    \centering

    \begin{tabular}{@{\hskip 0mm}c@{\hskip 0.5mm}c@{\hskip 2mm}c@{\hskip 0.5mm}c@{\hskip 0.5mm}c@{\hskip 0.5mm}c@{\hskip 0.5mm}c@{\hskip 0.5mm}}
         & & Base & Base + Multi-Task & \makecell{Base + Inference \\ Adaptation} & Ours + Light Stage & Ours \\

        \includegraphics[height=\ablationheight]{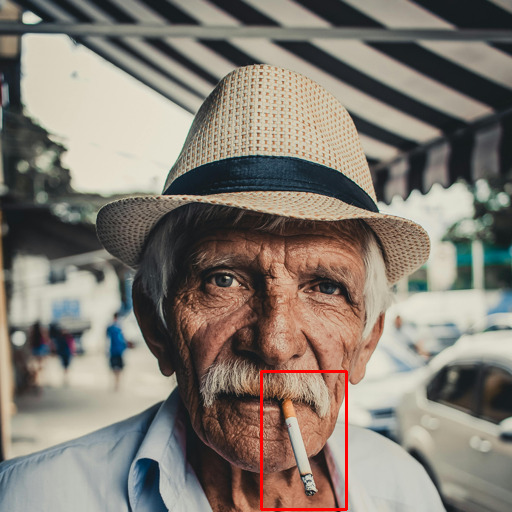} &
        \includegraphics[height=\ablationheight]{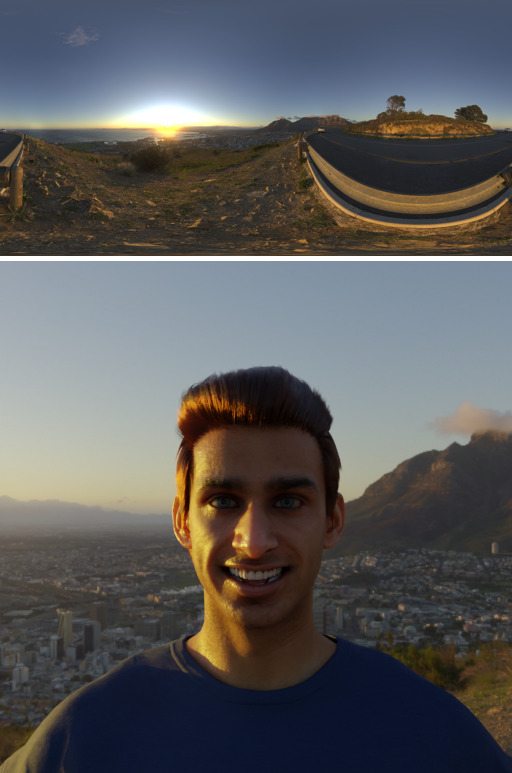} &
        \includegraphics[height=\ablationheight]{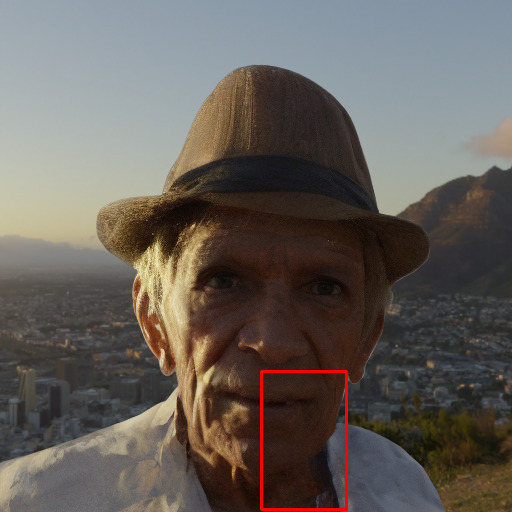} &
        \includegraphics[height=\ablationheight]{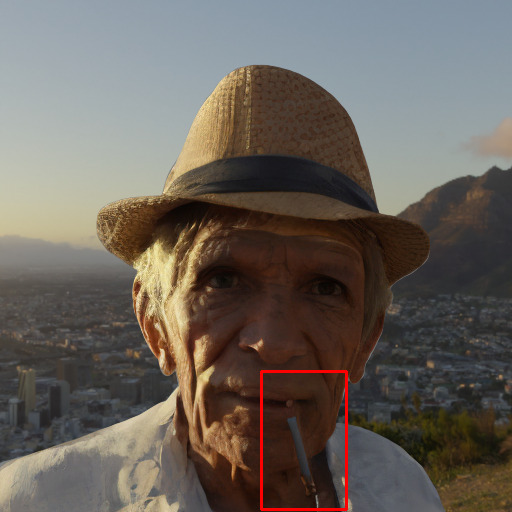} &
        \includegraphics[height=\ablationheight]{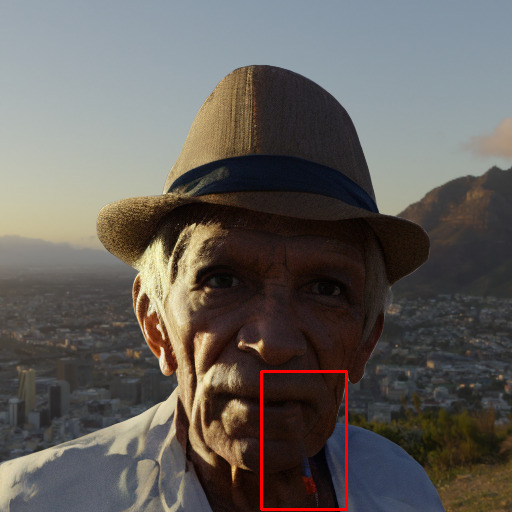} &
        \includegraphics[height=\ablationheight]{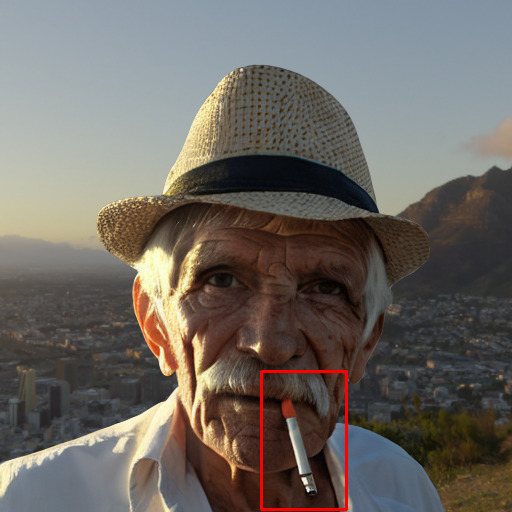} &
        \includegraphics[height=\ablationheight]{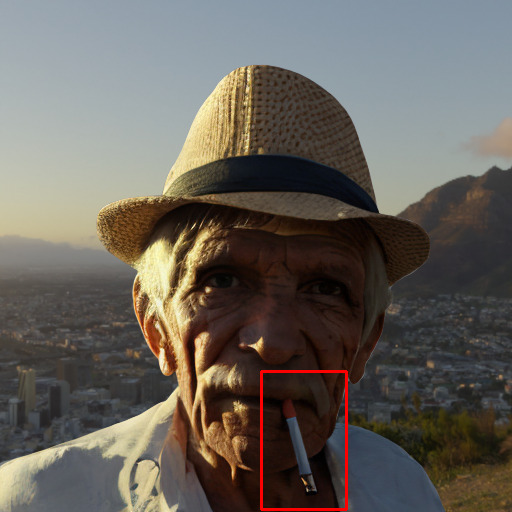} \\

        \includegraphics[height=\ablationheight]{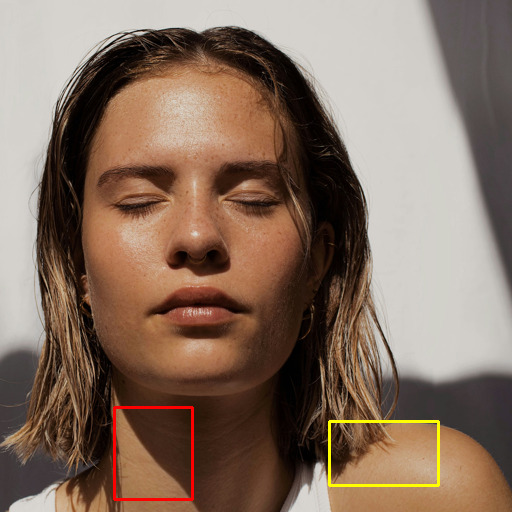} &
        \includegraphics[height=\ablationheight]{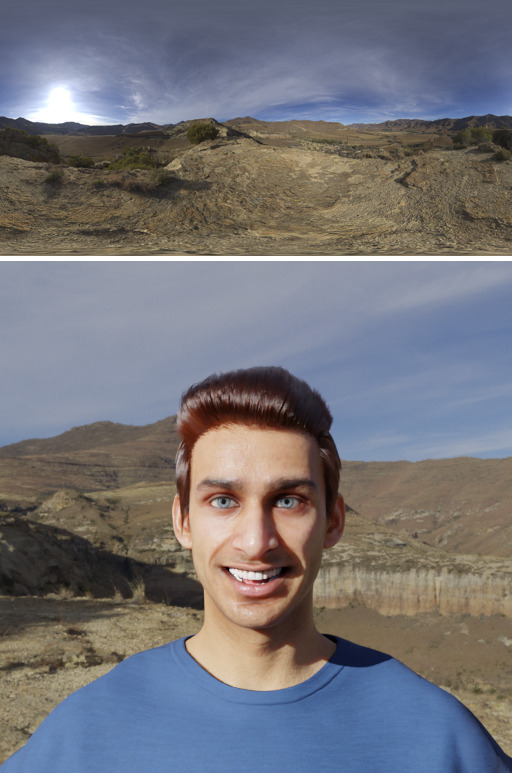} &
        \includegraphics[height=\ablationheight]{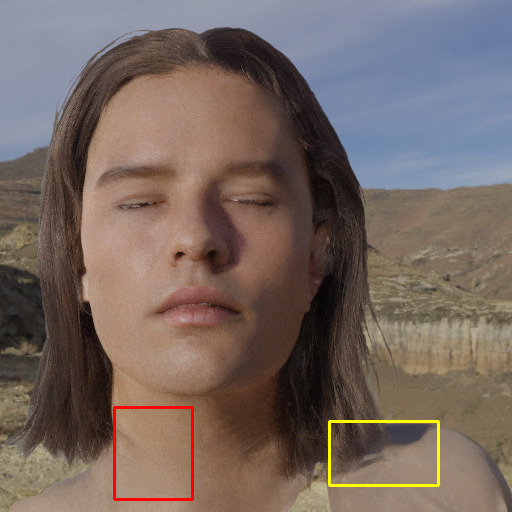} &
        \includegraphics[height=\ablationheight]{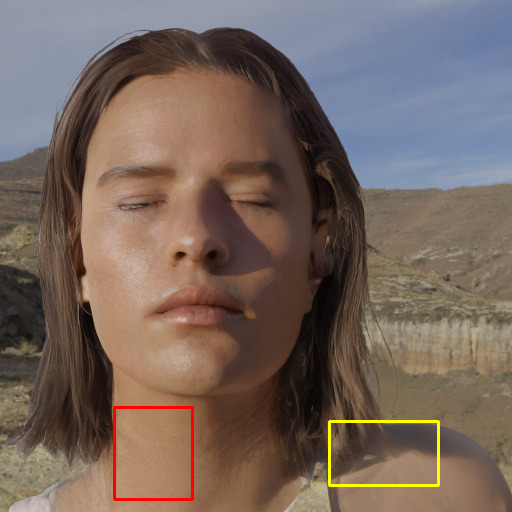} &
        \includegraphics[height=\ablationheight]{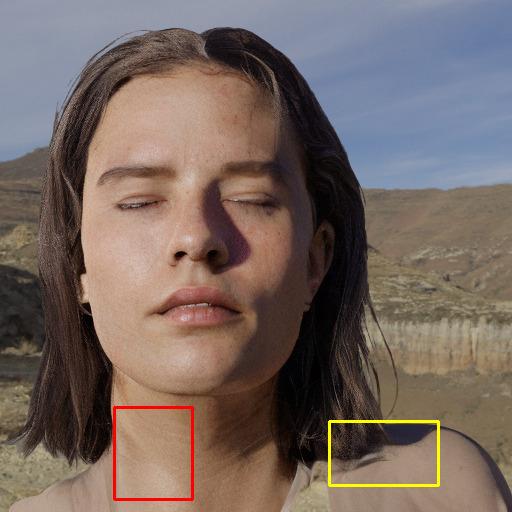} &
        \includegraphics[height=\ablationheight]{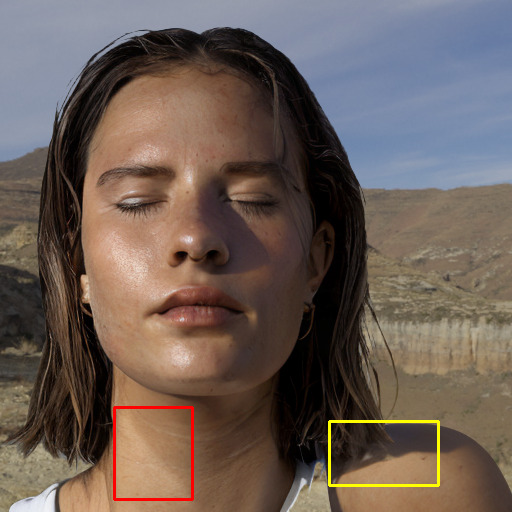} &
        \includegraphics[height=\ablationheight]{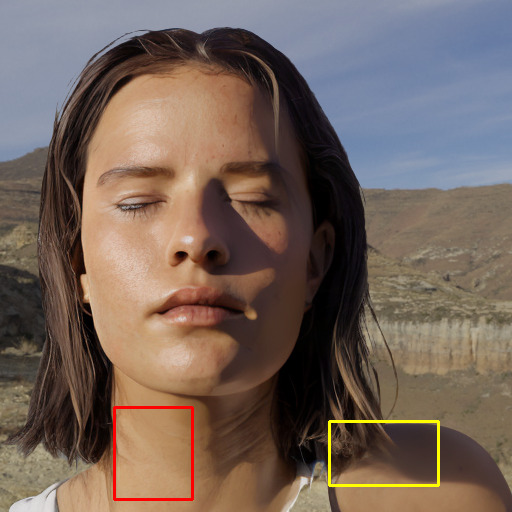} \\

    \end{tabular}
    \caption{\textit{Ablations}: We display the input portrait with its lighting condition and a reference image rendered in Blender (left). The \textit{Base} configuration fails to reproduce the portrait’s textures and alters its identity. In contrast, \textit{Base + Multi-Task} recovers some details, such as realistic skin tone (bottom row, yellow rectangle). The \textit{Base + Inference Adaptation} configuration struggles with unseen textures and accessories (e.g., the cigarette, top row, red rectangle) and produces unnatural textures for sleeveless skin (bottom row, yellow rectangle). Meanwhile, \textit{Ours + Light Stage} enhances details but inherits biases from Light Stage data and cannot remove strong shadows (neck region, bottom row, red rectangle). Finally, \textit{Ours} achieves plausible lighting, harmonizes well with the background, and preserves key details from the input portrait.}
    \label{fig:ablations}
\end{figure*}



\subsection{Implementation details}
\label{section:implementation}

We implement our model in PyTorch \cite{paszke2019pytorch} using 32 $\times$ 40GB A100 GPUs. We use a batch size of 192, a learning rate of $10^{-5}$, and the Adam \cite{diederik2014adam} optimizer. We train our model (and ablations) for 40K steps, which takes around 1 day. We initialized from the IC-Light \cite{iclight} checkpoint for background conditioned image relighting, which is fine-tuned based on Stable Diffusion 1.5~\cite{rombach2021highresolution}. We chose this particular checkpoint because we found it to be beneficial for learning our environment map based relighting model compare to a text-to-image checkpoint. We show more analysis and comparisons of this choice in supplemental material (see \cref{fig:with_and_without_iclight_ckpt} and \cref{tab:ablation_checkpoint}).
\subsection{Evaluation Results}
\label{section:quant}

We compare our method against state-of-the-art methods for portrait harmonization \cite{iclight}, portrait relighting \cite{kim2024switchlight} and object relighting \cite{zeng2024dilightnet, jin2024neural_gaffer} on both the synthetic and the light stage test set quantitatively (see \cref{tab:comparison}) and qualitatively (see \cref{fig:olat_comparison} and \cref{fig:comparison_all}). Quantitative evaluation shows that our method outperforms baselines on the synthetic test set and  performs comparably to state-of-the-art portrait relighting methods such as SwitchLight, on the Test Light Stage dataset. Even though our results do not always attain the highest PSNR, they display better visual relighting quality than baselines.


\begin{table}[!htbp]
\centering
\begin{tabular}{lccc}
\toprule
 & IC-Light & SwitchLight & Neural Gaffer \\
\midrule
Lighting & 0.92 & 0.56 & 0.65 \\
Quality & 0.57 & 0.64 & 0.73 \\
Identity & 0.52 & 0.70 & 0.65 \\
\bottomrule
\end{tabular}
\caption{\textit{User Study}: Preference rates indicate how often our method was preferred over baselines. For example, a rate of 0.92 under \textit{Lighting} means our method was preferred 92\% of the time over IC-Light. Based on 482 responses from 20 participants, our method consistently outperforms baselines in \textit{lighting}, \textit{image quality}, and \textit{subject identity}, since all preference rates exceed 0.5. This highlights superior image quality over relighting methods \cite{kim2024switchlight, jin2024neural_gaffer} and better lighting over harmonization methods \cite{iclight}.}

\label{tab:user_study}
\end{table}

We further conduct a user study (see \cref{tab:user_study}) to quantify human perceptual preference for relighting. For each pair (our method vs. a baseline), participants are asked three questions: (1)  which method has better \textbf{lighting} (2) which has better image \textbf{quality} (3) which better preserves \textbf{identity}. All questions are presented as a 2-alternative forced choice (2AFC). We collect 482 responses from 20 participants with diverse backgrounds, ranging from design to computer science. Results show that our methods outperforms baselines in perceived image lighting, quality, and identity preservation. Refer to the supplementary material for screenshots, \cref{fig:user_study_a} and \cref{fig:user_study_b}, showcasing the precise format of our user study.

\begin{table*}[!htbp]
\centering
\begin{tabularx}{\textwidth}{l|*{4}{>{\centering\arraybackslash}X}|*{4}{>{\centering\arraybackslash}X}}
\toprule
& \multicolumn{4}{c|}{Test Synthetic} & \multicolumn{4}{c}{Test Light Stage} \\
\midrule
Method & LPIPS$\downarrow$ & SSIM$\uparrow$ & PSNR$\uparrow$ &  FaceNet$\downarrow$ & LPIPS$\downarrow$ & SSIM$\uparrow$ & PSNR$\uparrow$ & FaceNet$\downarrow$ \\
\midrule

Base & 
0.066 & 0.942 & 29.131 & 0.193 &  
0.210 & 0.790 & 18.919 & 0.295  \\

Base + Multi-Task & 
0.066 & 0.942 & 29.049 & 0.196 &  
0.186 & 0.797 & 19.184 & 0.242 \\

Base + Inference Adaptation & \textbf{0.062} & \textbf{0.946} & \textbf{29.638} & \textbf{0.163} & 
0.178 & 0.810 & 19.484 & 0.179  \\
\midrule

Ours           & 0.063 & 0.945  & 29.572  & 0.165  & 0.165  & 0.813  & 19.698  & 0.173   \\

Ours + Light Stage & 
0.065 & 0.942 & 29.126 & 0.171 & 
\textbf{0.156} & \textbf{0.822} & \textbf{20.136} & \textbf{0.149} \\

\bottomrule
\end{tabularx}
\label{tab:ablations}
\caption{\textit{Ablations} highlight the contributions of each component i.e. \textit{Multi-Task training} and \textit{Inference-time Adaptation} (\cref{sec:train} and \cref{sec:inf} respectively). Adding Light Stage data during training improves performance on Light Stage Test set, and qualitatively improves details but brings lighting biases (See \cref{fig:ablations}).}
\end{table*}

\begin{figure}[!h]
    \includegraphics[width=0.475\textwidth]{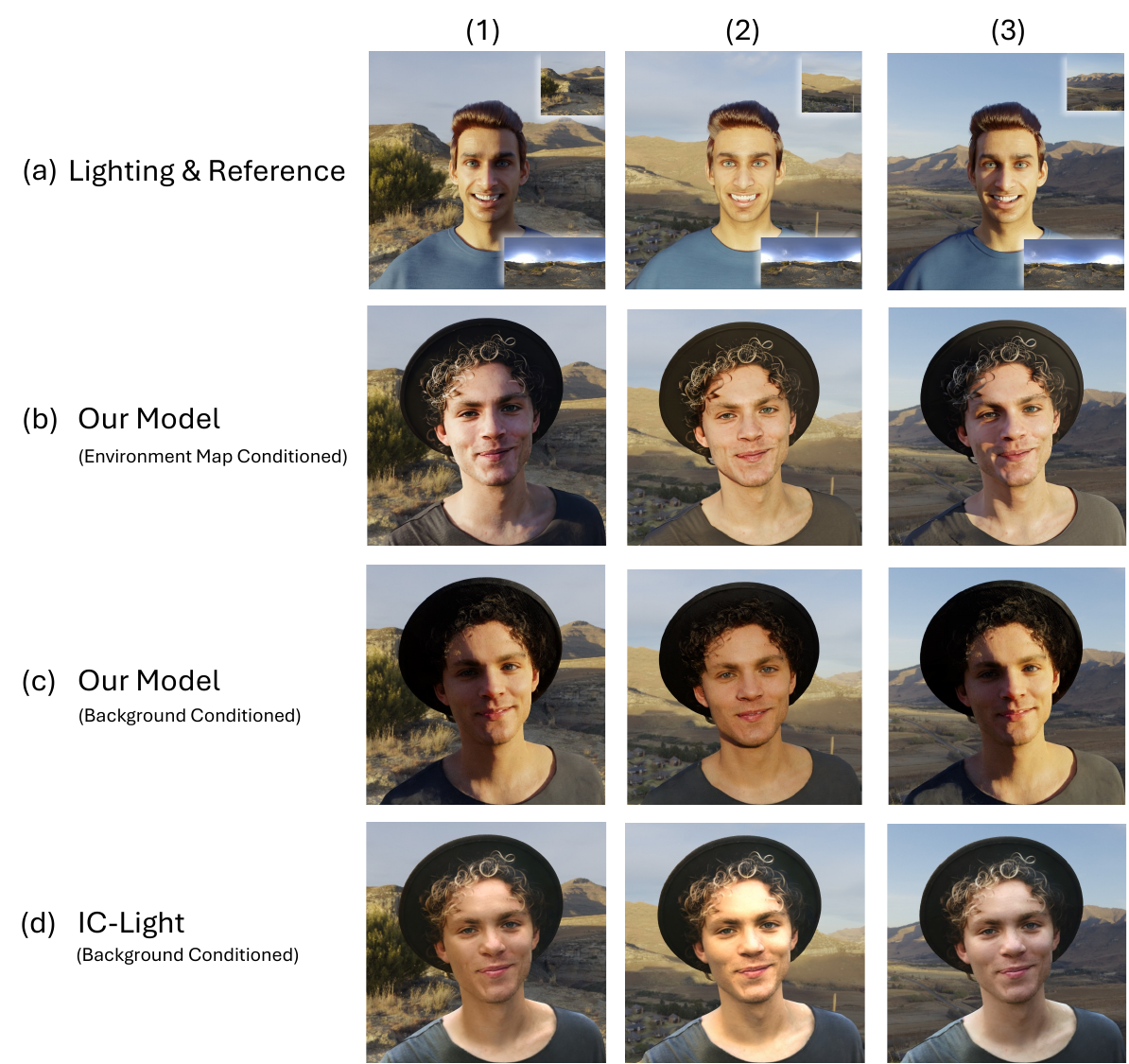}
    \caption{\textit{Background vs. Environment Map as Lighting Conditions}: The background provides limited lighting cues, leading the background-conditioned model to produce inaccurate lighting (note the wrong lighting direction in (1)-(c)). Even so, by utilizing our \textit{Synthetic Faces} dataset, the background-conditioned model is able to generate plausible lighting, characterized by strong cast shadows, whereas harmonization methods such as IC-Light \cite{iclight} fall short. See \cref{fig:comparison_all}, Row 3 for the input portrait.}
    \label{fig:bg_v_env}
\end{figure}

\subsection{Ablations} \label{subsec:ablations}

We conduct an ablation study to evaluate the contribution of our two key methods for domain adaptation: multi-task training (See \cref{sec:train}) and inference-time adaptation (See \cref{sec:inf}). 

We start with a \textit{Base} configuration that excludes both multi-task training and inference time adaptation. Next, we examine the individual impact of each component by separately adding multi-task training, denoted as \textit{Base + Multi-Task}, and inference time adaptation, denoted as \textit{Base + Inference Adaptation}.  Our full configuration, combining both techniques is referred to as \textit{Ours}. Finally, we explore the role of Light Stage data, by adding a fraction of it to each training batch, denoted as \textit{Ours + Light Stage}. Please refer to the supplementary material, \cref{sec:additional_results}, for more details on the Light Stage data.

\cref{fig:ablations} shows the effect of each configuration. \textit{Base} loses important details from the input and fails to produce textures in clothing or accessories. \textit{Base + Multi-Task} shows partial detail recovery, and \textit{Base + Inference Adaptation} enhances finer details by leveraging information present in the input portrait but still lacks photo-realism. \textit{Ours + Light Stage} addresses identity and texture issues but inherits lighting biases from the Light Stage dataset. For example, under strong sunlight, it yields oversaturated images (see \cref{fig:light_stage_v_switchlight}, in the supplementary material). Similar artifacts appear in other methods (e.g., SwitchLight) that are trained on Light Stage data. It also struggles to remove strong shadows, which are rarely present in Light Stage captures. Finally, \textit{Ours},  generates images with plausible lighting, that are well harmonized with background and preserve important details from the input portrait. These findings are corroborated by our quantitative evaluation in \cref{tab:ablations}.


\subsection{Environment map better than background}

We train two variants of our model, one using a background and the other using an environment map as lighting condition. We observe that while in many cases, the background-conditioned model produces plausible lighting and appears well harmonized with the background, when we continuously rotate the environment map, lighting inconsistencies appear. See \cref{fig:bg_v_env} for lighting inaccuracies in a background-conditioned method. Despite these, leveraging our synthetic dataset makes our background-conditioned model generate plausible self-occlusions, whereas harmonization methods such as \cite{iclight} fail in this use case.





\section{Limitations \& Discussion}  \label{section:limitations}
Despite the advances proposed by our method both in terms of simplicity and image quality, it bears some limitations. In particular, our rendering pipeline could achieve a higher level of realism if we specialized it for rendering humans. Of note, it does not model unseen occluders casting shadows on the subject's face, accessories such as hats, glasses, or even facial hair, which limits the diversity of lighting our method saw during training. Despite this, our method achieves great generalization capabilities. Furthermore, user editing of the light is cumbersome in the current representation; we could improve this aspect by proposing a parametric representation of the light, such as 3D point lights or spherical Gaussians, that is easier to understand and edit for users. Additional qualitative examples illustrating the limitations of our method are provided in the supplementary material, see \cref{fig:failure_cases}.



\section{Conclusion} 

We present SynthLight, a Portrait Relighting Diffusion model that relights in-the-wild images while garnering lighting supervision only from synthetic data. It underscores the potential of using synthetic data to achieve plausible portrait relighting, enabling interesting lighting effects such as strong cast shadows, catch light in the eyes, and inter-reflections. 

\section*{Acknowledgement}
We thank Weijie Lyu, Ziwen Chen, Haian Jin, Vikas Thamizharasan, Natalia Pacheco-Tallaj, Ryusuke Sugimoto and Christophe Bolduc for their insightful discussions and the many participants in our user study. We also thank Kalyan Sunkavalli and Nathan Carr for their support. 

\vspace{10mm}

{
    \small
    \bibliographystyle{ieeenat_fullname}
    \bibliography{main}

\begin{thebibliography}{61}
\providecommand{\natexlab}[1]{#1}
\providecommand{\url}[1]{\texttt{#1}}
\expandafter\ifx\csname urlstyle\endcsname\relax
  \providecommand{\doi}[1]{doi: #1}\else
  \providecommand{\doi}{doi: \begingroup \urlstyle{rm}\Url}\fi

\bibitem[Balaji et~al.(2022)Balaji, Nah, Huang, Vahdat, Song, Zhang, Kreis, Aittala, Aila, Laine, et~al.]{balaji2022ediff}
Yogesh Balaji, Seungjun Nah, Xun Huang, Arash Vahdat, Jiaming Song, Qinsheng Zhang, Karsten Kreis, Miika Aittala, Timo Aila, Samuli Laine, et~al.
\newblock ediff-i: Text-to-image diffusion models with an ensemble of expert denoisers.
\newblock \emph{arXiv preprint arXiv:2211.01324}, 2022.

\bibitem[Barron and Malik(2014)]{barron2014shape}
Jonathan~T Barron and Jitendra Malik.
\newblock Shape, illumination, and reflectance from shading.
\newblock \emph{IEEE transactions on pattern analysis and machine intelligence}, 37\penalty0 (8):\penalty0 1670--1687, 2014.

\bibitem[Bi et~al.(2021)Bi, Lombardi, Saito, Simon, Wei, Mcphail, Ramamoorthi, Sheikh, and Saragih]{bi2021deep}
Sai Bi, Stephen Lombardi, Shunsuke Saito, Tomas Simon, Shih-En Wei, Kevyn Mcphail, Ravi Ramamoorthi, Yaser Sheikh, and Jason Saragih.
\newblock Deep relightable appearance models for animatable faces.
\newblock \emph{ACM Transactions on Graphics (ToG)}, 40\penalty0 (4):\penalty0 1--15, 2021.

\bibitem[Blanz and Vetter(2023)]{blanz2023morphable}
Volker Blanz and Thomas Vetter.
\newblock A morphable model for the synthesis of 3d faces.
\newblock In \emph{Seminal Graphics Papers: Pushing the Boundaries, Volume 2}, pages 157--164. 2023.

\bibitem[Brooks et~al.(2023)Brooks, Holynski, and Efros]{brooks2023instructpix2pix}
Tim Brooks, Aleksander Holynski, and Alexei~A Efros.
\newblock Instructpix2pix: Learning to follow image editing instructions.
\newblock In \emph{Proceedings of the IEEE/CVF Conference on Computer Vision and Pattern Recognition}, pages 18392--18402, 2023.

\bibitem[Cai et~al.(2024)Cai, Jiang, Chen, Lai, Fu, Shi, and Gao]{cai2024real}
Ziqi Cai, Kaiwen Jiang, Shu-Yu Chen, Yu-Kun Lai, Hongbo Fu, Boxin Shi, and Lin Gao.
\newblock Real-time 3d-aware portrait video relighting.
\newblock In \emph{Proceedings of the IEEE/CVF Conference on Computer Vision and Pattern Recognition}, pages 6221--6231, 2024.

\bibitem[Cook and Torrance(1982)]{cook1982reflectance}
Robert~L Cook and Kenneth~E. Torrance.
\newblock A reflectance model for computer graphics.
\newblock \emph{ACM Transactions on Graphics (ToG)}, 1\penalty0 (1):\penalty0 7--24, 1982.

\bibitem[Debevec(2008)]{debevec2008rendering}
Paul Debevec.
\newblock Rendering synthetic objects into real scenes: Bridging traditional and image-based graphics with global illumination and high dynamic range photography.
\newblock In \emph{Acm siggraph 2008 classes}, pages 1--10. 2008.

\bibitem[Debevec et~al.(2000)Debevec, Hawkins, Tchou, Duiker, Sarokin, and Sagar]{debevec2000acquiring}
Paul Debevec, Tim Hawkins, Chris Tchou, Haarm-Pieter Duiker, Westley Sarokin, and Mark Sagar.
\newblock Acquiring the reflectance field of a human face.
\newblock In \emph{Proceedings of the 27th annual conference on Computer graphics and interactive techniques}, pages 145--156, 2000.

\bibitem[Dhariwal and Nichol(2021)]{dhariwal2021diffusion}
Prafulla Dhariwal and Alexander Nichol.
\newblock Diffusion models beat gans on image synthesis.
\newblock \emph{Advances in neural information processing systems}, 34:\penalty0 8780--8794, 2021.

\bibitem[Diederik(2014)]{diederik2014adam}
P~Kingma Diederik.
\newblock Adam: A method for stochastic optimization.
\newblock \emph{(No Title)}, 2014.

\bibitem[Donner and Jensen(2006)]{donner2006spectral}
Craig Donner and Henrik~Wann Jensen.
\newblock A spectral bssrdf for shading human skin.
\newblock \emph{Rendering techniques}, 2006:\penalty0 409--418, 2006.

\bibitem[Gal et~al.(2022)Gal, Alaluf, Atzmon, Patashnik, Bermano, Chechik, and Cohen-Or]{gal2022image}
Rinon Gal, Yuval Alaluf, Yuval Atzmon, Or Patashnik, Amit~H Bermano, Gal Chechik, and Daniel Cohen-Or.
\newblock An image is worth one word: Personalizing text-to-image generation using textual inversion.
\newblock \emph{arXiv preprint arXiv:2208.01618}, 2022.

\bibitem[He et~al.(2022)He, Chen, Xie, Li, Doll{\'a}r, and Girshick]{he2022masked}
Kaiming He, Xinlei Chen, Saining Xie, Yanghao Li, Piotr Doll{\'a}r, and Ross Girshick.
\newblock Masked autoencoders are scalable vision learners.
\newblock In \emph{Proceedings of the IEEE/CVF conference on computer vision and pattern recognition}, pages 16000--16009, 2022.

\bibitem[He et~al.(2024)He, Clausen, Ta{\c{s}}el, Ma, Pilarski, Xian, Rikker, Yu, Burgert, Yu, et~al.]{he2024diffrelight}
Mingming He, Pascal Clausen, Ahmet~Levent Ta{\c{s}}el, Li Ma, Oliver Pilarski, Wenqi Xian, Laszlo Rikker, Xueming Yu, Ryan Burgert, Ning Yu, et~al.
\newblock Diffrelight: Diffusion-based facial performance relighting.
\newblock \emph{arXiv preprint arXiv:2410.08188}, 2024.

\bibitem[Ho and Salimans(2022)]{ho2022classifier}
Jonathan Ho and Tim Salimans.
\newblock Classifier-free diffusion guidance.
\newblock \emph{arXiv preprint arXiv:2207.12598}, 2022.

\bibitem[Ho et~al.(2020)Ho, Jain, and Abbeel]{ho2020denoising}
Jonathan Ho, Ajay Jain, and Pieter Abbeel.
\newblock Denoising diffusion probabilistic models.
\newblock \emph{Advances in neural information processing systems}, 33:\penalty0 6840--6851, 2020.

\bibitem[Hu et~al.(2021)Hu, Shen, Wallis, Allen-Zhu, Li, Wang, Wang, and Chen]{hu2021lora}
Edward~J Hu, Yelong Shen, Phillip Wallis, Zeyuan Allen-Zhu, Yuanzhi Li, Shean Wang, Lu Wang, and Weizhu Chen.
\newblock Lora: Low-rank adaptation of large language models.
\newblock \emph{arXiv preprint arXiv:2106.09685}, 2021.

\bibitem[Jin et~al.(2024)Jin, Li, Luan, Xiangli, Bi, Zhang, Xu, Sun, and Snavely]{jin2024neural_gaffer}
Haian Jin, Yuan Li, Fujun Luan, Yuanbo Xiangli, Sai Bi, Kai Zhang, Zexiang Xu, Jin Sun, and Noah Snavely.
\newblock Neural gaffer: Relighting any object via diffusion, 2024.

\bibitem[Karras et~al.(2022)Karras, Aittala, Aila, and Laine]{karras2022elucidating}
Tero Karras, Miika Aittala, Timo Aila, and Samuli Laine.
\newblock Elucidating the design space of diffusion-based generative models.
\newblock \emph{Advances in neural information processing systems}, 35:\penalty0 26565--26577, 2022.

\bibitem[Karras et~al.(2024)Karras, Aittala, Lehtinen, Hellsten, Aila, and Laine]{karras2024analyzing}
Tero Karras, Miika Aittala, Jaakko Lehtinen, Janne Hellsten, Timo Aila, and Samuli Laine.
\newblock Analyzing and improving the training dynamics of diffusion models.
\newblock In \emph{Proceedings of the IEEE/CVF Conference on Computer Vision and Pattern Recognition}, pages 24174--24184, 2024.

\bibitem[Kim et~al.(2024)Kim, Jang, Yoon, Lee, Na, and Woo]{kim2024switchlight}
Hoon Kim, Minje Jang, Wonjun Yoon, Jisoo Lee, Donghyun Na, and Sanghyun Woo.
\newblock Switchlight: Co-design of physics-driven architecture and pre-training framework for human portrait relighting.
\newblock In \emph{Proceedings of the IEEE/CVF Conference on Computer Vision and Pattern Recognition}, pages 25096--25106, 2024.

\bibitem[Kim et~al.(2022)Kim, Rushmeier, Dorsey, Nowrouzezahrai, Syed, Jarosz, and Darke]{kim2022countering}
Theodore Kim, Holly Rushmeier, Julie Dorsey, Derek Nowrouzezahrai, Raqi Syed, Wojciech Jarosz, and AM Darke.
\newblock Countering racial bias in computer graphics research.
\newblock In \emph{ACM SIGGRAPH 2022 Talks}, pages 1--2. 2022.

\bibitem[Kynk{\"a}{\"a}nniemi et~al.(2024)Kynk{\"a}{\"a}nniemi, Aittala, Karras, Laine, Aila, and Lehtinen]{kynkaanniemi2024applying}
Tuomas Kynk{\"a}{\"a}nniemi, Miika Aittala, Tero Karras, Samuli Laine, Timo Aila, and Jaakko Lehtinen.
\newblock Applying guidance in a limited interval improves sample and distribution quality in diffusion models.
\newblock \emph{arXiv preprint arXiv:2404.07724}, 2024.

\bibitem[Liu et~al.(2024)Liu, Li, Wu, and Lee]{liu2024visual}
Haotian Liu, Chunyuan Li, Qingyang Wu, and Yong~Jae Lee.
\newblock Visual instruction tuning.
\newblock \emph{Advances in neural information processing systems}, 36, 2024.

\bibitem[Mashita et~al.(2011)Mashita, Mukaigawa, and Yagi]{mashita2011measuring}
Tomohiro Mashita, Yasuhiro Mukaigawa, and Yasushi Yagi.
\newblock Measuring and modeling of multi-layered subsurface scattering for human skin.
\newblock In \emph{Virtual and Mixed Reality-New Trends: International Conference, Virtual and Mixed Reality 2011, Held as Part of HCI International 2011, Orlando, FL, USA, July 9-14, 2011, Proceedings, Part I 4}, pages 335--344. Springer, 2011.

\bibitem[Nestmeyer et~al.(2020)Nestmeyer, Lalonde, Matthews, and Lehrmann]{nestmeyer2020learning}
Thomas Nestmeyer, Jean-Fran{\c{c}}ois Lalonde, Iain Matthews, and Andreas Lehrmann.
\newblock Learning physics-guided face relighting under directional light.
\newblock In \emph{Proceedings of the IEEE/CVF Conference on Computer Vision and Pattern Recognition}, pages 5124--5133, 2020.

\bibitem[Pandey et~al.(2021)Pandey, Orts-Escolano, Legendre, Haene, Bouaziz, Rhemann, Debevec, and Fanello]{pandey2021total}
Rohit Pandey, Sergio Orts-Escolano, Chloe Legendre, Christian Haene, Sofien Bouaziz, Christoph Rhemann, Paul~E Debevec, and Sean~Ryan Fanello.
\newblock Total relighting: learning to relight portraits for background replacement.
\newblock \emph{ACM Trans. Graph.}, 40\penalty0 (4):\penalty0 43--1, 2021.

\bibitem[Paris et~al.(2003)Paris, Sillion, and Quan]{paris2003lightweight}
Sylvain Paris, Fran{\c{c}}ois~X Sillion, and Long Quan.
\newblock Lightweight face relighting.
\newblock In \emph{11th Pacific Conference onComputer Graphics and Applications, 2003. Proceedings.}, pages 41--50. IEEE, 2003.

\bibitem[Paszke et~al.(2019)Paszke, Gross, Massa, Lerer, Bradbury, Chanan, Killeen, Lin, Gimelshein, Antiga, et~al.]{paszke2019pytorch}
Adam Paszke, Sam Gross, Francisco Massa, Adam Lerer, James Bradbury, Gregory Chanan, Trevor Killeen, Zeming Lin, Natalia Gimelshein, Luca Antiga, et~al.
\newblock Pytorch: An imperative style, high-performance deep learning library.
\newblock \emph{Advances in neural information processing systems}, 32, 2019.

\bibitem[Phong(1998)]{phong1998illumination}
Bui~Tuong Phong.
\newblock Illumination for computer generated pictures.
\newblock In \emph{Seminal graphics: pioneering efforts that shaped the field}, pages 95--101. 1998.

\bibitem[Rao et~al.(2024)Rao, Fox, Meka, BR, Zhan, Weyrich, Bickel, Pfister, Matusik, Elgharib, et~al.]{rao2024lite2relight}
Pramod Rao, Gereon Fox, Abhimitra Meka, Mallikarjun BR, Fangneng Zhan, Tim Weyrich, Bernd Bickel, Hanspeter Pfister, Wojciech Matusik, Mohamed Elgharib, et~al.
\newblock Lite2relight: 3d-aware single image portrait relighting.
\newblock In \emph{ACM SIGGRAPH 2024 Conference Papers}, pages 1--12, 2024.

\bibitem[Ren et~al.(2024)Ren, Xiong, Yoon, Shu, Zhang, Jung, Gerig, and Zhang]{ren2024relightful}
Mengwei Ren, Wei Xiong, Jae~Shin Yoon, Zhixin Shu, Jianming Zhang, HyunJoon Jung, Guido Gerig, and He Zhang.
\newblock Relightful harmonization: Lighting-aware portrait background replacement.
\newblock In \emph{Proceedings of the IEEE/CVF Conference on Computer Vision and Pattern Recognition}, pages 6452--6462, 2024.

\bibitem[Rombach et~al.(2021)Rombach, Blattmann, Lorenz, Esser, and Ommer]{rombach2021highresolution}
Robin Rombach, Andreas Blattmann, Dominik Lorenz, Patrick Esser, and Björn Ommer.
\newblock High-resolution image synthesis with latent diffusion models, 2021.

\bibitem[Ruiz et~al.(2023)Ruiz, Li, Jampani, Pritch, Rubinstein, and Aberman]{ruiz2023dreambooth}
Nataniel Ruiz, Yuanzhen Li, Varun Jampani, Yael Pritch, Michael Rubinstein, and Kfir Aberman.
\newblock Dreambooth: Fine tuning text-to-image diffusion models for subject-driven generation.
\newblock In \emph{Proceedings of the IEEE/CVF conference on computer vision and pattern recognition}, pages 22500--22510, 2023.

\bibitem[Saharia et~al.(2022{\natexlab{a}})Saharia, Chan, Chang, Lee, Ho, Salimans, Fleet, and Norouzi]{saharia2022palette}
Chitwan Saharia, William Chan, Huiwen Chang, Chris Lee, Jonathan Ho, Tim Salimans, David Fleet, and Mohammad Norouzi.
\newblock Palette: Image-to-image diffusion models.
\newblock In \emph{ACM SIGGRAPH 2022 conference proceedings}, pages 1--10, 2022{\natexlab{a}}.

\bibitem[Saharia et~al.(2022{\natexlab{b}})Saharia, Chan, Saxena, Li, Whang, Denton, Ghasemipour, Gontijo~Lopes, Karagol~Ayan, Salimans, et~al.]{saharia2022photorealistic}
Chitwan Saharia, William Chan, Saurabh Saxena, Lala Li, Jay Whang, Emily~L Denton, Kamyar Ghasemipour, Raphael Gontijo~Lopes, Burcu Karagol~Ayan, Tim Salimans, et~al.
\newblock Photorealistic text-to-image diffusion models with deep language understanding.
\newblock \emph{Advances in neural information processing systems}, 35:\penalty0 36479--36494, 2022{\natexlab{b}}.

\bibitem[Schroff et~al.(2015)Schroff, Kalenichenko, and Philbin]{schroff2015facenet}
Florian Schroff, Dmitry Kalenichenko, and James Philbin.
\newblock Facenet: A unified embedding for face recognition and clustering.
\newblock In \emph{Proceedings of the IEEE conference on computer vision and pattern recognition}, pages 815--823, 2015.

\bibitem[Schuhmann et~al.(2022)Schuhmann, Beaumont, Vencu, Gordon, Wightman, Cherti, Coombes, Katta, Mullis, Wortsman, et~al.]{schuhmann2022laion}
Christoph Schuhmann, Romain Beaumont, Richard Vencu, Cade Gordon, Ross Wightman, Mehdi Cherti, Theo Coombes, Aarush Katta, Clayton Mullis, Mitchell Wortsman, et~al.
\newblock Laion-5b: An open large-scale dataset for training next generation image-text models.
\newblock \emph{Advances in Neural Information Processing Systems}, 35:\penalty0 25278--25294, 2022.

\bibitem[Sengupta et~al.(2018)Sengupta, Kanazawa, Castillo, and Jacobs]{sengupta2018sfsnet}
Soumyadip Sengupta, Angjoo Kanazawa, Carlos~D Castillo, and David~W Jacobs.
\newblock Sfsnet: Learning shape, reflectance and illuminance of facesin the wild'.
\newblock In \emph{Proceedings of the IEEE conference on computer vision and pattern recognition}, pages 6296--6305, 2018.

\bibitem[Shih et~al.(2014)Shih, Paris, Barnes, Freeman, and Durand]{shih2014style}
YiChang Shih, Sylvain Paris, Connelly Barnes, William~T Freeman, and Fr{\'e}do Durand.
\newblock Style transfer for headshot portraits.
\newblock 2014.

\bibitem[Shu et~al.(2017{\natexlab{a}})Shu, Hadap, Shechtman, Sunkavalli, Paris, and Samaras]{shu2017portrait}
Zhixin Shu, Sunil Hadap, Eli Shechtman, Kalyan Sunkavalli, Sylvain Paris, and Dimitris Samaras.
\newblock Portrait lighting transfer using a mass transport approach.
\newblock \emph{ACM Transactions on Graphics (TOG)}, 36\penalty0 (4):\penalty0 1, 2017{\natexlab{a}}.

\bibitem[Shu et~al.(2017{\natexlab{b}})Shu, Yumer, Hadap, Sunkavalli, Shechtman, and Samaras]{shu2017neural}
Zhixin Shu, Ersin Yumer, Sunil Hadap, Kalyan Sunkavalli, Eli Shechtman, and Dimitris Samaras.
\newblock Neural face editing with intrinsic image disentangling.
\newblock In \emph{Proceedings of the IEEE conference on computer vision and pattern recognition}, pages 5541--5550, 2017{\natexlab{b}}.

\bibitem[Sohl-Dickstein et~al.(2015)Sohl-Dickstein, Weiss, Maheswaranathan, and Ganguli]{sohl2015deep}
Jascha Sohl-Dickstein, Eric Weiss, Niru Maheswaranathan, and Surya Ganguli.
\newblock Deep unsupervised learning using nonequilibrium thermodynamics.
\newblock In \emph{International conference on machine learning}, pages 2256--2265. PMLR, 2015.

\bibitem[Song et~al.(2020)Song, Meng, and Ermon]{song2020denoising}
Jiaming Song, Chenlin Meng, and Stefano Ermon.
\newblock Denoising diffusion implicit models.
\newblock \emph{arXiv preprint arXiv:2010.02502}, 2020.

\bibitem[Sun et~al.(2019)Sun, Barron, Tsai, Xu, Yu, Fyffe, Rhemann, Busch, Debevec, and Ramamoorthi]{sun2019single}
Tiancheng Sun, Jonathan~T Barron, Yun-Ta Tsai, Zexiang Xu, Xueming Yu, Graham Fyffe, Christoph Rhemann, Jay Busch, Paul Debevec, and Ravi Ramamoorthi.
\newblock Single image portrait relighting.
\newblock \emph{ACM Transactions on Graphics (TOG)}, 38\penalty0 (4):\penalty0 1--12, 2019.

\bibitem[Sun et~al.(2020)Sun, Xu, Zhang, Fanello, Rhemann, Debevec, Tsai, Barron, and Ramamoorthi]{sun2020light}
Tiancheng Sun, Zexiang Xu, Xiuming Zhang, Sean Fanello, Christoph Rhemann, Paul Debevec, Yun-Ta Tsai, Jonathan~T Barron, and Ravi Ramamoorthi.
\newblock Light stage super-resolution: continuous high-frequency relighting.
\newblock \emph{ACM Transactions on Graphics (TOG)}, 39\penalty0 (6):\penalty0 1--12, 2020.

\bibitem[Sun et~al.(2021)Sun, Lin, Bi, Xu, and Ramamoorthi]{sun2021nelf}
Tiancheng Sun, Kai-En Lin, Sai Bi, Zexiang Xu, and Ravi Ramamoorthi.
\newblock Nelf: Neural light-transport field for portrait view synthesis and relighting.
\newblock \emph{arXiv preprint arXiv:2107.12351}, 2021.

\bibitem[Tan et~al.(2022)Tan, Fanello, Meka, Orts-Escolano, Tang, Pandey, Taylor, Tan, and Zhang]{tan2022volux}
Feitong Tan, Sean Fanello, Abhimitra Meka, Sergio Orts-Escolano, Danhang Tang, Rohit Pandey, Jonathan Taylor, Ping Tan, and Yinda Zhang.
\newblock Volux-gan: A generative model for 3d face synthesis with hdri relighting.
\newblock In \emph{ACM SIGGRAPH 2022 Conference Proceedings}, pages 1--9, 2022.

\bibitem[Wang et~al.(2024)Wang, Leroy, Cabon, Chidlovskii, and Revaud]{wang2024dust3r}
Shuzhe Wang, Vincent Leroy, Yohann Cabon, Boris Chidlovskii, and Jerome Revaud.
\newblock Dust3r: Geometric 3d vision made easy.
\newblock In \emph{Proceedings of the IEEE/CVF Conference on Computer Vision and Pattern Recognition}, pages 20697--20709, 2024.

\bibitem[Wang et~al.(2023)Wang, Holynski, Zhang, and Zhang]{wang2023sunstage}
Yifan Wang, Aleksander Holynski, Xiuming Zhang, and Xuaner Zhang.
\newblock Sunstage: Portrait reconstruction and relighting using the sun as a light stage.
\newblock In \emph{Proceedings of the IEEE/CVF Conference on Computer Vision and Pattern Recognition}, pages 20792--20802, 2023.

\bibitem[Wang et~al.(2020)Wang, Yu, Lu, Wang, Qian, and Xu]{wang2020single}
Zhibo Wang, Xin Yu, Ming Lu, Quan Wang, Chen Qian, and Feng Xu.
\newblock Single image portrait relighting via explicit multiple reflectance channel modeling.
\newblock \emph{ACM Transactions on Graphics (ToG)}, 39\penalty0 (6):\penalty0 1--13, 2020.

\bibitem[Wood et~al.(2021)Wood, Baltru{\v{s}}aitis, Hewitt, Dziadzio, Cashman, and Shotton]{wood2021fake}
Erroll Wood, Tadas Baltru{\v{s}}aitis, Charlie Hewitt, Sebastian Dziadzio, Thomas~J Cashman, and Jamie Shotton.
\newblock Fake it till you make it: face analysis in the wild using synthetic data alone.
\newblock In \emph{Proceedings of the IEEE/CVF international conference on computer vision}, pages 3681--3691, 2021.

\bibitem[Xie et~al.(2023)Xie, Zhang, Lin, Hinz, and Zhang]{xie2023smartbrush}
Shaoan Xie, Zhifei Zhang, Zhe Lin, Tobias Hinz, and Kun Zhang.
\newblock Smartbrush: Text and shape guided object inpainting with diffusion model.
\newblock In \emph{Proceedings of the IEEE/CVF Conference on Computer Vision and Pattern Recognition}, pages 22428--22437, 2023.

\bibitem[Ye et~al.(2023)Ye, Zhang, Liu, Han, and Yang]{ye2023ip}
Hu Ye, Jun Zhang, Sibo Liu, Xiao Han, and Wei Yang.
\newblock Ip-adapter: Text compatible image prompt adapter for text-to-image diffusion models.
\newblock \emph{arXiv preprint arXiv:2308.06721}, 2023.

\bibitem[Yeh et~al.(2022)Yeh, Nagano, Khamis, Kautz, Liu, and Wang]{yeh2022learning}
Yu-Ying Yeh, Koki Nagano, Sameh Khamis, Jan Kautz, Ming-Yu Liu, and Ting-Chun Wang.
\newblock Learning to relight portrait images via a virtual light stage and synthetic-to-real adaptation.
\newblock \emph{ACM Transactions on Graphics (TOG)}, 41\penalty0 (6):\penalty0 1--21, 2022.

\bibitem[Zeng et~al.(2024)Zeng, Dong, Peers, Kong, Wu, and Tong]{zeng2024dilightnet}
Chong Zeng, Yue Dong, Pieter Peers, Youkang Kong, Hongzhi Wu, and Xin Tong.
\newblock Dilightnet: Fine-grained lighting control for diffusion-based image generation.
\newblock In \emph{ACM SIGGRAPH 2024 Conference Papers}, pages 1--12, 2024.

\bibitem[Zhang et~al.(2023)Zhang, Rao, and Agrawala]{zhang2023adding}
Lvmin Zhang, Anyi Rao, and Maneesh Agrawala.
\newblock Adding conditional control to text-to-image diffusion models.
\newblock In \emph{Proceedings of the IEEE/CVF International Conference on Computer Vision}, pages 3836--3847, 2023.

\bibitem[Zhang et~al.(2024)Zhang, Rao, and Agrawala]{iclight}
Lvmin Zhang, Anyi Rao, and Maneesh Agrawala.
\newblock Ic-light github page, 2024.

\bibitem[Zhang et~al.(2018)Zhang, Isola, Efros, Shechtman, and Wang]{zhang2018unreasonable}
Richard Zhang, Phillip Isola, Alexei~A Efros, Eli Shechtman, and Oliver Wang.
\newblock The unreasonable effectiveness of deep features as a perceptual metric.
\newblock In \emph{Proceedings of the IEEE conference on computer vision and pattern recognition}, pages 586--595, 2018.

\bibitem[Zhou et~al.(2019)Zhou, Hadap, Sunkavalli, and Jacobs]{zhou2019deep}
Hao Zhou, Sunil Hadap, Kalyan Sunkavalli, and David~W Jacobs.
\newblock Deep single-image portrait relighting.
\newblock In \emph{Proceedings of the IEEE/CVF international conference on computer vision}, pages 7194--7202, 2019.

\end{thebibliography}
}

\appendix
\clearpage

\maketitlesupplementary

\section{Additional Results}
\label{sec:additional_results}

We present additional results on input portraits from various stock websites such as Adobe Stock, Unsplash and Pexels as well as from our internal Light-Stage captures.

\begin{figure*}[htbp]
    \centering

    \begin{tabular}{c@{\hskip 0.2mm}c@{\hskip 0.1mm}c@{\hskip 0.1mm}c@{\hskip 0.1mm}c@{\hskip 0.1mm}c@{\hskip 0.1mm}c}

    \includegraphics[height=\extrafigureheight]{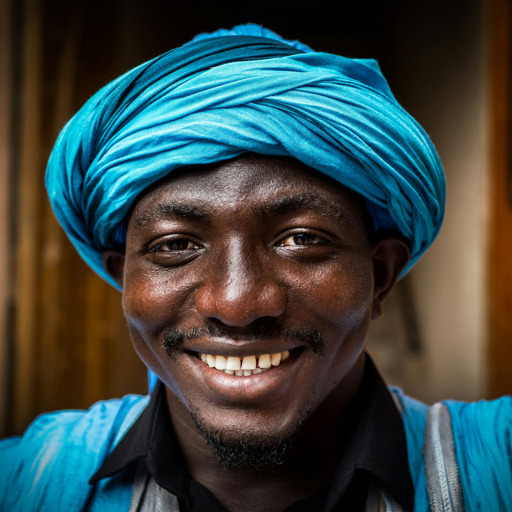} &
    \includegraphics[height=\extrafigureheight]{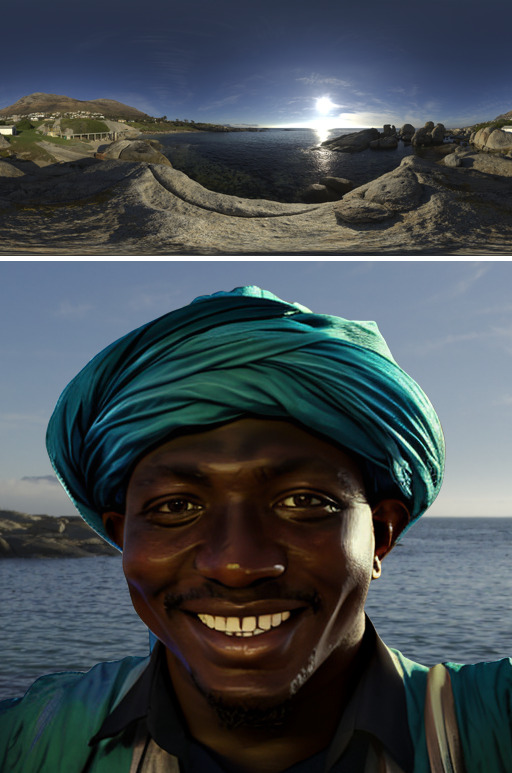} &
    \includegraphics[height=\extrafigureheight]{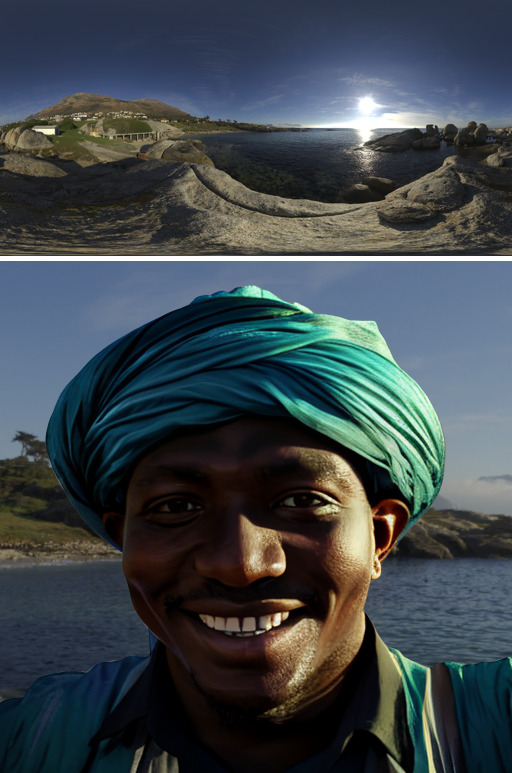} &
    \includegraphics[height=\extrafigureheight]{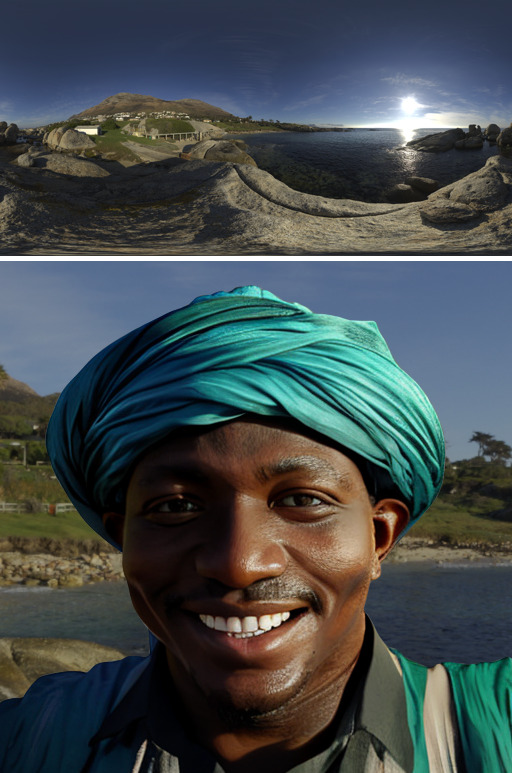} &
    \includegraphics[height=\extrafigureheight]{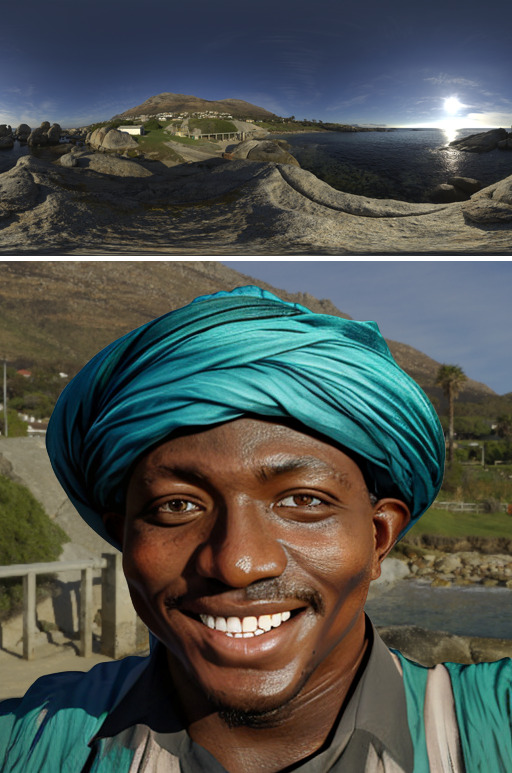} &
    \includegraphics[height=\extrafigureheight]{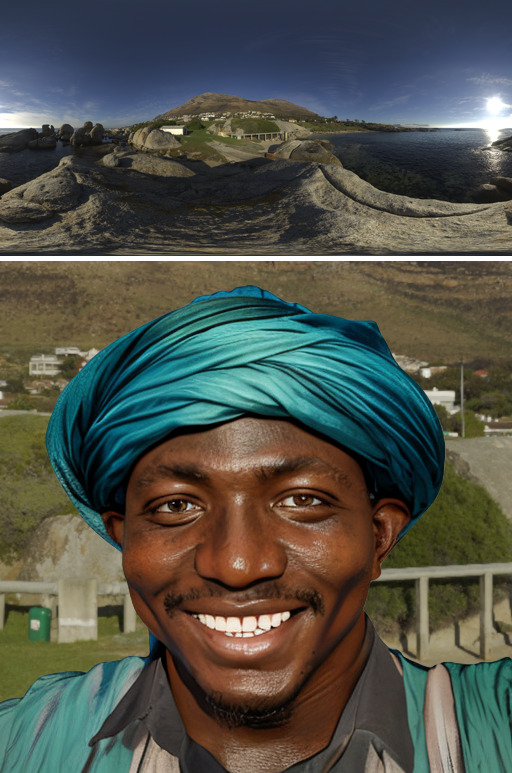} &
    \includegraphics[height=\extrafigureheight]{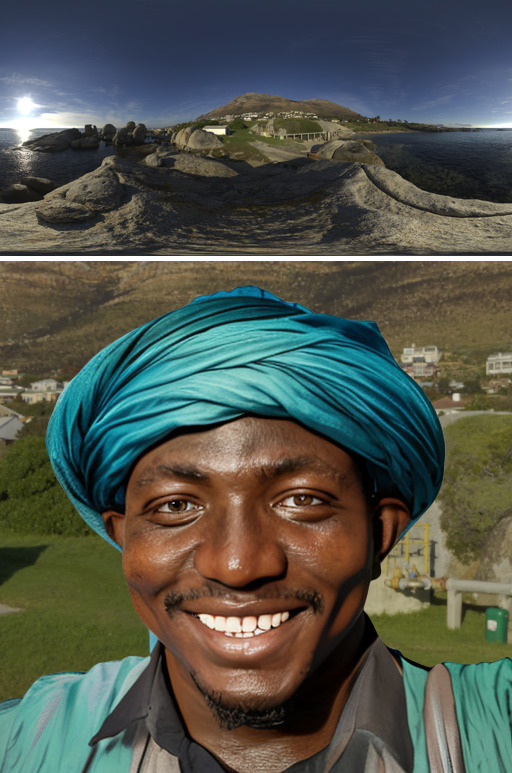} \\

    \multicolumn{1}{c}{} &
    \includegraphics[height=\extrafigureheight]{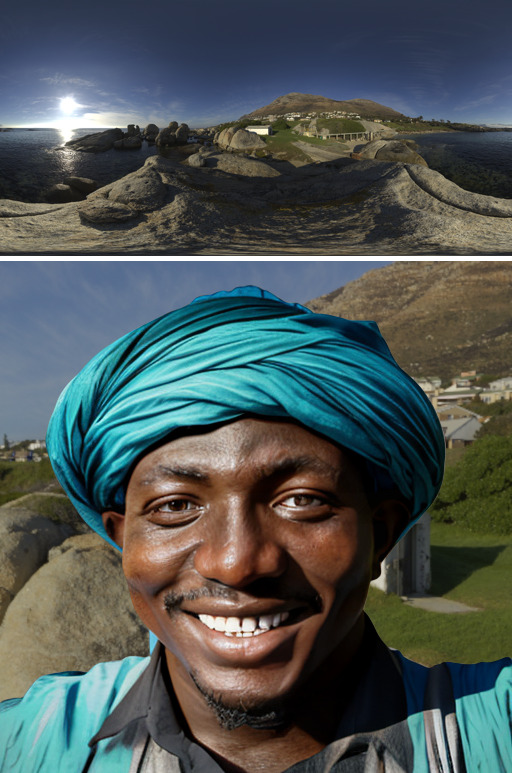} &
    \includegraphics[height=\extrafigureheight]{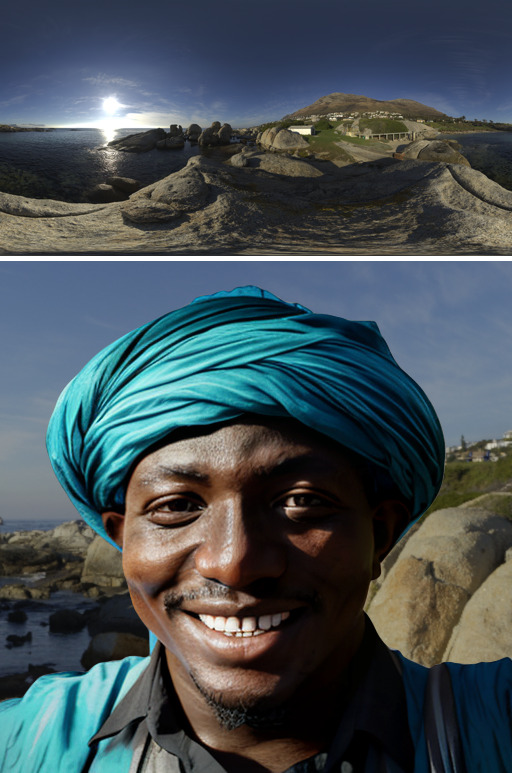} &
    \includegraphics[height=\extrafigureheight]{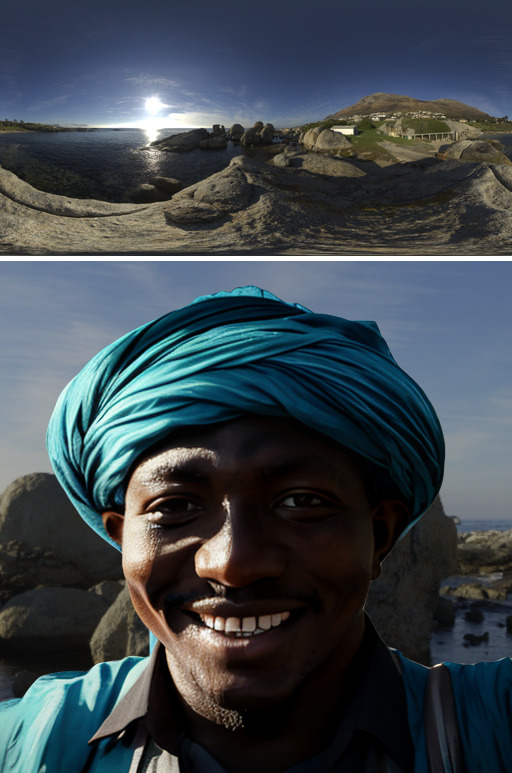} &
    \includegraphics[height=\extrafigureheight]{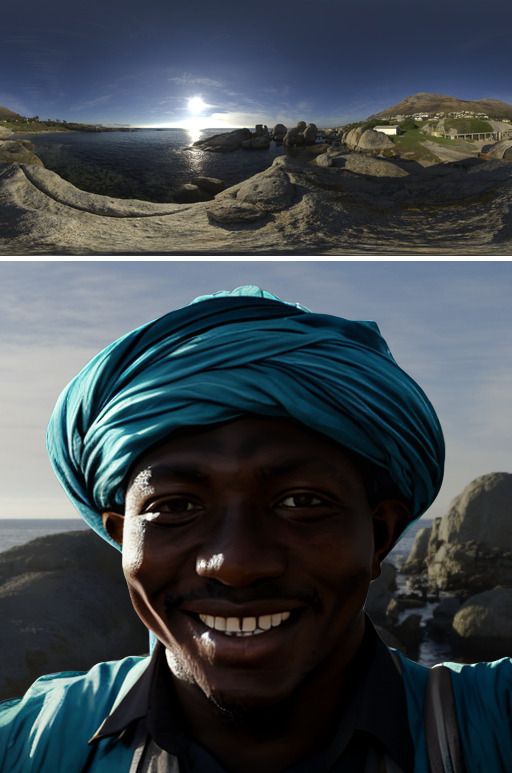} &
    \includegraphics[height=\extrafigureheight]{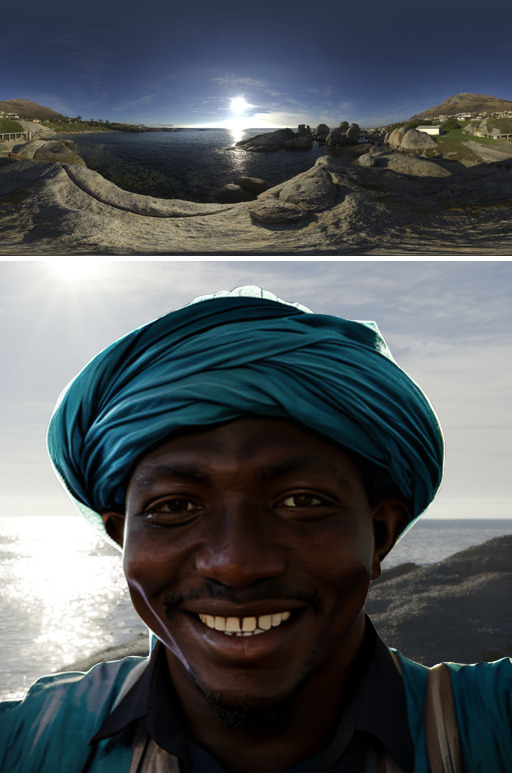} &
    \includegraphics[height=\extrafigureheight]{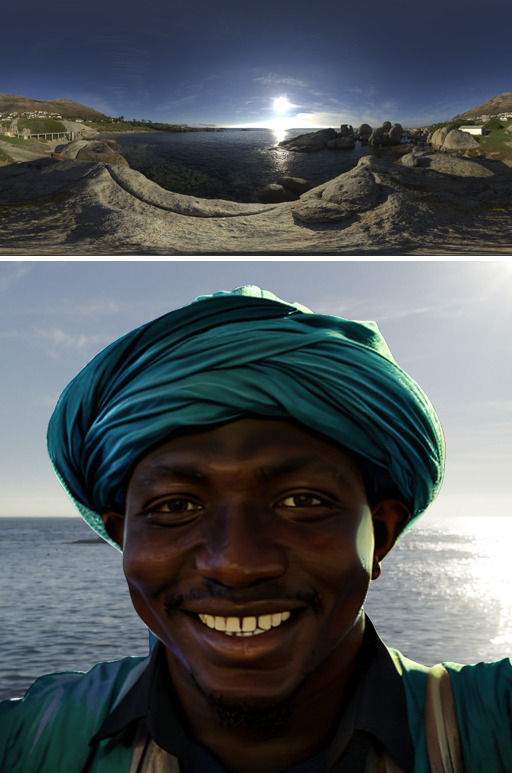} \\

    \includegraphics[height=\extrafigureheight]{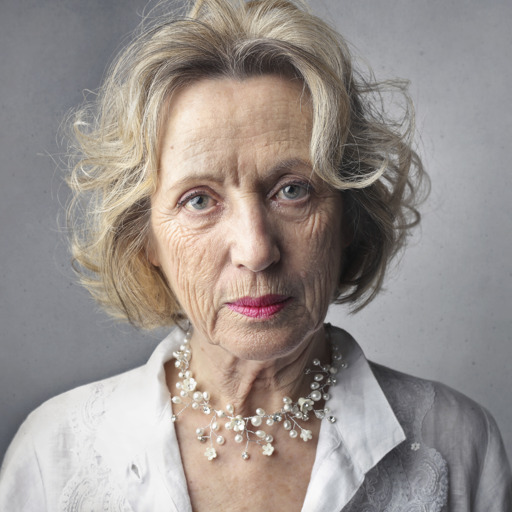} &
    \includegraphics[height=\extrafigureheight]{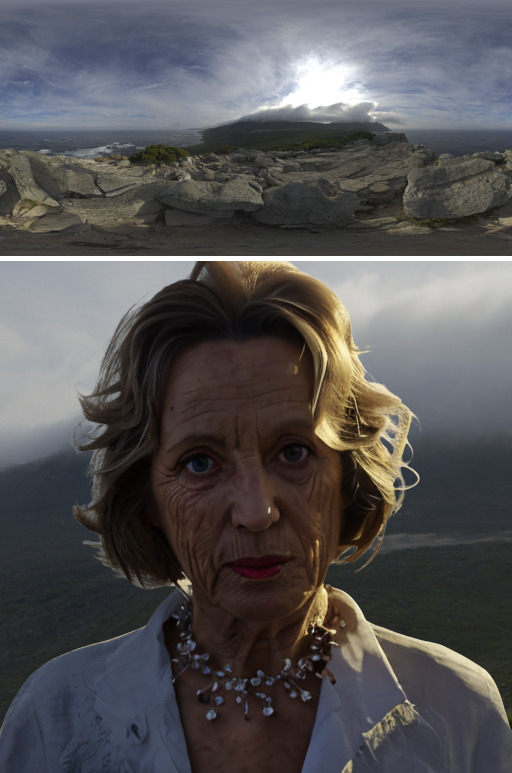} &
    \includegraphics[height=\extrafigureheight]{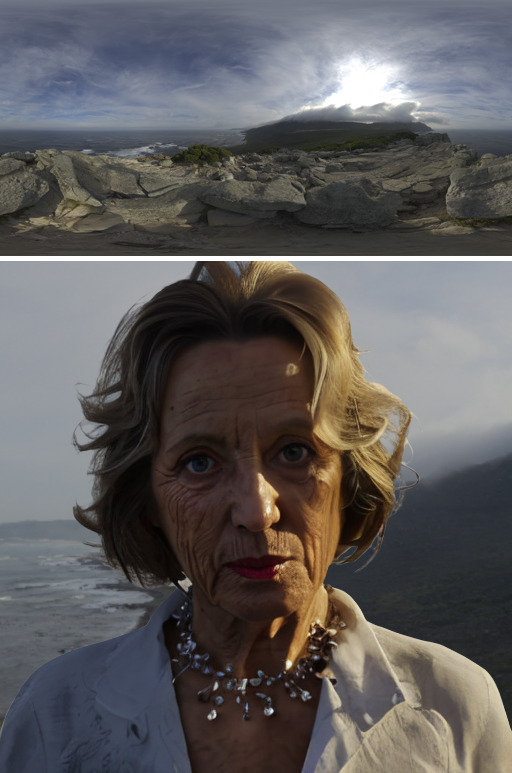} &
    \includegraphics[height=\extrafigureheight]{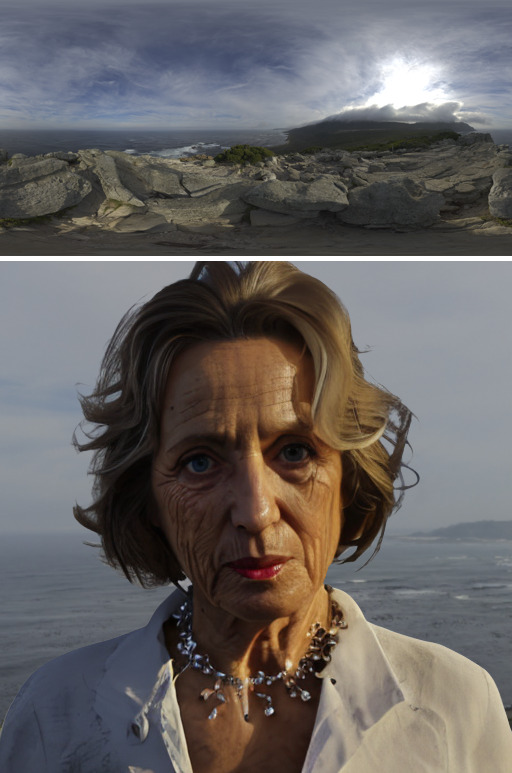} &
    \includegraphics[height=\extrafigureheight]{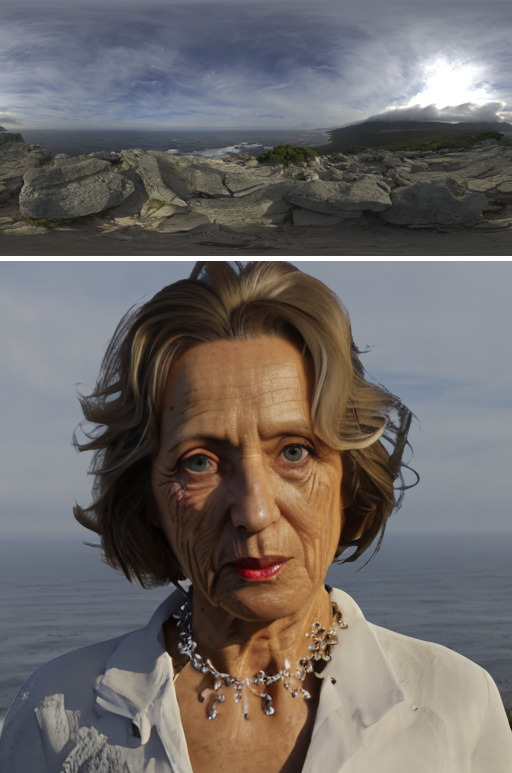} &
    \includegraphics[height=\extrafigureheight]{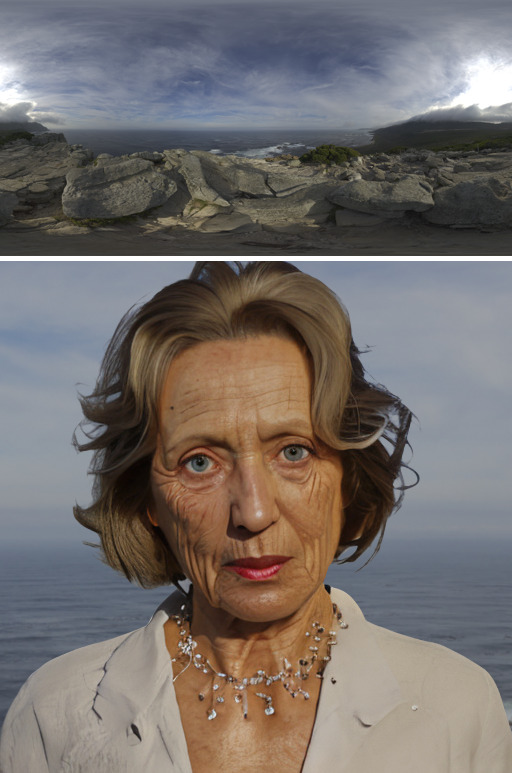} &
    \includegraphics[height=\extrafigureheight]{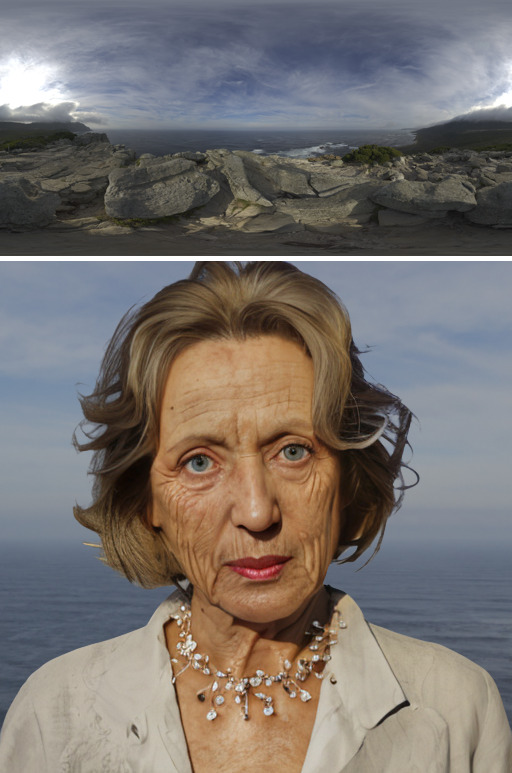} \\

    \multicolumn{1}{c}{} &
    \includegraphics[height=\extrafigureheight]{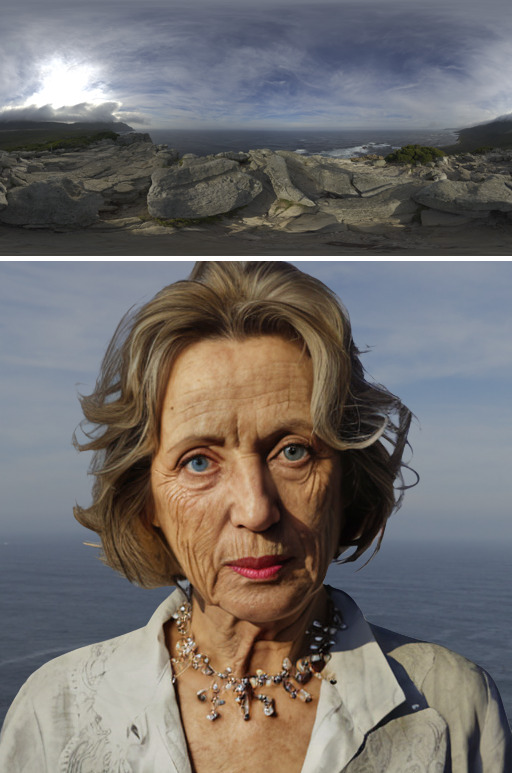} &
    \includegraphics[height=\extrafigureheight]{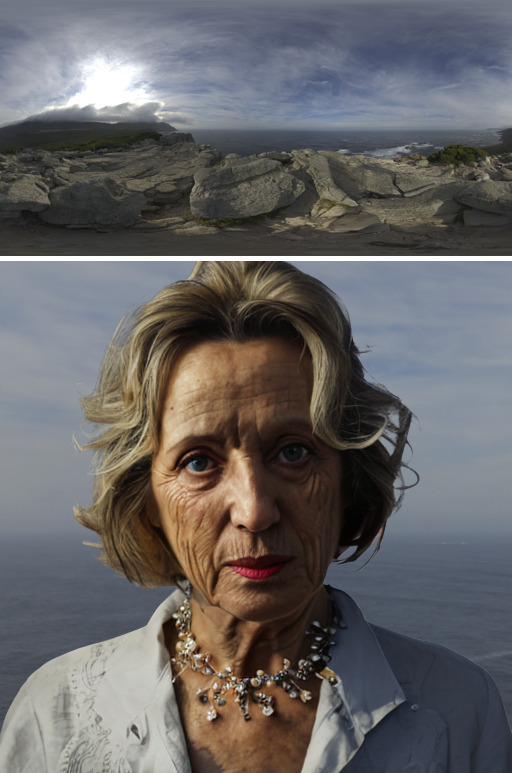} &
    \includegraphics[height=\extrafigureheight]{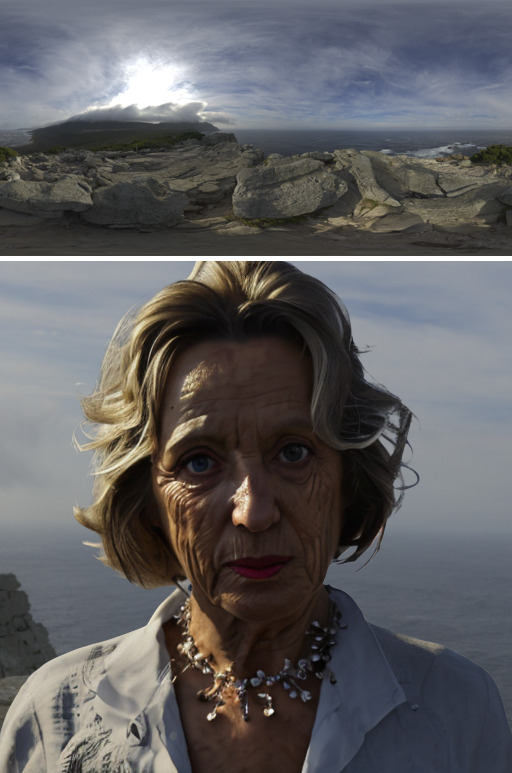} &
    \includegraphics[height=\extrafigureheight]{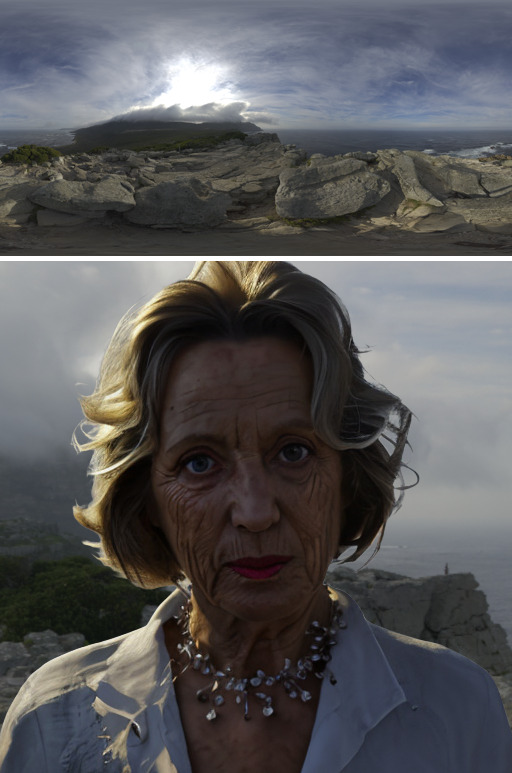} &
    \includegraphics[height=\extrafigureheight]{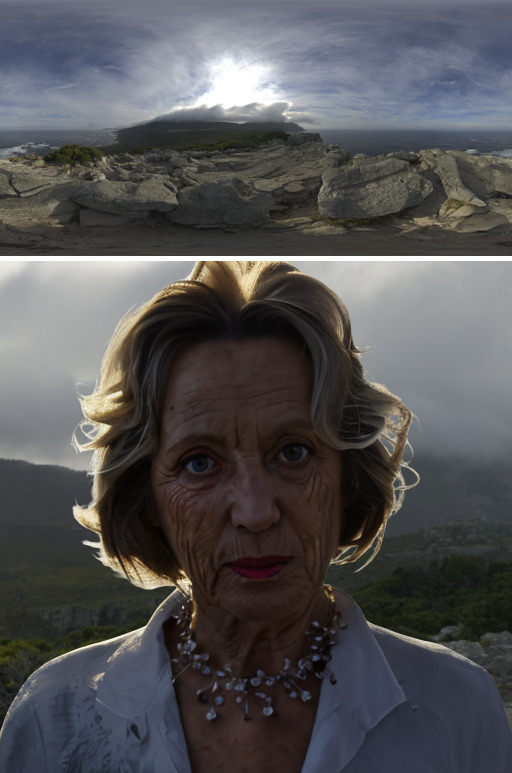} &
    \includegraphics[height=\extrafigureheight]{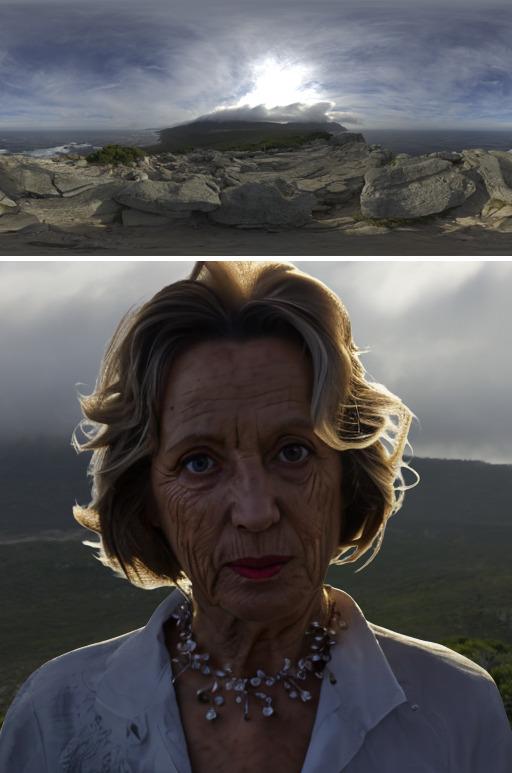} \\

    \includegraphics[height=\extrafigureheight]{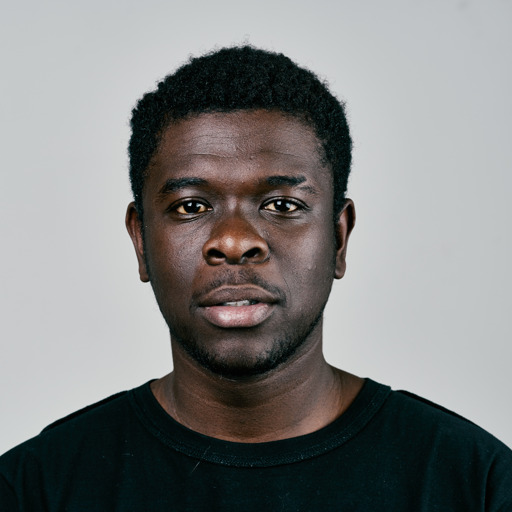} &
    \includegraphics[height=\extrafigureheight]{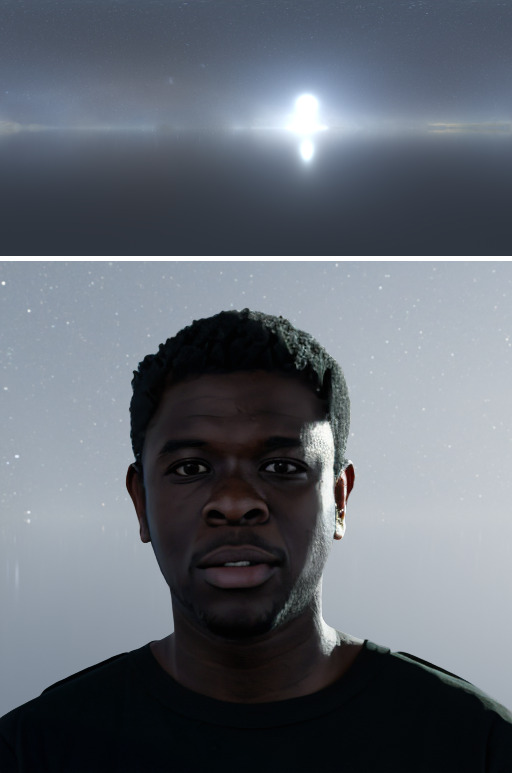} &
    \includegraphics[height=\extrafigureheight]{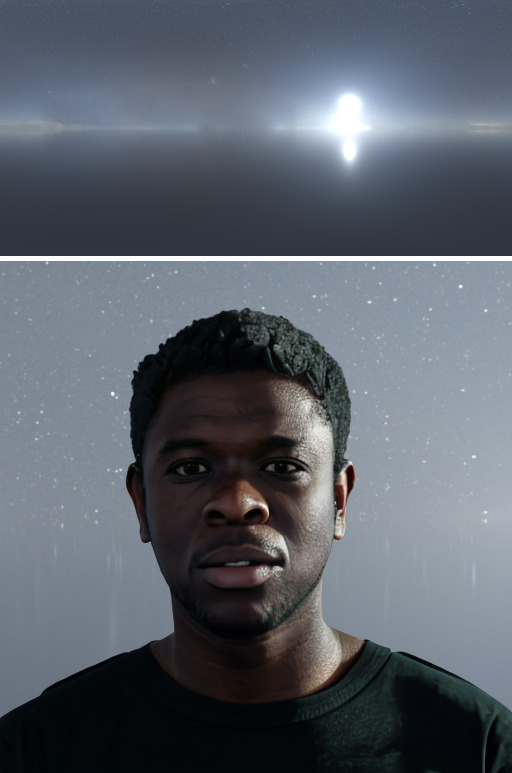} &
    \includegraphics[height=\extrafigureheight]{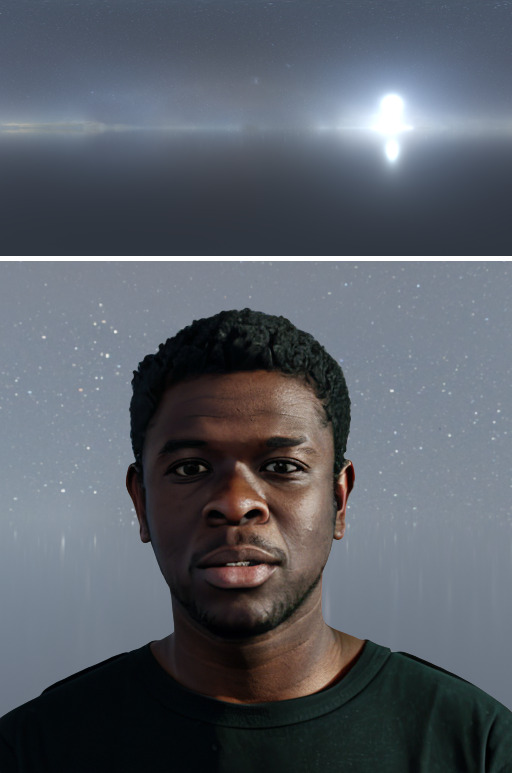} &
    \includegraphics[height=\extrafigureheight]{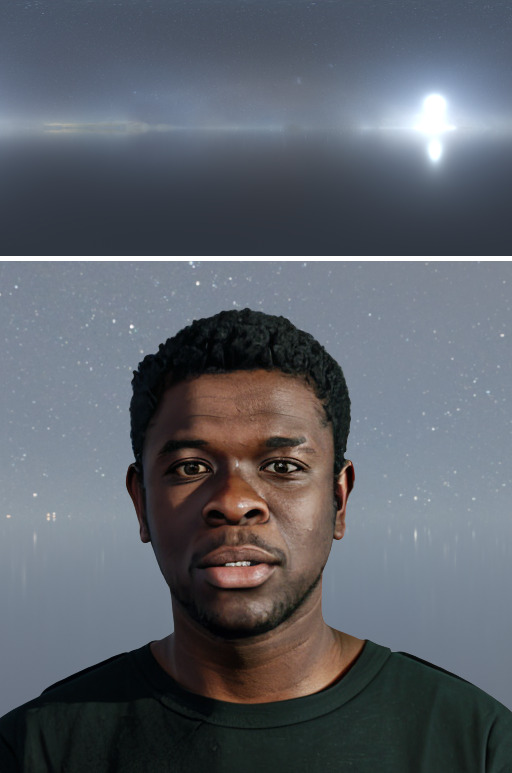} &
    \includegraphics[height=\extrafigureheight]{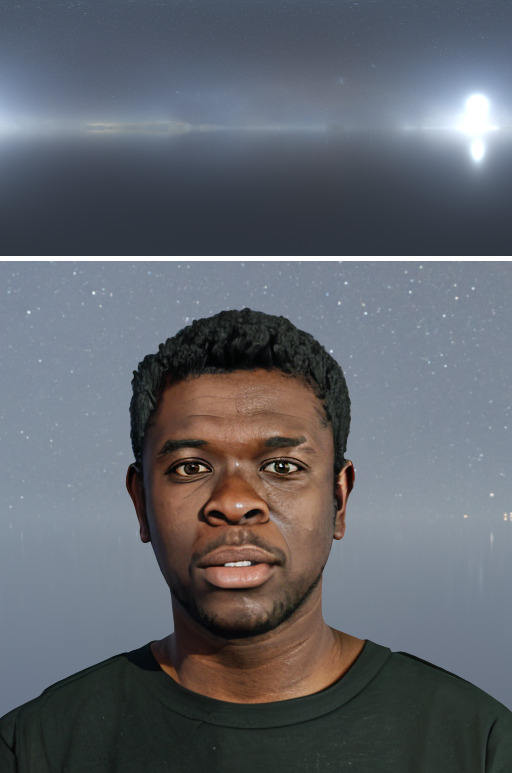} &
    \includegraphics[height=\extrafigureheight]{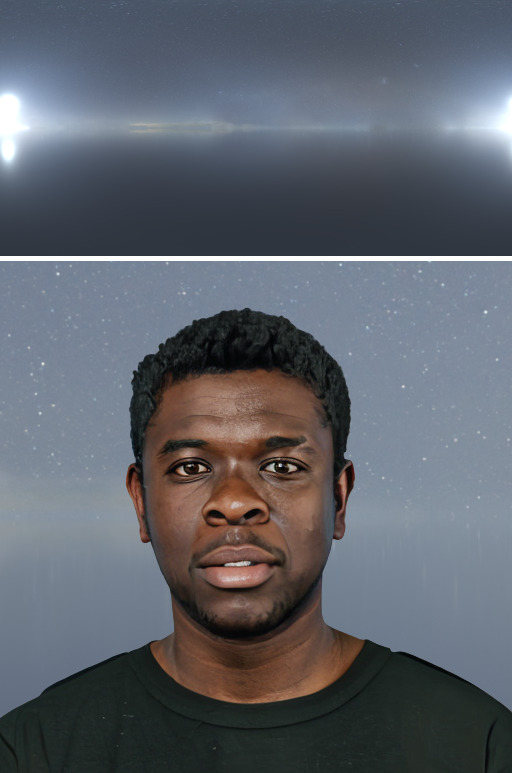} \\

    \multicolumn{1}{c}{} &
    \includegraphics[height=\extrafigureheight]{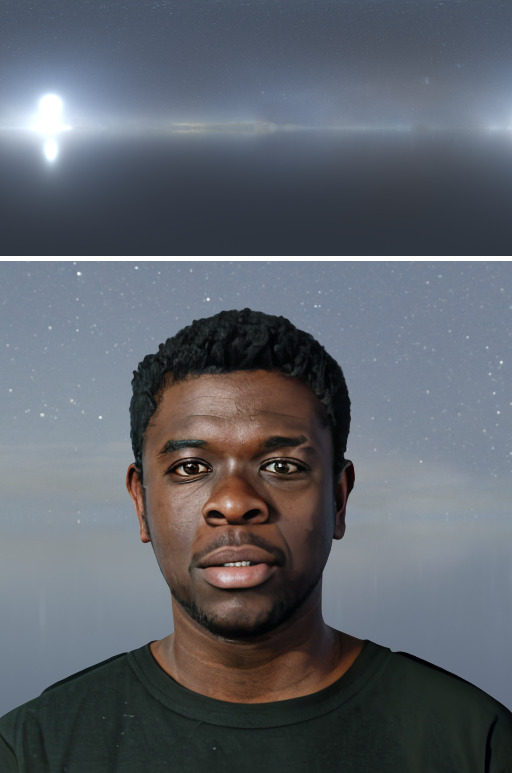} &
    \includegraphics[height=\extrafigureheight]{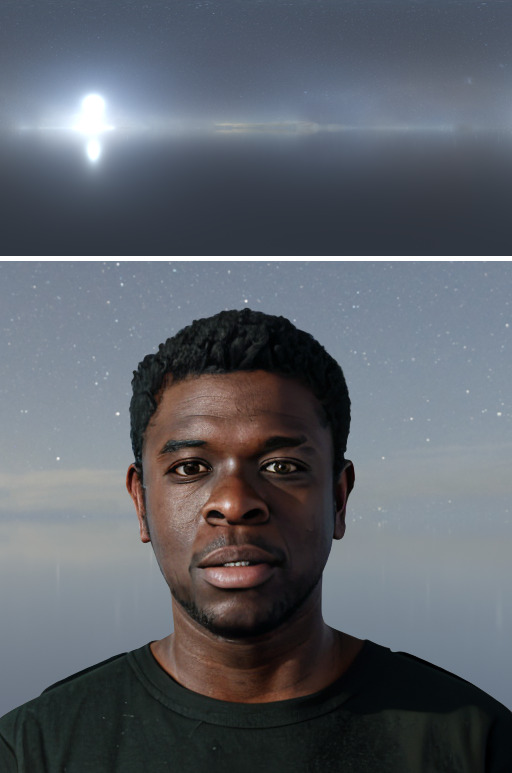} &
    \includegraphics[height=\extrafigureheight]{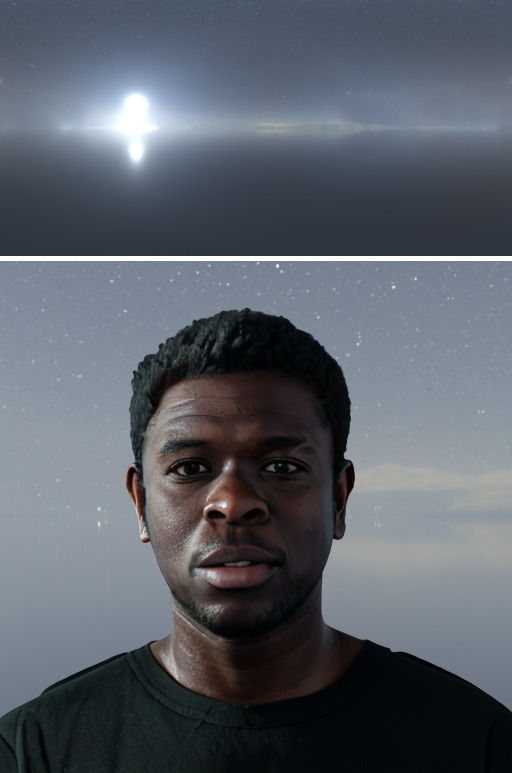} &
    \includegraphics[height=\extrafigureheight]{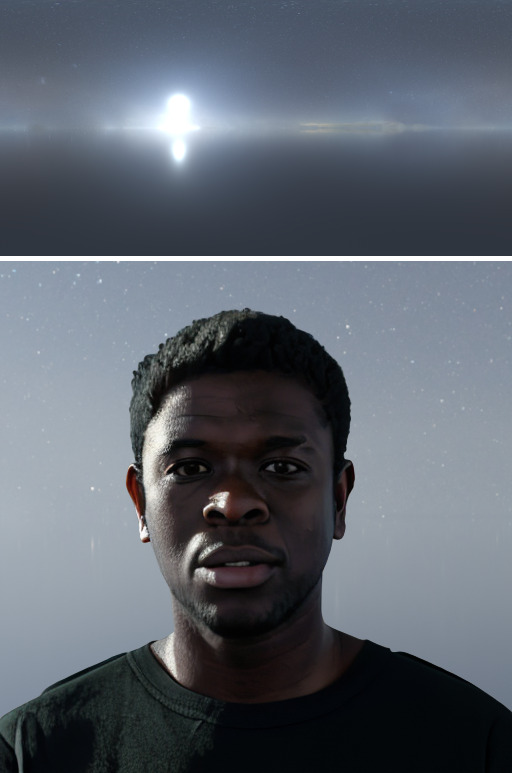} &
    \includegraphics[height=\extrafigureheight]{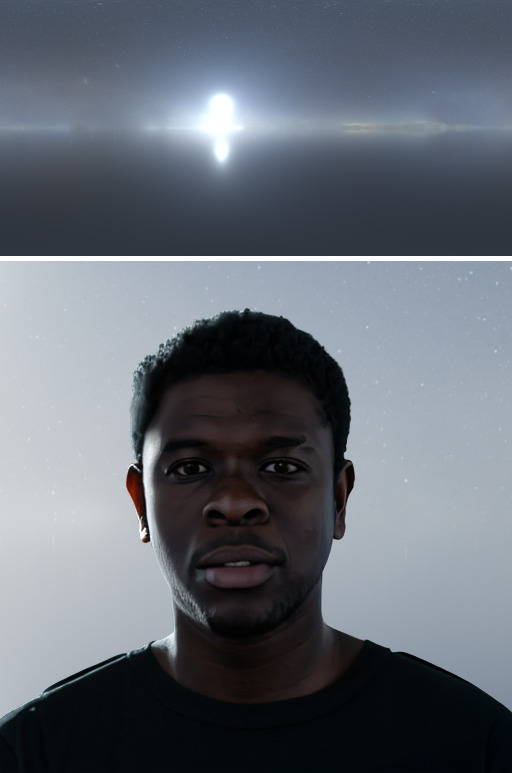} &
    \includegraphics[height=\extrafigureheight]{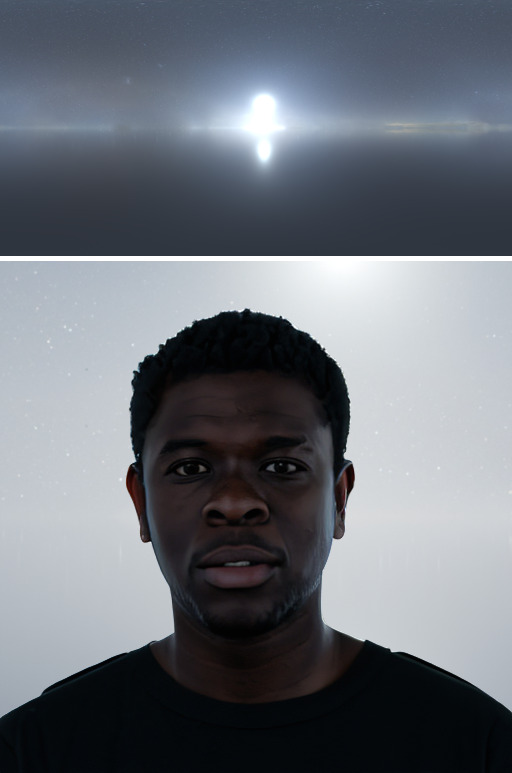} \\

    \end{tabular}
    \vspace{-3mm}
    \caption{In order to demonstrate portrait lighting effects in the presence of strong sunlight such as strong cast shadows by facial features, rim-effects in hair and specular highlights, we show in-the-wild portraits relit using outdoor environment maps.}
    \label{fig:outdoor}
\end{figure*}

\begin{figure*}[htbp]
    \centering

    \begin{tabular}{c@{\hskip 0.2mm}c@{\hskip 0.1mm}c@{\hskip 0.1mm}c@{\hskip 0.1mm}c@{\hskip 0.1mm}c@{\hskip 0.1mm}c}

    \includegraphics[height=\extrafigureheight]{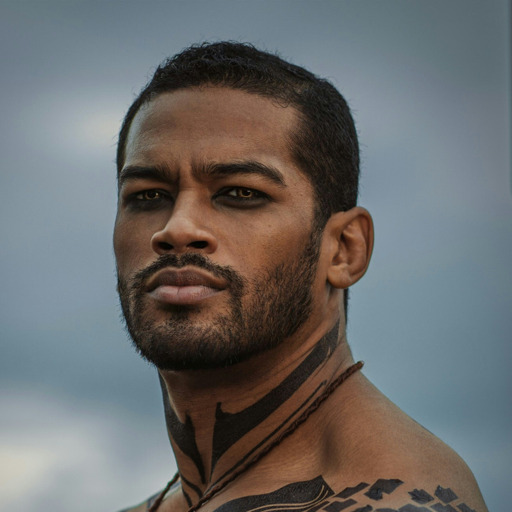} & 
    \includegraphics[height=\extrafigureheight]{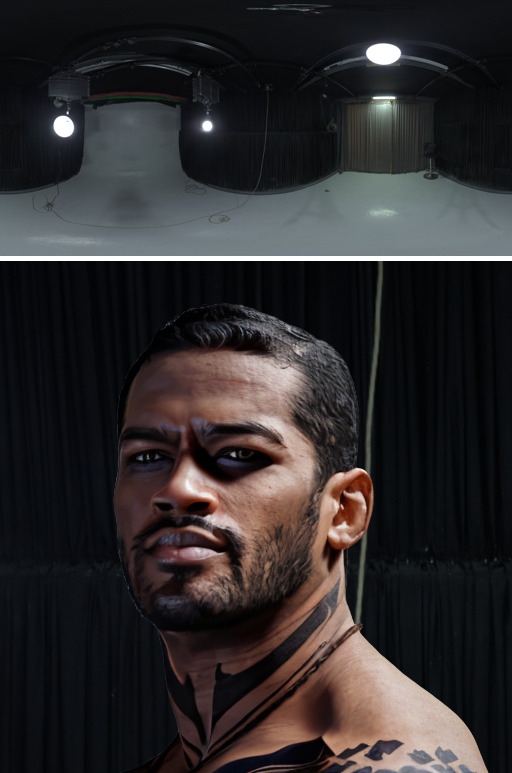} &
    \includegraphics[height=\extrafigureheight]{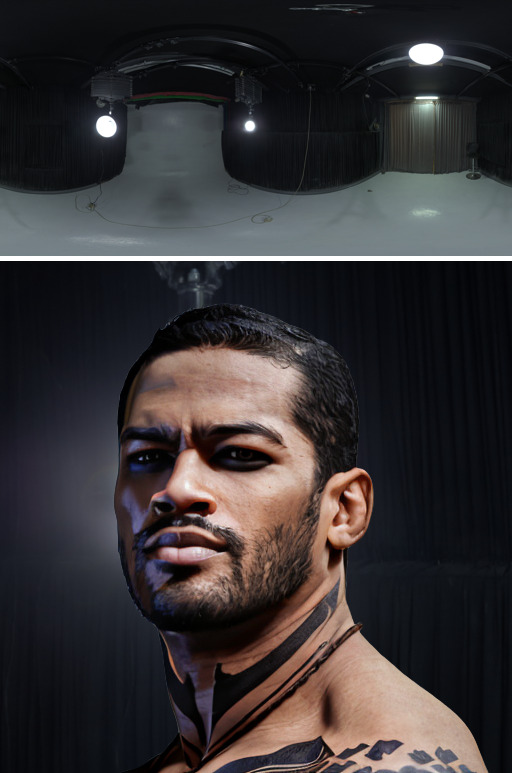} &
    \includegraphics[height=\extrafigureheight]{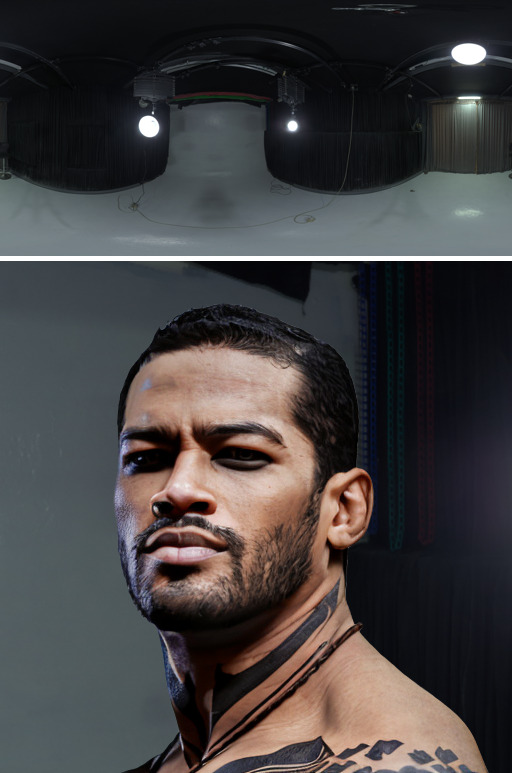} &
    \includegraphics[height=\extrafigureheight]{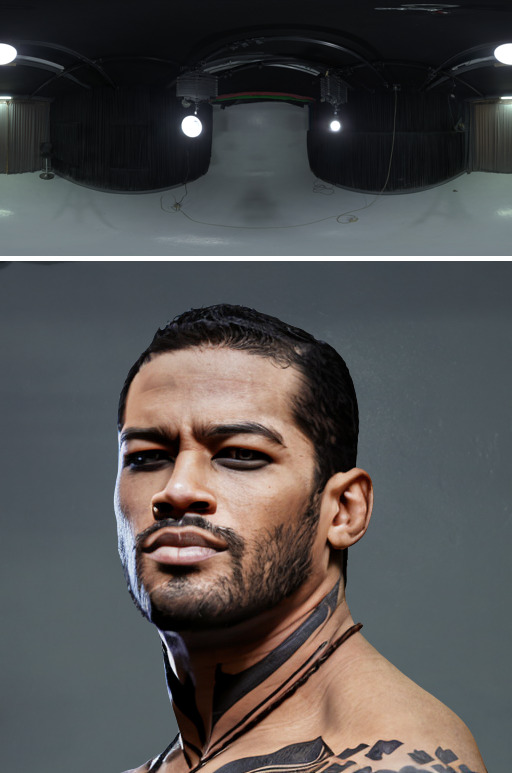} &
    \includegraphics[height=\extrafigureheight]{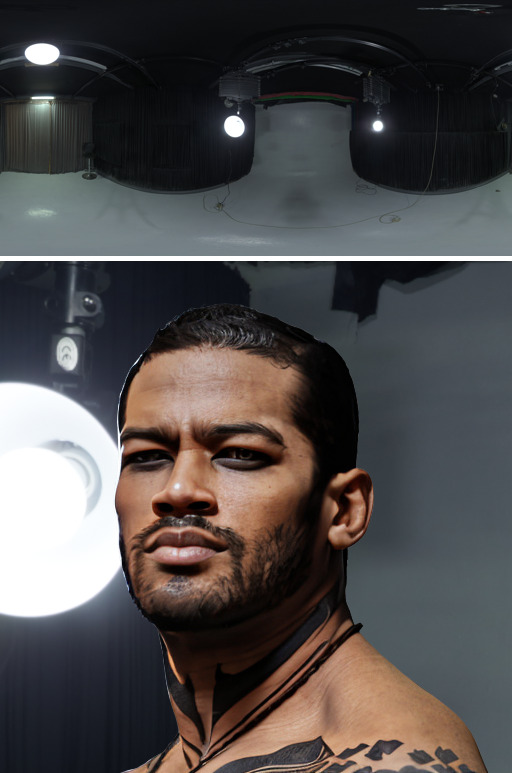} &
    \includegraphics[height=\extrafigureheight]{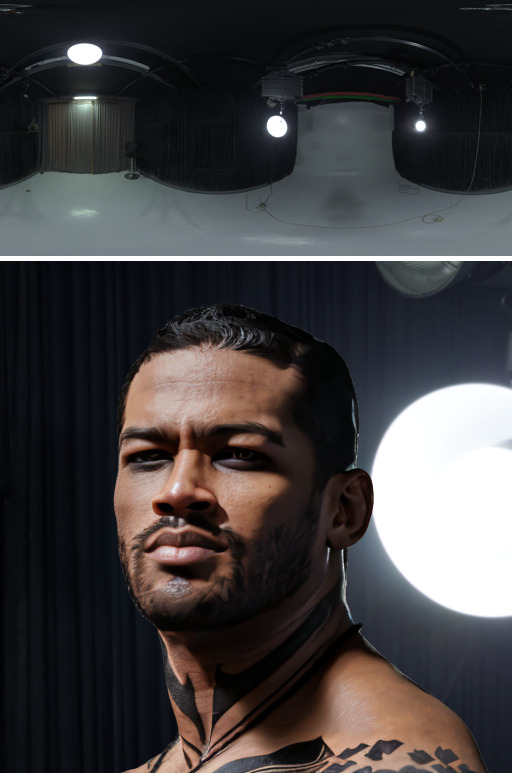} \\

    \multicolumn{1}{c}{} &
    \includegraphics[height=\extrafigureheight]{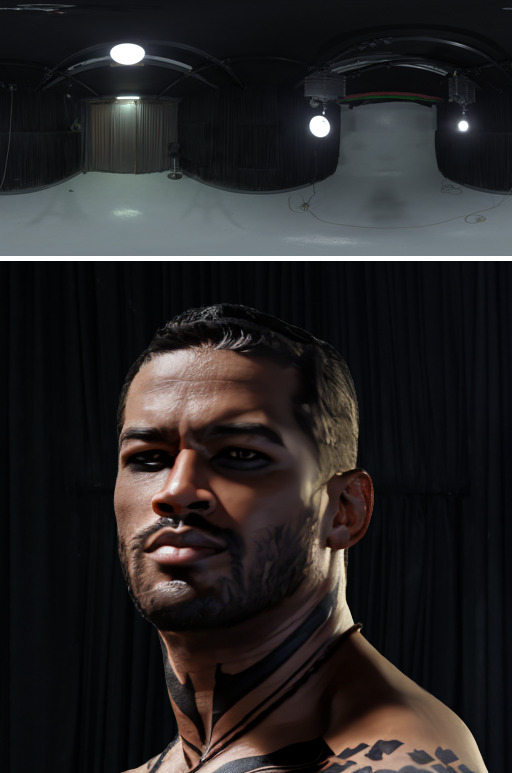} &
    \includegraphics[height=\extrafigureheight]{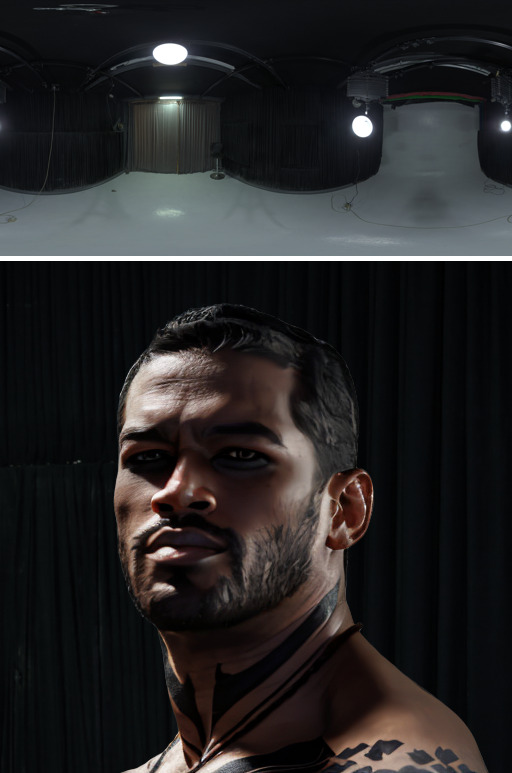} &
    \includegraphics[height=\extrafigureheight]{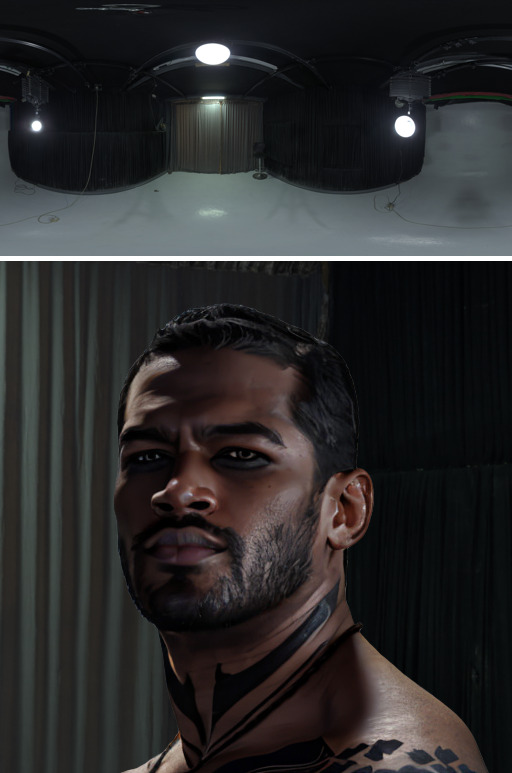} &
    \includegraphics[height=\extrafigureheight]{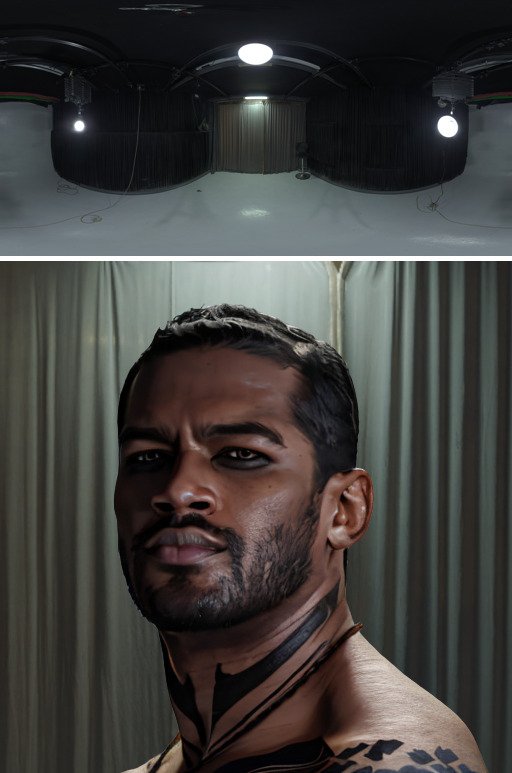} & 
    \includegraphics[height=\extrafigureheight]{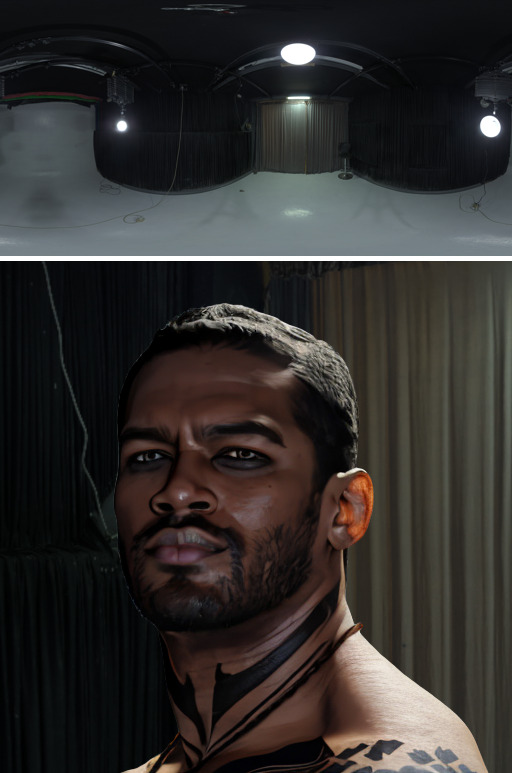} & 
    \includegraphics[height=\extrafigureheight]{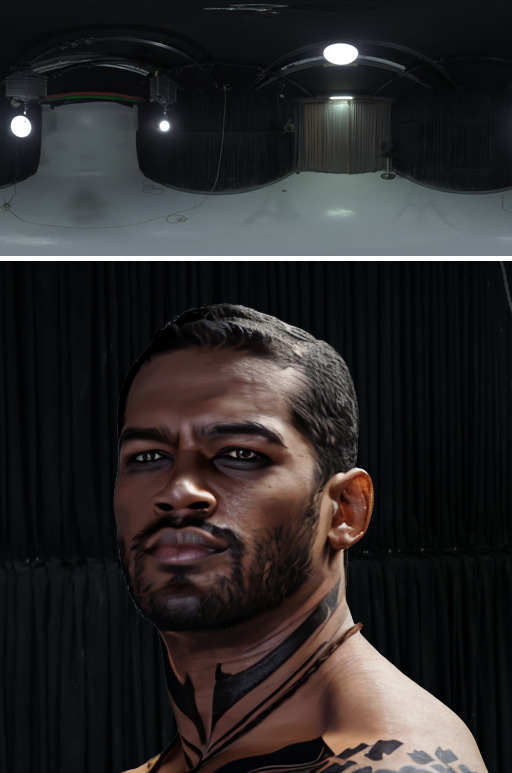} \\

    \includegraphics[height=\extrafigureheight]{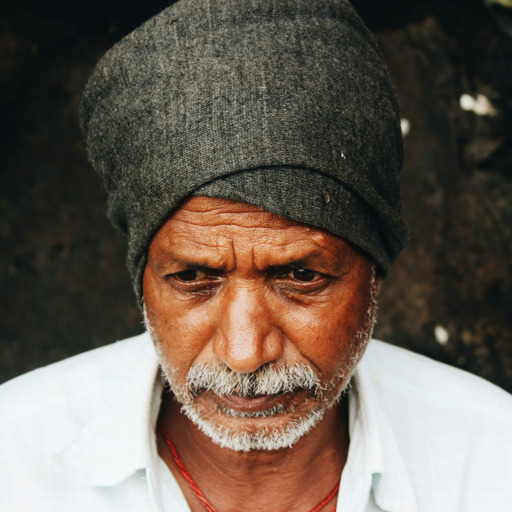} & 
    \includegraphics[height=\extrafigureheight]{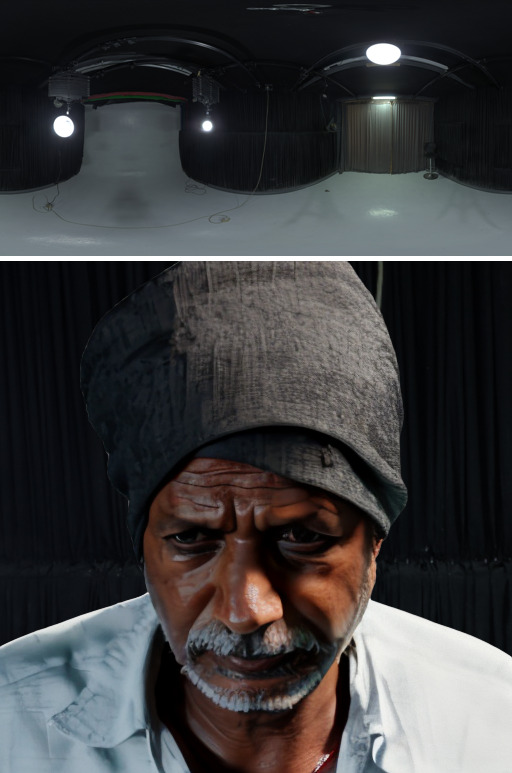} &
    \includegraphics[height=\extrafigureheight]{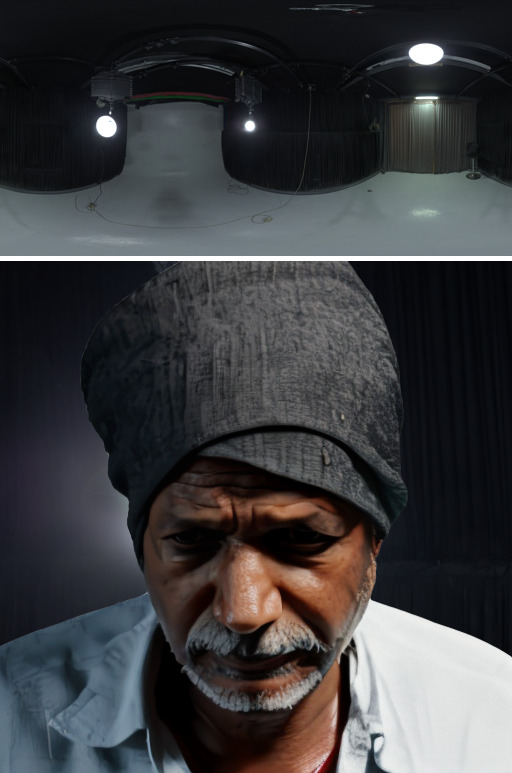} &
    \includegraphics[height=\extrafigureheight]{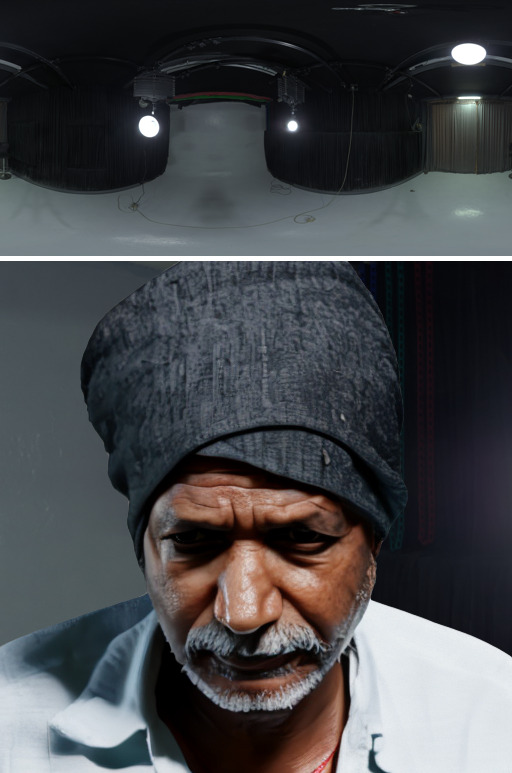} &
    \includegraphics[height=\extrafigureheight]{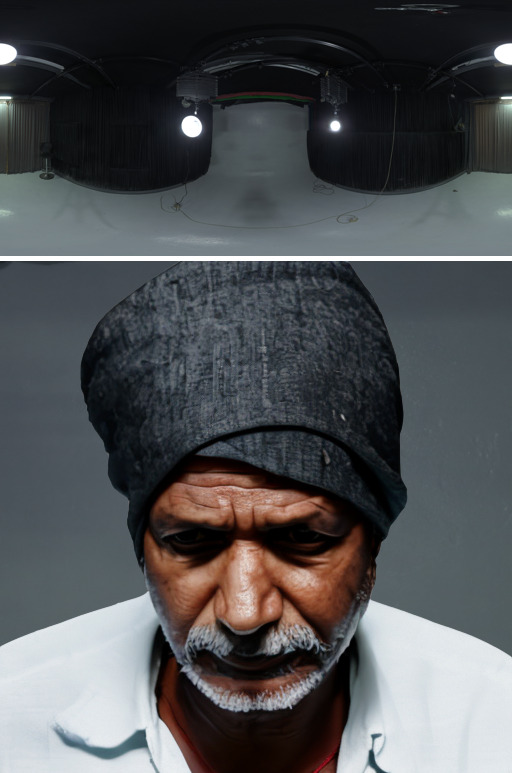} &
    \includegraphics[height=\extrafigureheight]{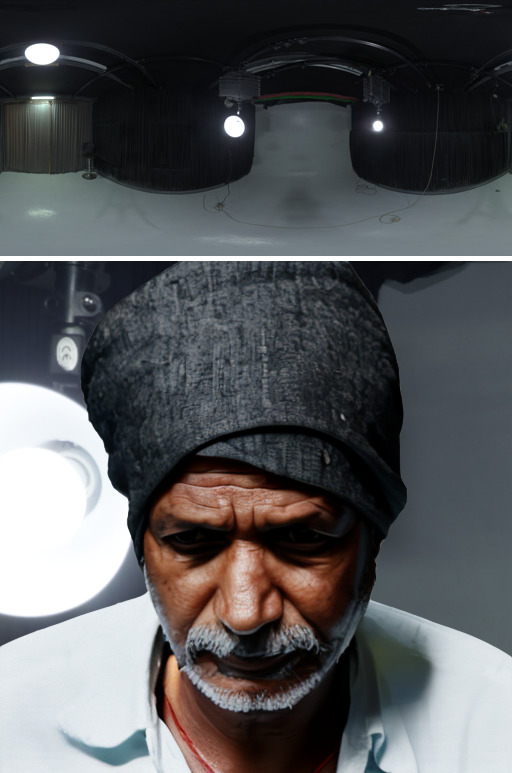} &
    \includegraphics[height=\extrafigureheight]{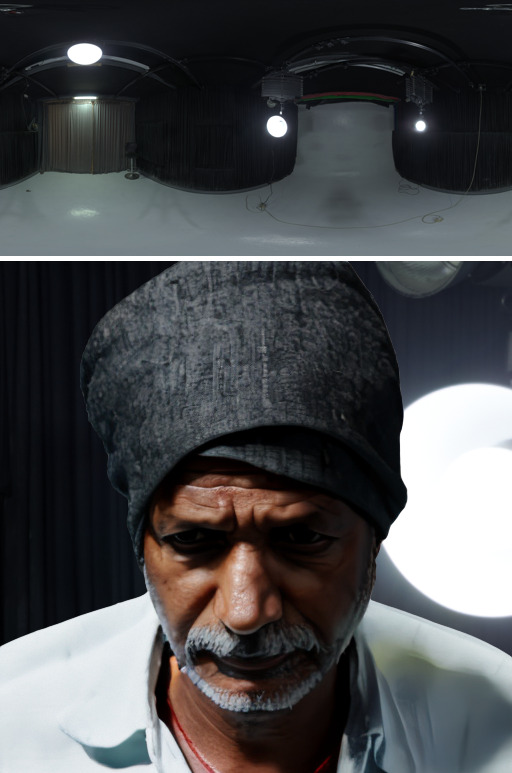} \\

    \multicolumn{1}{c}{} &
    \includegraphics[height=\extrafigureheight]{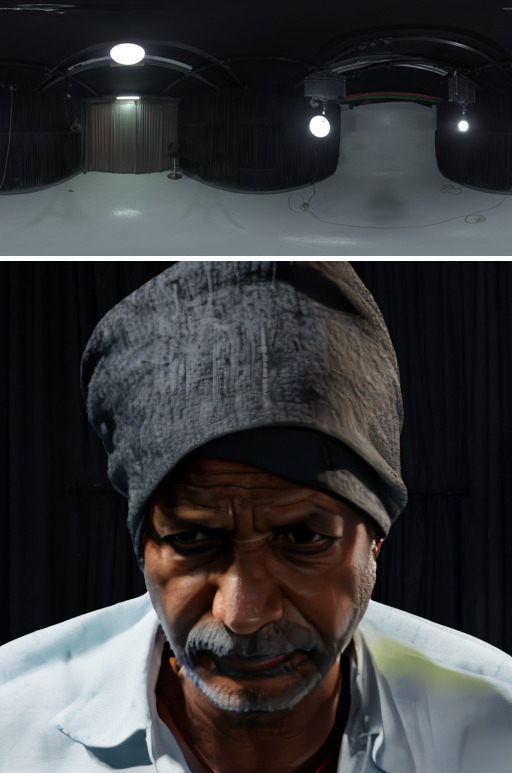} &
    \includegraphics[height=\extrafigureheight]{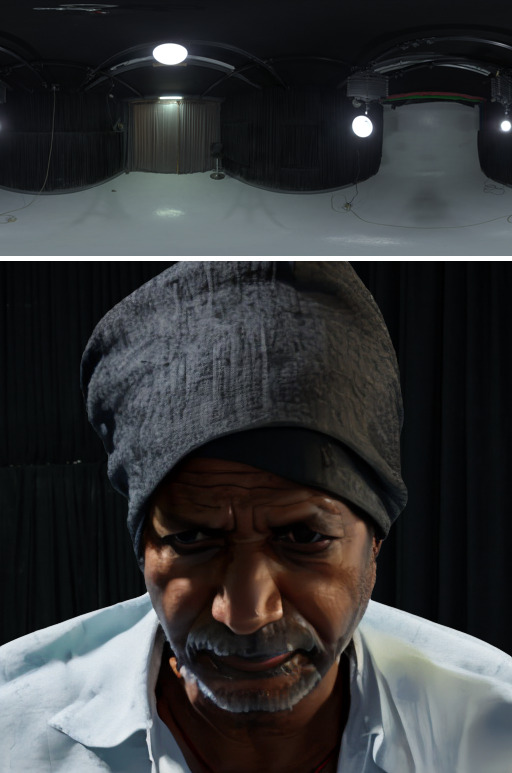} &
    \includegraphics[height=\extrafigureheight]{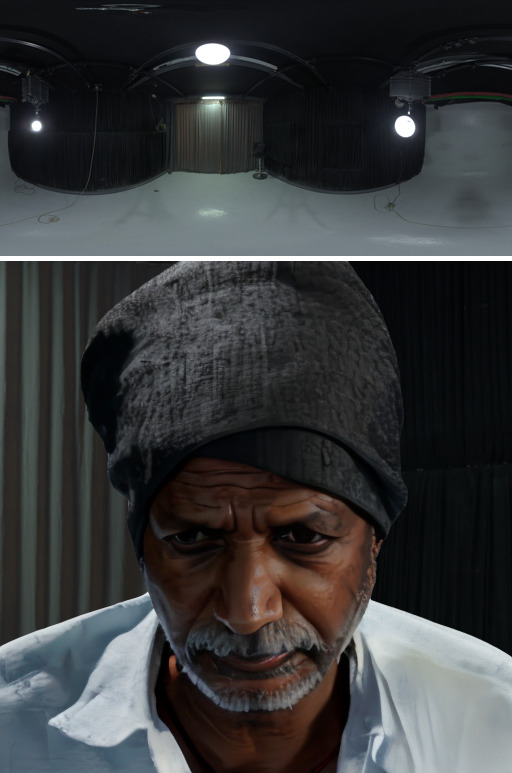} &
    \includegraphics[height=\extrafigureheight]{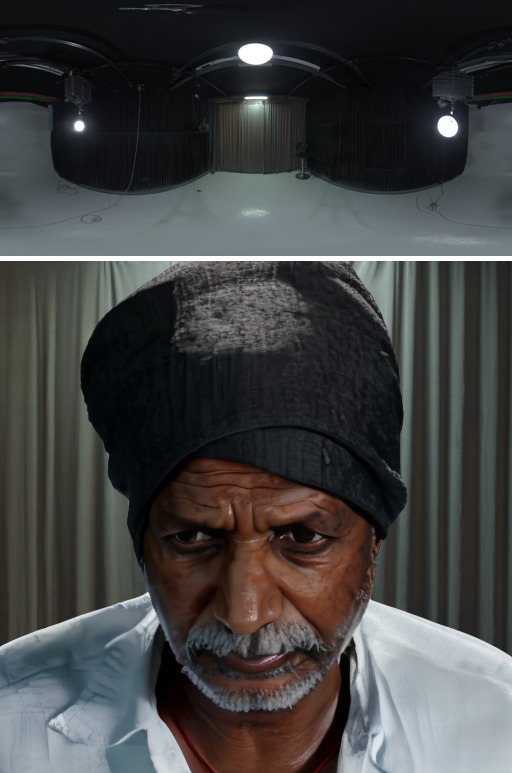} & 
    \includegraphics[height=\extrafigureheight]{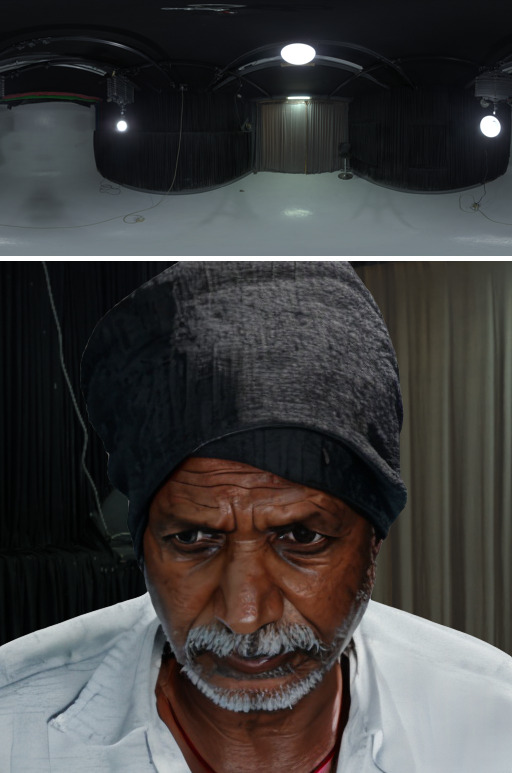} & 
    \includegraphics[height=\extrafigureheight]{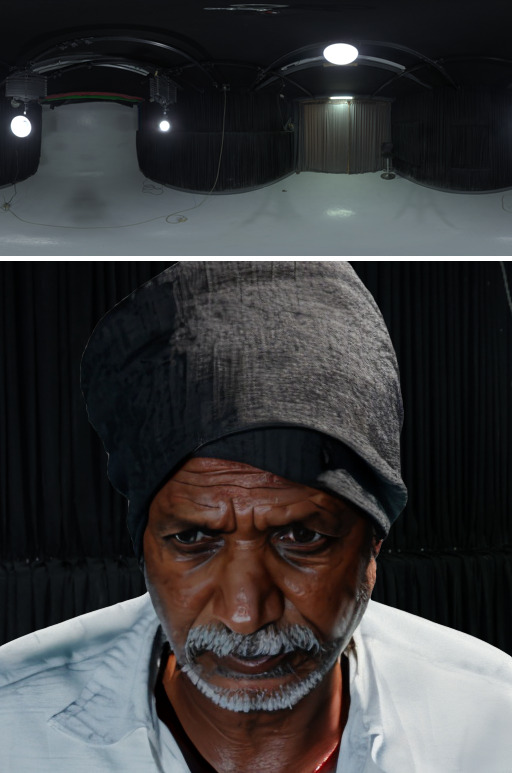} \\

    \includegraphics[height=\extrafigureheight]{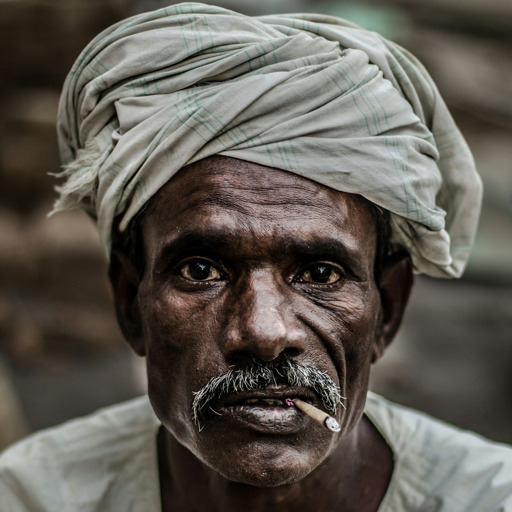} & 
    \includegraphics[height=\extrafigureheight]{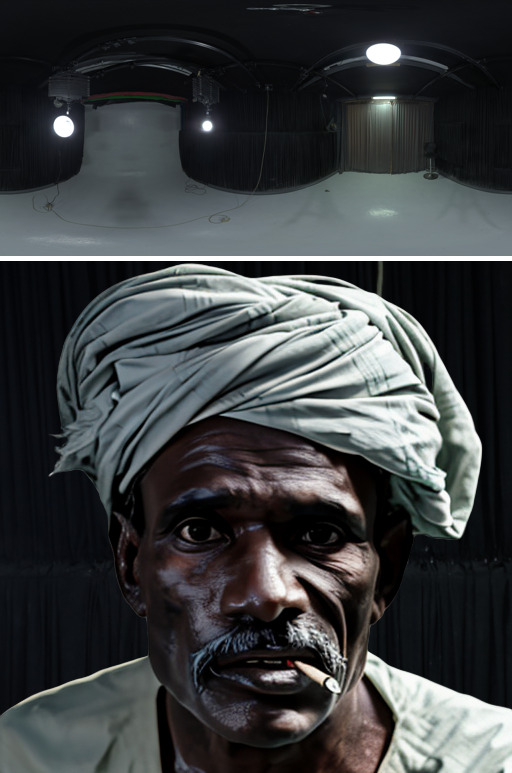} &
    \includegraphics[height=\extrafigureheight]{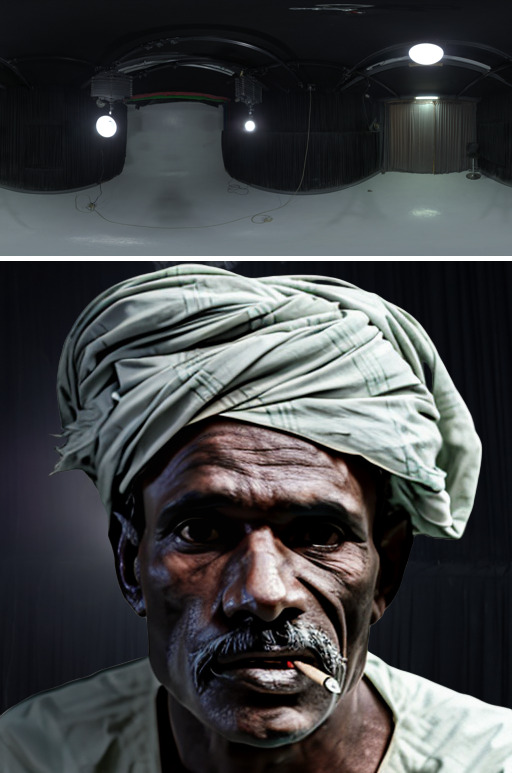} &
    \includegraphics[height=\extrafigureheight]{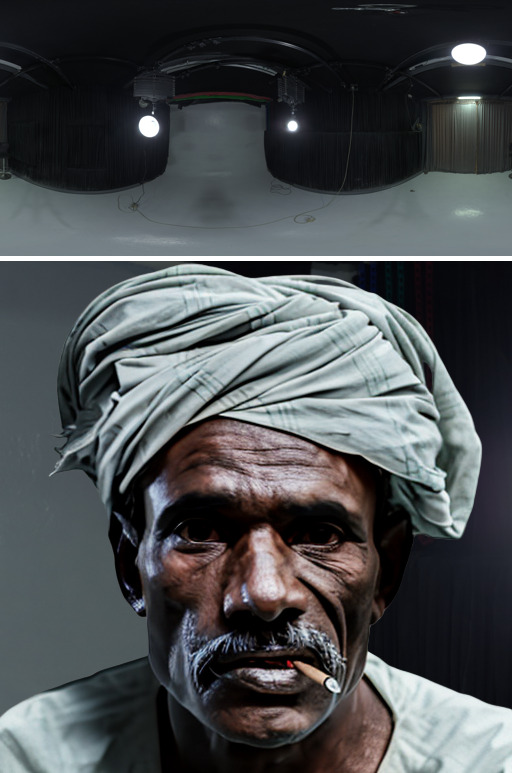} &
    \includegraphics[height=\extrafigureheight]{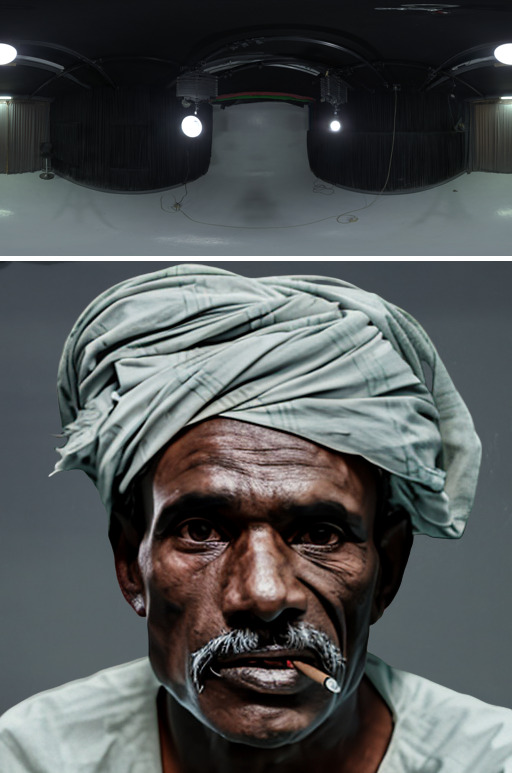} &
    \includegraphics[height=\extrafigureheight]{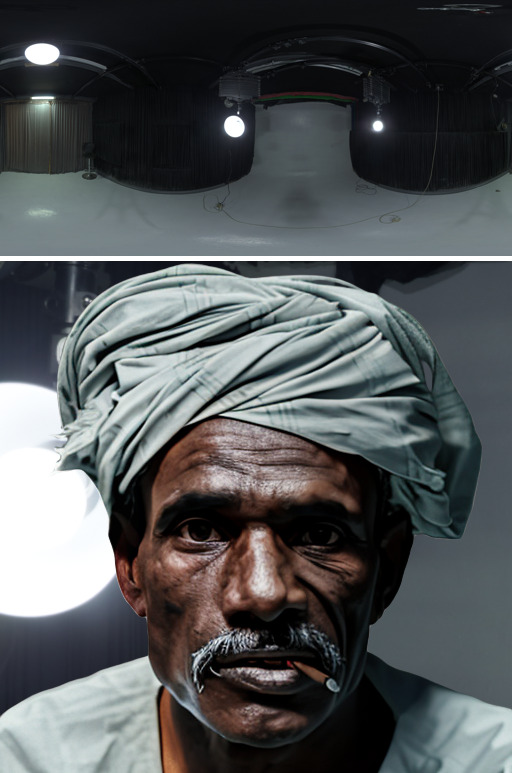} &
    \includegraphics[height=\extrafigureheight]{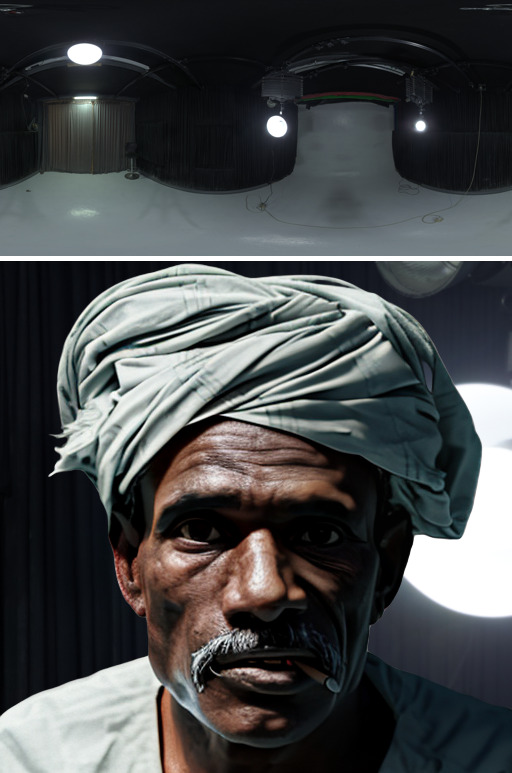} \\

    \multicolumn{1}{c}{} &
    \includegraphics[height=\extrafigureheight]{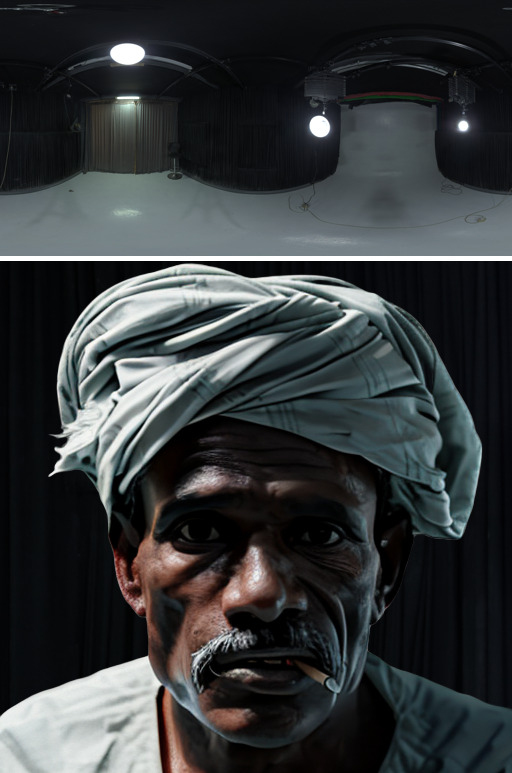} &
    \includegraphics[height=\extrafigureheight]{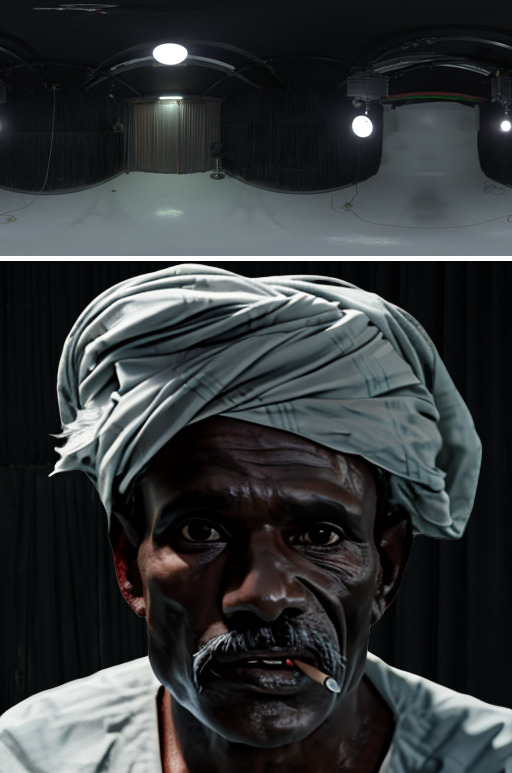} &
    \includegraphics[height=\extrafigureheight]{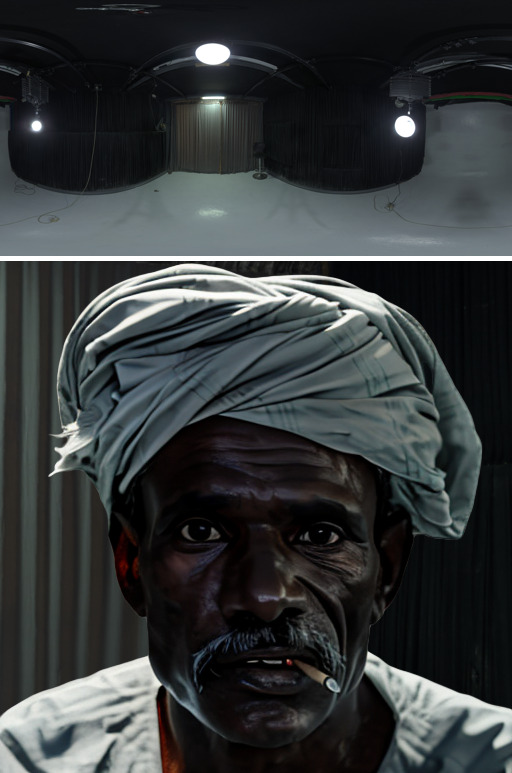} &
    \includegraphics[height=\extrafigureheight]{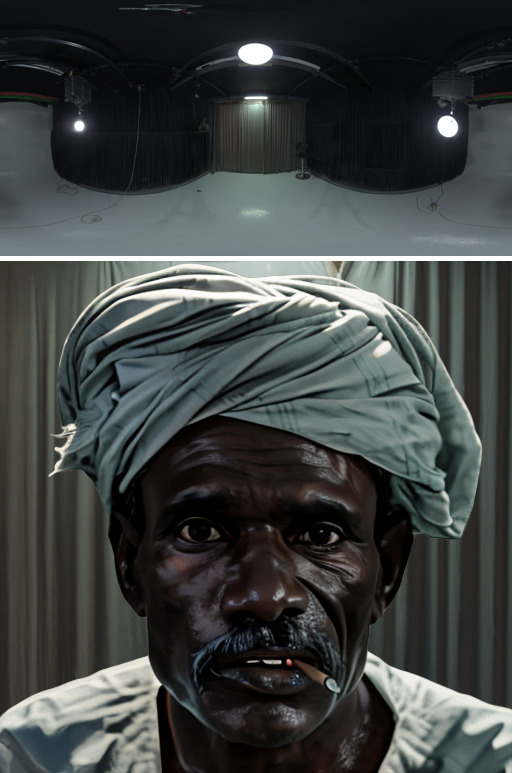} & 
    \includegraphics[height=\extrafigureheight]{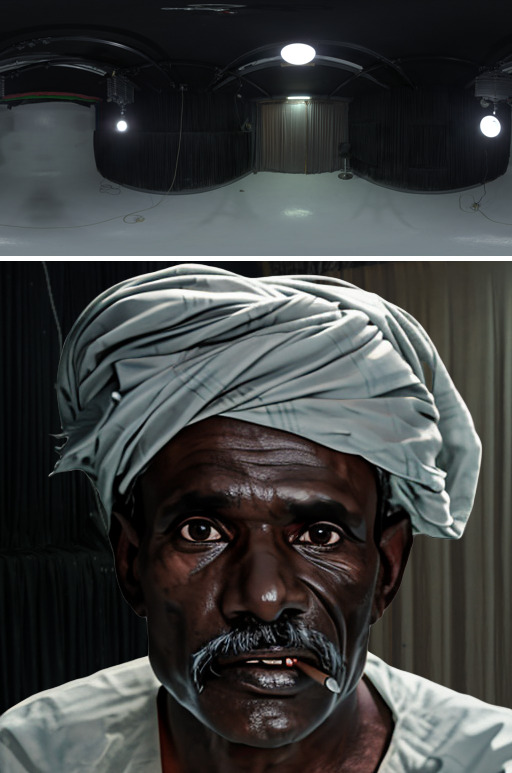} & 
    \includegraphics[height=\extrafigureheight]{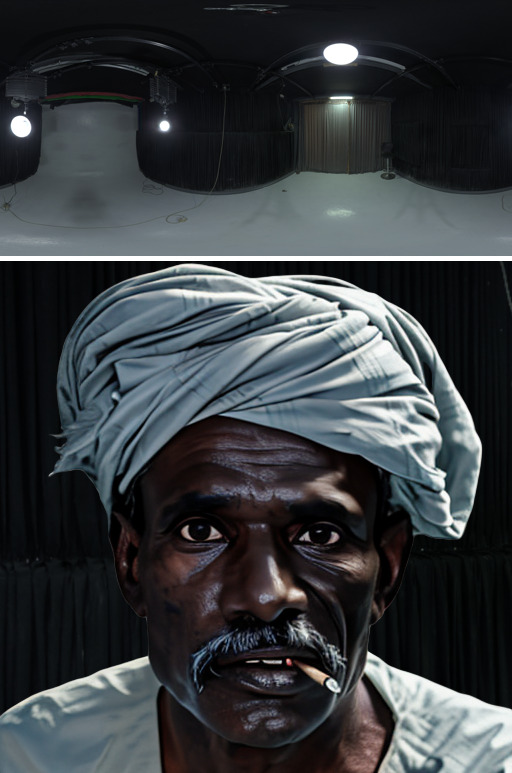} \\

    \end{tabular}
    \vspace{-3mm}
    \caption{To demonstrate SynthLight's ability to enhance portraits with studio-style lighting, we present in-the-wild portraits relit using a studio environment map, where the studio lights accentuate prominent features such as facial contours and expressions.}
    \label{fig:studio}
\end{figure*}

\begin{figure*}[htbp]
    \centering

    \begin{tabular}{c@{\hskip 0.2mm}c@{\hskip 0.1mm}c@{\hskip 0.1mm}c@{\hskip 0.1mm}c@{\hskip 0.1mm}c@{\hskip 0.1mm}c}

    \includegraphics[height=\extrafigureheight]{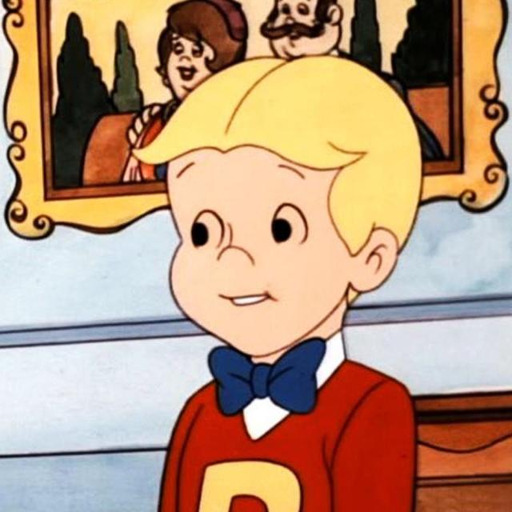} & 
    \includegraphics[height=\extrafigureheight]{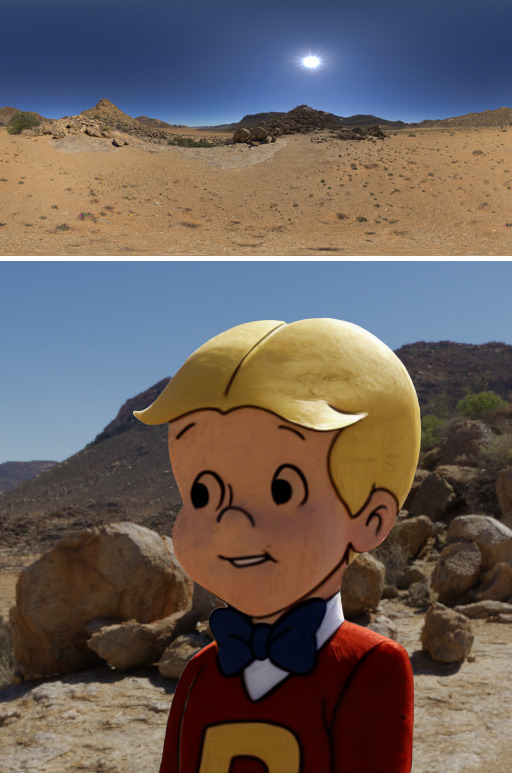} &
    \includegraphics[height=\extrafigureheight]{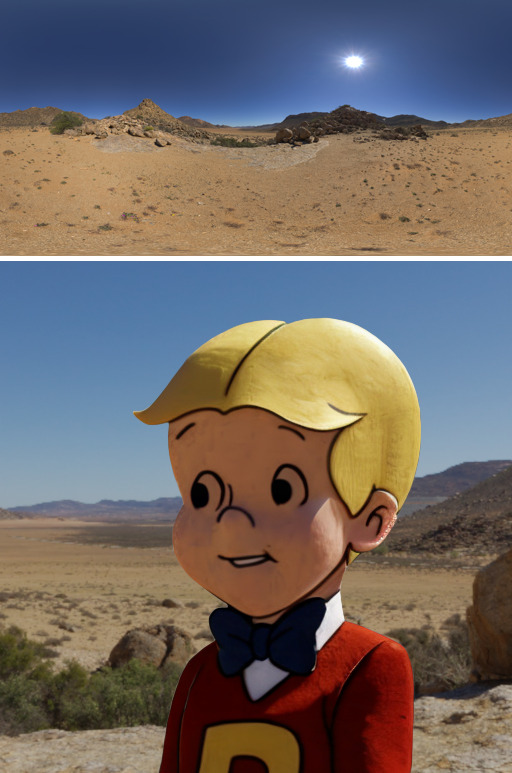} &
    \includegraphics[height=\extrafigureheight]{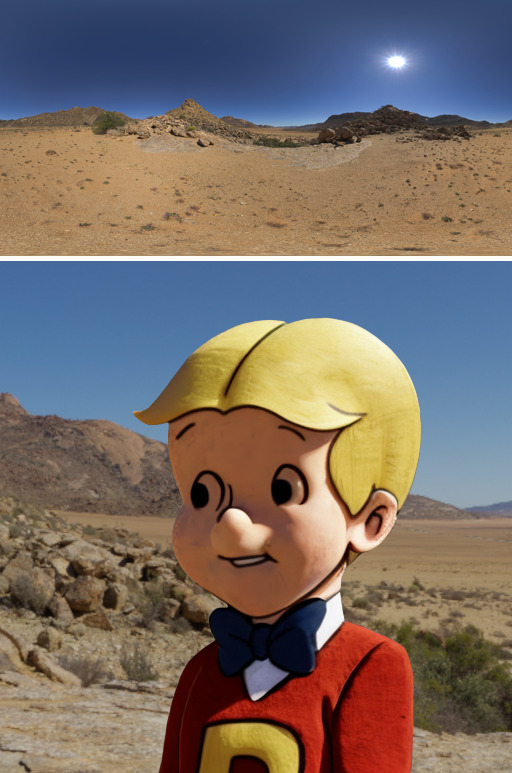} &
    \includegraphics[height=\extrafigureheight]{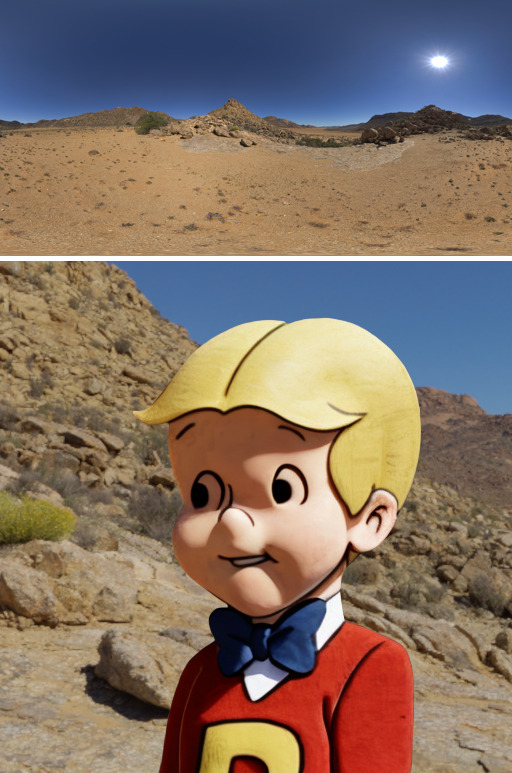} &
    \includegraphics[height=\extrafigureheight]{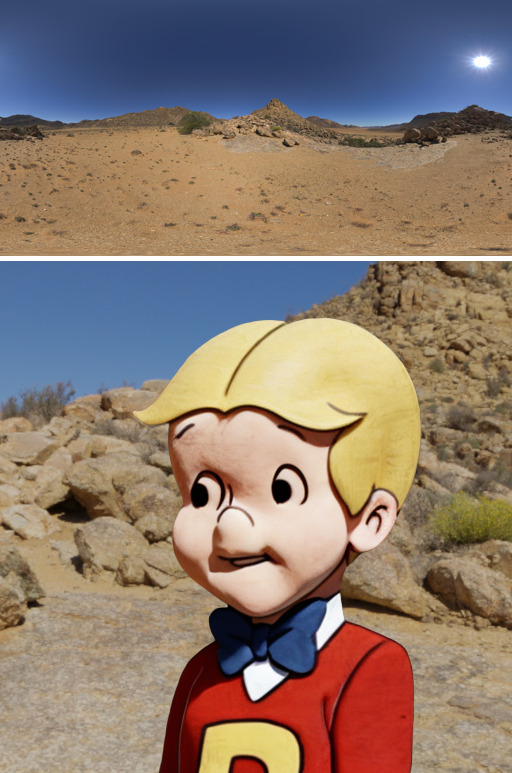} &
    \includegraphics[height=\extrafigureheight]{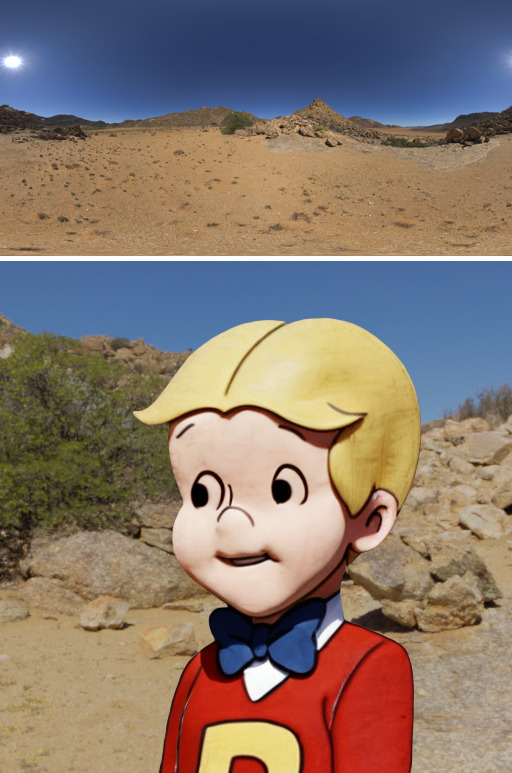} \\

    \multicolumn{1}{c}{} &
    \includegraphics[height=\extrafigureheight]{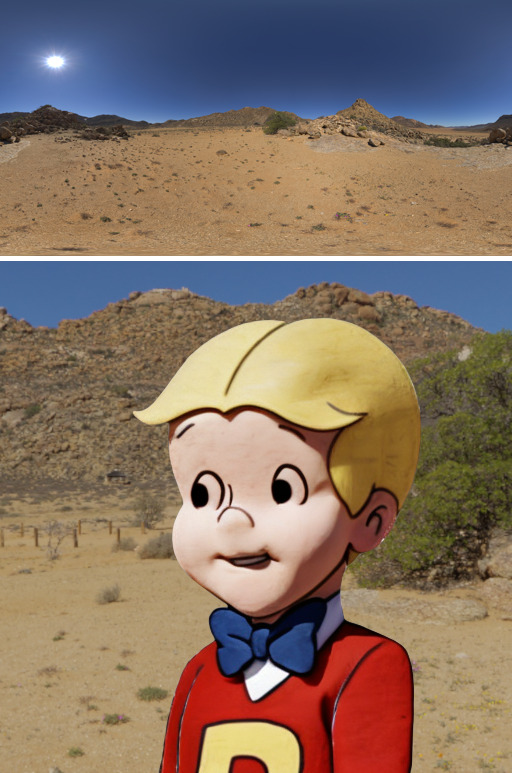} &
    \includegraphics[height=\extrafigureheight]{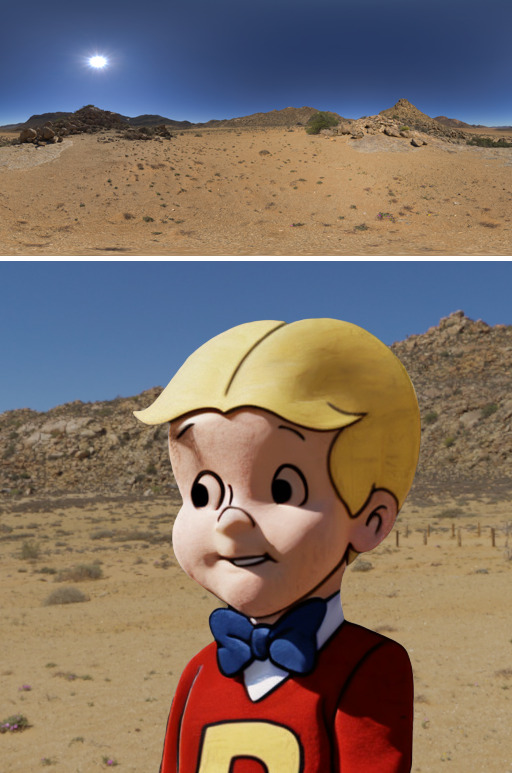} &
    \includegraphics[height=\extrafigureheight]{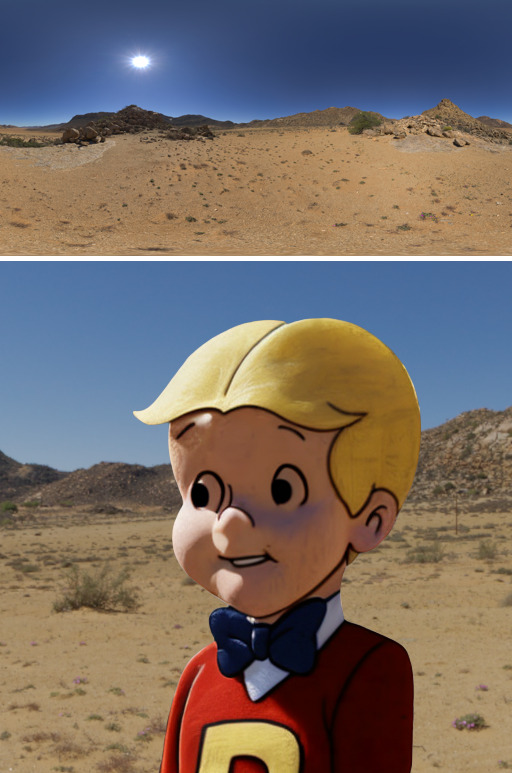} &
    \includegraphics[height=\extrafigureheight]{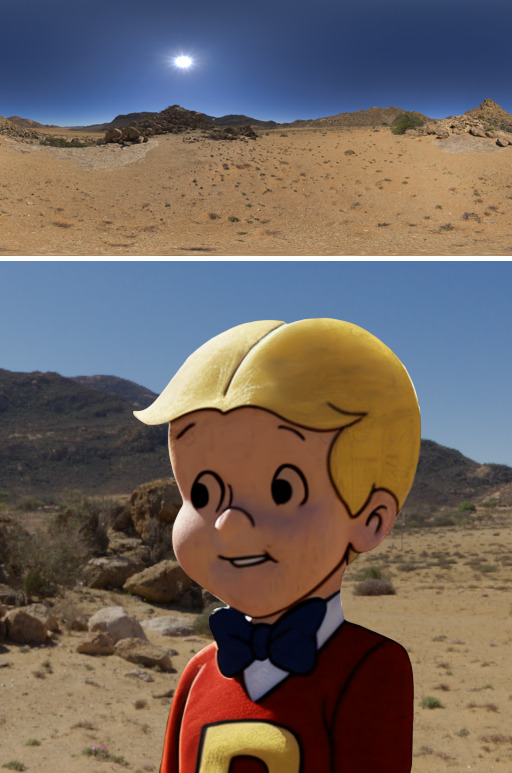} & 
    \includegraphics[height=\extrafigureheight]{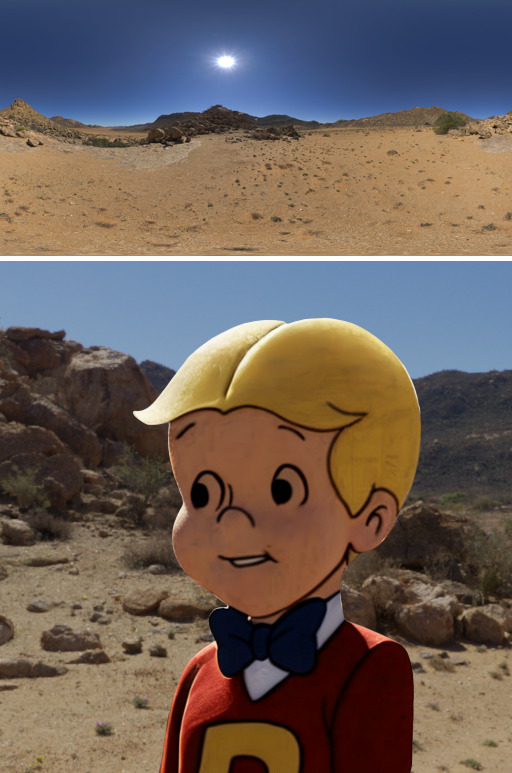} & 
    \includegraphics[height=\extrafigureheight]{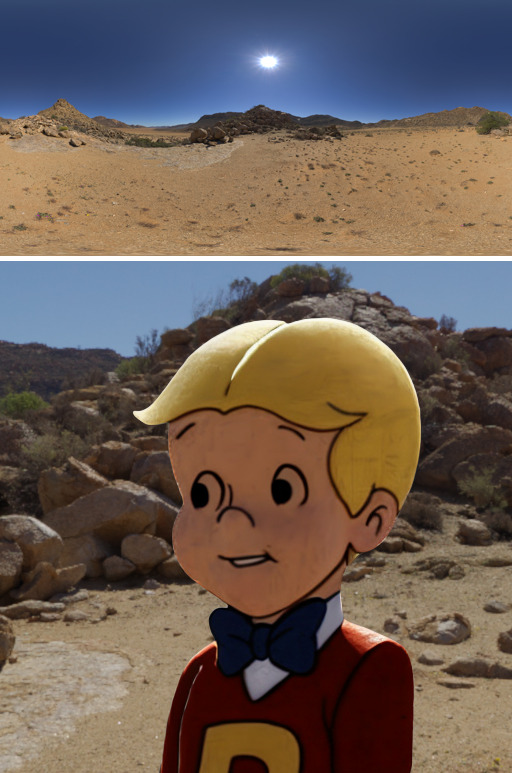} \\

    \includegraphics[height=\extrafigureheight]{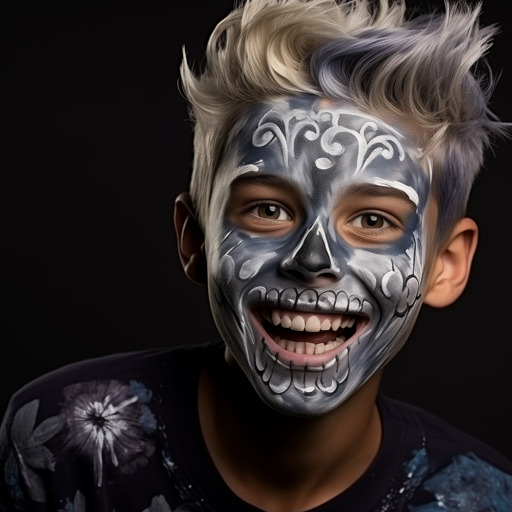} & 
    \includegraphics[height=\extrafigureheight]{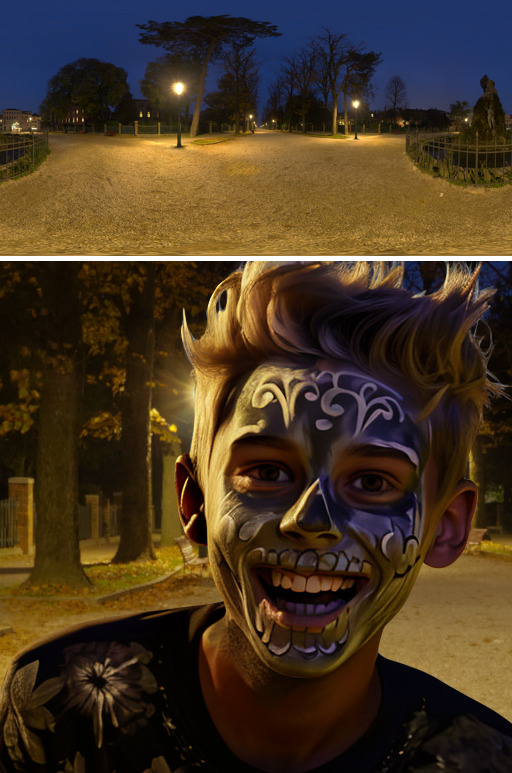} &
    \includegraphics[height=\extrafigureheight]{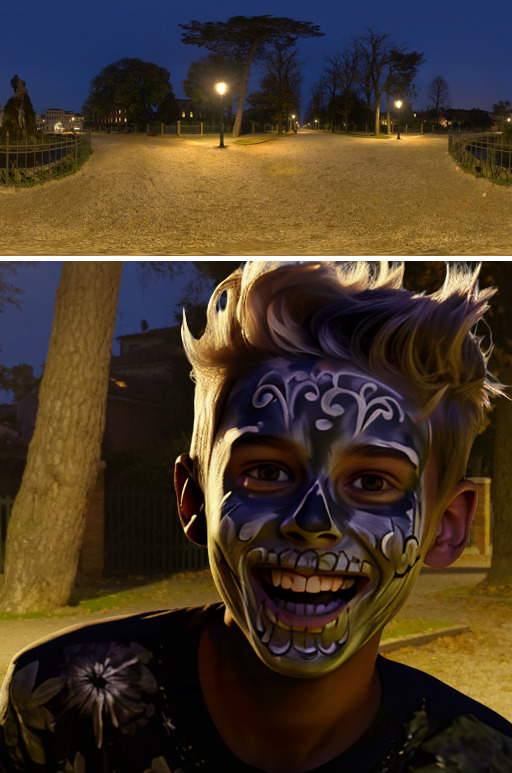} &
    \includegraphics[height=\extrafigureheight]{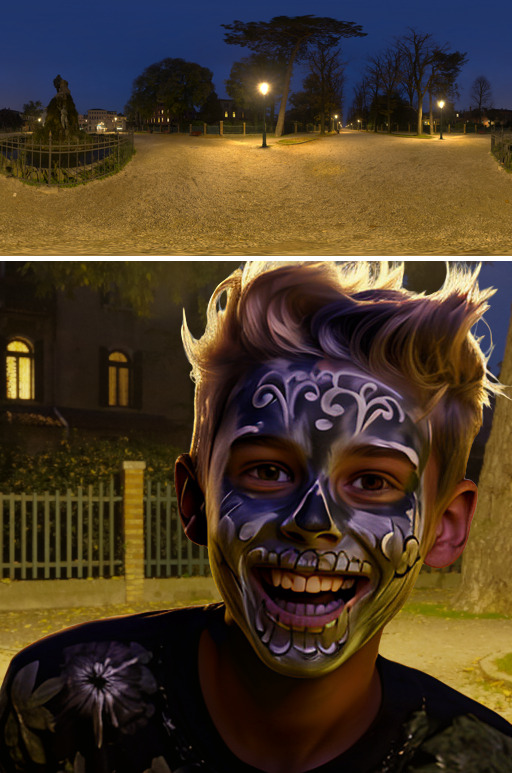} &
    \includegraphics[height=\extrafigureheight]{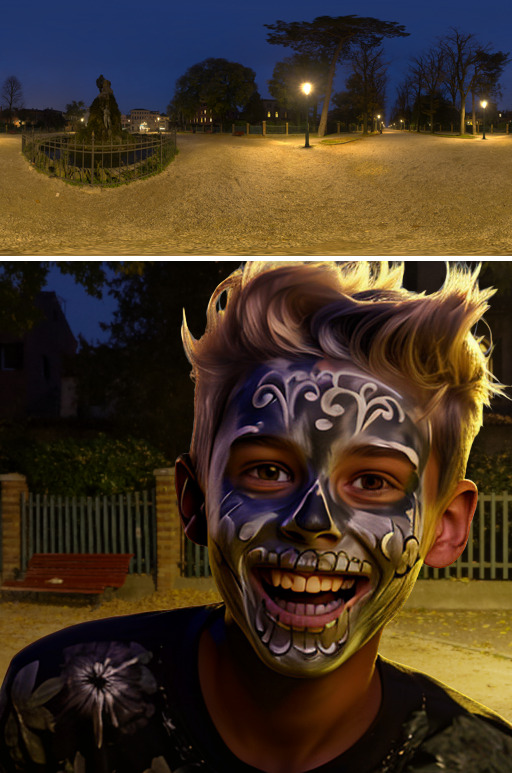} &
    \includegraphics[height=\extrafigureheight]{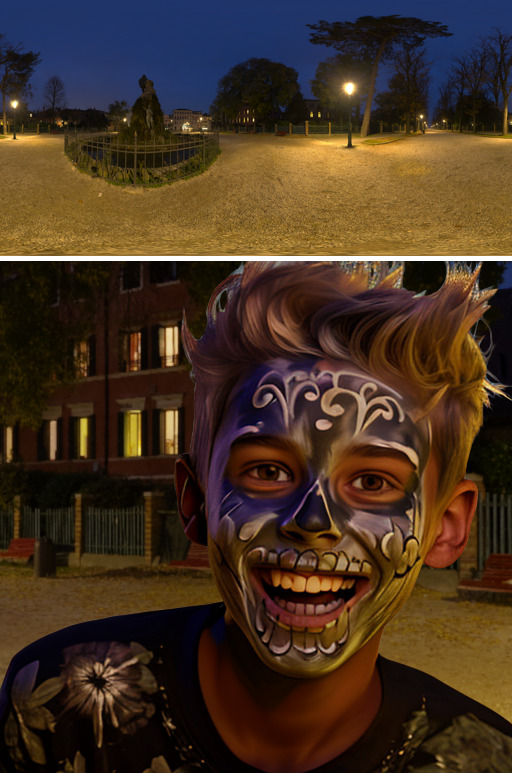} &
    \includegraphics[height=\extrafigureheight]{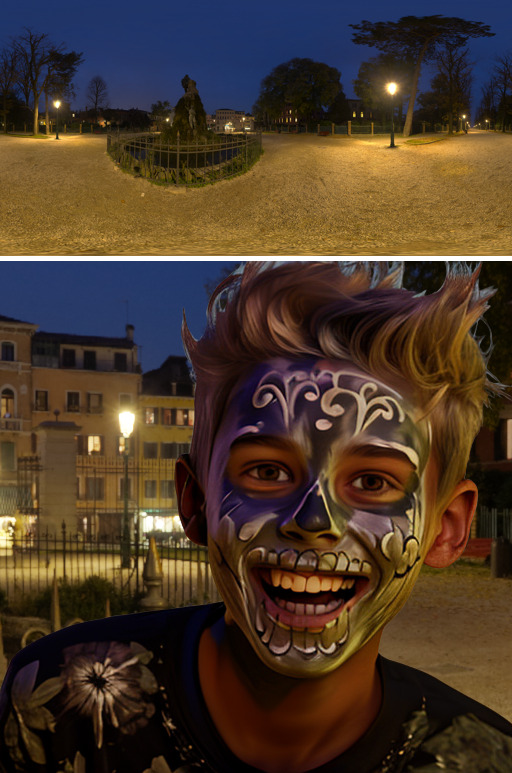} \\

    \multicolumn{1}{c}{} &
    \includegraphics[height=\extrafigureheight]{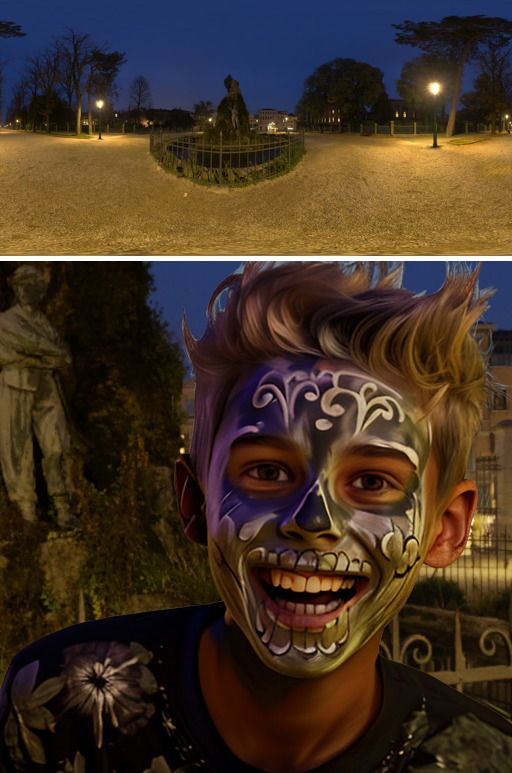} &
    \includegraphics[height=\extrafigureheight]{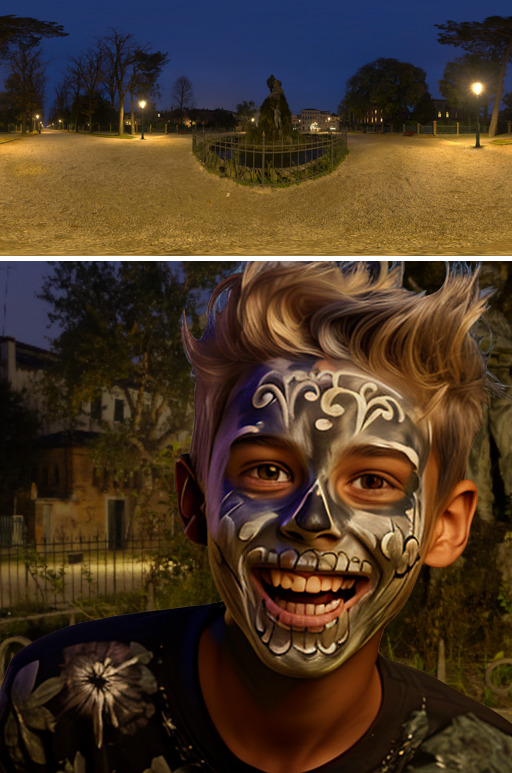} &
    \includegraphics[height=\extrafigureheight]{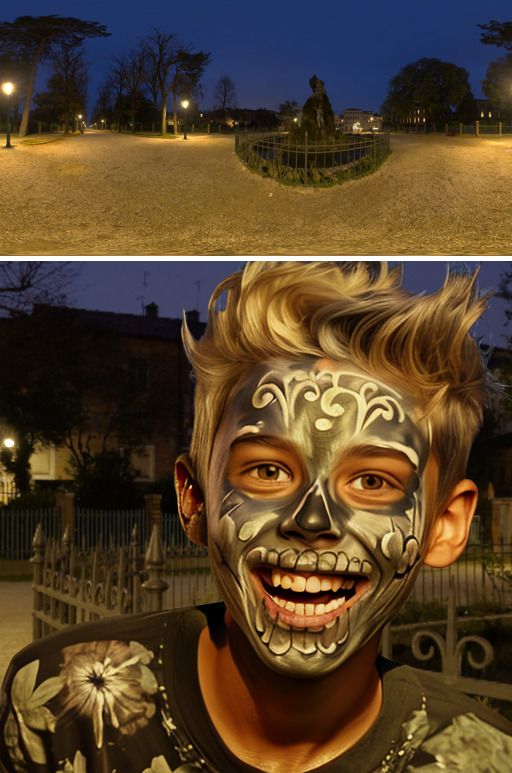} &
    \includegraphics[height=\extrafigureheight]{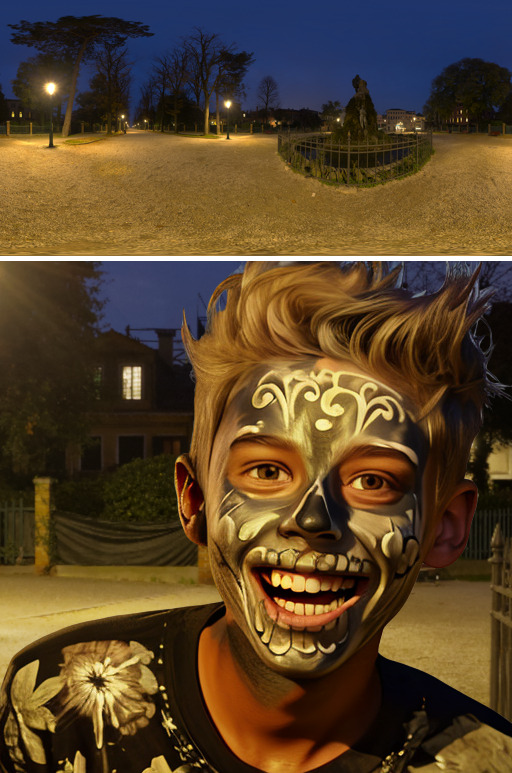} & 
    \includegraphics[height=\extrafigureheight]{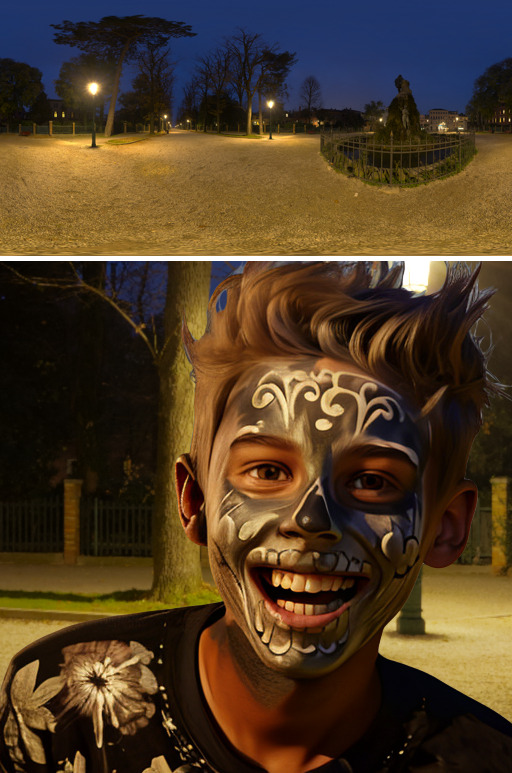} & 
    \includegraphics[height=\extrafigureheight]{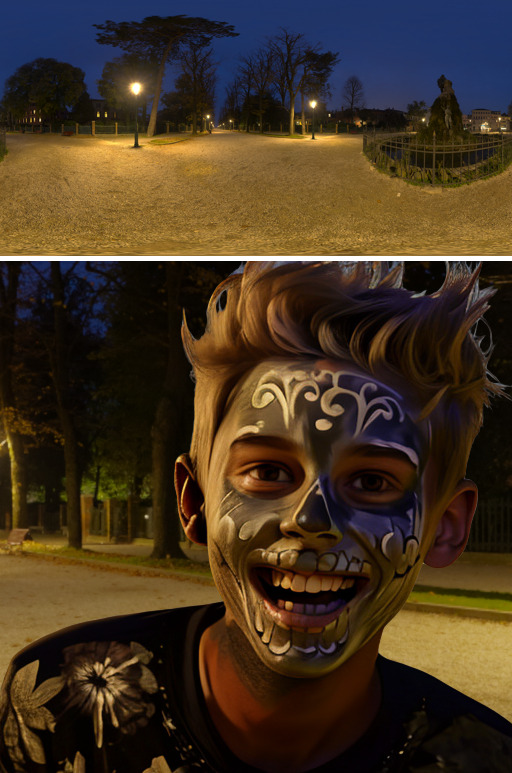} \\

    \vspace{2mm}

    \includegraphics[height=\extrafigureheight]{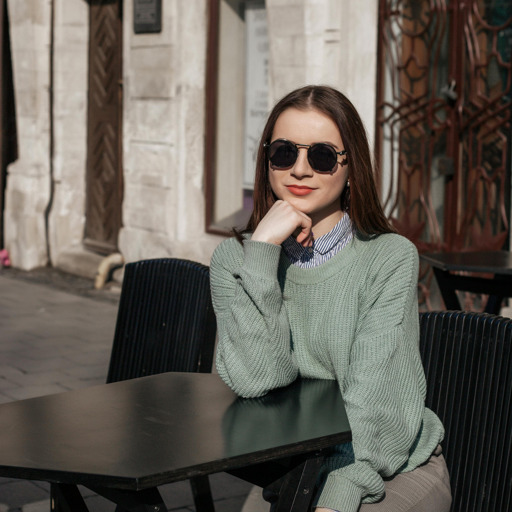} & 
    \includegraphics[height=\extrafigureheight]{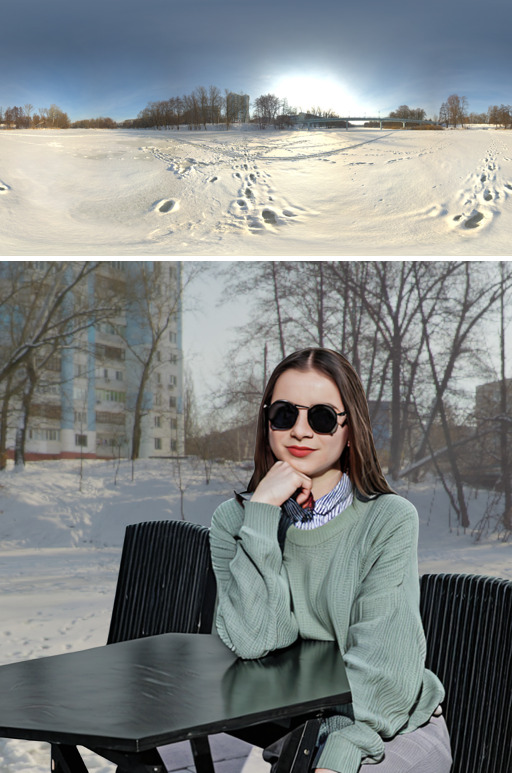} &
    \includegraphics[height=\extrafigureheight]{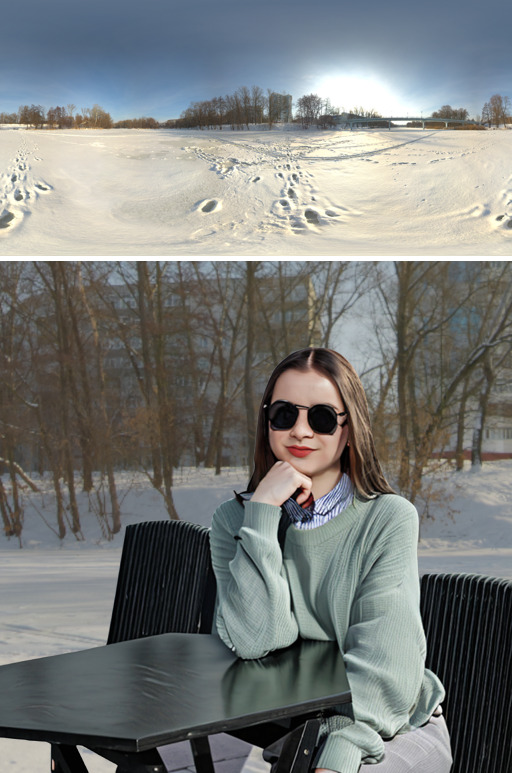} &
    \includegraphics[height=\extrafigureheight]{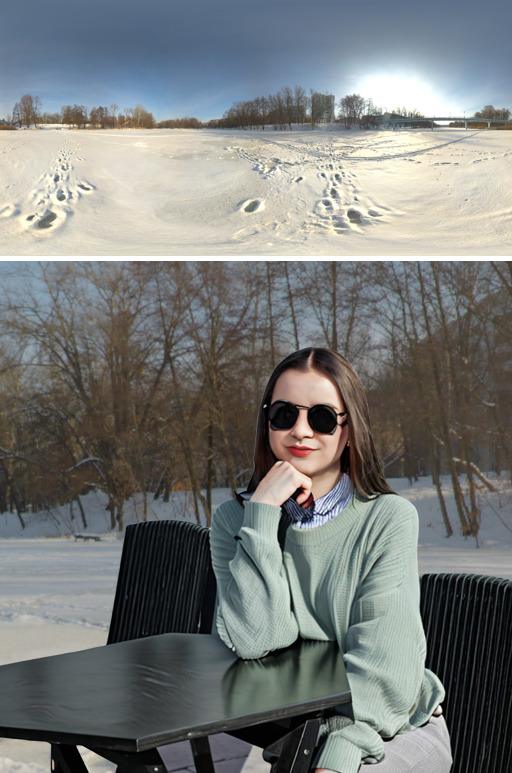} &
    \includegraphics[height=\extrafigureheight]{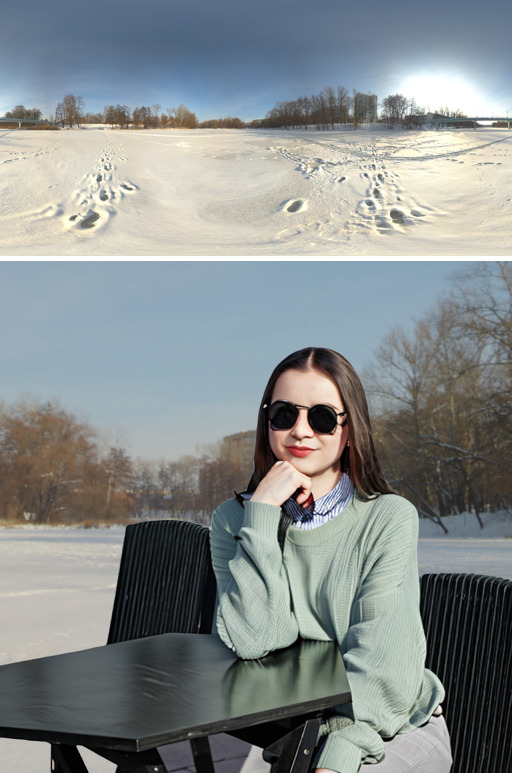} &
    \includegraphics[height=\extrafigureheight]{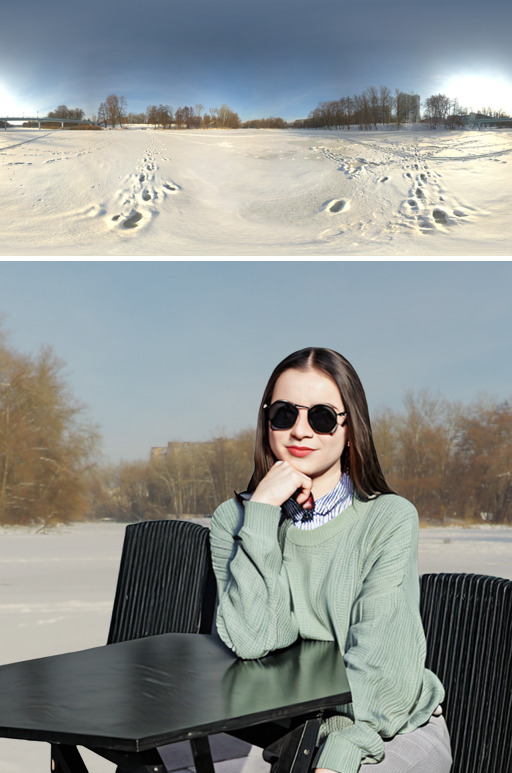} &
    \includegraphics[height=\extrafigureheight]{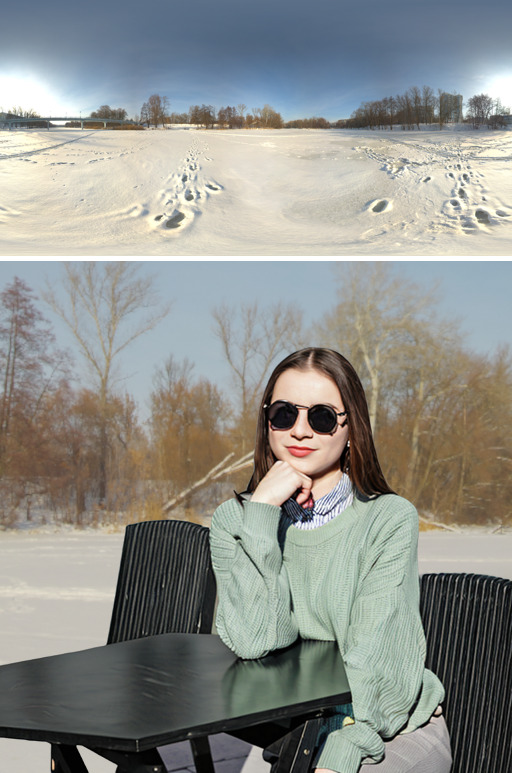} \\

    \multicolumn{1}{c}{} &
    \includegraphics[height=\extrafigureheight]{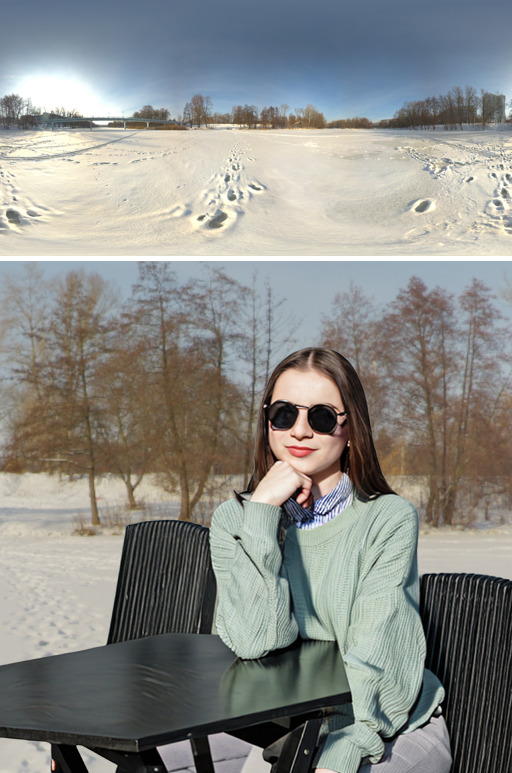} &
    \includegraphics[height=\extrafigureheight]{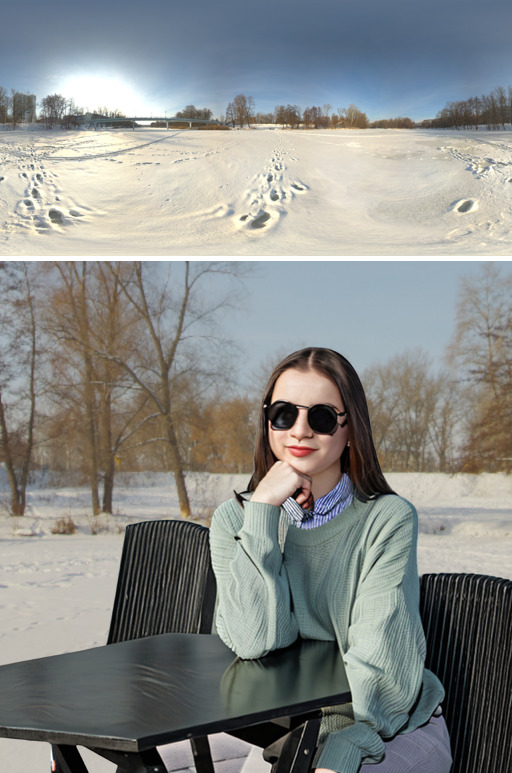} &
    \includegraphics[height=\extrafigureheight]{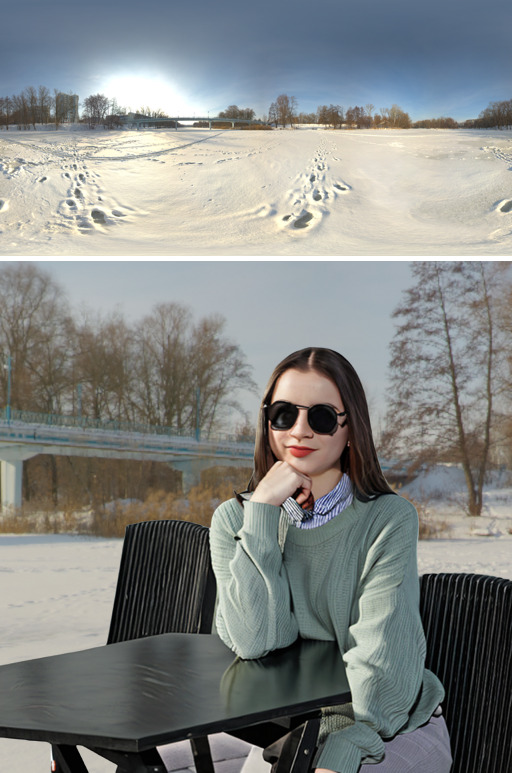} &
    \includegraphics[height=\extrafigureheight]{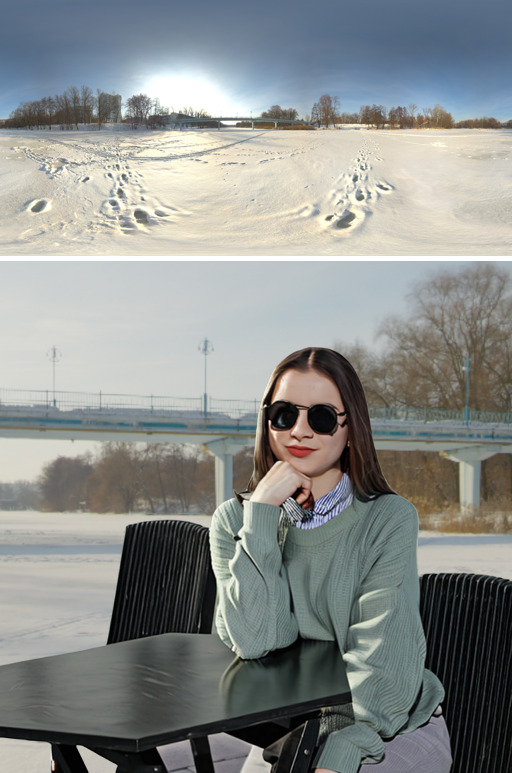} & 
    \includegraphics[height=\extrafigureheight]{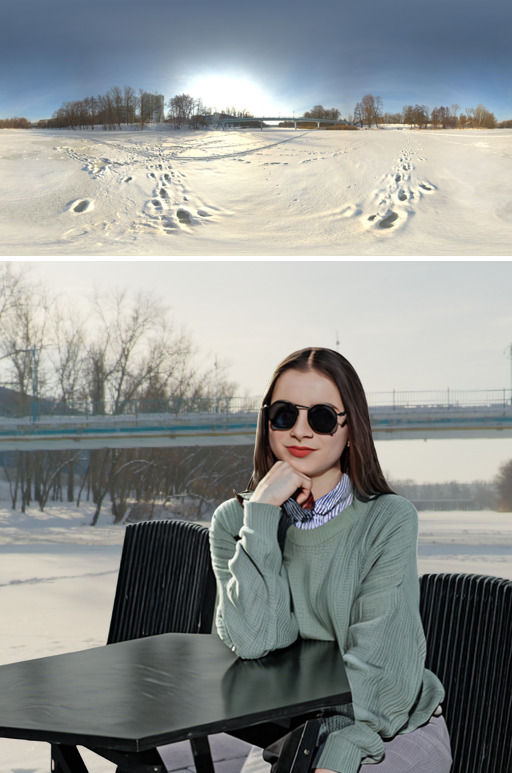} & 
    \includegraphics[height=\extrafigureheight]{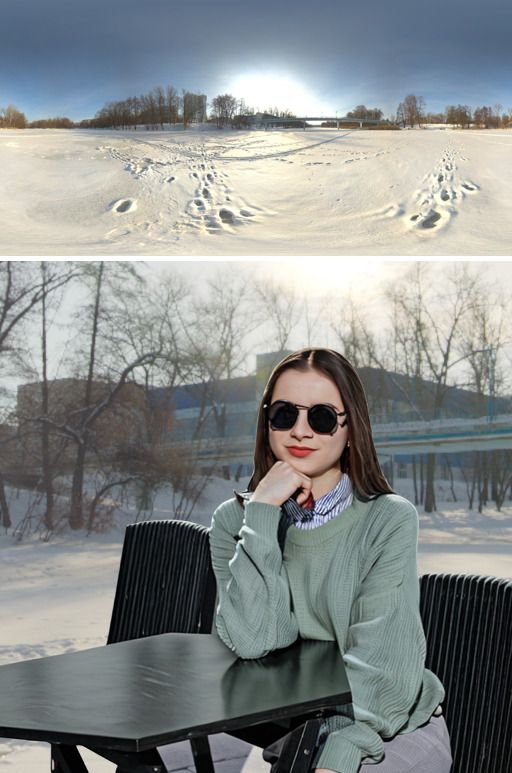} \\

    \end{tabular}
    \vspace{-3mm}
    \caption{We show challenging in-the-wild portraits featuring 2D cartoon characters, child wearing face paint and a full body portrait, demonstrating that our method can generalize beyond the synthetic dataset seen during training.}
    \label{fig:challenging}
\end{figure*}

\vspace{-4mm}
\paragraph{In-the-wild Test Portraits} We demonstrate portrait relighting in the presence of strong sunlight to produce effects such as strong cast shadow from facial features, rim-effects in hair and specular highlights in \cref{fig:outdoor}. In \cref{fig:studio}, we demonstrate applying a studio environment map on in-the-wild test portraits to accentuate prominent features such as facial contours and expressions in the portraits. In \cref{fig:challenging}, we showcase that SynthLight generalises to several to several challenging cases such as a 2D cartoon, a boy with face paint and a full body portrait, beyond the diversity present in the synthetic training data.

\begin{figure*}[!htbp]
    \centering
    \begin{tabular}{@{\hskip 0.0mm}c@{\hskip 0.1mm}c@{\hskip 0.1mm}c@{\hskip 0.1mm}c@{\hskip 0.0mm}}

        \includegraphics[height=\specialeffectheight]{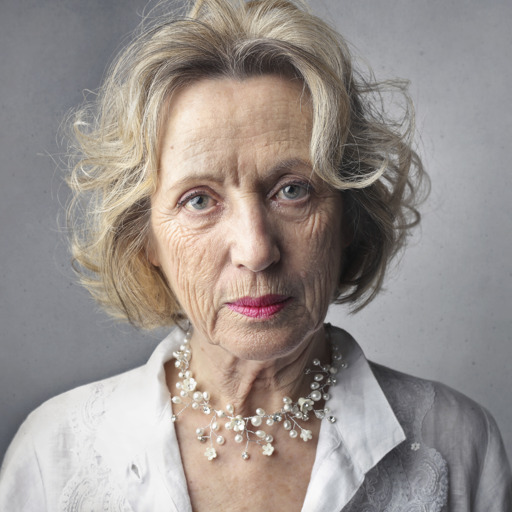} &
        \includegraphics[height=\specialeffectheight]{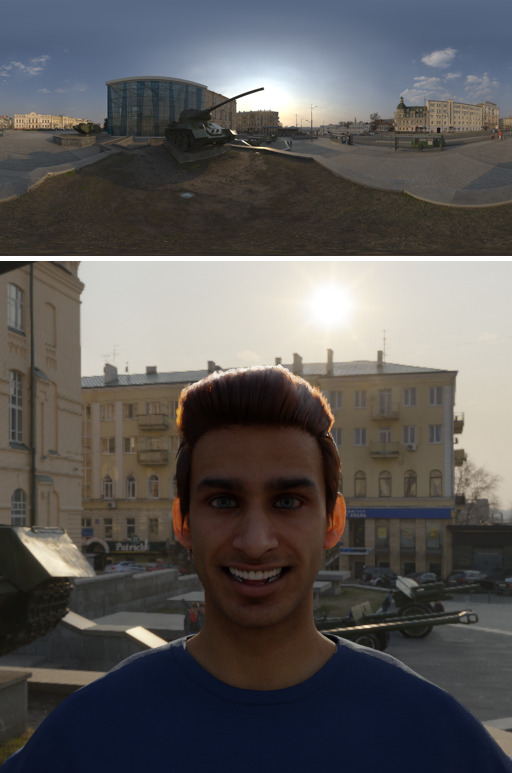} &
        \includegraphics[height=\specialeffectheight]{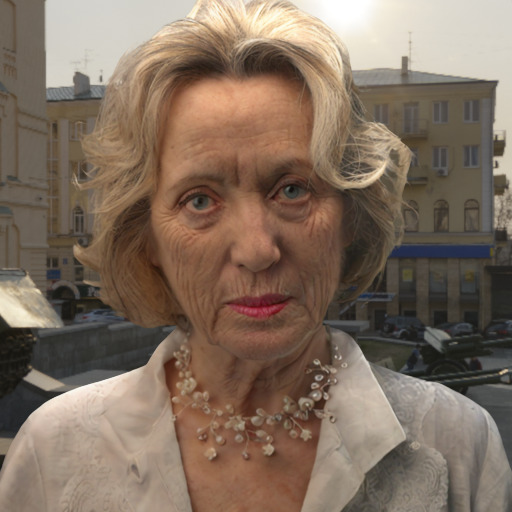} &
        \includegraphics[height=\specialeffectheight]{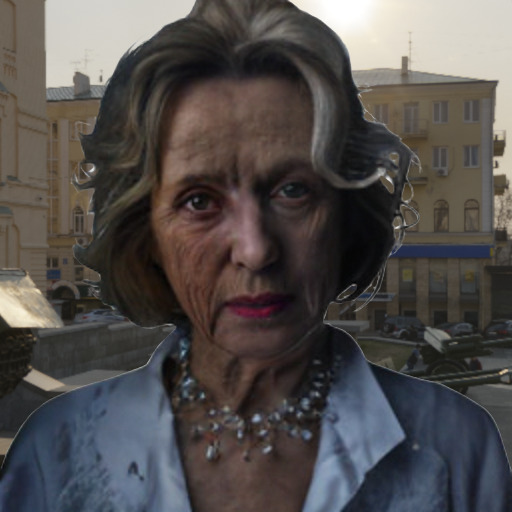} \\

        & & IC-Light & Neural Gaffer \\

        & &
        \includegraphics[height=\specialeffectheight]{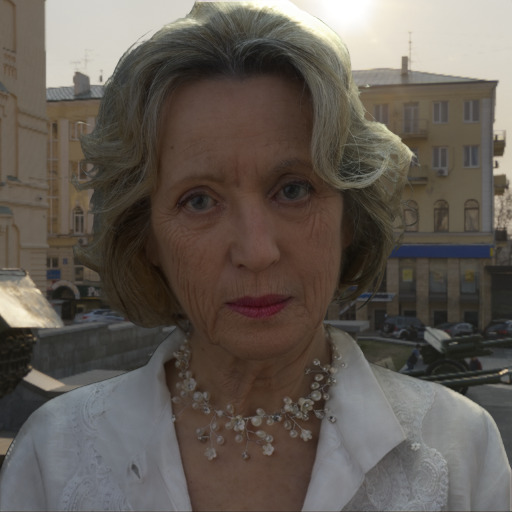} &
        \includegraphics[height=\specialeffectheight]{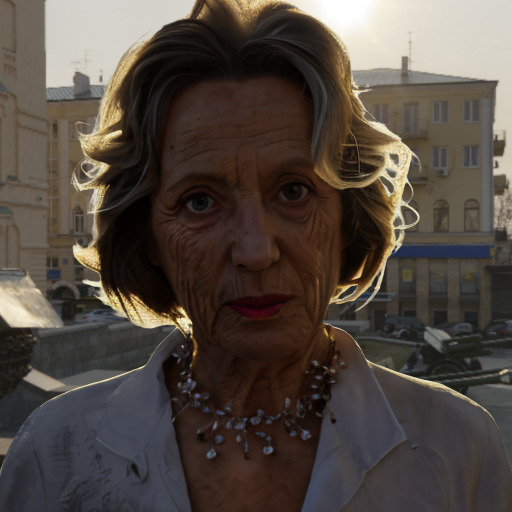} \\

        & & SwitchLight & Ours \\

        \vspace{-4mm} \\

        \includegraphics[height=\specialeffectheight]{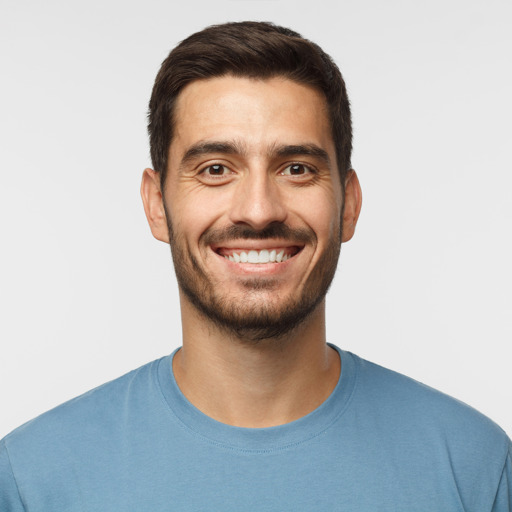} &
        \includegraphics[height=\specialeffectheight]{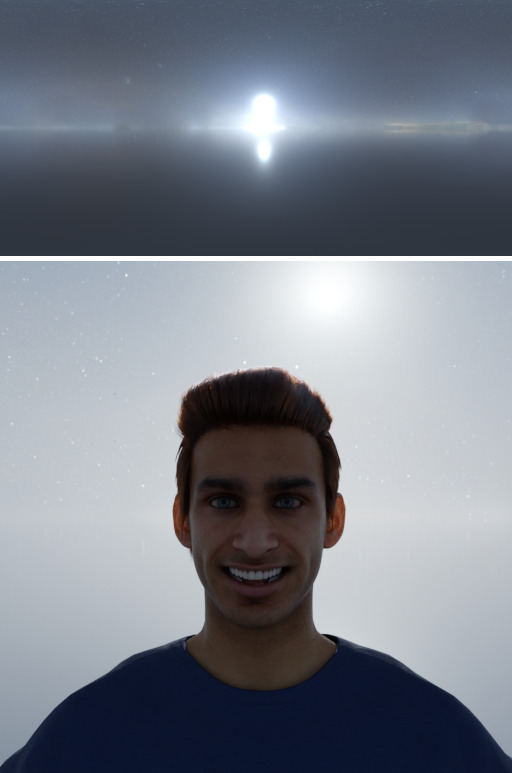} &
        \includegraphics[height=\specialeffectheight]{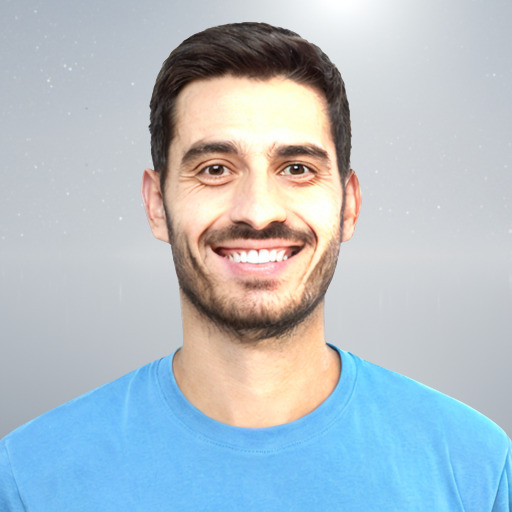} &
        \includegraphics[height=\specialeffectheight]{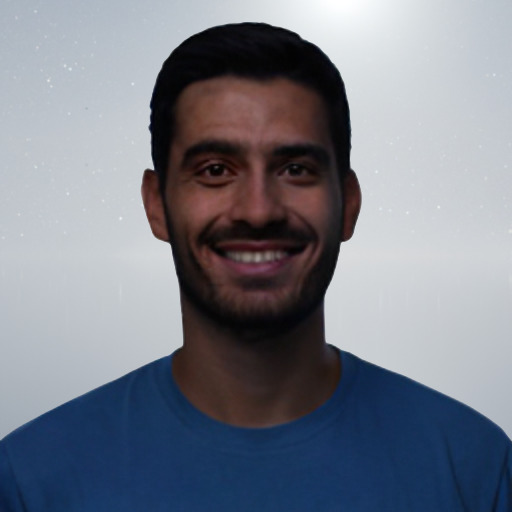} \\

        & & IC-Light & Neural Gaffer \\

        & &
        \includegraphics[height=\specialeffectheight]{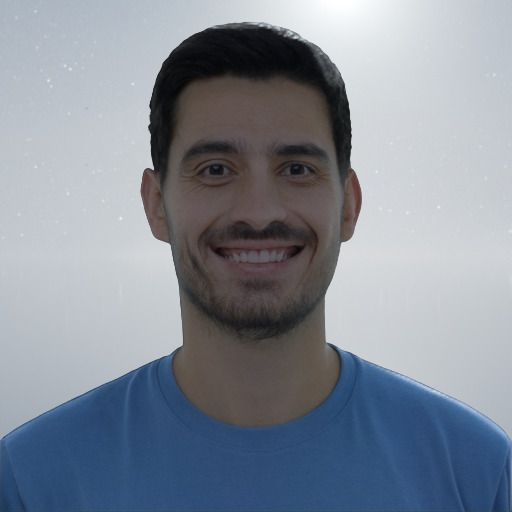} &
        \includegraphics[height=\specialeffectheight]{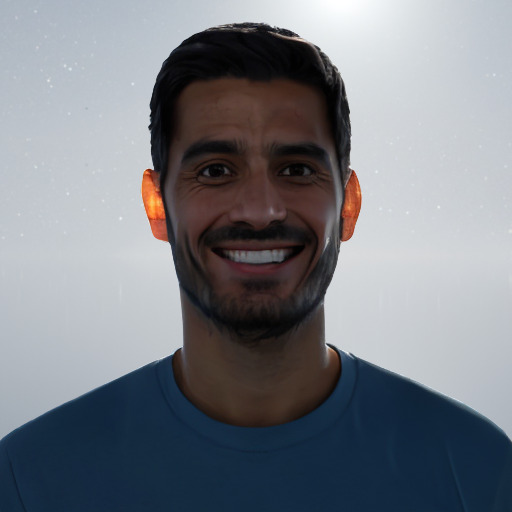} \\

        & & SwitchLight & Ours \\

        \vspace{5mm} \\

    \end{tabular}
    \vspace{-2mm}
    \caption{We show the input portrait, the environment map used to relight and a reference synthetic data rendering from Blender (left) and results from our method and baselines (right). SynthLight achieves lighting effects such as rim-light on hair (top) and subsurface scattering in ears (bottom). These cannot be generated by baselines.}
    \label{fig:comparison_all_extra_special}
\end{figure*}

\begin{figure*}[!htbp]
    \centering
    \begin{tabular}{@{\hskip 0.0mm}c@{\hskip 0.1mm}c@{\hskip 0.1mm}c@{\hskip 0.1mm}c@{\hskip 0.0mm}}

        \includegraphics[height=\specialeffectheight]{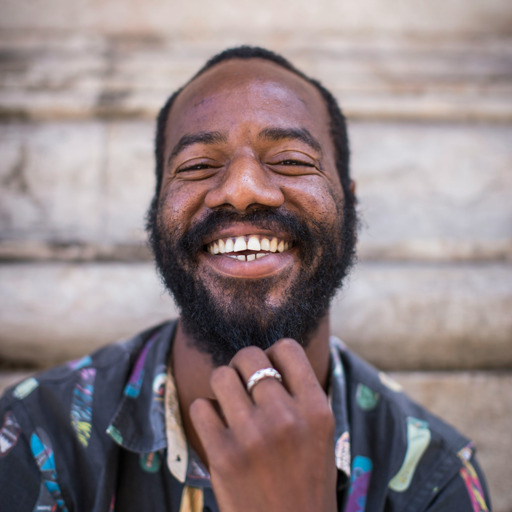} &
        \includegraphics[height=\specialeffectheight]{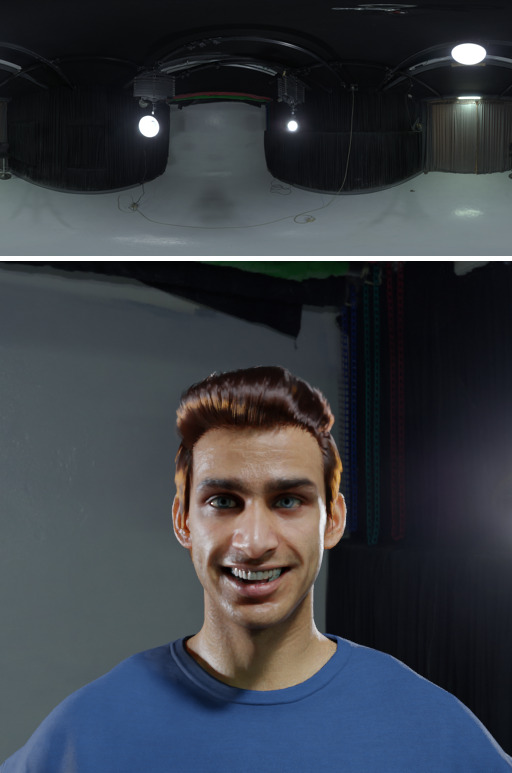} &
        \includegraphics[height=\specialeffectheight]{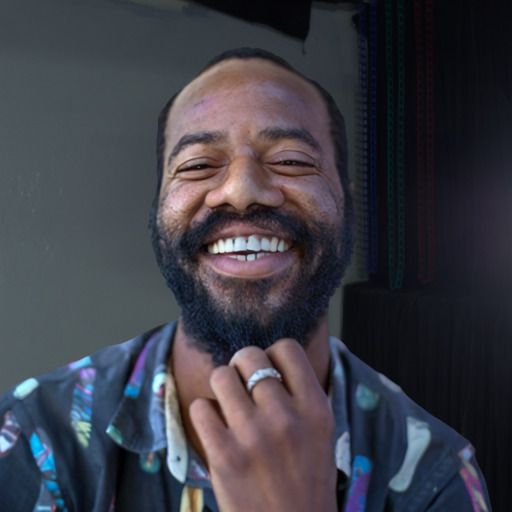} &
        \includegraphics[height=\specialeffectheight]{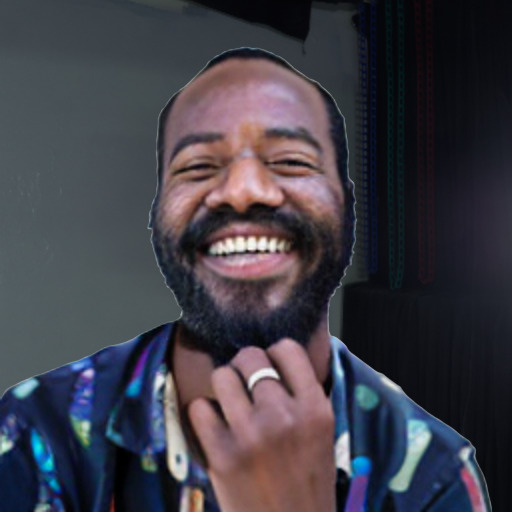} \\

        & & IC-Light & Neural Gaffer \\

        & &
        \includegraphics[height=\specialeffectheight]{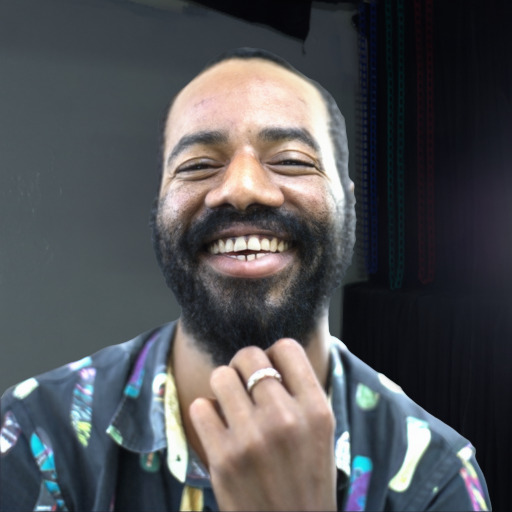} &
        \includegraphics[height=\specialeffectheight]{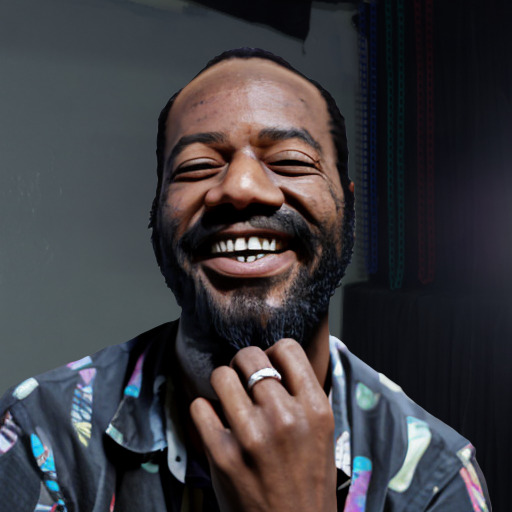} \\

        & & SwitchLight & Ours \\
        
    \end{tabular}
    \vspace{-2mm}
    \caption{We highlight lighting effects that our method achieves in contrast to baselines such as specular highlights in response to lighting direction.}
    \label{fig:comparison_all_extra_special_another}
\end{figure*}

\vspace{-4mm}
\paragraph{Comparison with Baselines} 
We evaluate SynthLight against several baseline methods on in-the-wild portraits. As shown in \cref{fig:comparison_all_extra_special}, SynthLight achieves lighting effects, such as the rim-light effect in hair and subsurface scattering in the ears. Additionally, \cref{fig:comparison_all_extra_special_another} illustrates specular highlights on darker skin tones, a capability not replicated by baseline methods. 

These limitations in baselines can be attributed to the nature of the underlying methods. For instance, IC-Light, being an image harmonization technique, is not trained on physically based rendered data and hence struggles with achieving these effects. Surprisingly, even relighting approaches, such as Neural Gaffer and SwitchLight fall short. While Neural Gaffer is trained on rendered images, it is not explicitly trained on human facial data, leading to limited effectiveness in such scenarios. Even SwitchLight, despite leveraging Light Stage data, does not capture these intricate lighting effects.

\begin{figure*}[!htbp]
    \centering
    \begin{tabular}{@{\hskip 0.0mm}c@{\hskip 0.1mm}c@{\hskip 0.1mm}c@{\hskip 0.1mm}c@{\hskip 0.0mm}}

        \includegraphics[height=\specialeffectheight]{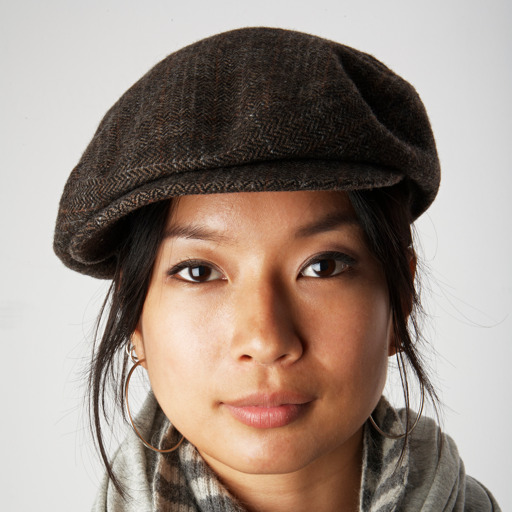} &
        \includegraphics[height=\specialeffectheight]{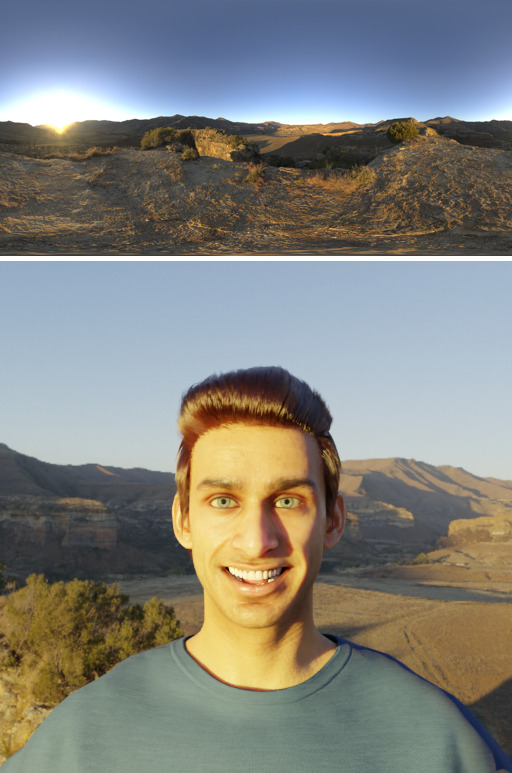} &
        \includegraphics[height=\specialeffectheight]{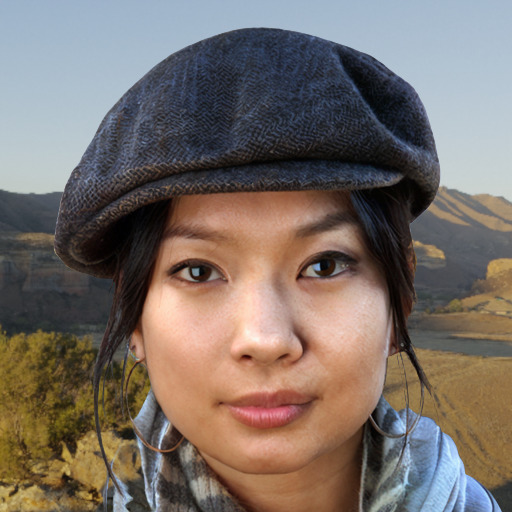} &
        \includegraphics[height=\specialeffectheight]{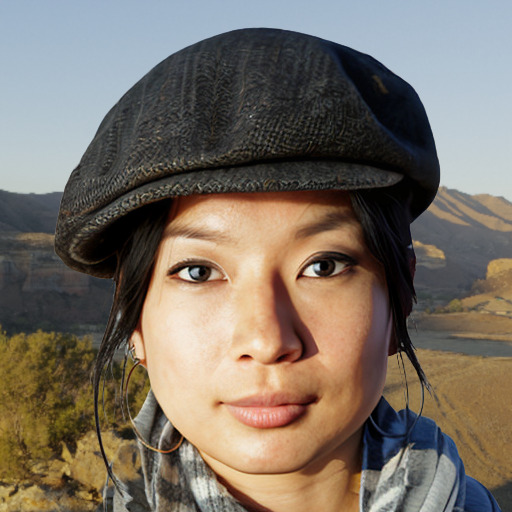} \\

        \includegraphics[height=\specialeffectheight]{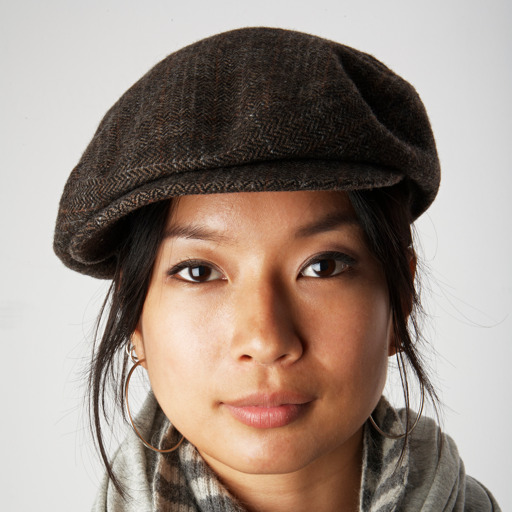} &
        \includegraphics[height=\specialeffectheight]{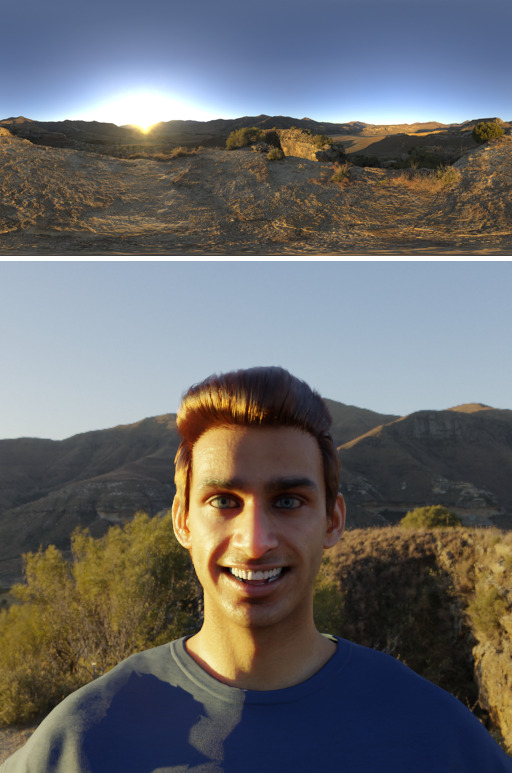} &
        \includegraphics[height=\specialeffectheight]{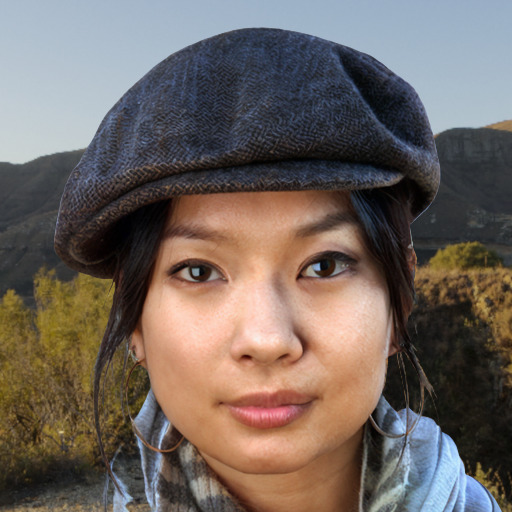} &
        \includegraphics[height=\specialeffectheight]{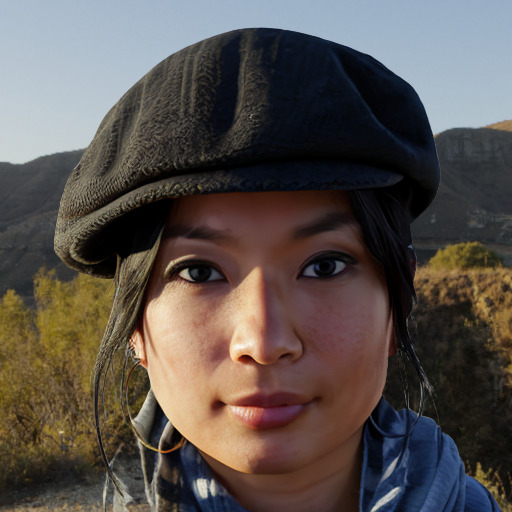} \\

        \includegraphics[height=\specialeffectheight]{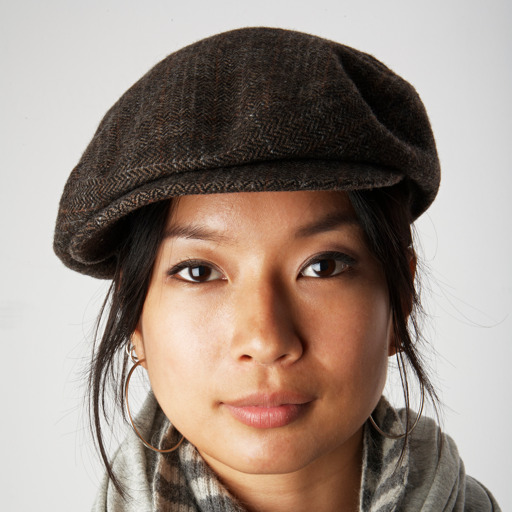} &
        \includegraphics[height=\specialeffectheight]{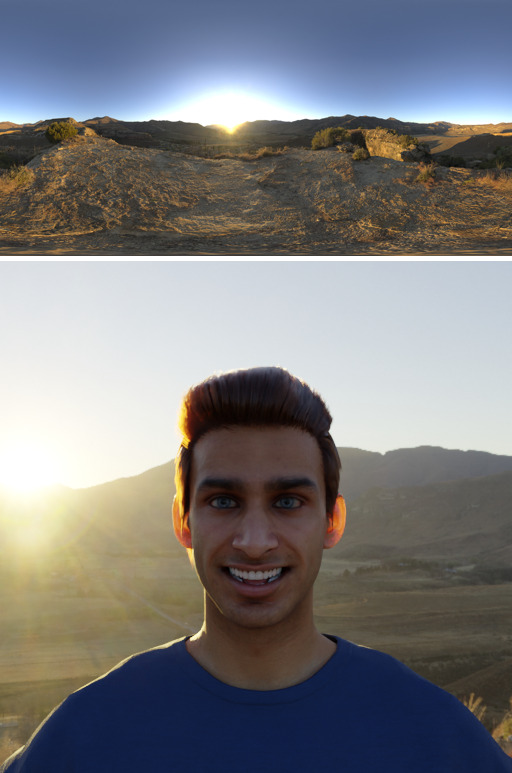} &
        \includegraphics[height=\specialeffectheight]{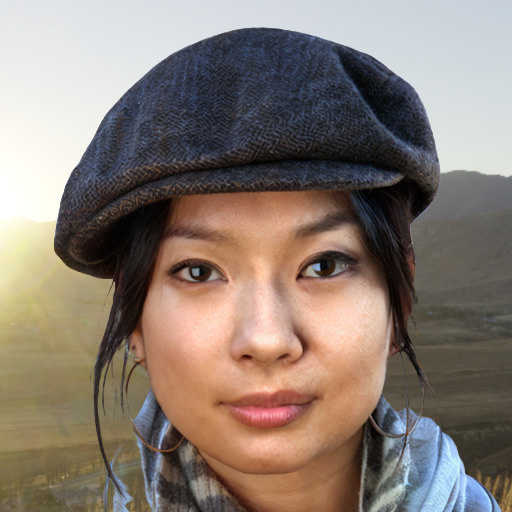} &
        \includegraphics[height=\specialeffectheight]{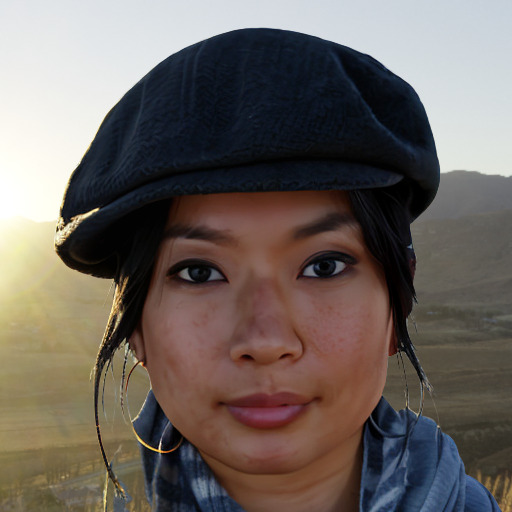} \\
        
        & & Without Finetune (IC-Light) & With Finetune (Ours) \\
        
    \end{tabular}
    \vspace{-2mm}
    \caption{We show the input portrait, the environment map used to relight and a reference synthetic data rendering from Blender (left) and results from our method and ablations (right). We demonstrate the impact of fine-tuning with our synthetic dataset. The base model, IC-Light, without this fine-tuning, is unable to relight images using an environment map.}
    \label{fig:with_and_without_finetune}
\end{figure*}

\begin{figure*}[!htbp]
    \centering
    \begin{tabular}{@{\hskip 0.0mm}c@{\hskip 0.1mm}c@{\hskip 0.1mm}c@{\hskip 0.1mm}c@{\hskip 0.0mm}}

        \includegraphics[height=\specialeffectheight]{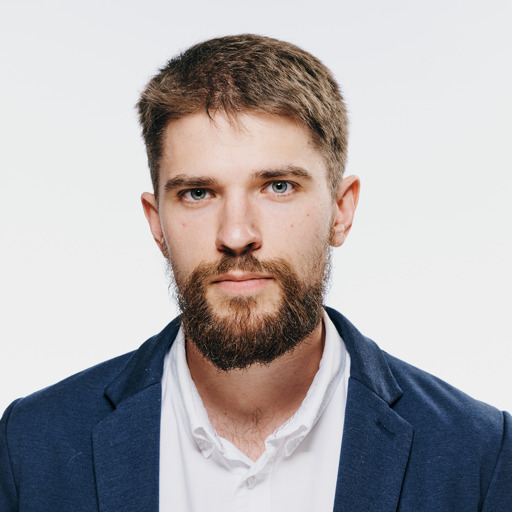} &
        \includegraphics[height=\specialeffectheight]{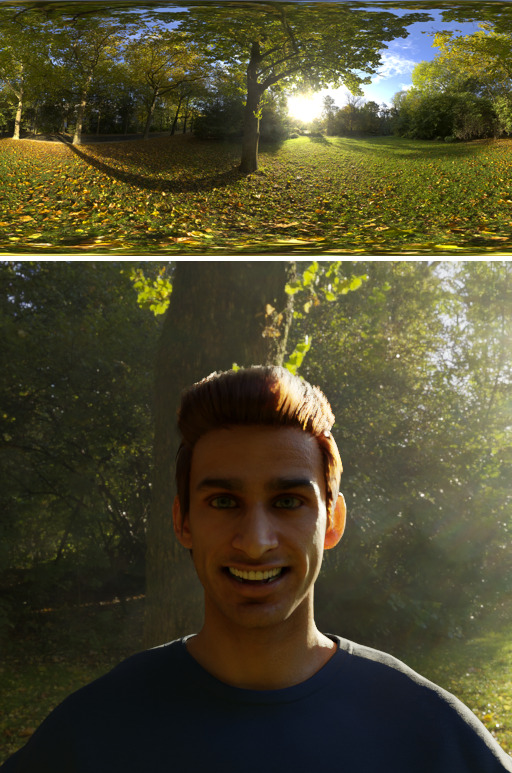} &
        \includegraphics[height=\specialeffectheight]{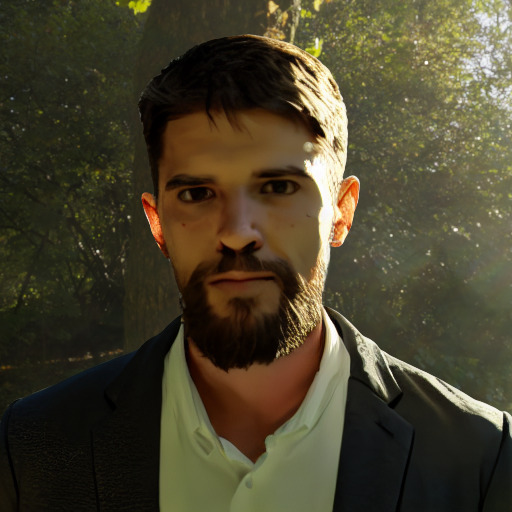} &
        \includegraphics[height=\specialeffectheight]{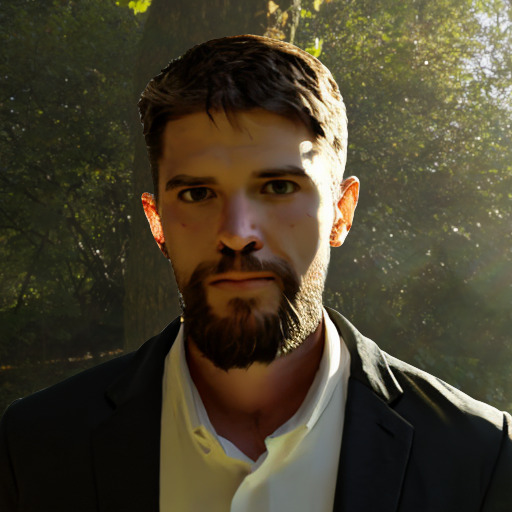} \\

        \includegraphics[height=\specialeffectheight]{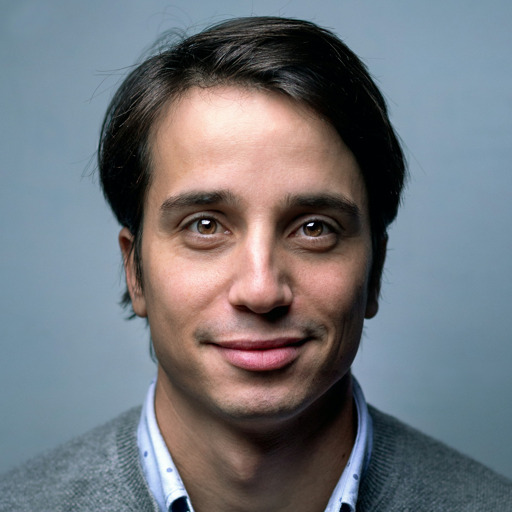} &
        \includegraphics[height=\specialeffectheight]{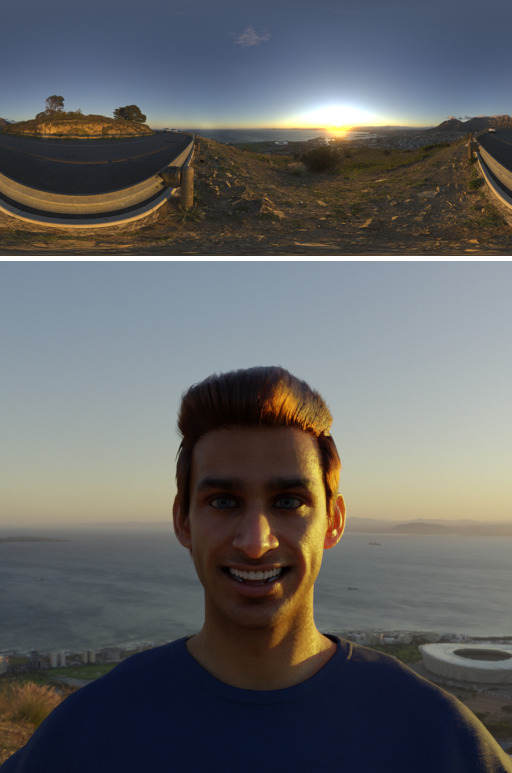} &
        \includegraphics[height=\specialeffectheight]{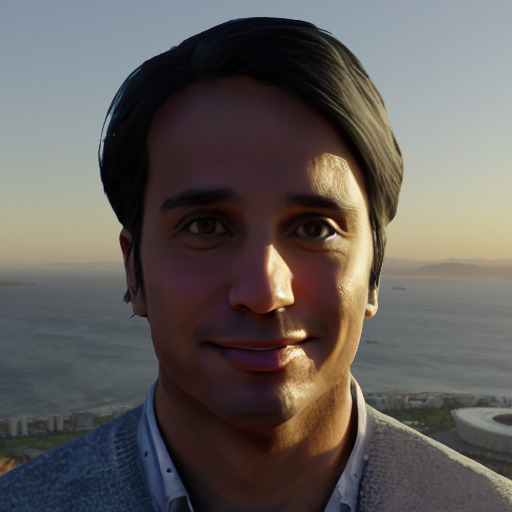} &
        \includegraphics[height=\specialeffectheight]{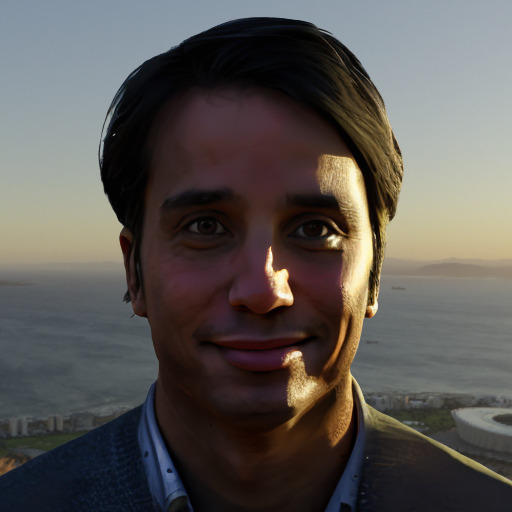} \\
        
        & & \makecell{Finetuning \\ with IC-Light \\ initialization} & \makecell{Finetuning \\ with SD 1.5 \\ initialization} \\
        
    \end{tabular}
    \vspace{-2mm}
    \caption{We show the input portrait, the environment map used for relighting, and a reference synthetic data rendering from Blender (left). On the right, we present results with IC-Light and SD 1.5 initialization for finetuning on our synthetic dataset. We note that while IC-Light initialization yields slightly better performance on our Light Stage Test set, both are comparable in terms of visual quality and achieve realistic lighting effects such as shadows and subsurface scattering.}
    \label{fig:with_and_without_iclight_ckpt}
\end{figure*}

\begin{figure*}[!htbp]
    \centering

    \begin{tabular}{@{\hskip 0mm}c@{\hskip 2mm}c@{\hskip 0.5mm}c@{\hskip 0.5mm}c@{\hskip 0.5mm}c@{\hskip 0.5mm}c@{\hskip 0.5mm}c@{\hskip 0.5mm}c}
        & & DiLightNet & IC-Light & Neural Gaffer & Total Relighting & SwitchLight & Ours \\

        \includegraphics[height=2.2cm]{figures/comparison_figure_all/row_02/col_00.jpg} &
        \includegraphics[height=2.2cm]{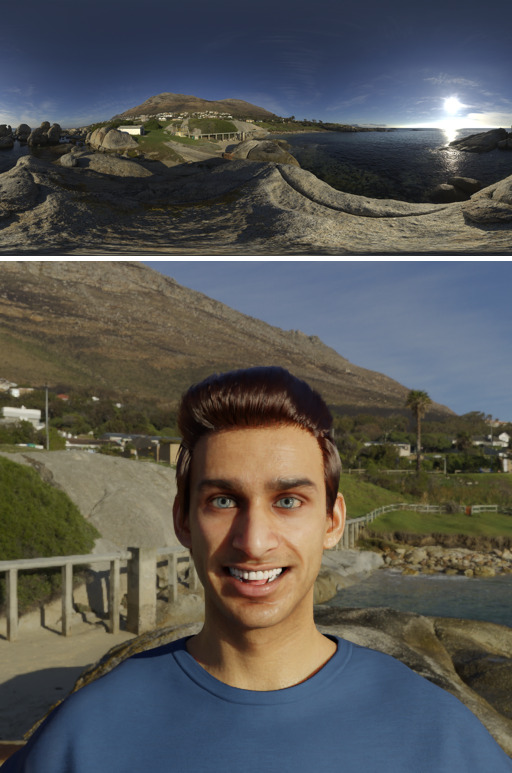} &
        \includegraphics[height=2.2cm]{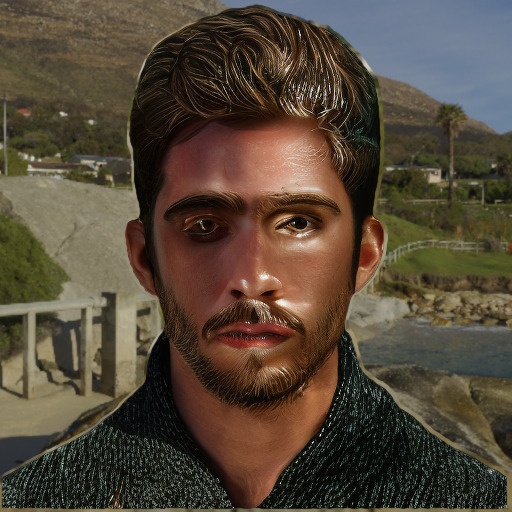} &
        \includegraphics[height=2.2cm]{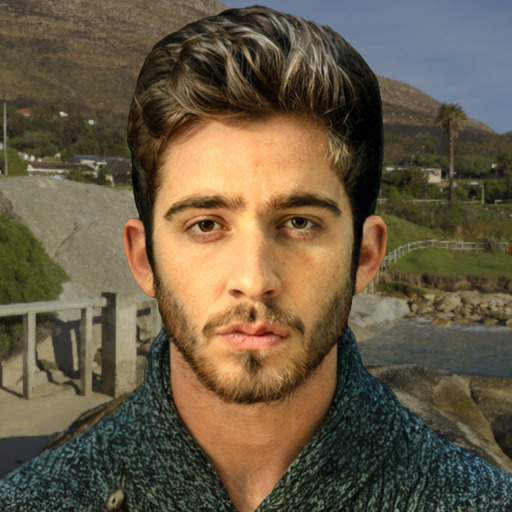} &
        \includegraphics[height=2.2cm]{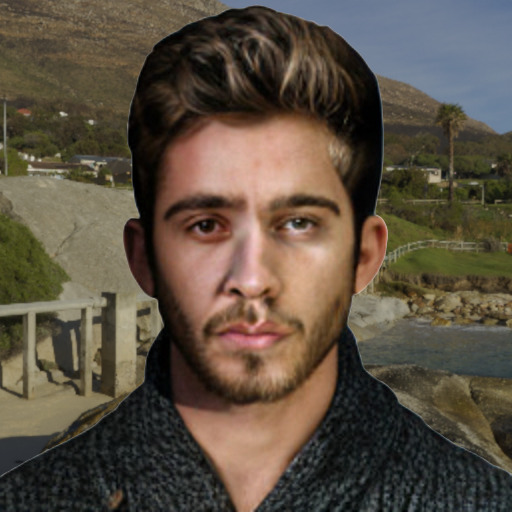} &
        \includegraphics[height=2.2cm]{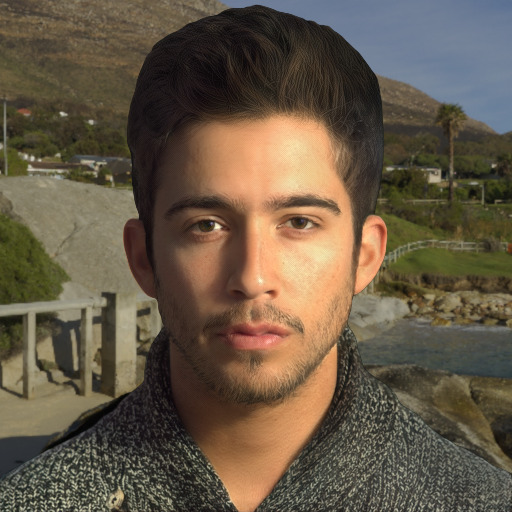} &
        \includegraphics[height=2.2cm]{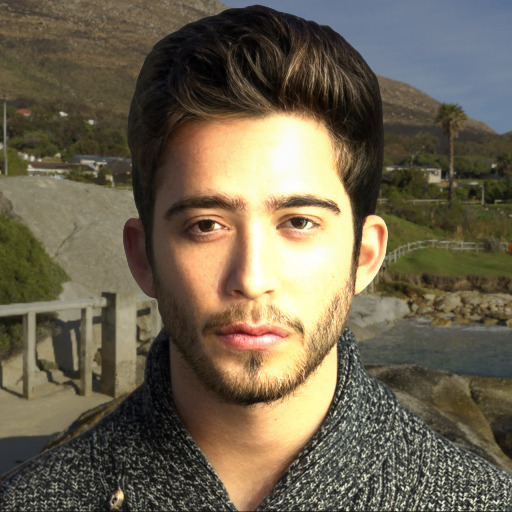} &
        \includegraphics[height=2.2cm]{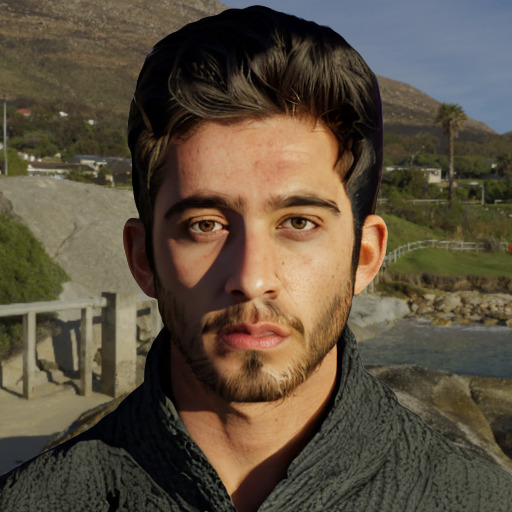} \\

        \includegraphics[height=2.2cm]{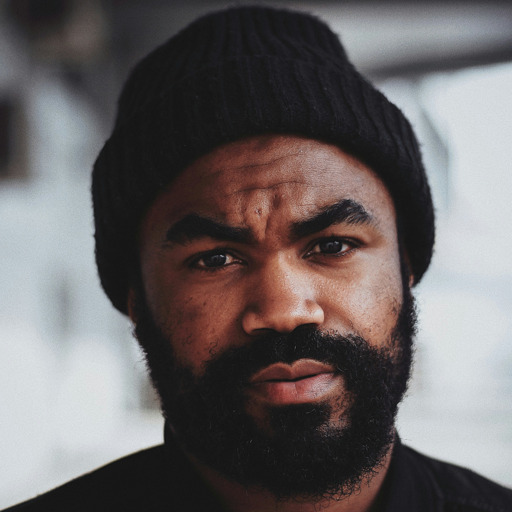} &
        \includegraphics[height=2.2cm]{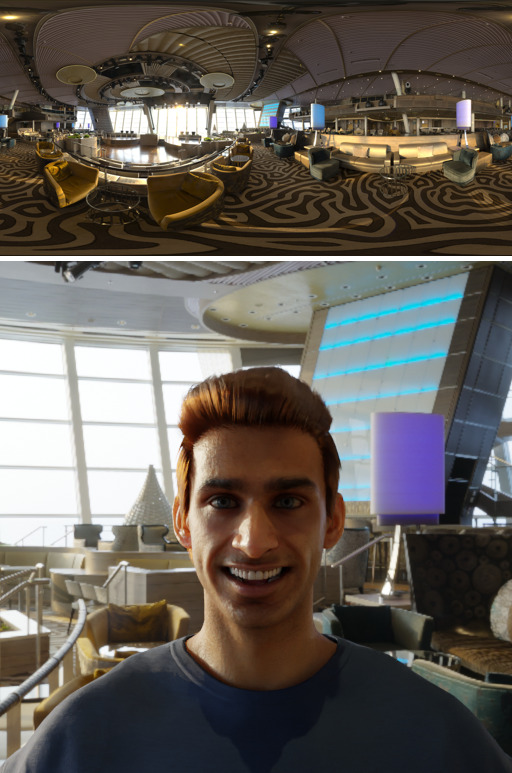} &
        \includegraphics[height=2.2cm]{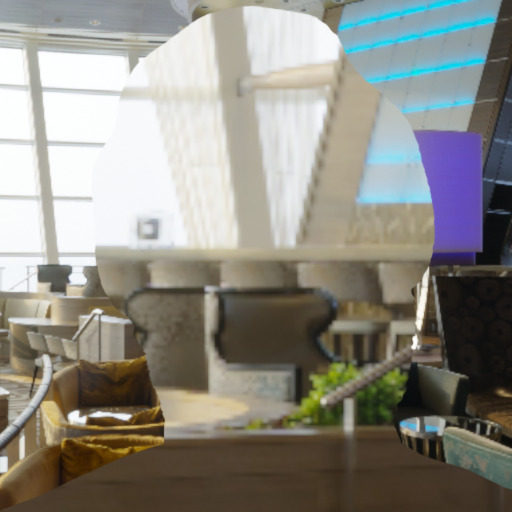} &
        \includegraphics[height=2.2cm]{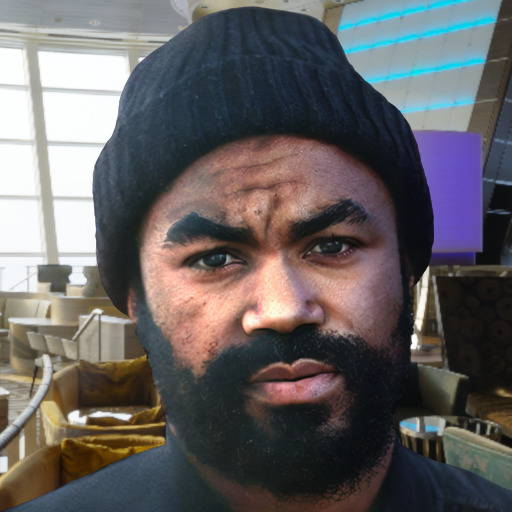} &
        \includegraphics[height=2.2cm]{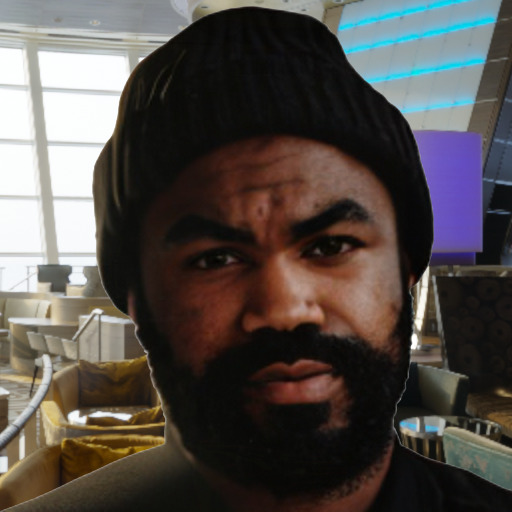} &
        \includegraphics[height=2.2cm]{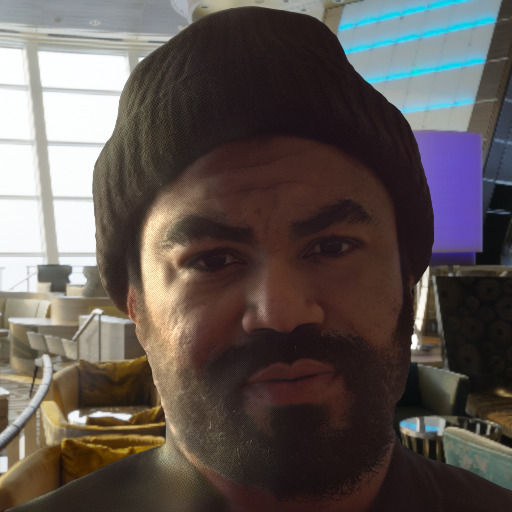} &
        \includegraphics[height=2.2cm]{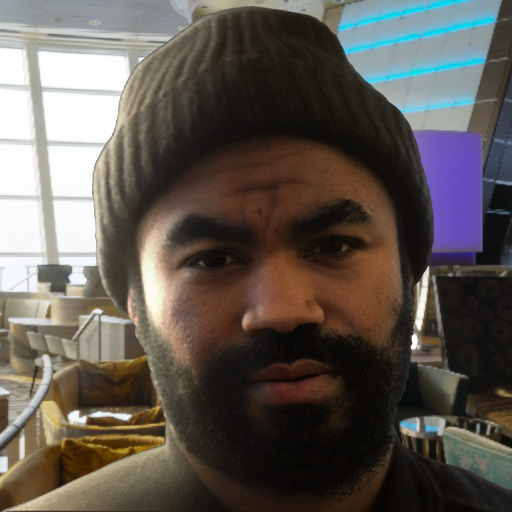} &
        \includegraphics[height=2.2cm]{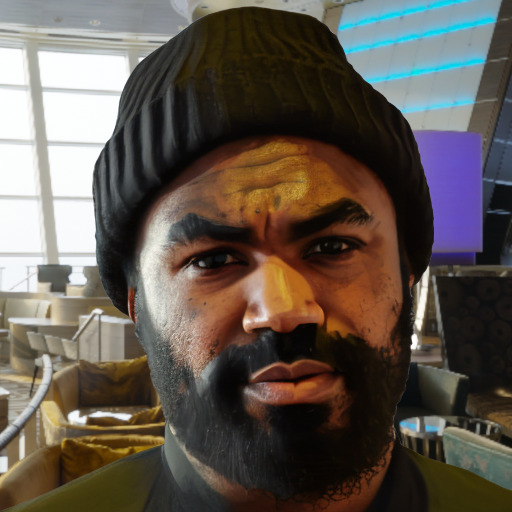} \\

        \includegraphics[height=2.2cm]{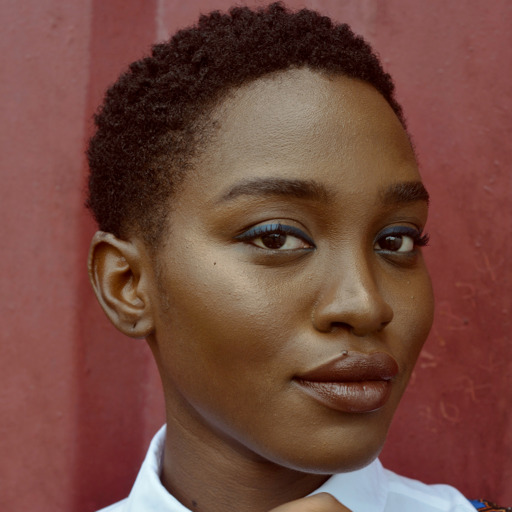} &
        \includegraphics[height=2.2cm]{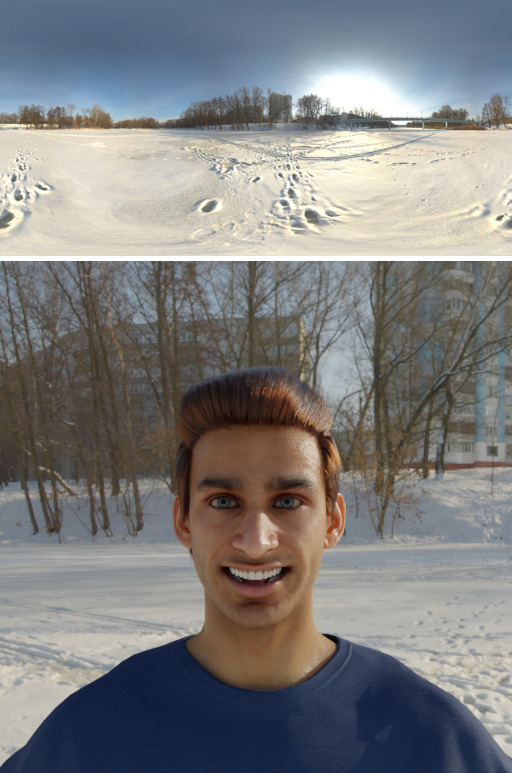} &
        \includegraphics[height=2.2cm]{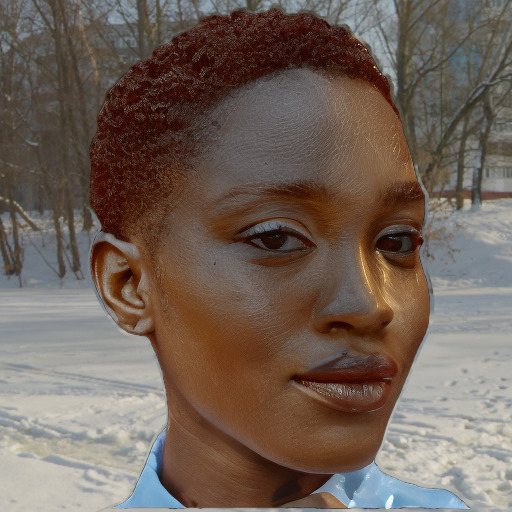} &
        \includegraphics[height=2.2cm]{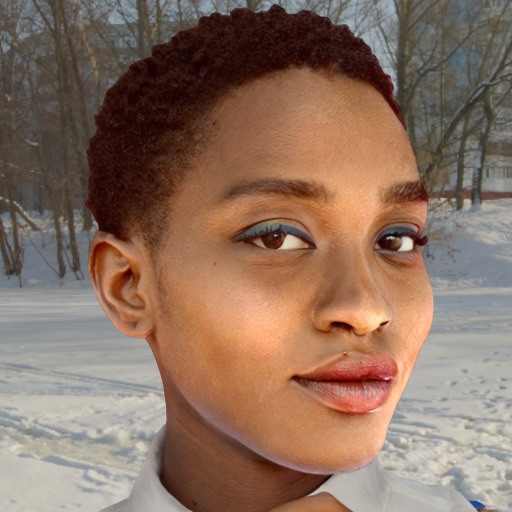} &
        \includegraphics[height=2.2cm]{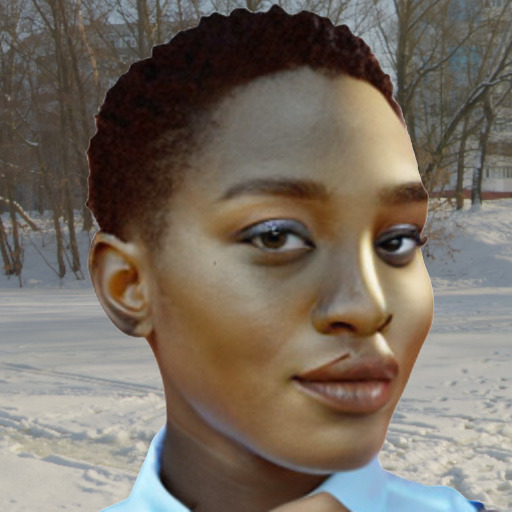} &
        \includegraphics[height=2.2cm]{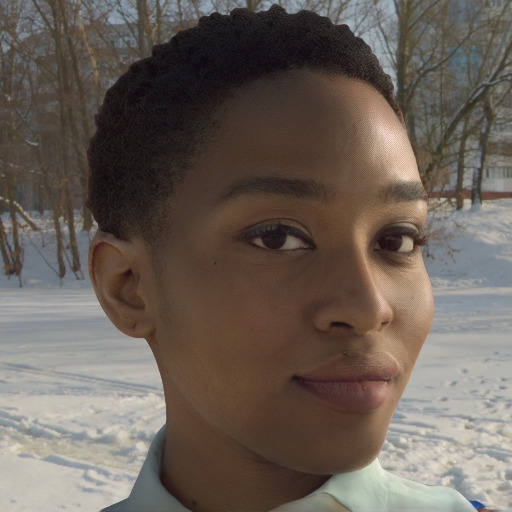} &
        \includegraphics[height=2.2cm]{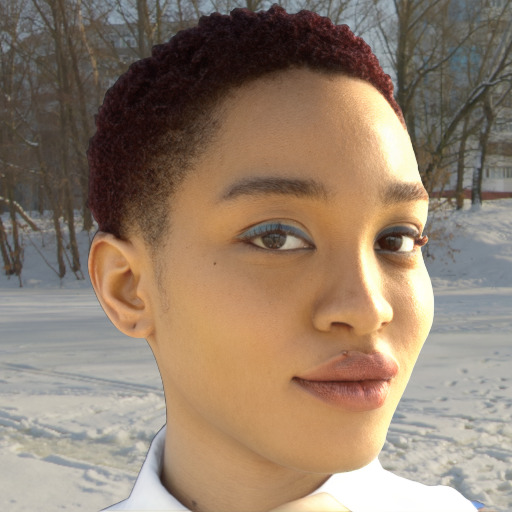} &
        \includegraphics[height=2.2cm]{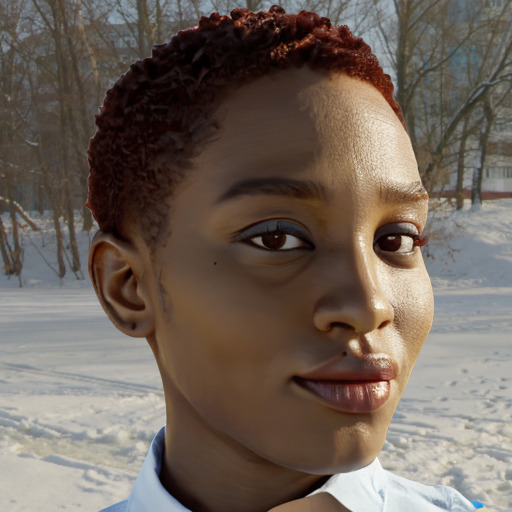} \\

        \includegraphics[height=2.2cm]{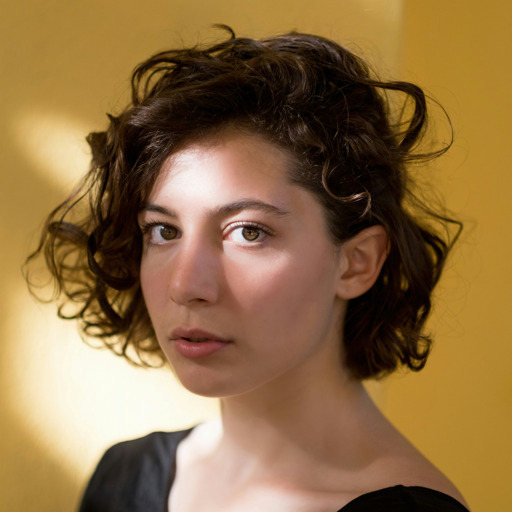} &
        \includegraphics[height=2.2cm]{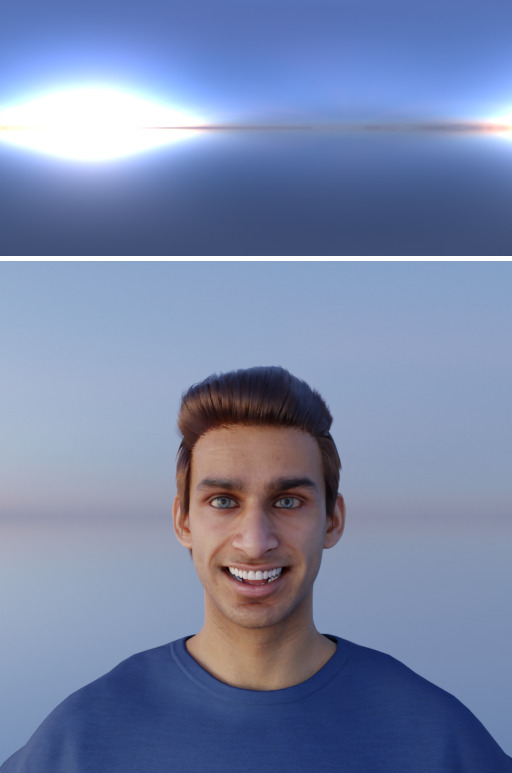} &
        \includegraphics[height=2.2cm]{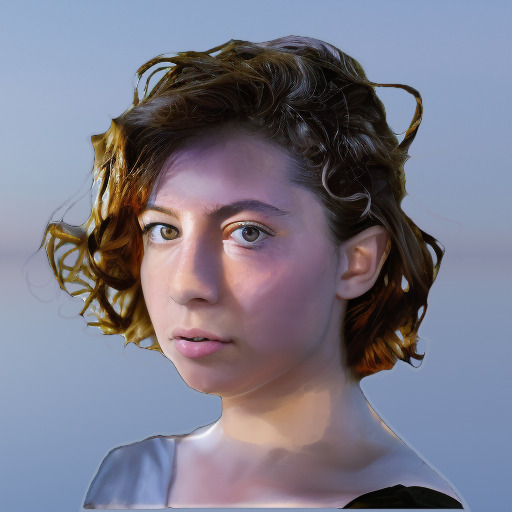} &
        \includegraphics[height=2.2cm]{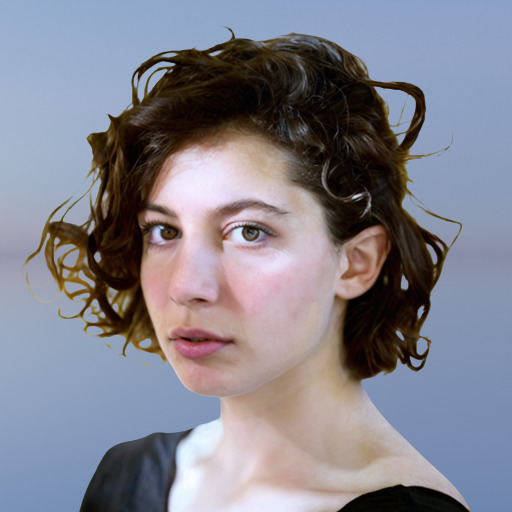} &
        \includegraphics[height=2.2cm]{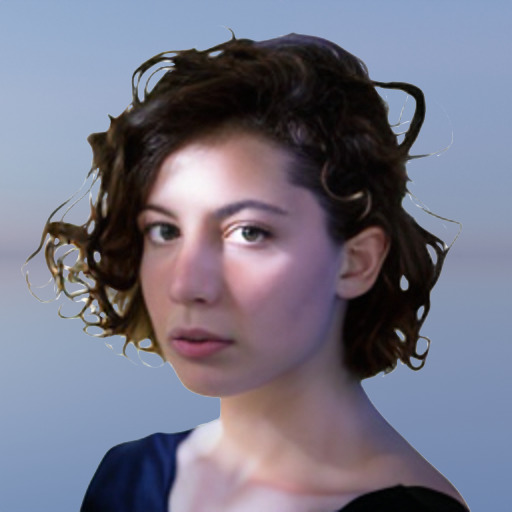} &
        \includegraphics[height=2.2cm]{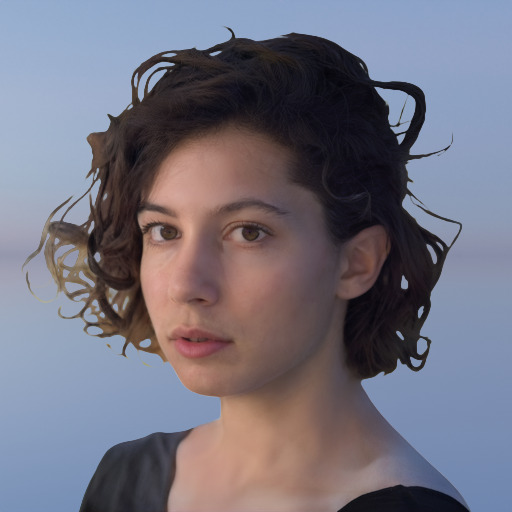} &
        \includegraphics[height=2.2cm]{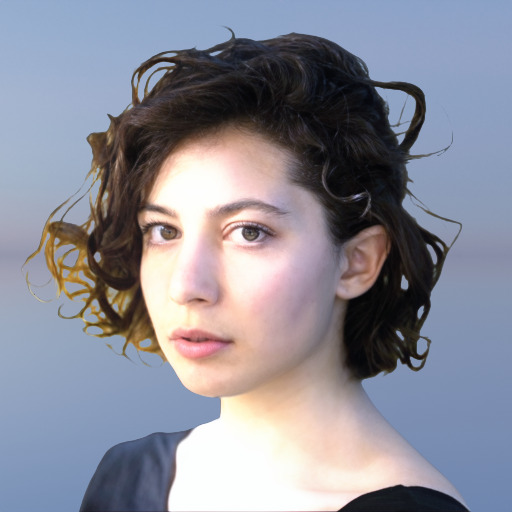} &
        \includegraphics[height=2.2cm]{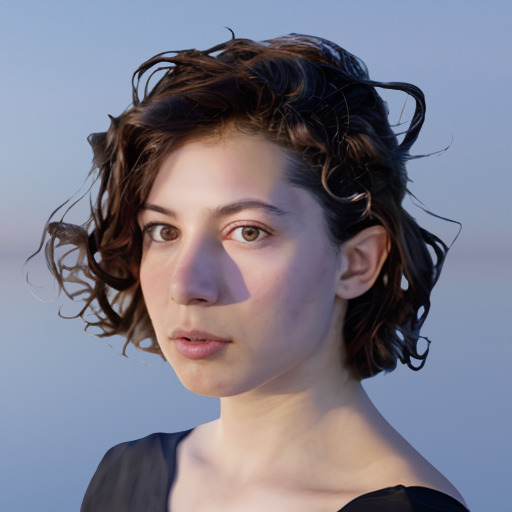} \\

        \includegraphics[height=2.2cm]{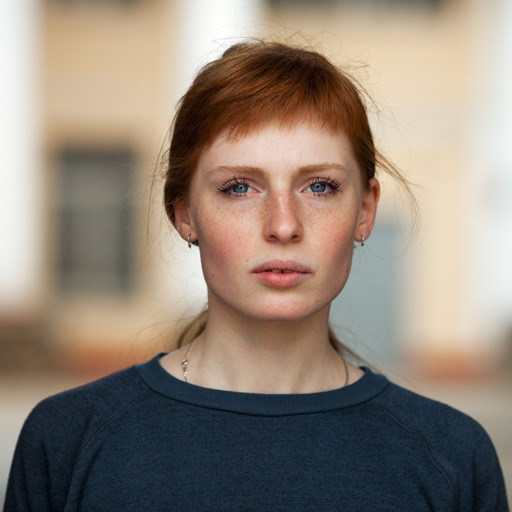} &
        \includegraphics[height=2.2cm]{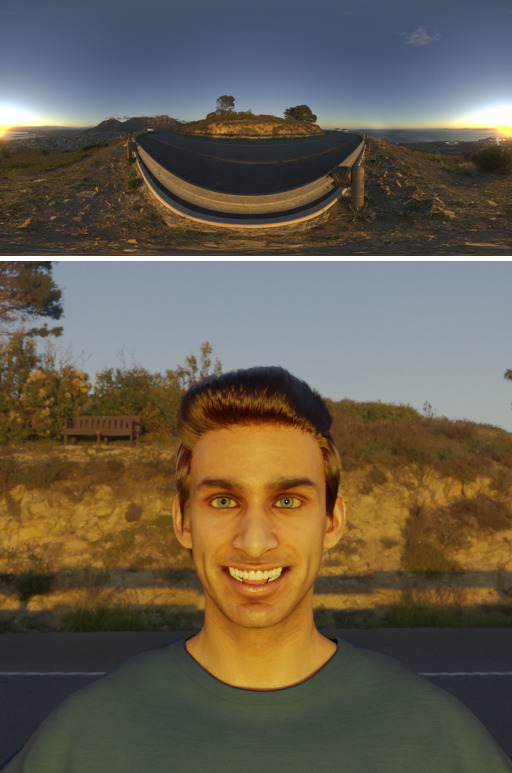} &
        \includegraphics[height=2.2cm]{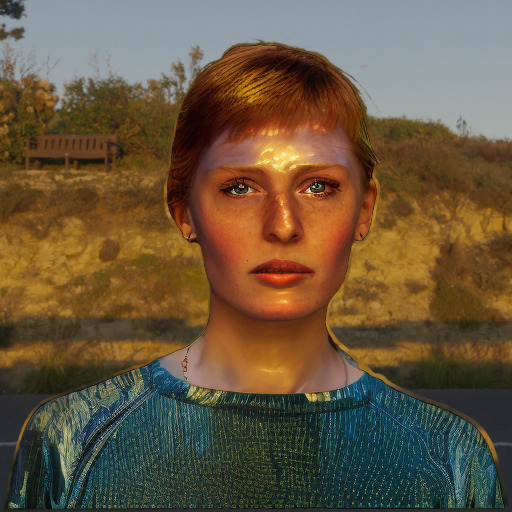} &
        \includegraphics[height=2.2cm]{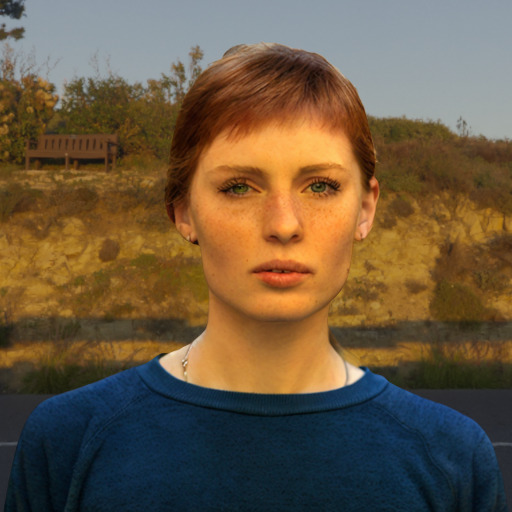} &
        \includegraphics[height=2.2cm]{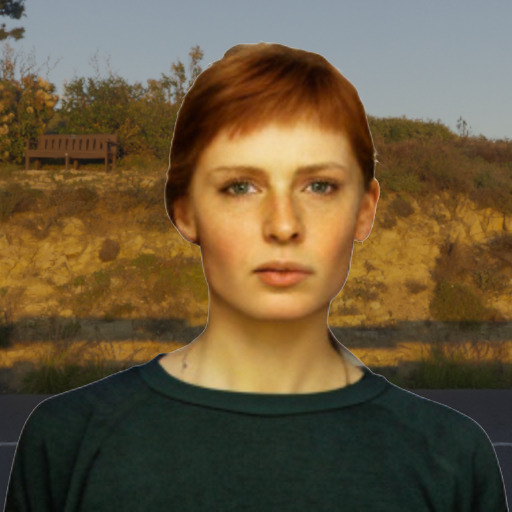} &
        \includegraphics[height=2.2cm]{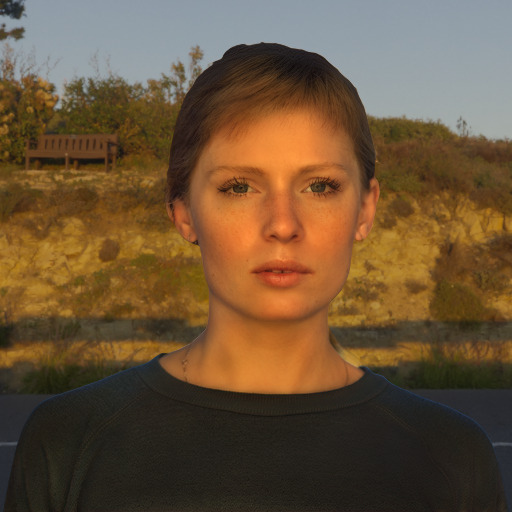} &
        \includegraphics[height=2.2cm]{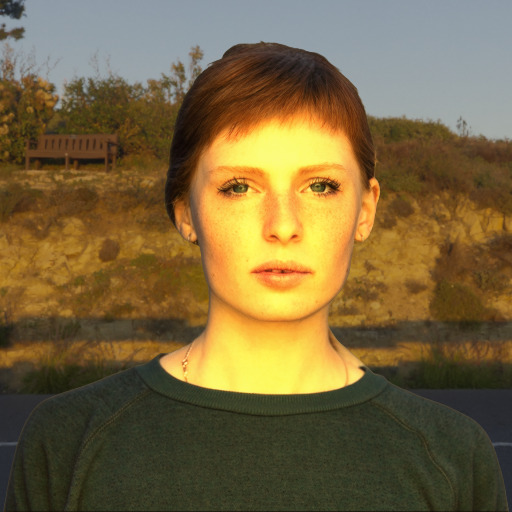} &
        \includegraphics[height=2.2cm]{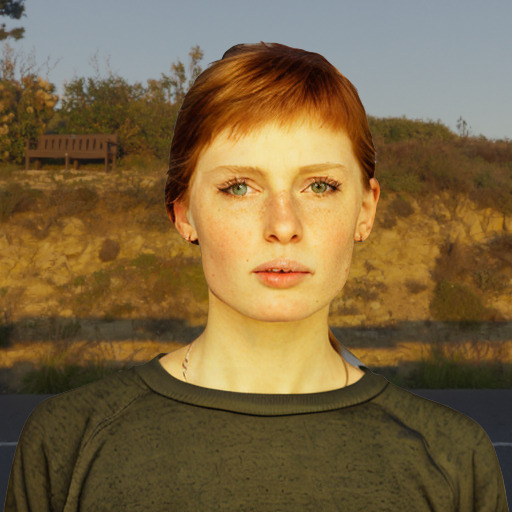} \\

    \end{tabular}
    \vspace{-2mm}
    \caption{We show additional comparisons against baselines, illustrating, that unlike baselines, our method produces accurate lighting, that matches given reference, while preserving identity and maintaining high visual quality.}
    \label{fig:comparison_all_extra}
\end{figure*}

\begin{figure*}[!htbp]
    \centering
    \begin{tabular}{@{\hskip 0.0mm}c@{\hskip 0.1mm}c@{\hskip 0.1mm}c@{\hskip 0.1mm}c@{\hskip 0.0mm}}

        \includegraphics[height=\specialeffectheight]{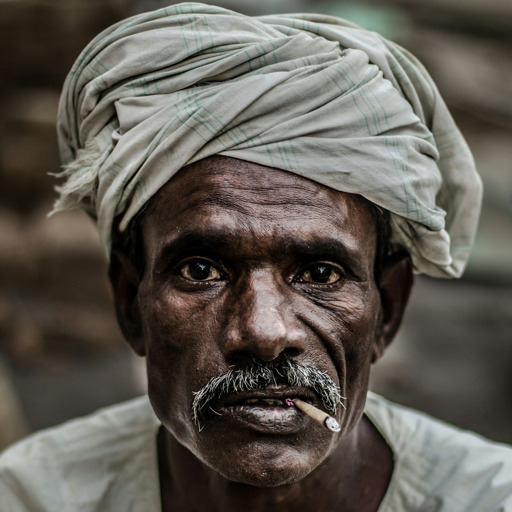} &
        \includegraphics[height=\specialeffectheight]{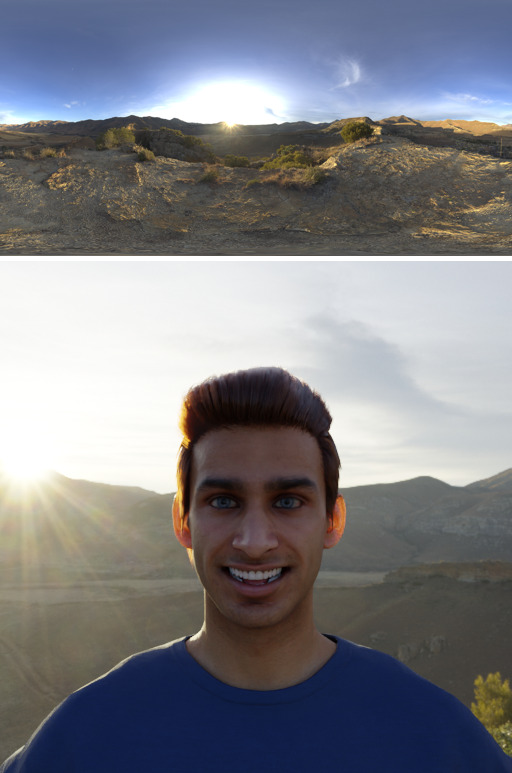} &
        \includegraphics[height=\specialeffectheight]{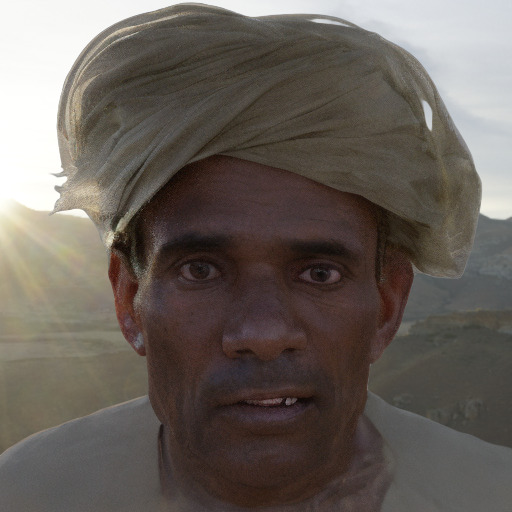} &
        \includegraphics[height=\specialeffectheight]{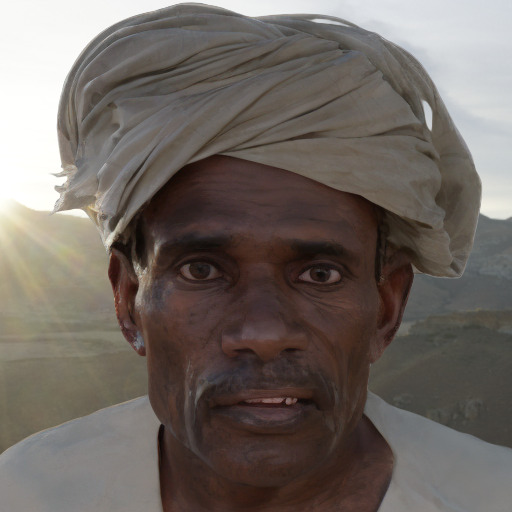} \\

        & & Base & Base + Multi-Task \\

        & & 
        \includegraphics[height=\specialeffectheight]{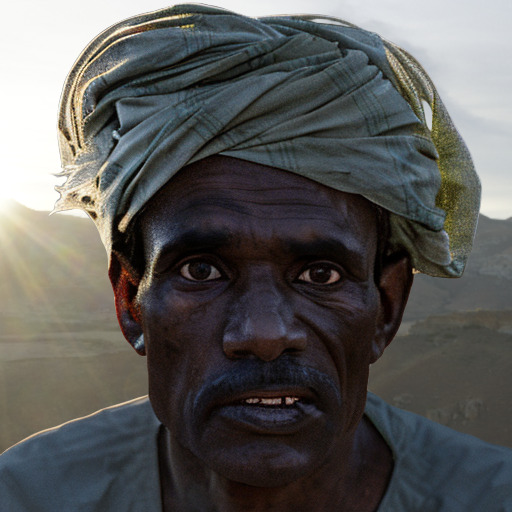} &
        \includegraphics[height=\specialeffectheight]{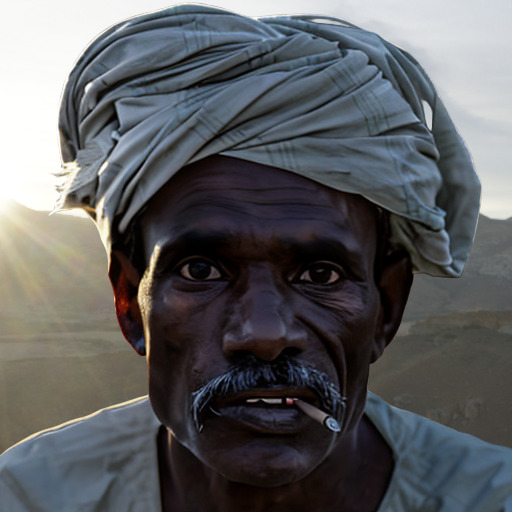} \\ 

        & & \makecell{Base + \\ Inference-time \\ Adaptation} & Ours \\

        \vspace{2mm}
        \includegraphics[height=\specialeffectheight]{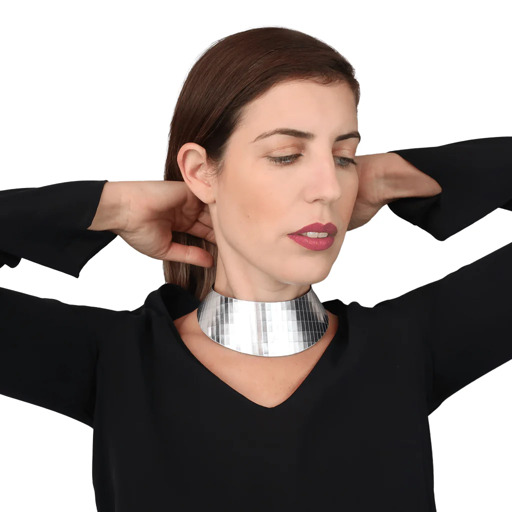} &
        \includegraphics[height=\specialeffectheight]{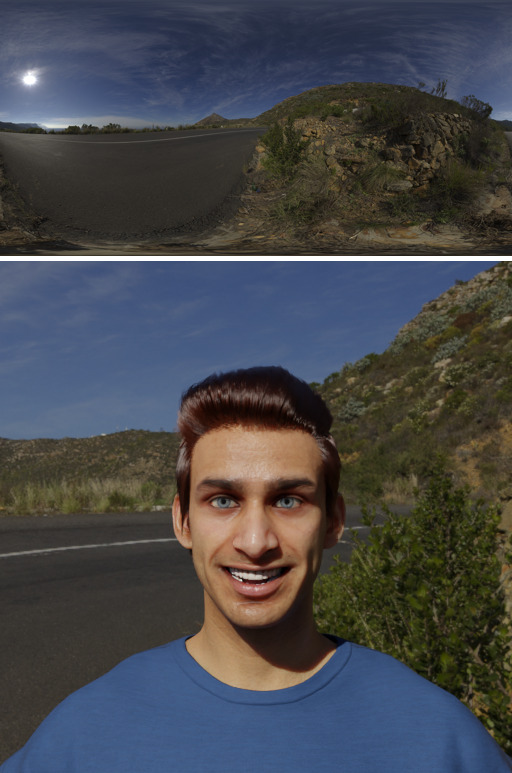} &
        \includegraphics[height=\specialeffectheight]{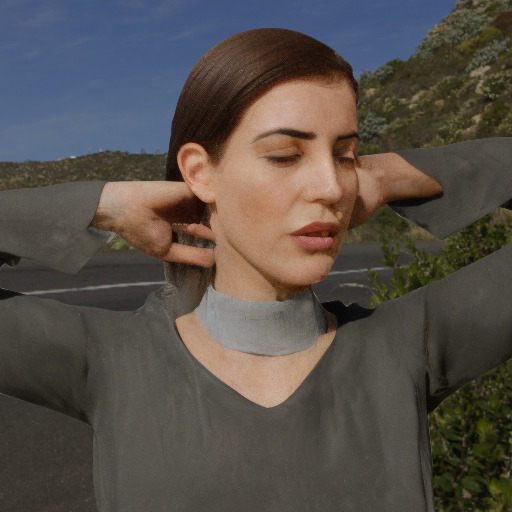} &
        \includegraphics[height=\specialeffectheight]{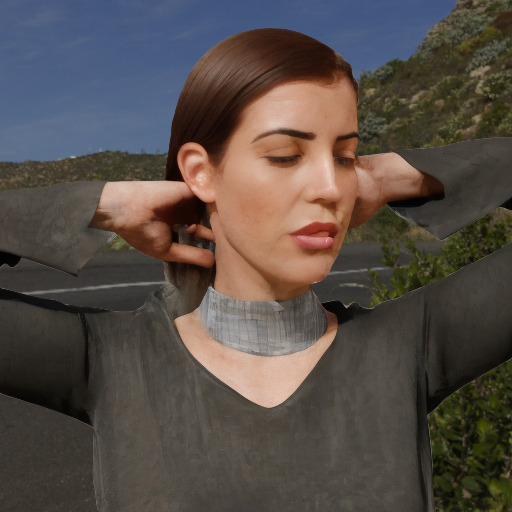} \\

        & & Base & Base + Multi-Task \\

        & & 
        \includegraphics[height=\specialeffectheight]{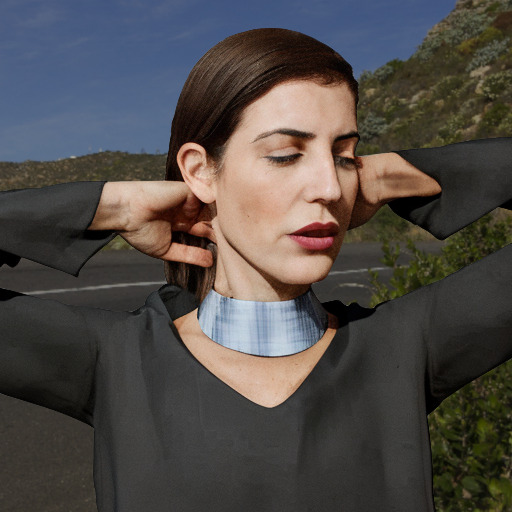} &
        \includegraphics[height=\specialeffectheight]{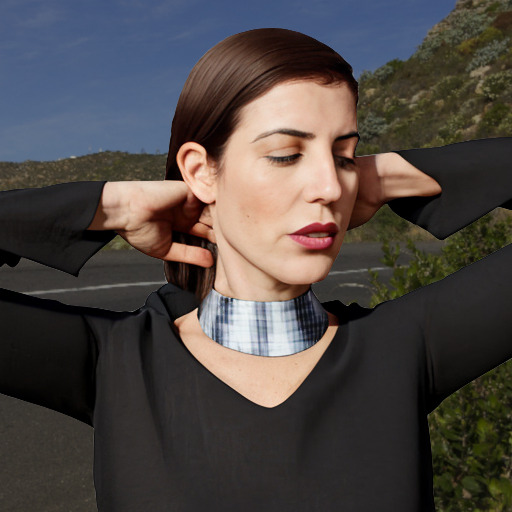} \\ 

        & & \makecell{Base + \\ Inference-time \\ Adaptation} & Ours \\

    \end{tabular}
    \vspace{-2mm}
    \caption{We show the input portrait, the environment map used to relight and a reference synthetic data rendering from Blender (left) and results from our method and ablations (right). Examples show the contributions of each component in our proposed method. The \textit{Base} model struggles with identity preservation and detail reproduction. \textit{Base + Multitask} and \textit{Base + Inference-Time Adaptation} improve detail recovery but fail to replicate complex features like accessories and textures. Our method successfully preserves identity and reproduces intricate details, such as the cigarette (top) and specularity of the necklace (bottom).}
    \label{fig:ablations_extra}
\end{figure*}

\vspace{-4mm}
\paragraph{Ablations} \cref{fig:ablations_extra} showcases additional examples from our ablation study, illustrating the contribution of each component to the final qualitative results. The \textit{Base} model struggles with identity preservation and fails to capture key details present in the input portrait. Adding either \textit{Base + Multi-Task} or \textit{Base + Inference Adaptation} improves detail recovery but remains insufficient for reproducing complex accessories, materials, and textures. For example, in \cref{fig:ablations_extra}, the cigarette in the input portrait (top) and the specularity of the choker necklace or the accurate dress color (bottom) are not faithfully replicated. In contrast, our method successfully addresses these challenges, achieving superior results.

\begin{figure*}[!htbp]
    \centering

    \begin{tabular}{@{\hskip 0mm}c@{\hskip 2mm}c@{\hskip 0.5mm}c@{\hskip 0.5mm}c@{\hskip 0.5mm}c}

        \includegraphics[height=3.65cm]{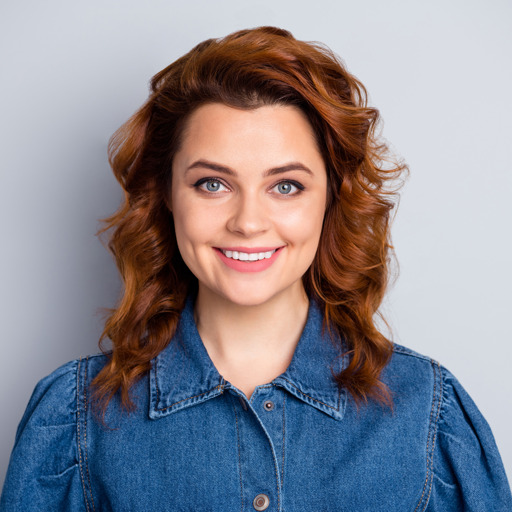} &
        \includegraphics[height=3.65cm]{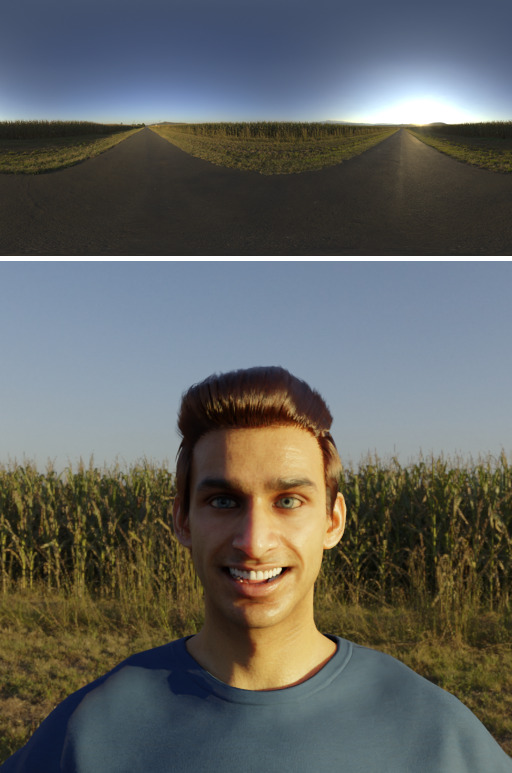} &
        \includegraphics[height=3.65cm]{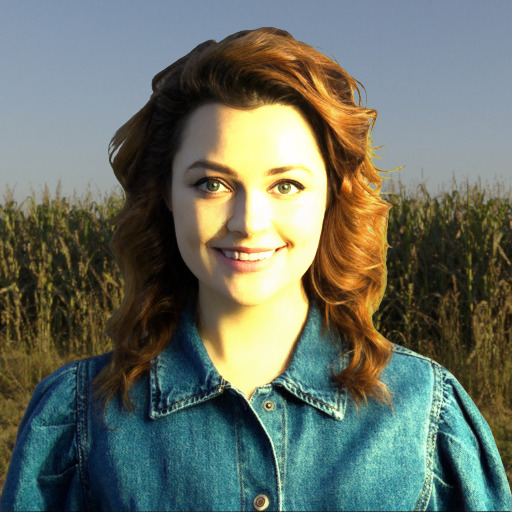} &
        \includegraphics[height=3.65cm]{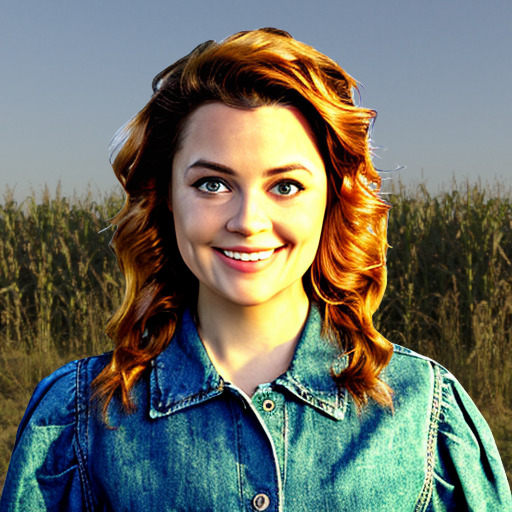} &
        \includegraphics[height=3.65cm]{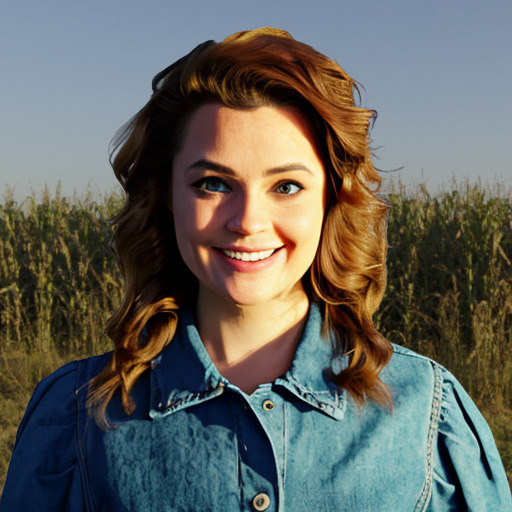} \\

        \includegraphics[height=3.65cm]{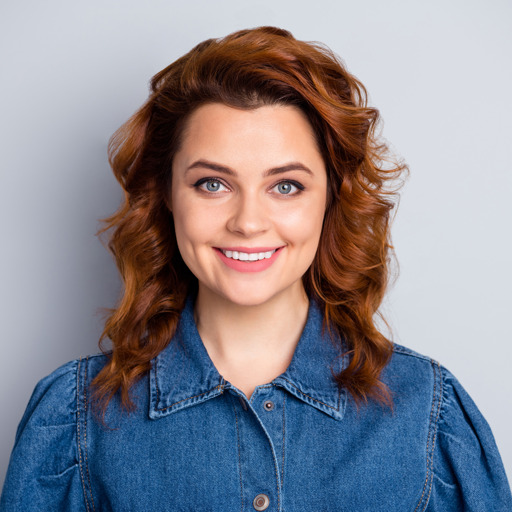} &
        \includegraphics[height=3.65cm]{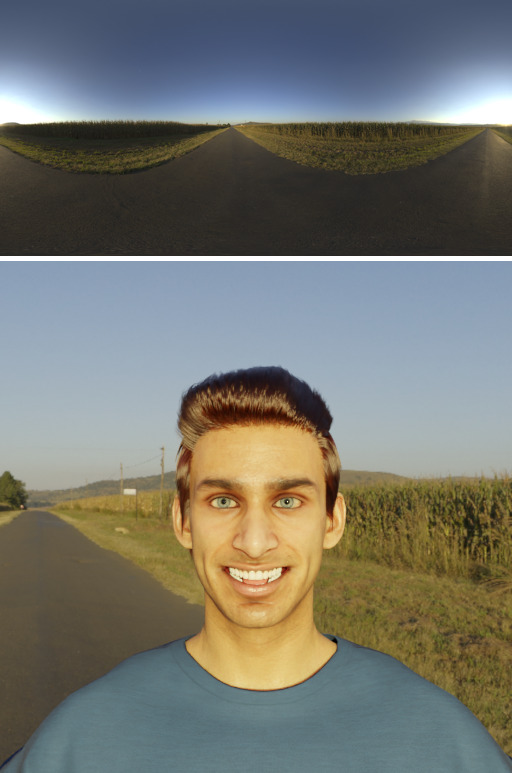} &
        \includegraphics[height=3.65cm]{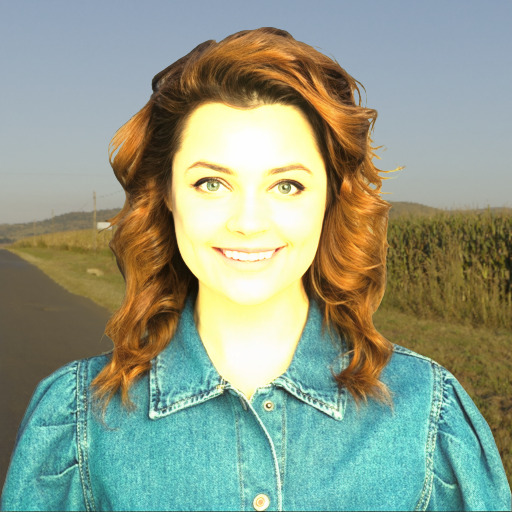} &
        \includegraphics[height=3.65cm]{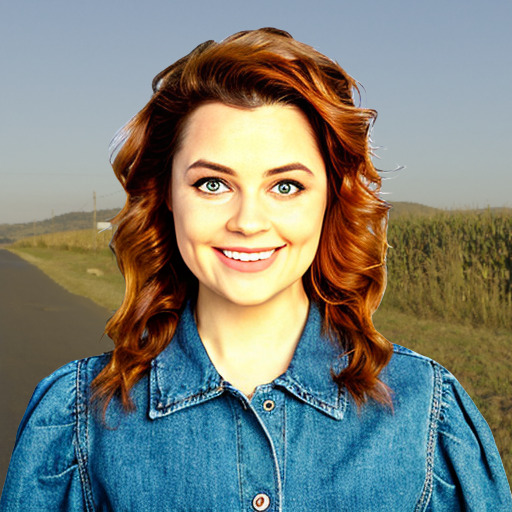} &
        \includegraphics[height=3.65cm]{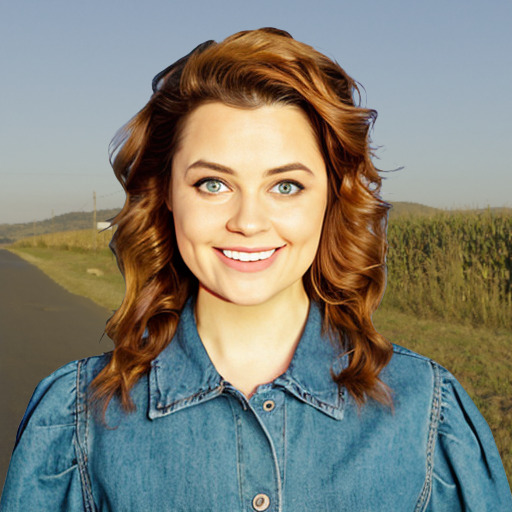} \\

        & & SwitchLight & Ours + Light Stage & Ours \\

    \end{tabular}
    \vspace{-2mm}
    \caption{Overexposure issues due to Light Stage data. SwitchLight, trained purely on Light Stage data, suffers from severe overexposure and unnatural skin tones. \textit{Ours + Light Stage} reduces this issue but retains some artifacts. \textit{Ours}, trained on synthetic data alone, avoids these problems entirely.}
    \label{fig:light_stage_v_switchlight}
\end{figure*}

We train an additional model, \textit{Ours + Light Stage}, where Light Stage-rendered data is combined with the synthetic dataset for relighting. The Light Stage data is same as in \cite{ren2024relightful}, and consists of roughly 6000 light stage captures, rendered under 100 environment maps. \cref{fig:light_stage_v_switchlight} illustrates a spectrum of overexposure issues. Models trained purely on Light Stage data, such as SwitchLight, often suffer from severe overexposure, resulting in unnatural yellowish skin tones. \textit{Ours + Light Stage} reduces this issue due to the inclusion of physically-based rendered synthetic data, though some overexposure persists. In contrast, our method trained exclusively on physically-based rendered synthetic data avoids this problem entirely, producing natural and balanced skin tones.

\begin{table*}[!htbp]
\centering
\begin{tabularx}{\textwidth}{l|*{4}{>{\centering\arraybackslash}X}|*{4}{>{\centering\arraybackslash}X}}
\toprule
& \multicolumn{4}{c|}{Test Synthetic} & \multicolumn{4}{c}{Test Light Stage} \\
\midrule
Method & LPIPS$\downarrow$ & SSIM$\uparrow$ & PSNR$\uparrow$ &  FN$\downarrow$ & LPIPS$\downarrow$ & SSIM$\uparrow$ & PSNR$\uparrow$ & FN$\downarrow$ \\
\midrule

Ours (init SD 1.5)  & 0.061  & 0.945  & 30.002  & 0.143     & 0.177  & 0.808  & 19.317  & 0.188 \\
Ours (init IC-Light)         & \textbf{0.057}  &\textbf{0.948}  & \textbf{30.268}  &\textbf{0.125}     & \textbf{0.165}  & \textbf{0.813}  & \textbf{19.698}  & \textbf{0.173}   \\

\bottomrule
\end{tabularx}
\caption{\textit{Ablating initial checkpoint}: We evaluate our method, initialized with IC-Light, against initialization with SD 1.5. All tables in both main paper and supplementary, including non-inference specific ablations, are generated with classifier-free guidance parameters, $\lambda_T = 2$, $\lambda_I = 3$. See main paper for detailed descriptions of them.}
\label{tab:ablation_checkpoint}
\end{table*}

\vspace{-4mm}
\paragraph{Comparison with Background-Conditioned Models}

\begin{figure*}[!h]
    \centering
    \begin{tabular}{@{}c@{}c@{\hskip 0.5mm}c@{\hskip 0.5mm}c@{\hskip 0.5mm}c@{}}
    \raisebox{16mm}{\makecell[c]{Reference}} &
    \includegraphics[width=3.6cm]{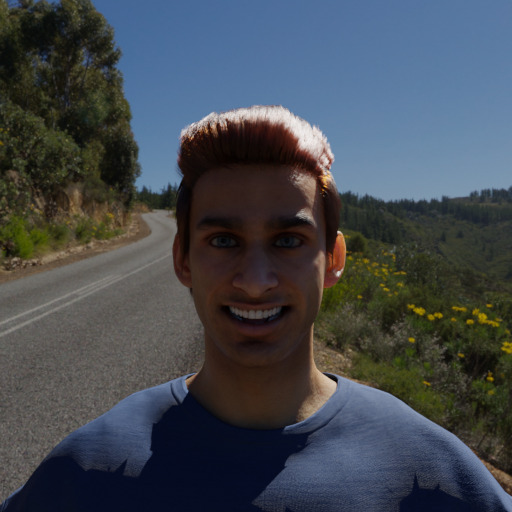} &
    \includegraphics[width=3.6cm]{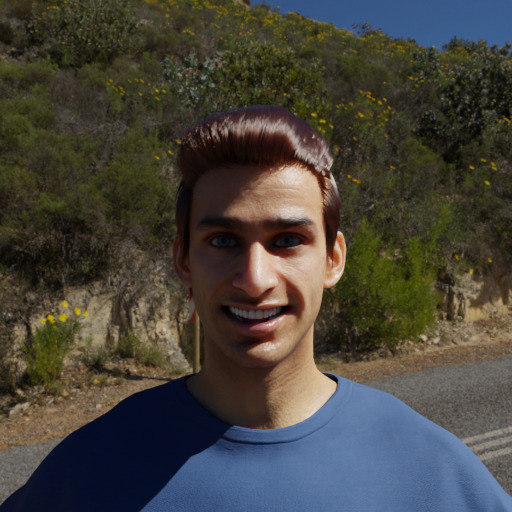} &
    \includegraphics[width=3.6cm]{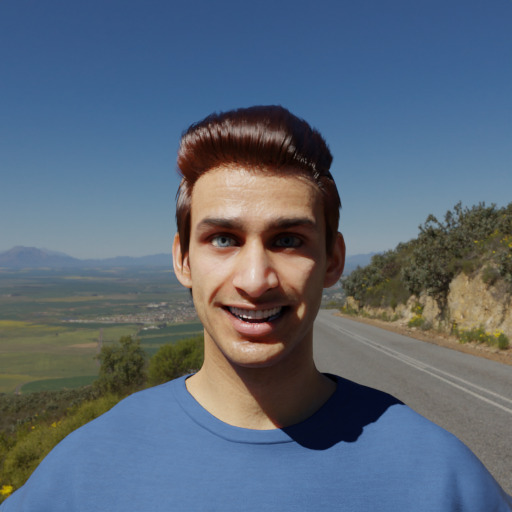} &
    \includegraphics[width=3.6cm]{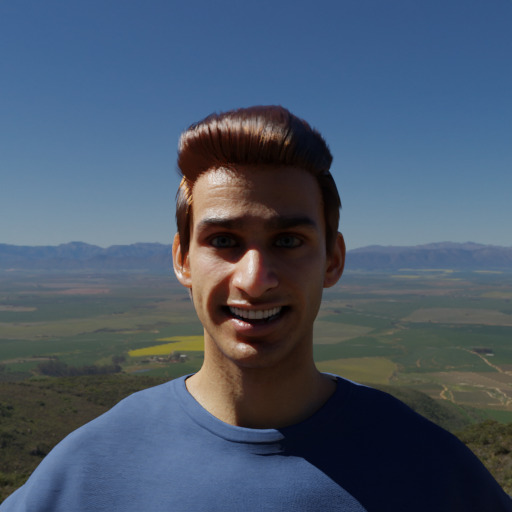} \\
    \raisebox{16mm}{\makecell[c]{SynthLight}} &
    \includegraphics[width=3.6cm]{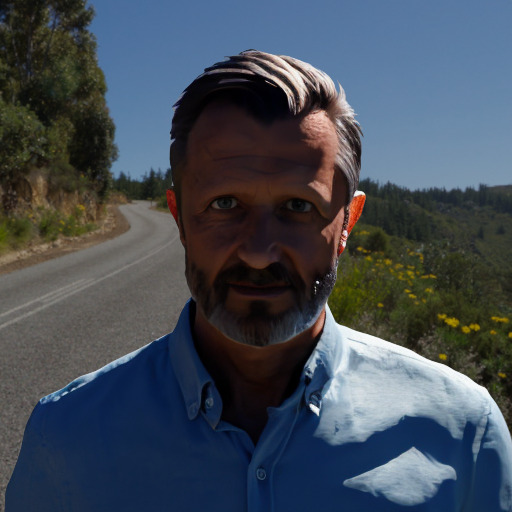} &
    \includegraphics[width=3.6cm]{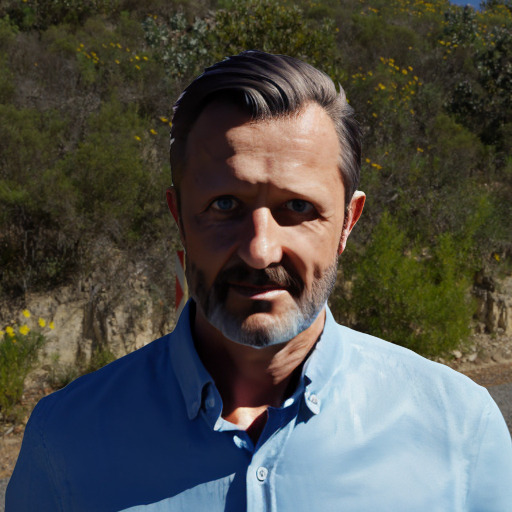} &
    \includegraphics[width=3.6cm]{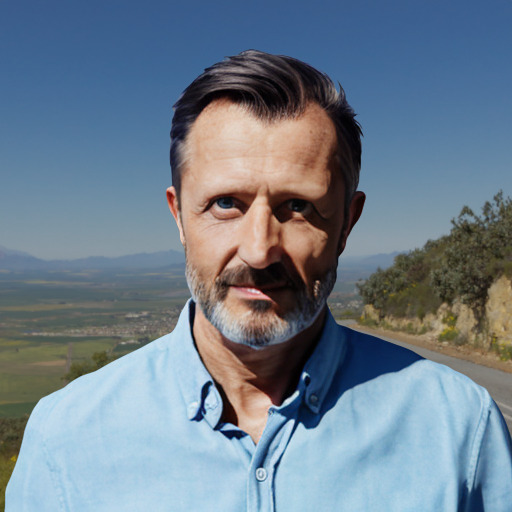} &
    \includegraphics[width=3.6cm]{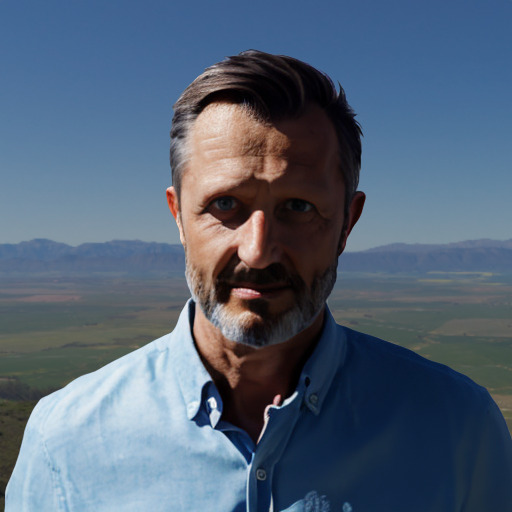} \\
    \raisebox{16mm}{\makecell[c]{Background \\ Conditioned \\ Model}} &
    \includegraphics[width=3.6cm]{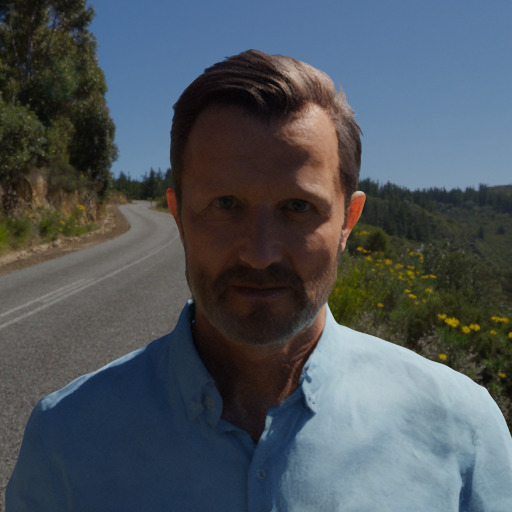} &
    \includegraphics[width=3.6cm]{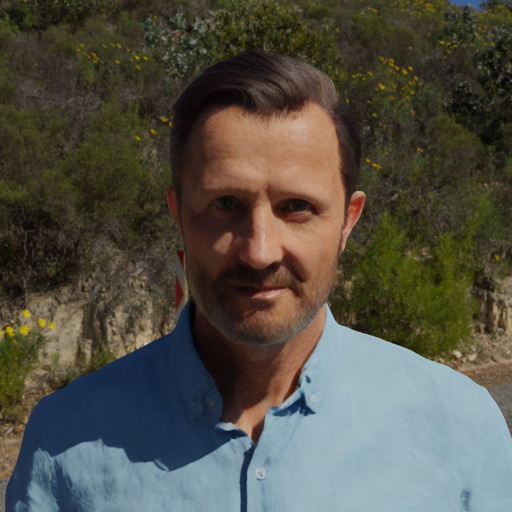} &
    \includegraphics[width=3.6cm]{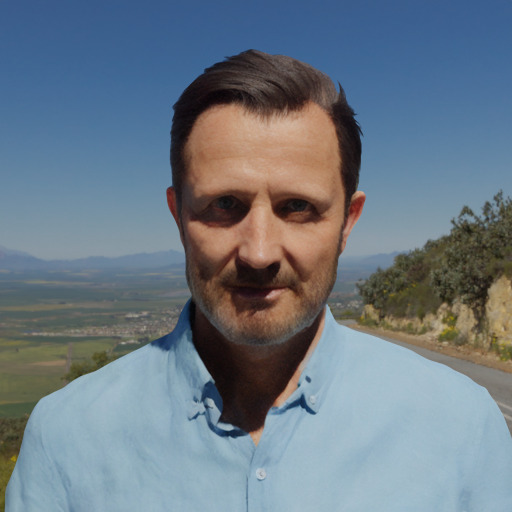} &
    \includegraphics[width=3.6cm]{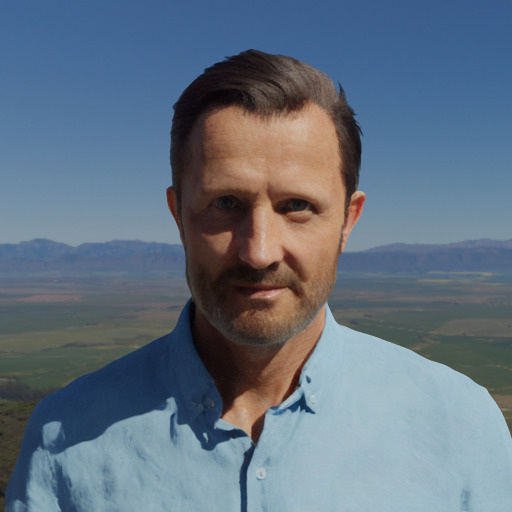} \\
    \raisebox{16mm}{\makecell[c]{IC-Light}} &
    \includegraphics[width=3.6cm]{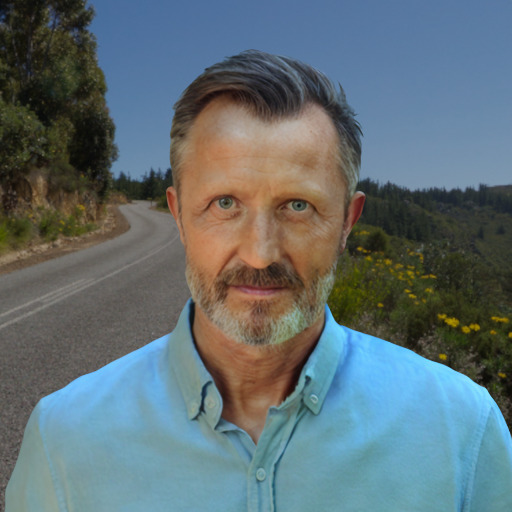} &
    \includegraphics[width=3.6cm]{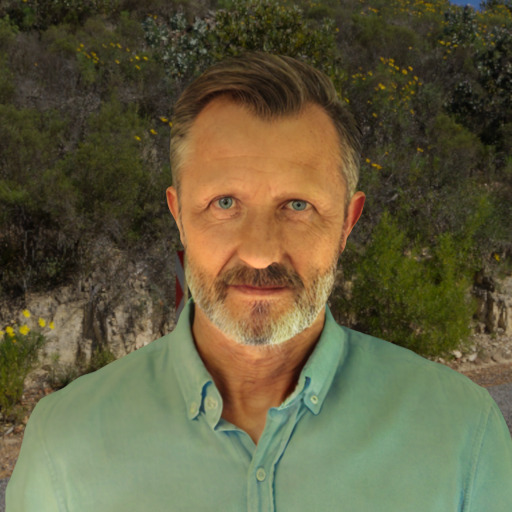} &
    \includegraphics[width=3.6cm]{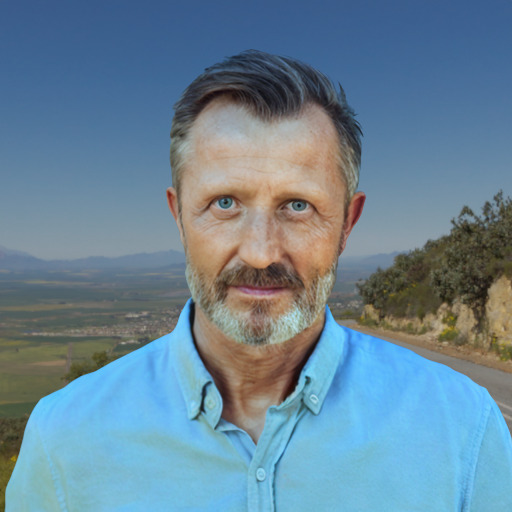} &
    \includegraphics[width=3.6cm]{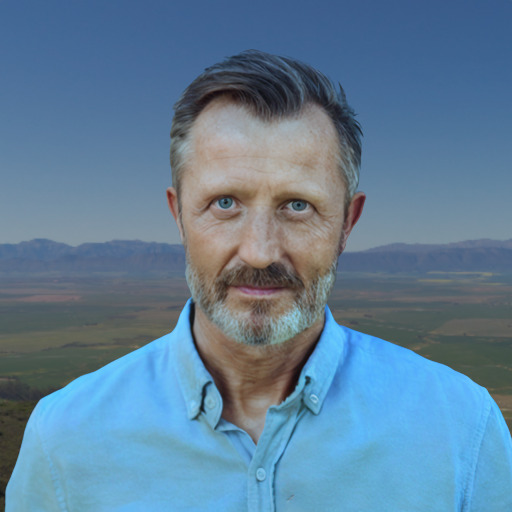} \\
    \end{tabular}

    \caption{\textit{Background vs Environment Map as Lighting Condition}: We compare SynthLight with a background conditioned model and IC-Light and show a reference model rendered in blender (top row). Background contains insufficient lighting cues, causing a background conditioned model to generate inaccurate lighting (columns 3-4). By leveraging our synthetic dataset, the background conditioned model can still generate lighting effects like strong cast shadows, whereas harmonization methods, for example, IC-Light can neither reproduce these effects or relight accurately.}
    \label{fig:bg_v_env_extra}
\end{figure*}

In \cref{fig:bg_v_env_extra}, we compare SynthLight, trained on our synthetic physically-based rendered data using environment maps with comprehensive 360° lighting information, to a background-conditioned variant of SynthLight, and IC-Light. SynthLight excels at capturing nuanced lighting effects, such as cast shadows and self-occlusion, due to its precise environmental lighting inputs. The background-conditioned model, while able to generate these lighting effects, generates inaccurate lighting. IC-Light, an image harmonisation method, neither generates these effects nor generates accurate lighting.

\section{Dataset}
\label{sec:appendix_dataset}

\begin{figure*}[!htbp]
    \centering
    \begin{tabular}{@{\hskip 0mm}c@{\hskip 0.5mm}cc@{\hskip 0.5mm}cc@{\hskip 0.5mm}cc@{\hskip 0.5mm}c@{\hskip 0mm}}
        \includegraphics[width=0.11\textwidth]{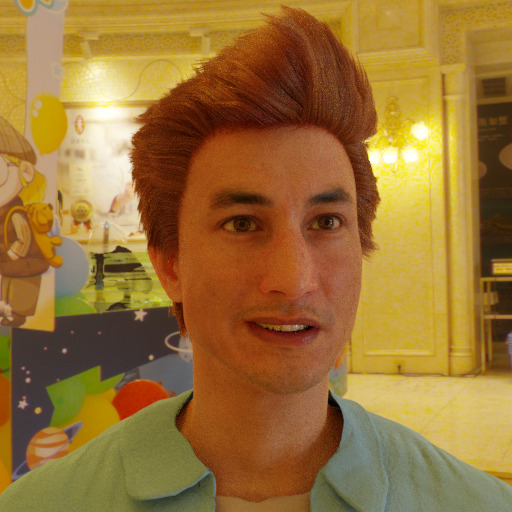} &
        \includegraphics[width=0.11\textwidth]{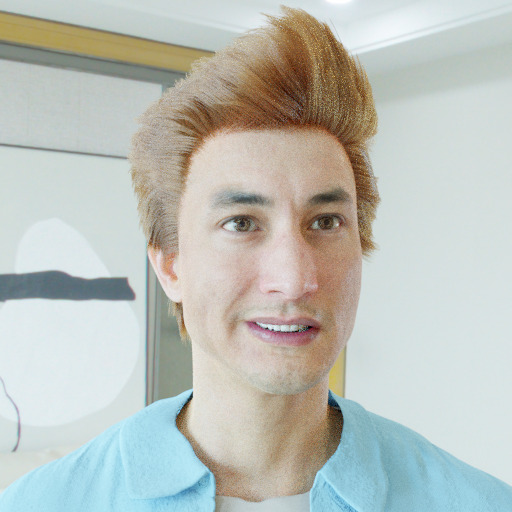} &
        \includegraphics[width=0.11\textwidth]{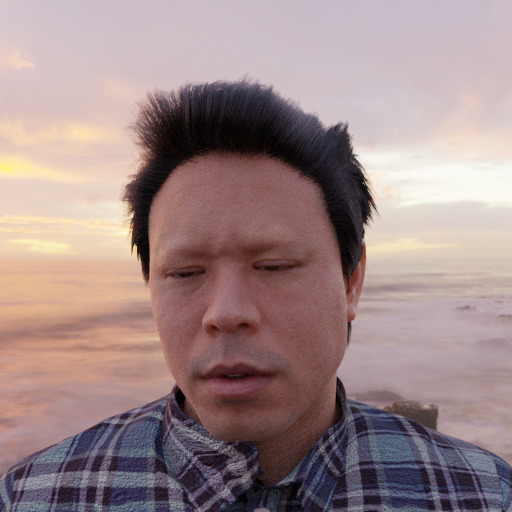} &
        \includegraphics[width=0.11\textwidth]{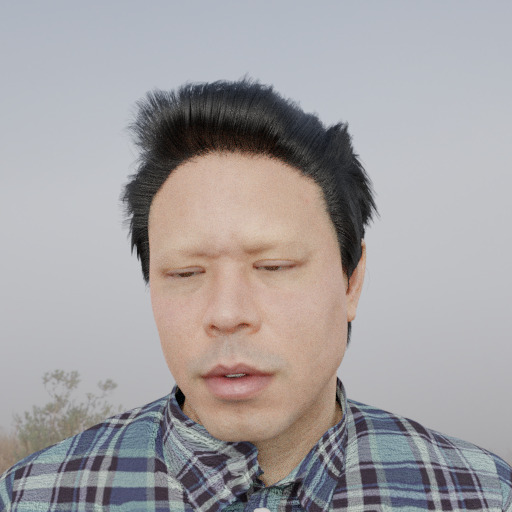} &
        \includegraphics[width=0.11\textwidth]{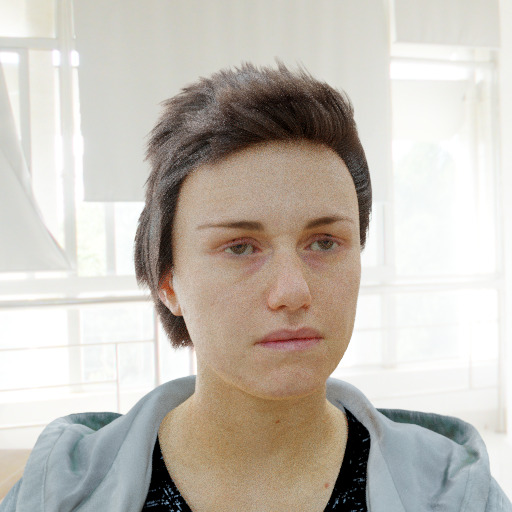} &
        \includegraphics[width=0.11\textwidth]{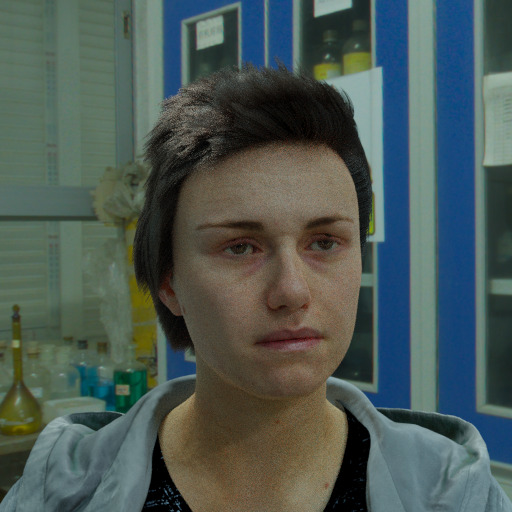} &
        \includegraphics[width=0.11\textwidth]{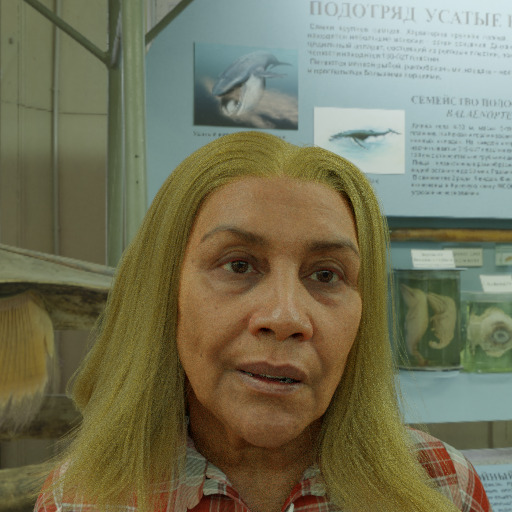} &
        \includegraphics[width=0.11\textwidth]{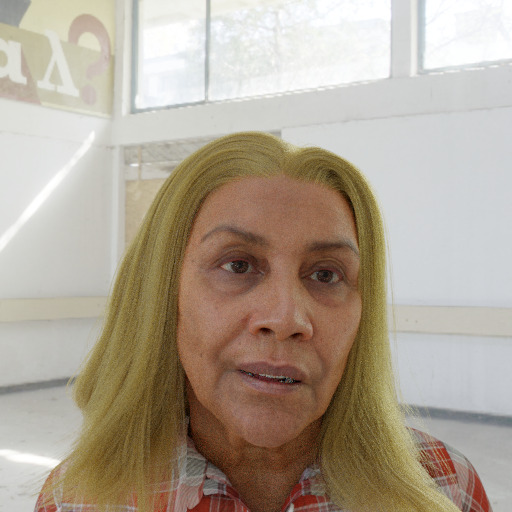} \\

        \includegraphics[width=0.11\textwidth]{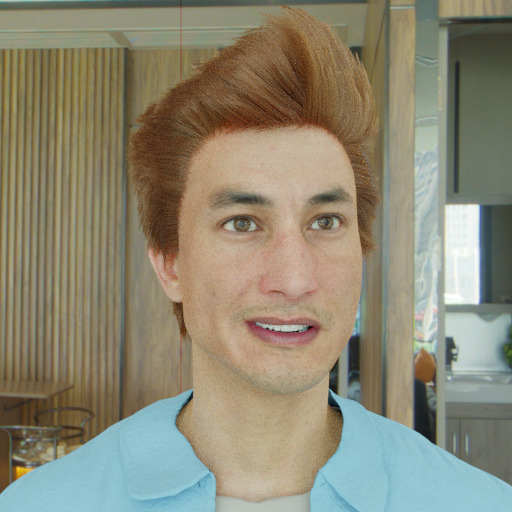} &
        \includegraphics[width=0.11\textwidth]{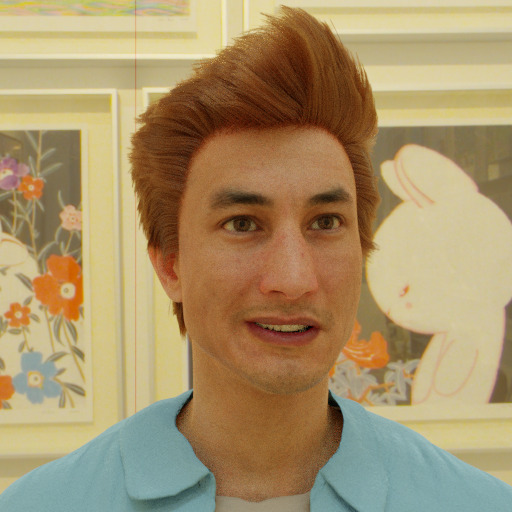} &
        \includegraphics[width=0.11\textwidth]{figures/syn_data_samples/row_11/img_02.jpg} &
        \includegraphics[width=0.11\textwidth]{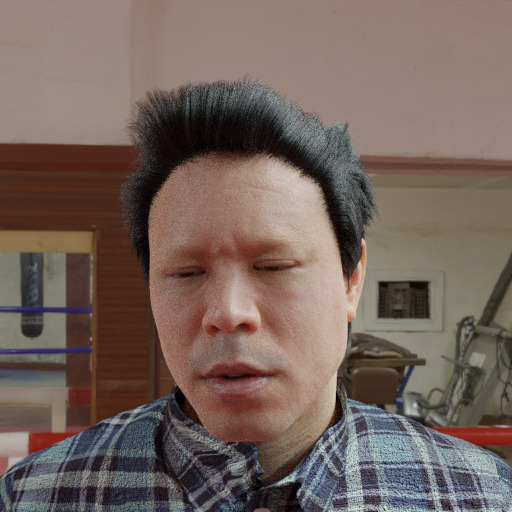} &
        \includegraphics[width=0.11\textwidth]{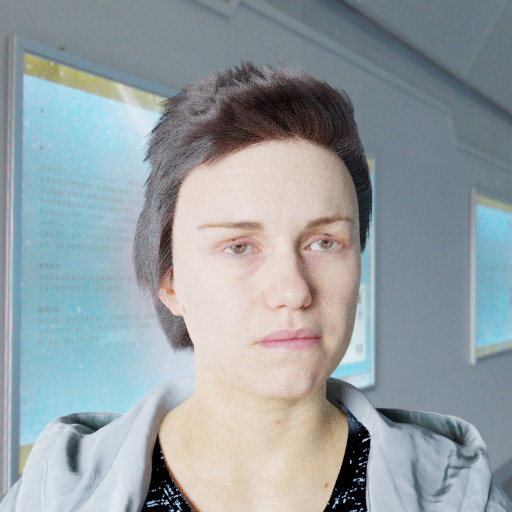} &
        \includegraphics[width=0.11\textwidth]{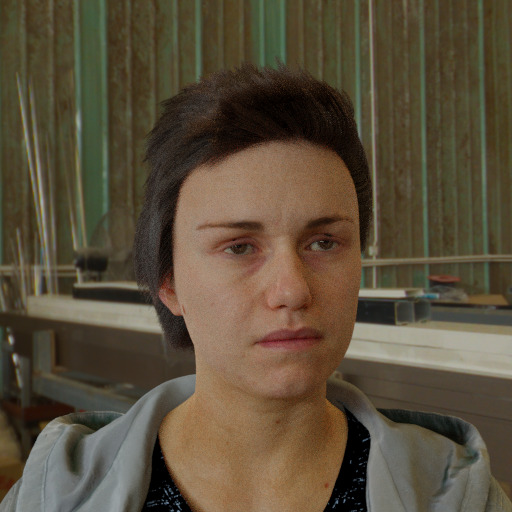} &
        \includegraphics[width=0.11\textwidth]{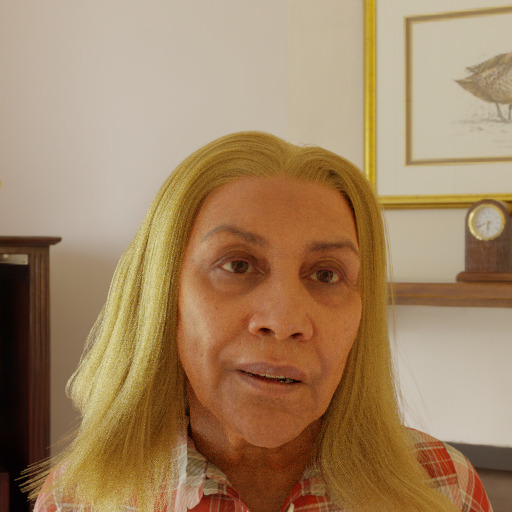} &
        \includegraphics[width=0.11\textwidth]{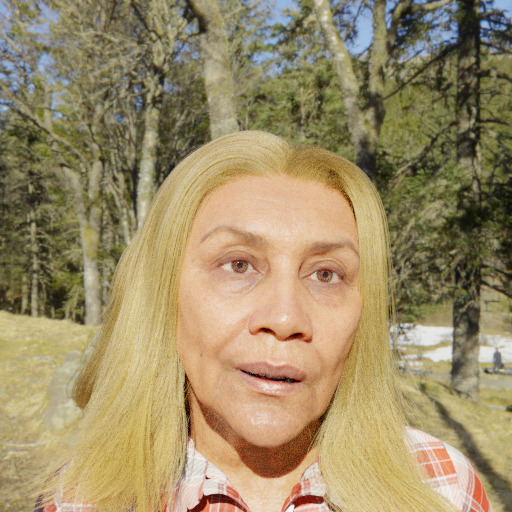} \\

        \includegraphics[width=0.11\textwidth]{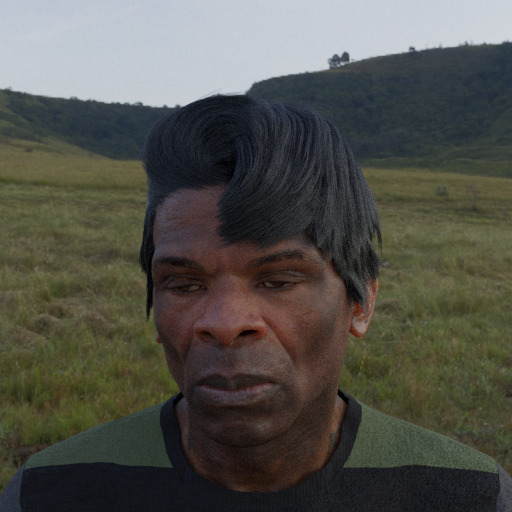} &
        \includegraphics[width=0.11\textwidth]{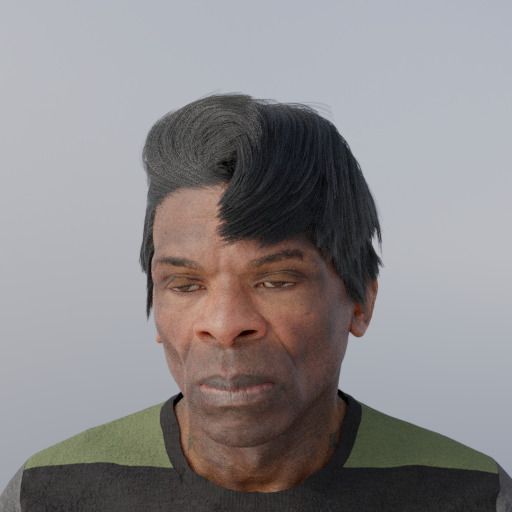} &
        \includegraphics[width=0.11\textwidth]{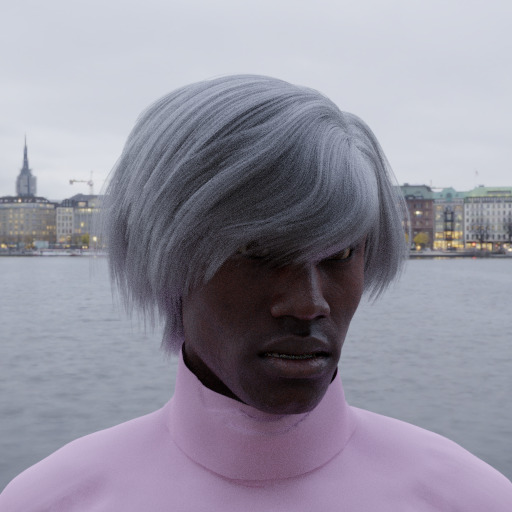} &
        \includegraphics[width=0.11\textwidth]{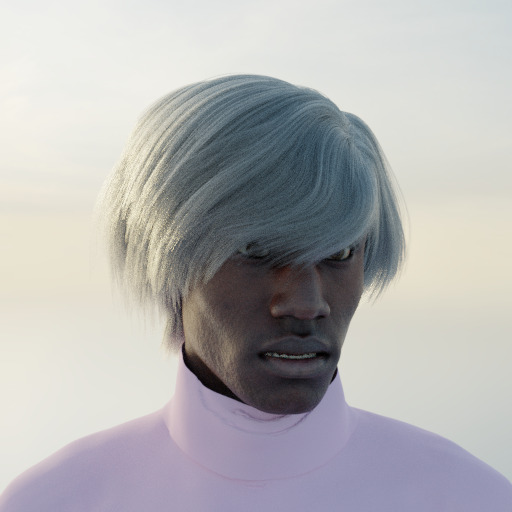} &
        \includegraphics[width=0.11\textwidth]{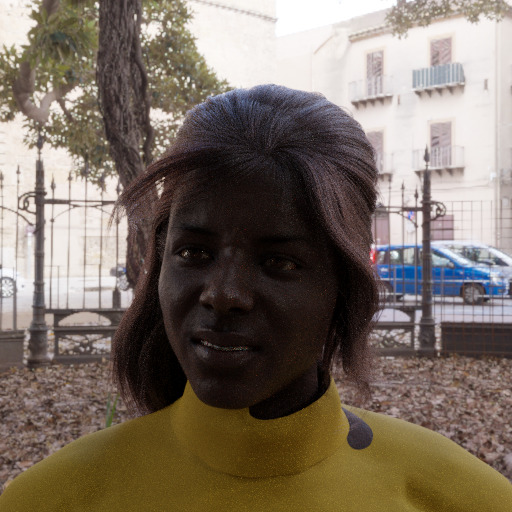} &
        \includegraphics[width=0.11\textwidth]{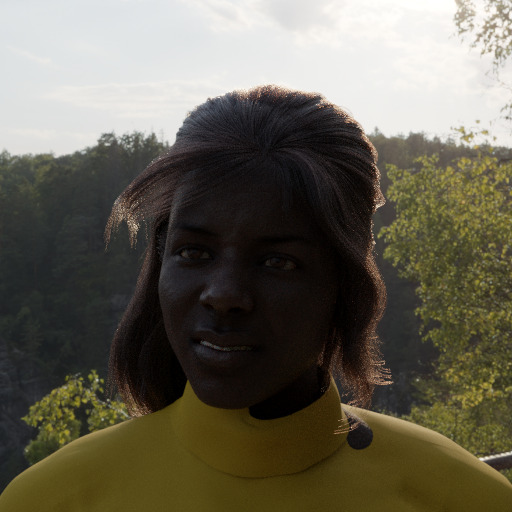} &
        \includegraphics[width=0.11\textwidth]{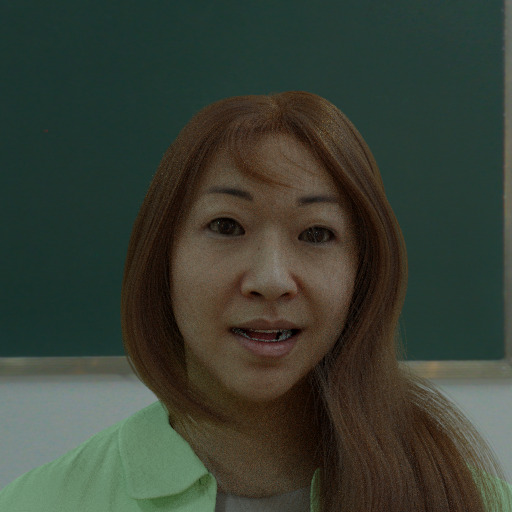} &
        \includegraphics[width=0.11\textwidth]{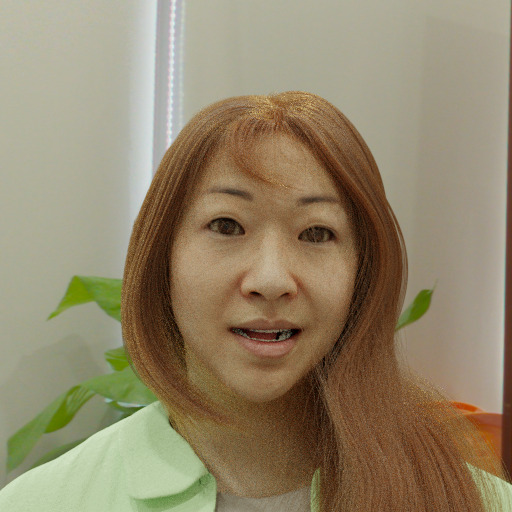} \\

        \includegraphics[width=0.11\textwidth]{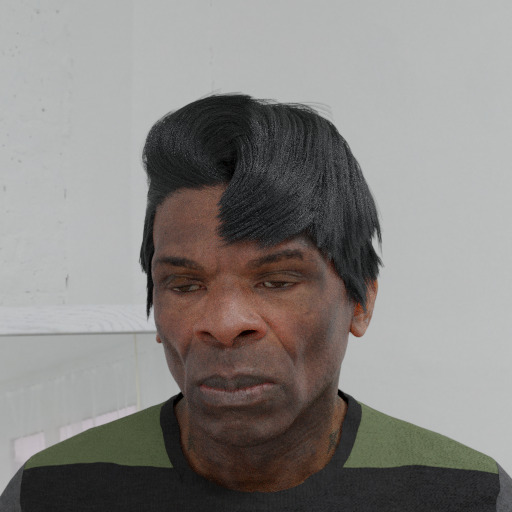} &
        \includegraphics[width=0.11\textwidth]{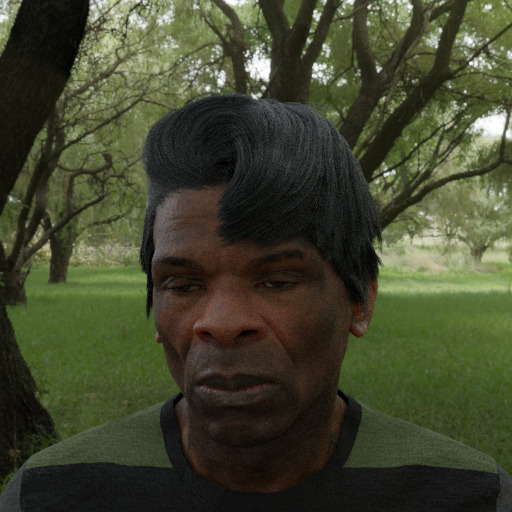} &
        \includegraphics[width=0.11\textwidth]{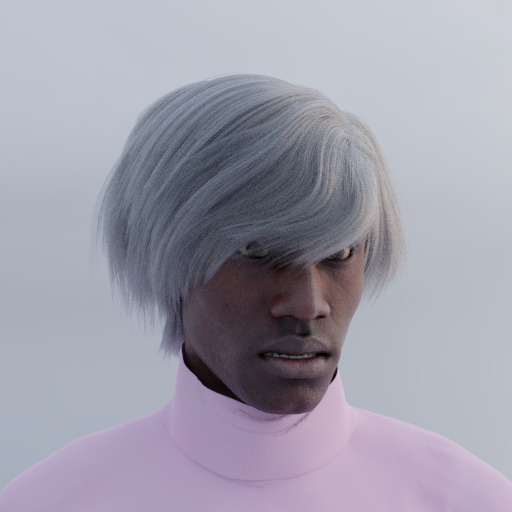} &
        \includegraphics[width=0.11\textwidth]{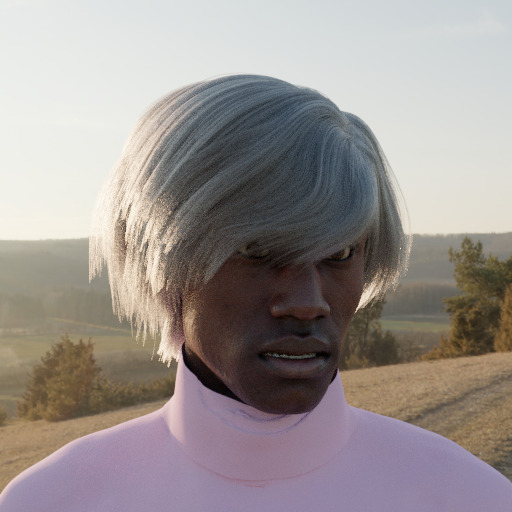} &
        \includegraphics[width=0.11\textwidth]{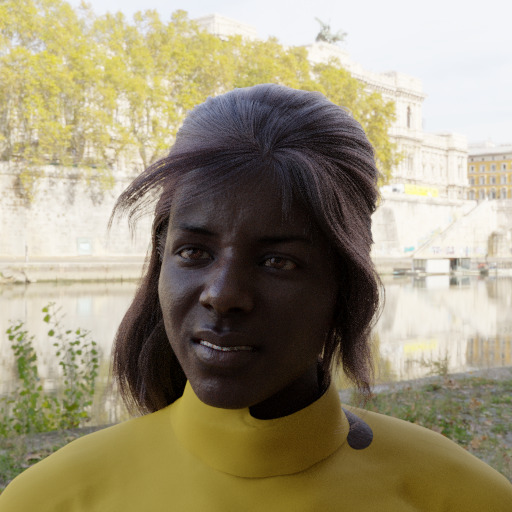} &
        \includegraphics[width=0.11\textwidth]{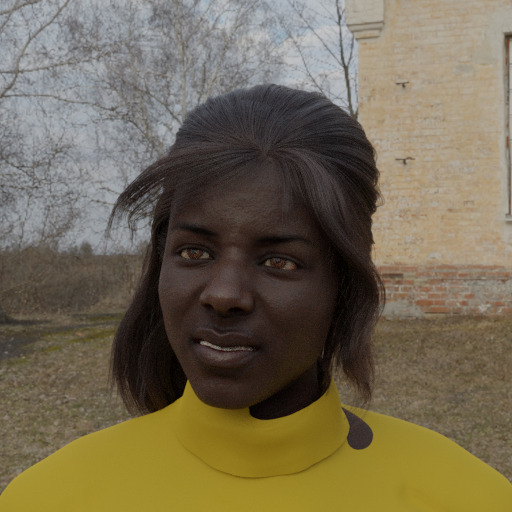} &
        \includegraphics[width=0.11\textwidth]{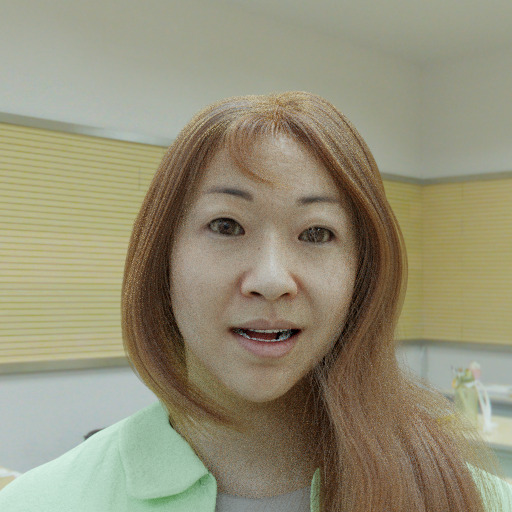} &
        \includegraphics[width=0.11\textwidth]{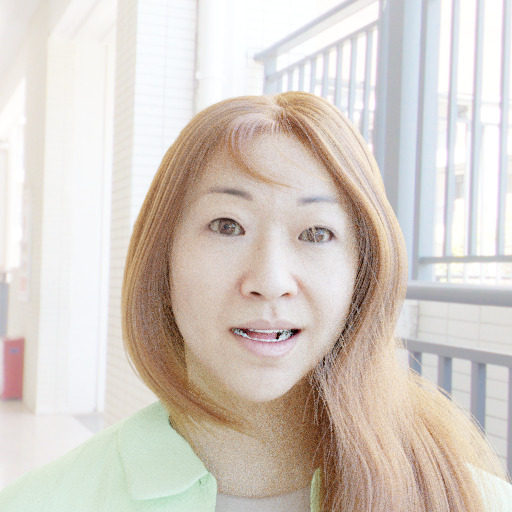} \\
    \end{tabular}
    \caption{\textit{More examples from synthetic dataset}: Each group of four represents a subject rendered under four different lighting conditions.}
    \label{fig:more_examples_dataset}
\end{figure*}

\paragraph{Synthetic Dataset} In \cref{fig:more_examples_dataset} we show more examples from our synthetic dataset of subjects rendered under different environment maps. Each group of 4 visualizes a subject rendered under 4 lighting conditions, highlighting variety across race and gender.   

\vspace{-4mm}
\paragraph{LAION Data Filtration} We filter a subset of LAION by first running a face detector. Since this results in a large number of false positives, we additionally curate a set of query phrases whose matching images we seek to avoid. We filter the set of images further by evaluating the CLIP score of each image against the query words and retaining only those images whose CLIP score is below a threshold. Emperically, we set this threshold to 0.15. 

\section{Additional Implementation Details}

\paragraph{Network Architecture} The inputs to SynthLight are a portrait image and an environment map, both with a resolution of \(512 \times 512\). The environment map is transformed from high-dynamic range to low-dynamic range through the following sequence of operations: clipping, normalization, and exponentiation by \(\frac{1}{2.2}\). These inputs are encoded into latents of shape \(64 \times 64 \times 4\) using the VAE from Stable Diffusion.

SynthLight extends Stable Diffusion 1.5 by adding 8 additional channels to the first convolutional layer of the U-Net, yielding a total of 12 channels (4 each for the denoising latent, input portrait, and environment map). The weights for these extra channels are initialized to 0.

\paragraph{Training and Inference} We evaluate the performance of training with SD 1.5 initialization compared to IC-Light initialization (see \cref{tab:ablation_checkpoint} and \cref{fig:with_and_without_iclight_ckpt}). While IC-Light initialization yields slightly better test set performance—prompting us to report it as our primary method—our approach is not reliant on IC-Light. As shown in \cref{fig:with_and_without_iclight_ckpt}, even without IC-Light, our method generates advanced lighting effects, such as strong cast shadows and subsurface scattering in the ear. Conversely, without our training and inference procedures, IC-Light alone cannot produce the nuanced lighting effects (e.g. rim-effects, subsurface scattering and specular highlights) as illustrated in \cref{fig:comparison_all_extra_special} and \cref{fig:comparison_all_extra_special_another}.

During training, a foreground mask is applied to the input portrait. Each condition—input portrait, environment map, and text prompt—is randomly dropped with a probability of 0.1. For inference, classifier-free guidance is applied with \(\lambda_I = 3\) and \(\lambda_T = 2\), using the prompt "A nice person."

\vspace{-4mm}   
\paragraph{Ablation Details} \textit{Base} serves as the baseline model, trained solely on the synthetic dataset. During inference, it omits \textit{inference time adaptation}, meaning no classifier-free guidance is applied to the input portrait. \textit{Base + Multi-Task} incorporates additional training with LAION data using a text-to-image task, where the input portrait and environment maps are randomly dropped. The relighting and text-to-image tasks are mixed in a 7:3 ratio. \textit{Base + Inference time Adaptation} applies classifier-free guidance on input portrait, while keeping the same training configuration as \textit{Base}. Finally, \textit{Ours} combines both strategies. We train an additional model where Light Stage-rendered data complement the synthetic dataset for relighting -- \textit{Ours + Light Stage}. 

\section{User Study}

\begin{figure*}[!htbp]
    \centering

    \begin{tabular}{@{\hskip 0mm}c@{\hskip 0mm}}
        \includegraphics[width=0.8\textwidth]{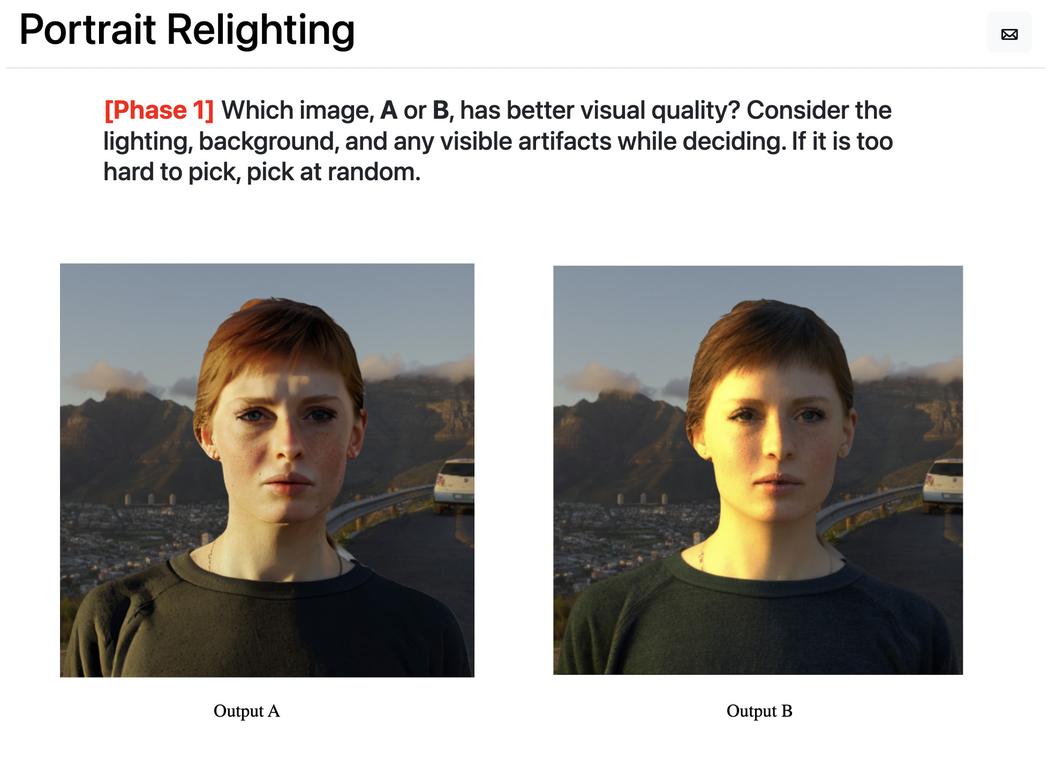} \\
        \includegraphics[width=0.8\textwidth]{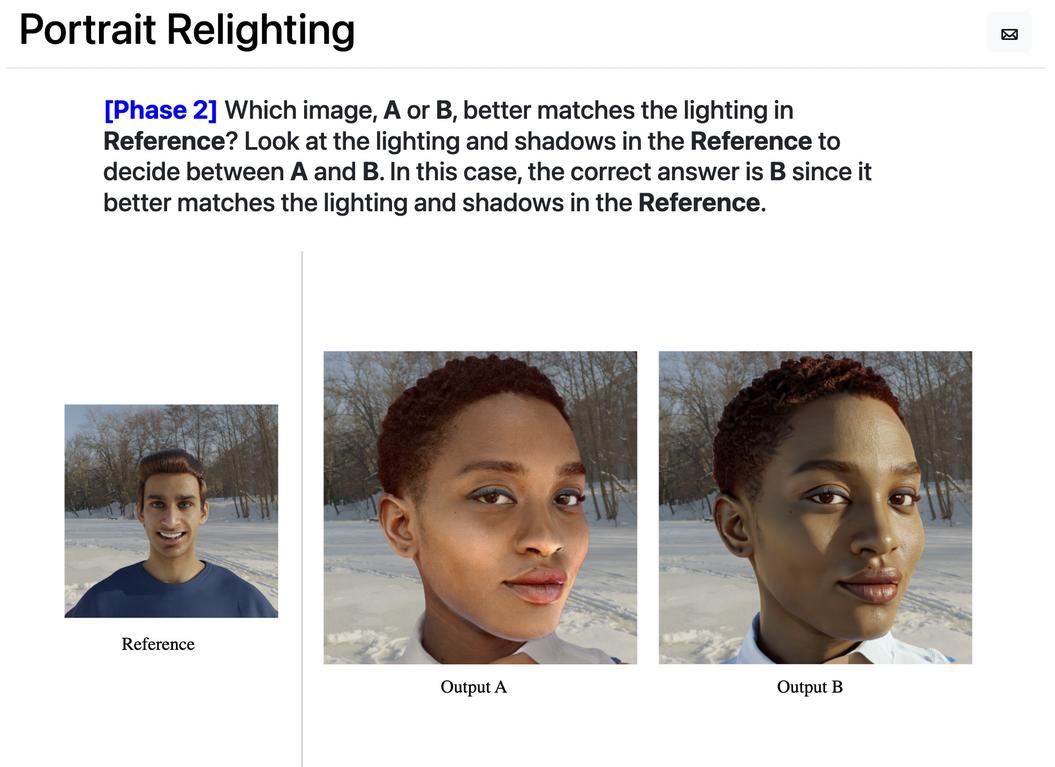} \\
    \end{tabular}
    \vspace{-4mm}
    \caption{\textit{User Study}: We ask users to pick between our method and baseline on visual quality of image (top) and lighting, with a given reference (bottom).}
    \label{fig:user_study_a}
\end{figure*}

\begin{figure*}[!htbp]
    \centering
    \begin{tabular}{@{\hskip 0mm}c@{\hskip 0mm}}
        \includegraphics[width=0.8\textwidth]{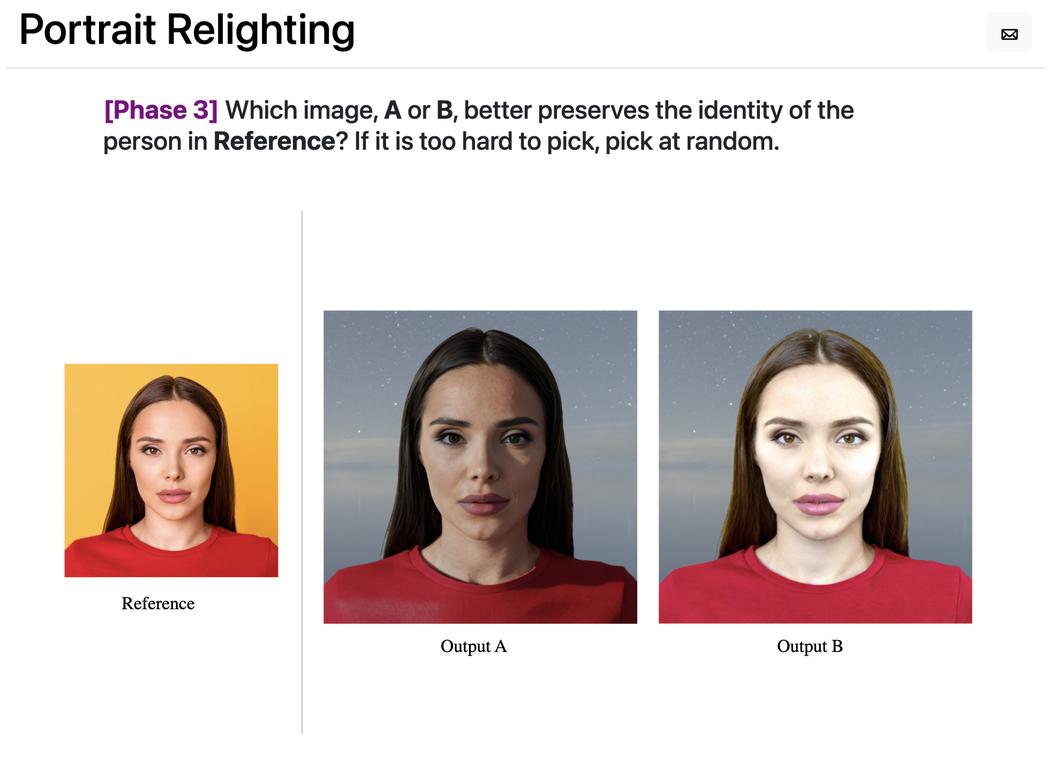} \\
    \end{tabular}
    \vspace{-4mm}
    \caption{\textit{User Study}: We ask users to judge identity preservation by providing a reference identity and asking them to select between our method and baseline.}
    \label{fig:user_study_b}
\end{figure*}

We provide additional details about our user study. Screenshots illustrating the setup can be found in \cref{fig:user_study_a} and \cref{fig:user_study_b}. The user study is conducted in three phases, with each phase focusing on a specific aspect of evaluation:

\vspace{-4mm}
\paragraph{Phase 1: Visual Quality}
In the first phase, participants are asked to specify their preference between our method and the baseline in terms of \textit{visual quality}. Each comparison is presented as a two-option forced choice.

\vspace{-4mm}
\paragraph{Phase 2: Lighting}
In the second phase, participants evaluate the \textit{lighting} of the renderings. To aid their judgment, we provide a synthetic reference rendered in Blender under the same environment map. This phase also uses a two-option forced choice format.

\vspace{-4mm}
\paragraph{Phase 3: Identity}
In the final phase, participants assess the \textit{identity} of the renderings. A reference input portrait is provided, and users judge which option better preserves the subject's identity. As with the previous phases, this is conducted as a two-option forced choice task.

\vspace{-4mm}
\paragraph{General Instructions}
Participants are instructed to choose at random if making a selection is too difficult. At the beginning of each phase, a tutorial question is presented, where the answer is obvious. For example, in these cases:
\begin{itemize}
    \item One example has severe degradation in visual quality.
    \item The lighting in one example is clearly incorrect.
    \item One rendering fails to match the reference identity.
\end{itemize}
The correct answer and the reasoning are explained to participants to familiarize them with the task.

\paragraph{Study Statistics}
The study consists of 30 questions in total, including three tutorial questions (one per phase). Participants can opt to exit the study at any time. In total, we collected 482 responses from 20 participants over a one-week period.

\section{Limitations}

\begin{figure*}[htbp]
    \centering

    \begin{subfigure}[b]{\textwidth}
        \centering
        \includegraphics[width=0.45\textwidth]{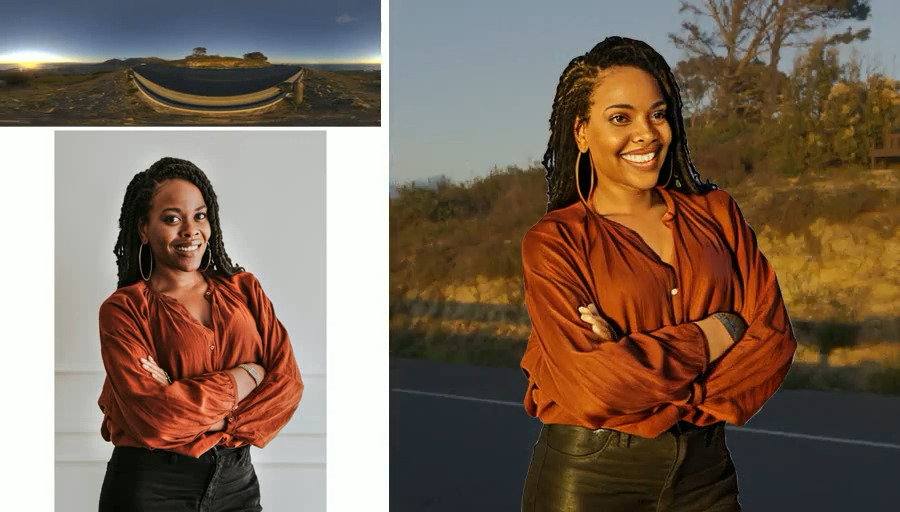}
        \hfill
        \includegraphics[width=0.45\textwidth]{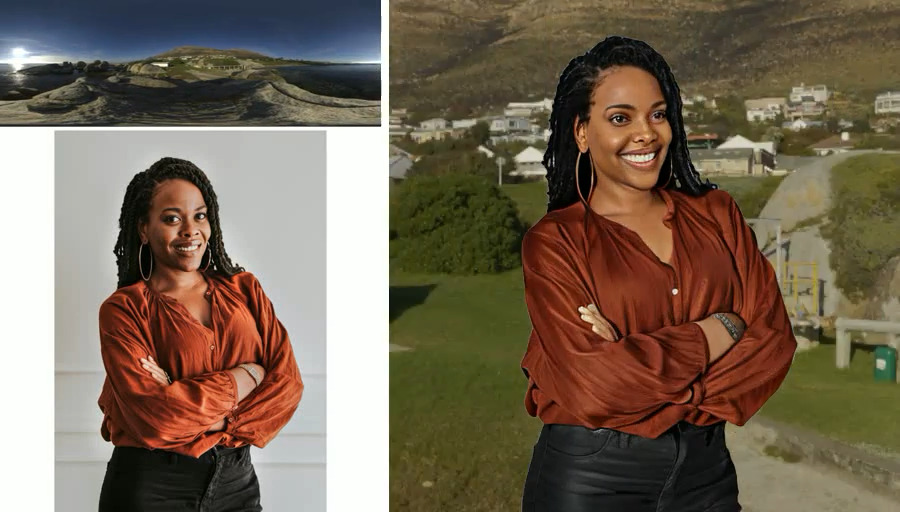}
        \label{fig:failure_a}
        \caption{We observe \textit{minor detail loss} in facial features, such as the eyes, arising from limited camera pose diversity and Stable Diffusion 1.5's VAE artifacts. Mitigations include using improved VAEs (e.g., Flux, Stable Diffusion 3) and enhancing pose variation in the dataset.}
    \end{subfigure}
    \vskip\baselineskip 

    \begin{subfigure}[b]{\textwidth}
        \centering
        \includegraphics[width=0.45\textwidth]{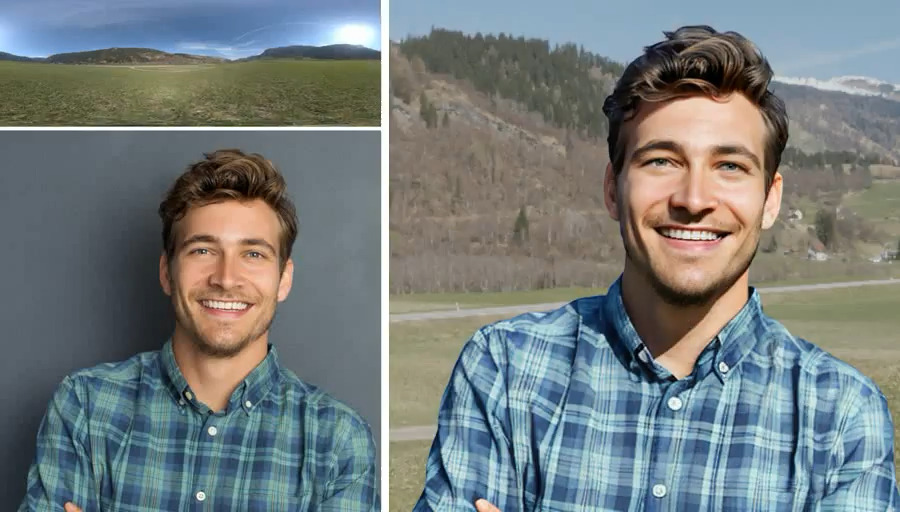}
        \hfill
        \includegraphics[width=0.45\textwidth]{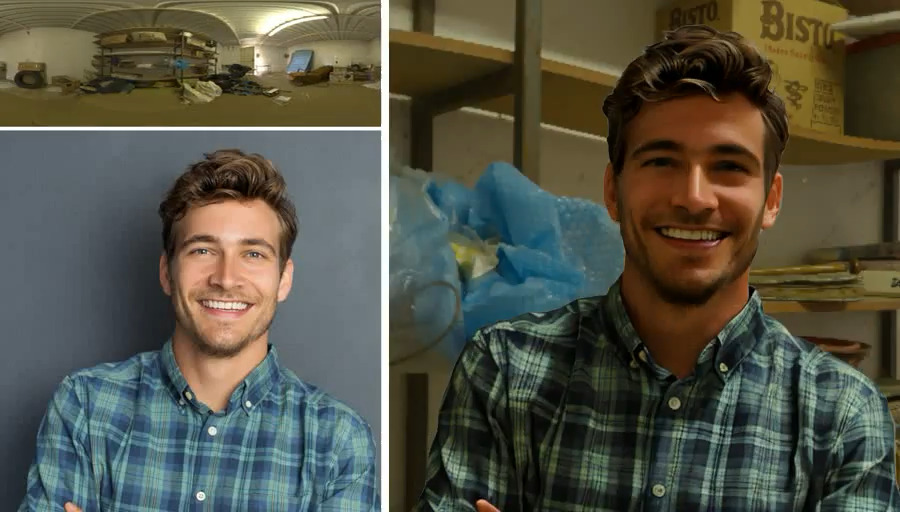}
        \label{fig:failure_b}
    \end{subfigure}
    
    \caption{Limitations of our method include minor detail loss in full-body portraits and inaccuracies in cloth texture.}
    \label{fig:failure_cases}
\end{figure*}

\cref{fig:failure_cases} highlights some limitations observed with our method. We notice minor loss of detail, particularly in small or intricate facial features. This can be attributed to limited camera pose diversity in our synthetic dataset, i.e. headshot-only renderings, and the reliance on Stable Diffusion 1.5, which causes our method to inherit image reconstruction artifacts from Stable Diffusion's VAE. These issues can be mitigated by leveraging larger models with with better VAEs, such as those in Flux or Stable Diffusion 3, and incorporating greater camera pose variation in our synthetic dataset.

\cref{fig:failure_cases} illustrate another failure mode where our method struggles with accurately capturing cloth textures. While this limitation is rare, it arises from the restricted range of materials and textures used for clothing in the synthetic dataset. Expanding the diversity and quality of the dataset’s cloth-related materials could effectively address this issue.

\end{document}